\documentclass[12pt,a4paper]{article} 
\usepackage{amsfonts, amsmath, amssymb}
\usepackage{longtable}
\usepackage{url}
\usepackage[pdftex]{graphicx}
\usepackage[usenames,dvipsnames,svgnames]{xcolor}

\newtheorem{proposition}{Proposition}
\newtheorem{definition}{Definition}
\newtheorem{lemma}{Lemma}
\newtheorem{theorem}{Theorem}

\newcommand{\comment}[1]{}

\begin{document}

%: Title -----------------------------------------------
\thispagestyle{empty} {\vspace*{-2.8cm} \hspace*{7.5cm}
\includegraphics[width=75mm,viewport=0 0 235
85,clip]{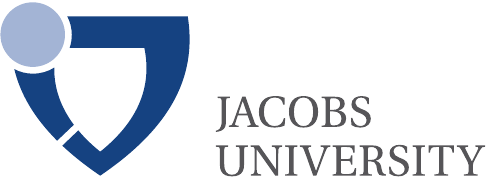}} \vspace{6.5cm}

% Author
{\noindent \large \textsf{Herbert Jaeger}} \\ \\

% Title
{\noindent \LARGE \bf \textsf{Controlling Recurrent Neural Networks by Conceptors}} \\ \vfill

% Report number
{\noindent \LARGE \textsf{Technical Report No. 31}} \\
% Month and year (e.g. January 2005)
{\noindent \textsf{March 2014}} \\ {\hspace*{-3.17cm}
\rule[3mm]{\textwidth}{0.75pt}} \\ {\LARGE \textsf{School of
Engineering and Science}} \\
% ----------------------------------------------------

\newpage \thispagestyle{empty}

{ 
\noindent 
\huge {\bf Controlling Recurrent Neural Networks by Conceptors (Revision 4)} \\

\normalsize 

\vspace{1cm} 

\noindent 
{\bf Herbert Jaeger}\\ \\ {\it Original affiliation: Jacobs University Bremen\\ School of
Engineering and Science\\ Campus Ring, 28759 Bremen, Germany\\ \\
Current affiliation (since 2019): University of Groningen\\
Department of AI and Groningen Cognitive Systems and Materials Center
(CogniGron)\\
Nijenborgh 9, 9747 AG Groningen, Netherlands\\
email: h.jaeger@rug.nl\\
web: https://www.ai.rug.nl/minds
\\
}

\begin{abstract} 
  The human brain is a dynamical system whose extremely complex
  sensor-driven neural processes give rise to conceptual, logical
  cognition. Understanding the interplay between nonlinear neural
  dynamics and concept-level cognition remains a major scientific
  challenge. Here I propose a mechanism of neurodynamical
  organization, called \emph{conceptors}, which unites nonlinear dynamics
  with basic principles of conceptual abstraction and logic. It becomes possible to
  learn, store, abstract, focus, morph, generalize, de-noise
  and recognize a large number of dynamical patterns within a
  single neural system; novel patterns can be added
  without interfering with previously acquired ones; neural noise is
  automatically filtered. Conceptors help explaining how
  conceptual-level information processing emerges naturally and
  robustly in neural systems, and remove a number of roadblocks in the theory and
   applications of recurrent neural networks.  
\end{abstract}

\newpage

\noindent{\bf Changelog for Revision 1:} Besides correcting trivial typos, the following edits / corrections have been made
after the first publication on arXiv, leading to the revision 1
version on arXiv:

\begin{itemize}
\item Apr 7, 2014: updated algorithm description for
  incremental memory management by documenting $W^{\mbox{\scriptsize
  out}}$ component.  
\item Jun 15, 2014: added
  missing parentheses in $C$ update formula given at beginning of
  Section 3.14.
\item Oct 11, 2014: added missing Frobenious norms in similarity
formulas (3) and (10) (pointed out by Dennis Hamester)
\item May 2016: re-wrote and extended the section on incremental
memory management (Section \ref{subsec:memmanage})
\item July 2016: deleted the subsubsection ``To be or not to be an
attractor'' from Section  3.14.3 because this material is repeated later in the
text in more detail. 
\end{itemize}

\noindent{\bf Changelog for Revision 2:}

\begin{itemize}
\item April 22, 2017: Proposition 12 was removed - it was wrong. 
\end{itemize}

\noindent{\bf Changelog for Revision 4:}
November 17, 2024: There were a number of inconsistencies concerning
the use of terminology 'input simulation weights' versus 'input
recreation weights', where text wording was at odds with notation in
formulas. This has been corrected. I was alerted to these problems by
my Master student Otto Bervoets. Furthermore, two notes pointing out
recent developments with regards to using conceptors for continual
deep learning and a formal conceptor logic were added.

\newpage
\paragraph*{Notes on the Structure of this Report.} This report
introduces several novel analytical concepts describing neural
dynamics; develops the corresponding mathematical theory under aspects
of linear algebra, dynamical systems theory, and formal logic;
introduces a number of novel learning, adaptation and control
algorithms for recurrent neural networks; demonstrates these in a
number of case studies; proposes biologically (not too im-)plausible
realizations of the dynamical mechanisms; and discusses relationships
to other work. Said shortly, it's long. Not all parts
will be of interest to all readers. In order to facilitate navigation
through the text and selection of relevant components, I start with an
overview section which gives an intuitive explanation of the novel
concepts and informal sketches of the main results and demonstrations
(Section 1). After this overview, the material is presented in detail,
starting with an introduction (Section 2) which relates this
contribution to other research. The main part is Section 3, where I
systematically develop the theory and algorithms, interspersed with
 simulation demos. A graphical dependency map for this
section is given at the beginning of Section 3. The technical documentation of the
computer simulations is provided in Section 4, and mathematical proofs
are collected in Section 5. The detailed presentation in Sections 2 --
5 is self-contained. Reading the overview in Section 1 may be helpful but is not
necessary for reading these sections. For convenience some figures
from the overview section are repeated in Section 3. 

\paragraph*{Acknowledgements.} The work described in this report was
partly funded through the European FP7 project AMARSi
(www.amarsi-project.eu). The author is indebted to Dr. Mathieu Galtier
and Dr. Manjunath Ghandi for careful proofreading (not an easy task). 

\newpage

\tableofcontents

\newpage

\section{Overview}\label{secOverview}

\paragraph*{Scientific context.} Research on brains and cognition unfolds in two directions.
\emph{Top-down} oriented research starts from the ``higher'' levels of
cognitive performance, like rational reasoning, conceptual knowledge
representation,  command of language.
These phenomena are typically described in symbolic formalisms developed in
mathematical logic, artificial intelligence (AI), computer science and
linguistics. In the \emph{bottom-up} direction, one departs from
``low-level'' sensor data processing and motor control, using the
analytical tools offered by dynamical systems theory, signal
processing and control theory, statistics and information theory. The
human brain obviously has found a way to implement high-level logical
reasoning on the basis of low-level neuro-dynamical processes.  How this
is possible, and how the top-down and bottom-up research directions
can be united,  has largely remained an open question despite
long-standing efforts in  neural networks research and
computational neuroscience
\cite{Palm80,Rabinovich08,Friston05,Abbott08,Gerstneretal12,Grossberg13},
machine learning \cite{GedeonArathorn07,HintonSalakhutdinov06},
robotics \cite{Brooks89,PfeiferScheier98}, artificial intelligence
\cite{Pollack90,vanderVeldedeKamps06,Baderetal08,Borgesetal11},
dynamical systems modeling of cognitive processes
\cite{SchoenerKelso88,SmithThelen93a,Gelder98}, cognitive science and
linguistics \cite{Drescher91,Shastri99a}, or cognitive neuroscience
\cite{Andersonetal04,Eliasmithetal12}. 

\paragraph*{Summary of contribution.} Here I establish a fresh view on the neuro-symbolic integration
problem. I show how dynamical neural activation patterns can be
characterized by certain neural filters which I call
\emph{conceptors}. Conceptors derive naturally from the following key
observation. When a recurrent neural network (RNN) is actively
generating, or is passively being driven by different dynamical
patterns (say $a,b,c,\ldots$), its neural states populate different
regions $R_a, R_b, R_c,\ldots$ of neural state space. These regions
are characteristic of the respective patterns. For these regions,
neural filters $C_a, C_b, C_c,\ldots$ (the conceptors) can be
incrementally learnt. A conceptor $C_x$ representing a pattern $x$ can
then be invoked after learning to constrain the neural dynamics to the
state region $R_x$, and the network will select and re-generate
pattern $x$. Learnt conceptors can be blended, combined by Boolean operations,
specialized or abstracted in various ways, yielding novel patterns on
the fly.  
The logical operations on conceptors admit a rigorous semantical
interpretation; conceptors can be arranged in conceptual hierarchies
which are structured like semantic networks known from artificial
intelligence. Conceptors can be economically implemented by single
neurons (addressing patterns by neurons, leading to explicit command
over pattern generation), or they may self-organize spontaneously and
quickly upon the presentation of cue patterns (content-addressing,
leading to pattern imitation). Conceptors can also be employed to
``allocate free memory space'' when new patterns are learnt and stored
in long-term memory, enabling incremental life-long learning without
the danger of freshly learnt patterns disrupting already acquired
ones. Conceptors are robust against neural noise and parameter
variations. The basic mechanisms are  generic and can be
realized in any kind of dynamical neural network. All taken together,
conceptors offer a principled, transparent, and computationally 
efficient account of how neural dynamics can self-organize in
conceptual structures.

\paragraph*{Going bottom-up: from neural dynamics to conceptors.}

The neural model system in this report are standard
 recurrent neural networks (RNNs, Figure \ref{fig1Main}
{\bf A}) whose dynamics is mathematically described be the state
 update equations
\begin{eqnarray*}
x(n+1) & = & \tanh(W^\ast\,x(n) + W^{\mbox{\scriptsize in}}\,p(n)), \\
y(n) & = & W^{\mbox{\scriptsize out}} x(n).
\end{eqnarray*}
Time here progresses in unit steps $n = 1, 2, \ldots$. The network
consists of $N$  neurons (typically in the order of a hundred in
this report), whose activations $x_1(n),\ldots, x_N(n)$ at time $n$
are collected in an $N$-dimensional \emph{state vector} $x(n)$. The
neurons are linked by random synaptic connections, whose strengths are
collected in a \emph{weight matrix} $W^\ast$ of size $N \times N$. An
input signal $p(n)$ is fed to the network through synaptic input
connections assembled in the \emph{input weight} matrix
$W^{\mbox{\scriptsize in}}$. The ``S-shaped'' function $\tanh$
squashes the neuronal activation values into a range between $-1$ and
$1$. The second equation  specifies that an \emph{ouput
  signal} $y(n)$ can be read from the network activation state $x(n)$
by means of \emph{output weights}  $W^{\mbox{\scriptsize
    out}}$.  These weights are pre-computed such that the output signal
$y(n)$ just  repeats the input signal $p(n)$. The output
signal plays no functional role in what follows; it merely serves as a
convenient 1-dimensional observer of the high-dimensional network
dynamics. 

The network-internal neuron-to-neuron connections $W^\ast$ are created
at random. This will lead to the existence of cyclic (``recurrent'')
connection pathways inside the network. Neural activation can
reverberate inside the network along these cyclic pathways. The
network therefore can autonomously generate
complex neurodynamical patterns even when it receives no input.
Following the terminology of the \emph{reservoir computing}
\cite{JaegerHaas04,Appeltantetal11}, I refer to such randomly
connected neural networks as \emph{reservoirs}.

 \begin{figure}[htbp]
  \center
  \includegraphics[width=100 mm]{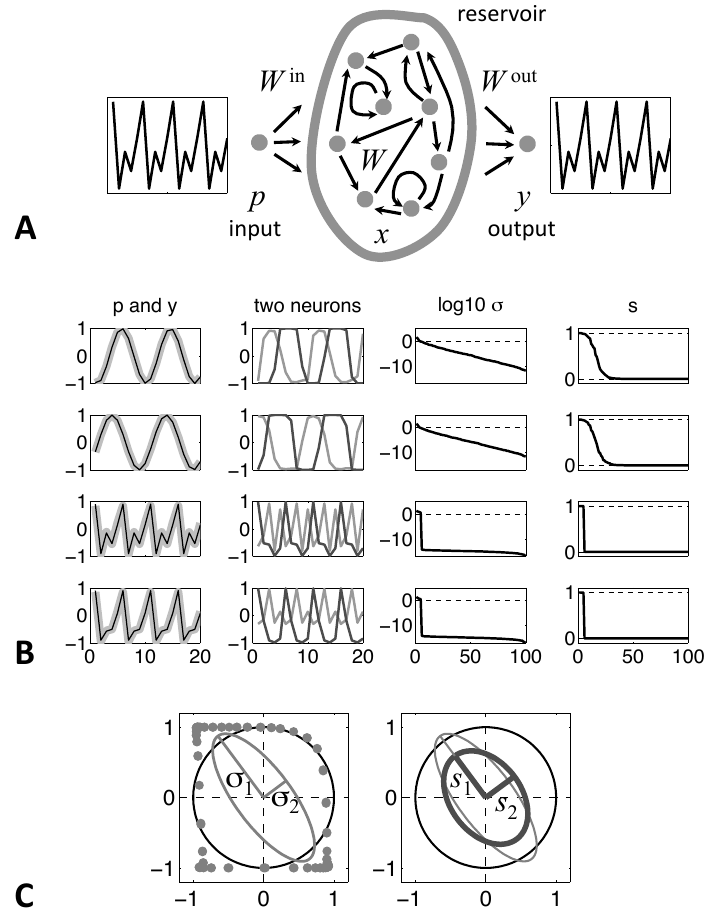}
  \caption{Deriving conceptors from network dynamics. {\bf A.}
    Network layout. Arrows indicate synaptic links.  {\bf B.} Driving
    the reservoir with four different input patterns.  Left panels: 20
    timesteps of input pattern $p(n)$ (black thin line) and
    conceptor-controlled output $y(n)$ (bold light gray).  Second
    column: 20 timesteps of traces $x_i(n), x_j(n)$ of two randomly
    picked reservoir neurons. Third column: the singular values
    $\sigma_i$ of the reservoir state correlation matrix $R$ in
    logarithmic scale.  Last column: the singular values $s_i$ of the
    conceptors $C$ in linear plotting scale. {\bf C.}  From pattern to
    conceptor. Left: plots of value pairs $x_i(n),x_j(n)$ (dots) of
    the two neurons shown in first row of {\bf B} and the resulting
    ellipse with axis lengths $\sigma_1, \sigma_2$.  Right: from $R$
    (thin light gray) to conceptor $C$ (bold dark gray) by normalizing
    axis lengths $\sigma_1, \sigma_2$ to $s_1, s_2$.  }
  \label{fig1Main}
  \end{figure}
  
  For the sake of introducing conceptors by way of an example,
  consider a reservoir with $N = 100$ neurons. I drive this system
  with a simple sinewave input $p(n)$ (first panel in first row in
  Fig.\ \ref{fig1Main} {\bf B}). The reservoir becomes entrained to
  this input, each neuron showing individual variations thereof (Fig.\ 
  \ref{fig1Main} {\bf B} second panel). The resulting reservoir state
  sequence $x(1), x(2),\ldots$ can be represented as a cloud of points
  in the 100-dimensional reservoir state space. The dots in the first
  panel of Fig.\ \ref{fig1Main} {\bf C} show a 2-dimensional projection of this
  point cloud. By a statistical method known as principal component
  analysis, the shape of this point cloud can be captured by an
  $N$-dimensional ellipsoid whose main axes point in the main
  scattering directions of the point cloud.  This ellipsoid is a
  geometrical representation of the \emph{correlation matrix} $R$ of
  the state points. The lengths $\sigma_1,\ldots,\sigma_N$ of the
  ellipsoid axes are known as the \emph{singular values} of $R$. The
  directions and lengths of these axes provide a succinct
  characterization of the geometry of the state point cloud. The $N =
  100$ lengths $\sigma_i$ resulting in this example are log-plotted in
  Fig.\ \ref{fig1Main} {\bf B}, third column, revealing an exponential
  fall-off in this case.

As a next step, these lengths $\sigma_i$ are normalized to become
$s_i = \sigma_i / (\sigma_i + \alpha^{-2})$, where $\alpha \geq 0$ is
a design parameter that I call \emph{aperture}. This normalization
ensures that all $s_i$ are not larger than 1 (last column in Fig.\ 
\ref{fig1Main} {\bf B}).  A new
ellipsoid is obtained (Fig.\ \ref{fig1Main} {\bf C} right) which is
located inside the unit sphere.  The normalized ellipsoid can be
described by a $N$-dimensional matrix $C$, which I call a
\emph{conceptor} matrix. $C$ can be directly expressed in terms of $R$
by $C = R(R+ \alpha^{-2}I)^{-1}$, where $I$ is the identity matrix.
  
When a different driving pattern $p$ is used, the shape of the state
point cloud, and subsequently the conceptor matrix $C$, will be
characteristically different. In the example, I drove the reservoir
with four patterns $p^1$ -- $p^4$ (rows in Fig.\ \ref{fig1Main}{\bf
  B}). The first two patterns were sines of slightly different
frequencies, the last two patterns were minor variations of a
5-periodic random pattern. The conceptors derived from the two sine
patterns differ considerably from the conceptors induced by the two
5-periodic patterns (last column in Fig.\ \ref{fig1Main}{\bf B}).
Within each of these two pairs, the conceptor differences are too
small to become visible in the plots.
  
  There is an instructive alternative way to define conceptors. Given
  a sequence of reservoir states $x(1),\ldots,x(L)$, the conceptor $C$
  which characterizes this state point cloud is the unique matrix
  which minimizes the cost function $\sum_{n=1,\ldots,L}\|x(n) - Cx(n)
  \|^2 / L + \alpha^{-2} \|C \|^2$, where $\|C \|^2$ is the sum of all
  squared matrix entries. The first term in this cost would become
  minimal if $C$ were the identity map, the second term would become
  minimal if $C$ would be the all-zero map. The aperture $\alpha$
  strikes a balance between these two competing cost components. For
  increasing apertures, $C$ will tend toward the
  identity matrix $I$; for shrinking apertures it will come out closer  to the zero matrix. In the terminology of machine learning,
  $C$ is hereby defined as a \emph{regularized identity map}. The explicit solution to
  this minimization problem is again given by the  
  formula $C = R\,(R + \alpha^{-2}I)^{-1}$. 

Summing up: if a reservoir is driven by a pattern $p(n)$, a conceptor
matrix $C$ can be obtained from the driven reservoir states $x(n)$ as the
regularized identity map on these states. $C$ can be likewise seen as
a normalized ellipsoid characterization of the shape of the $x(n)$ point
cloud.  I  write
$C(p,\alpha)$ to denote a conceptor derived from a pattern $p$ using
aperture $\alpha$, or $C(R,\alpha)$ to denote that $C$ was obtained
from a state correlation matrix $R$. 

\paragraph*{Loading a reservoir.} With the aid of conceptors a
reservoir can re-generate a number of different patterns
$p^1,\ldots,p^K$ that it has previously been driven with. For this to
work, these patterns have to be learnt by the reservoir in a special
sense, which I call \emph{loading} a reservoir with patterns.  The
loading procedure works as follows.  First, drive the reservoir with
the patterns $p^1,\ldots,p^K$ in turn, collecting reservoir states
$x^j(n)$ (where $j = 1,\ldots,K$). Then, recompute the reservoir
connection weights $W^\ast$ into $W$ such that $W$ optimally balances
between the following two goals. First, $W$ should be such that
$W\,x^j(n) \approx W^\ast\,x^j(n) + W^{\mbox{\scriptsize in}} p^j(n)$
for all times $n$ and patterns $j$. That is, $W$ should allow the
reservoir to ``internalize'' the driving input in the absence of the
same. I call such $W$ an \emph{input internalizing} matrix.  Second, $W$ should be such that the weights collected in this
matrix become as small as possible.  Technically this
compromise-seeking learning task amounts to computing what is known as
a regularized linear regression, a standard and simple computational
task.  This idea of ``internalizing'' a driven dynamics into a
reservoir has been independently (re-)introduced under different names
and for a variety of purposes (\emph{self-prediction}
\cite{MayerBrowne04}, \emph{equilibration} \cite{Jaeger10b},
\emph{reservoir regularization} \cite{ReinhartSteil10},
\emph{self-sensing networks} \cite{SussilloAbbott12}, \emph{innate
  training} \cite{LajeBuonomano13}) and appears to be a fundamental
RNN adaptation principle.

\paragraph*{Going top-down: from conceptors to neural dynamics.} 

Assume that conceptors $C^j = C(p^j,\alpha)$ have been derived for
patterns $p^1,\ldots,p^K$, and that these patterns have been loaded
into the reservoir, replacing the original random weights $W^\ast$ by
$W$. Intuitively, the loaded reservoir, when it is run using $x(n+1) =
\tanh(W\,x(n))$ (no input!) should behave exactly as when it was driven
with input earlier, because $W$ has been trained such that $W\,x(n)
\approx W^\ast \,x(n) + W^{\mbox{\scriptsize in}} p^j(n)$. In fact, if
only a single pattern had been loaded, the loaded reservoir would
readily re-generate it. But if more than one patter had been loaded,
the autonomous (input-free) update $x(n+1) = \tanh(W\,x(n))$ will lead
to an entirely unpredictable dynamics: the network can't ``decide''
which of the loaded patterns it should re-generate! This is where
conceptors come in. The reservoir dynamics is filtered through $C^j$.
This is effected by using the augmented update rule $x(n+1) =
C^j\,\tanh(W\,x(n))$. By virtue of inserting $C^j$ into the feedback
loop, the reservoir states become clipped to fall within the ellipsoid
associated with $C^j$. As a result, the pattern $p^j$ will be
re-generated:  when the reservoir is observed through
the previously trained output weights, one gets $y(n) =
W^{\mbox{\scriptsize out}}\,x(n) \approx p^j(n)$. The first column of
panels in Fig.\ \ref{fig1Main} {\bf B} shows an overlay of the four
autonomously re-generated patterns $y(n)$ with the original drivers
$p^j$ used in that example.  The recovery of the originals is quite
accurate (mean square errors 3.3e-05, 1.4e-05, 0.0040, 0.0019 for the
four loaded patterns). Note that the first two and the last two
patterns are rather similar to each other. The filtering afforded by
the respective conceptors is ``sharp'' enough to separate these twin
pairs.  I will later demonstrate that in this way a remarkably large
number of patterns can be faithfully re-generated by a single
reservoir.

\paragraph*{Morphing and generalization.} 

Given a reservoir loaded with $K$ patterns $p^j$, the associated
conceptors $C^j$ can be linearly combined by creating mixture
conceptors $M = \mu^1 C^1 + \ldots + \mu^K C^K$, where the mixing
coefficients $\mu^j$ must sum to 1. When the reservoir is run under
the control of such a \emph{morphed} conceptor $M$, the resulting
generated pattern is a morph between the original ``pure'' patterns
$p^j$. If all $\mu^j$ are non-negative, the morph can be considered an
\emph{interpolation} between the pure patterns; if some $\mu^j$ are
negative, the morph \emph{extrapolates} beyond the loaded pure
patterns. I demonstrate this with  the four patterns
used in the example above, setting $\mu^1 = (1-a)b, \mu^2 = ab, \mu^3
= (1-a)(1-b), \mu^4 = a(1-b)$, and letting $a,b$ vary from $-0.5$ to
$1.5$ in increments of $0.25$. Fig.\ \ref{figMorphMain} shows plots of
observer signals $y(n)$ obtained when the reservoir is generating
patterns under the control of these morphed conceptors. The innermost
5 by 5 panels show interpolations between the four pure patterns, all
other panels show extrapolations.

In machine learning terms, both interpolation and extrapolation are
cases of \emph{generalization}. A standard opinion in the field states
that generalization by interpolation is what one may expect from
learning algorithms, while extrapolation beyond the training data is
 hard to achieve. 

Morphing and generalizing dynamical patterns is a common but
nontrivial task for training motor patterns in robots. It typically requires training
demonstrations of numerous interpolating patterns
\cite{ReinhartSteil08,Coatesetal08,Lukicetal12}. Conceptor-based
pattern morphing  appears
promising for flexible robot motor pattern learning from a very small
number of demonstrations.

 \begin{figure}[htb]
  \center
  \includegraphics[width=120 mm]{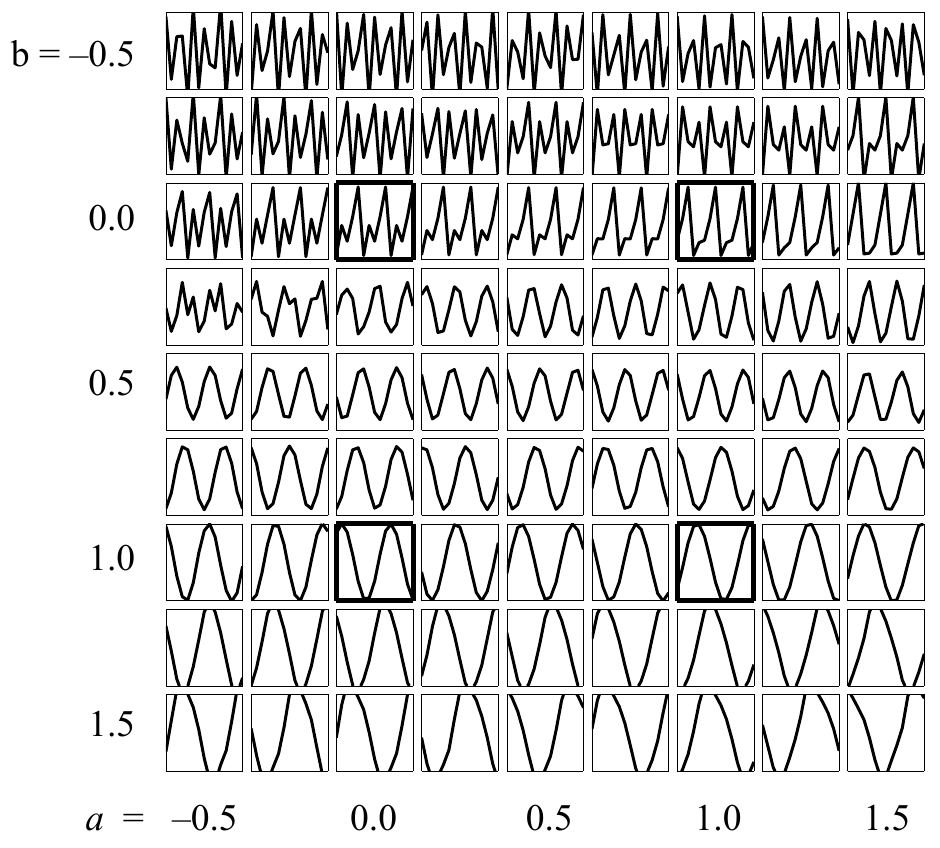}
  \caption{Morphing between, and generalizing beyond, four loaded
    patterns. Each panel shows a 15-step autonomously generated
    pattern (plot range between $-1$ and $+1$). Panels with bold
    frames: the four loaded prototype patterns  (same
    patterns as in Fig.\ \ref{fig1Main} {\bf B}.) }
  \label{figMorphMain}
  \end{figure}

\paragraph*{Aperture adaptation.}

Choosing the aperture $\alpha$ appropriately is crucial for
re-generating patterns in a stable and accurate way.  To demonstrate
this, I loaded a 500-neuron reservoir with signals $p^1$ -- $p^4$ derived from four
classical chaotic attractors: the Lorenz, R\"{o}ssler, Mackey-Glass,
and H\'{e}non attractors. Note that it used to be a challenging task
to make an RNN learn any single of these attractors
\cite{JaegerHaas04}; to my knowledge, training a single RNN to generate
several different chaotic attractors has not been attempted
before. After loading the reservoir, the re-generation was tested using
conceptors $C(p^j,\alpha)$ where for each attractor pattern $p^j$ a number of
different values for $\alpha$ were tried. Fig.\ \ref{figChaosApMain}
{\bf A} shows the resulting re-generated patterns for five apertures
for the Lorenz attractor. When the aperture is too small, the
 reservoir-conceptor feedback loop becomes
too constrained and the produced patterns de-differentiate. When  the
aperture is too large, the feedback loop  becomes over-excited.

\begin{figure}[htb]
  \center
{\bf A}  \includegraphics[width=140 mm]{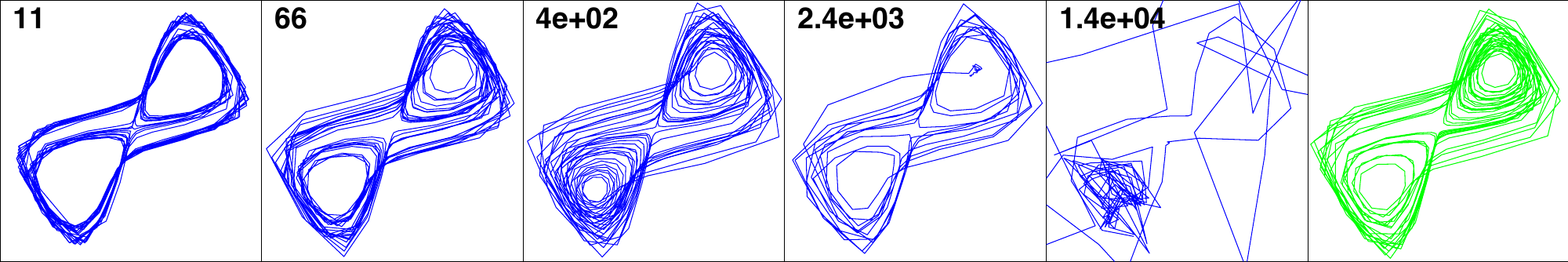}\\
\vspace{3mm}
{\bf B}  \includegraphics[width=140 mm]{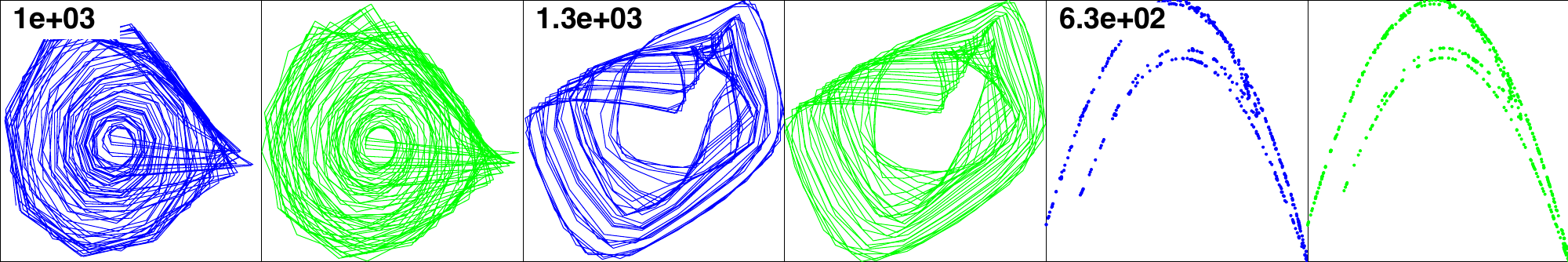}\\
\vspace{3mm}
{\bf C}  \includegraphics[width=140 mm]{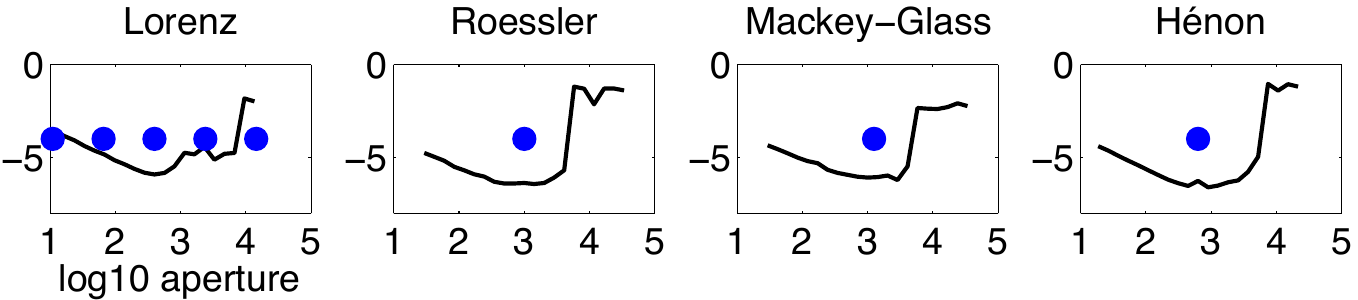}
  \caption{Aperture adaptation for re-generating four chaotic
    attractors. {\bf A} Lorenz attractor. Five versions re-generated
    with different apertures (values inserted in panels) and original
    attractor (green). {\bf B} Best re-generations of the
    other three attractors (from left to right: R\"{o}ssler,
    Mackey-Glass, and H\'enon, originals in green). {\bf C} Log10 of the attenuation criterion plotted
    against the log10 of aperture. Dots mark the apertures used for
    plots in {\bf A} and {\bf B}. }
  \label{figChaosApMain}
  \end{figure}

  An optimal aperture can be found by experimentation, but this will
  not be an option in many engineering applications or in biological
  neural systems. An intrinsic criterion for optimizing $\alpha$ is
  afforded by a quantity that I call \emph{attenuation}: the damping
  ratio which the conceptor imposes on the reservoir signal. Fig.\ 
  \ref{figChaosApMain} {\bf C} plots the attenuation against the
  aperture for the four chaotic signals. The minimum of this curve
  marks a good aperture value: when the conceptor dampens out a minimal
  fraction of the reservoir signal, conceptor and reservoir are in
  good ``resonance''. The chaotic attractor re-generations shown in
  Fig.\ \ref{figChaosApMain} {\bf B} were obtained by using this
  minimum-attenuation criterion.
  
  The aperture range which yields visibly good attractor
  re-generations in this demonstration spans about one order of
  magnitude. With further refinements (zeroing small singular values
  in conceptors is particularly effective), the viable aperture range
  can be expanded to about three orders of magnitude. While setting
  the aperture right is generally important, fine-tuning is
  unnecessary.

\paragraph*{Boolean operations and conceptor abstraction.} 

Assume that a reservoir is driven by a pattern $r$ which consists of
randomly alternating epochs of two patterns $p$ and $q$. If one
doesn't know which of the two patterns is active at a given time, all
one can say is that the pattern $r$ currently is $p$ OR it is $q$. Let
$C(R_p,1), C(R_q,1), C(R_r,1)$ be conceptors derived from the two
partial patterns $p, q$ and the ``OR'' pattern $r$, respectively. Then
it holds that $C(R_r,1) = C((R_p + R_q)/2, 1)$. Dropping the division
by 2, this motivates to define an OR (mathematical notation: $\vee$)
operation on conceptors $C_1(R_1,1), C_2(R_2,1)$ by putting $C_1 \vee
C_2 := (R_1 + R_2)(R_1 + R_2 + I)^{-1}$. The logical operations NOT
($\neg$) and AND ($\wedge$) can be defined along similar lines.  Fig.\ 
\ref{figBooleansMain} shows two-dimensional examples of  applying
the three operations.

\begin{figure}[htb]
  \center
\includegraphics[width=140 mm]{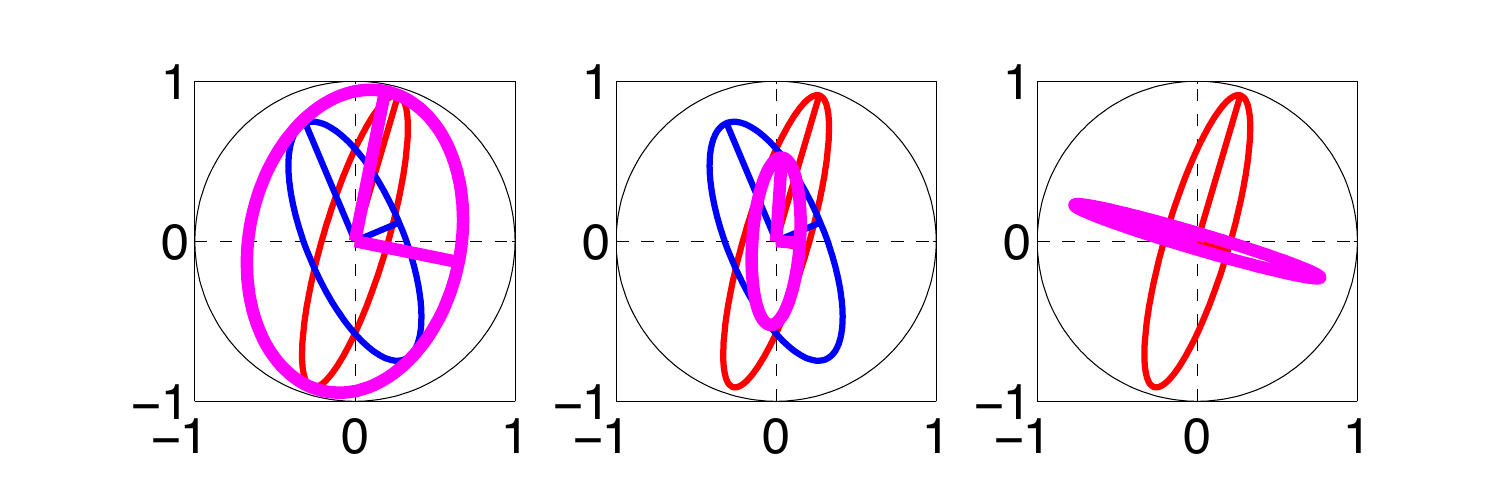}
  \caption{Boolean operations on conceptors. Red/blue (thin) ellipses represent
    source conceptors $C_1,C_2$. Magenta (thick) ellipses show $C_1 \vee C_2$,
    $C_1 \wedge C_2$, $\neg C_1$ (from left to right). }
  \label{figBooleansMain}
  \end{figure}
  
  Boolean logic is the mathematical theory of $\vee, \wedge, \neg$.
  Many laws of Boolean logic also hold for the $\vee, \wedge, \neg$
  operations on conceptors: the laws of associativity, commutativity,
  double negation, de Morgan's rules, some absorption rules.
  Furthermore, numerous simple laws connect aperture adaptation to
  Boolean operations. Last but not least, by defining $C_1 \leq C_2$ if
  and only if there exists a conceptor $B$ such that $C_2 = C_1 \vee B$, an
  \emph{abstraction ordering} is created on the set of all conceptors
  of  dimension $N$.

\paragraph*{Neural memory management.} Boolean conceptor
operations afford unprecedented  flexibility of organizing
and controlling the nonlinear dynamics of recurrent neural
networks. Here I demonstrate how a sequence of patterns $p^1,
p^2,\ldots$ can be \emph{incrementally} loaded into a reservoir, such
that (i) loading a new pattern $p^{j+1}$ does not interfere with
previously loaded $p^1,\ldots, p^j$; (ii) if a new pattern $p^{j+1}$
is similar to already loaded ones, the redundancies are automatically
detected and exploited, saving memory capacity; (iii) the amount of
still  ``free'' memory space can be logged. 

Let $C^j$ be the conceptor associated with pattern $p^j$. Three ideas
are combined to implement the memory management scheme.  First, keep
track of the ``already used'' memory space by maintaining a conceptor
$A^j = C^1 \vee \ldots \vee C^j$.  The sum of all singular values of
$A^j$, divided by the reservoir size, gives a number that ranges
between 0 and 1. It is an indicator of the portion of reservoir
``space'' which has been used up by loading $C^1,\ldots,C^j$, and I
call it the \emph{quota} claimed by $C^1, \ldots, C^j$.  Second,
characterize what is ``new'' about $C^{j+1}$ (not being already
represented by previously loaded patterns) by considering the
conceptor $N^{j+1} = C^{j+1} \setminus A^j$. The \emph{logical
  difference} operator $\setminus$ can be re-written as $A \setminus B
= A \,\wedge \,\neg B$. Third, load only that which is new about
$C^{j+1}$ into the still unclaimed reservoir space, that is, into
$\neg A^j$. These three ideas can be straightforwardly turned into a
modification of the basic pattern loading algorithm.

For a demonstration, I created a series of periodic patterns $p^1,
p^2, \ldots$ whose integer period lengths were picked randomly
between 3 and 15, some of these patterns being sines, others random
patterns. These patterns were incrementally loaded in a 100-neuron
reservoir, one by one. Fig.\ \ref{memManMain} shows the result. Since patterns $j = 5, 6, 7$ were identical replicas of
patterns $j = 1, 2, 3$, no additional quota space was consumed when these
patterns were (re-)loaded. Accuracy was measured by the \emph{normalized root mean square
  error} (NRMSE). The NRMSE jumps
from very small values to a high value when the last pattern is
loaded; the quota of 0.99 at this point indicates that the reservoir
is ``full''. The re-generation testing and NRMSE computation was done
after all patterns had been loaded. An attempt to load further
patterns would be unsuccessful, but it also would not harm the
re-generation quality of the already loaded ones.

\begin{figure}[htbp]
\center
\includegraphics[width=145 mm]{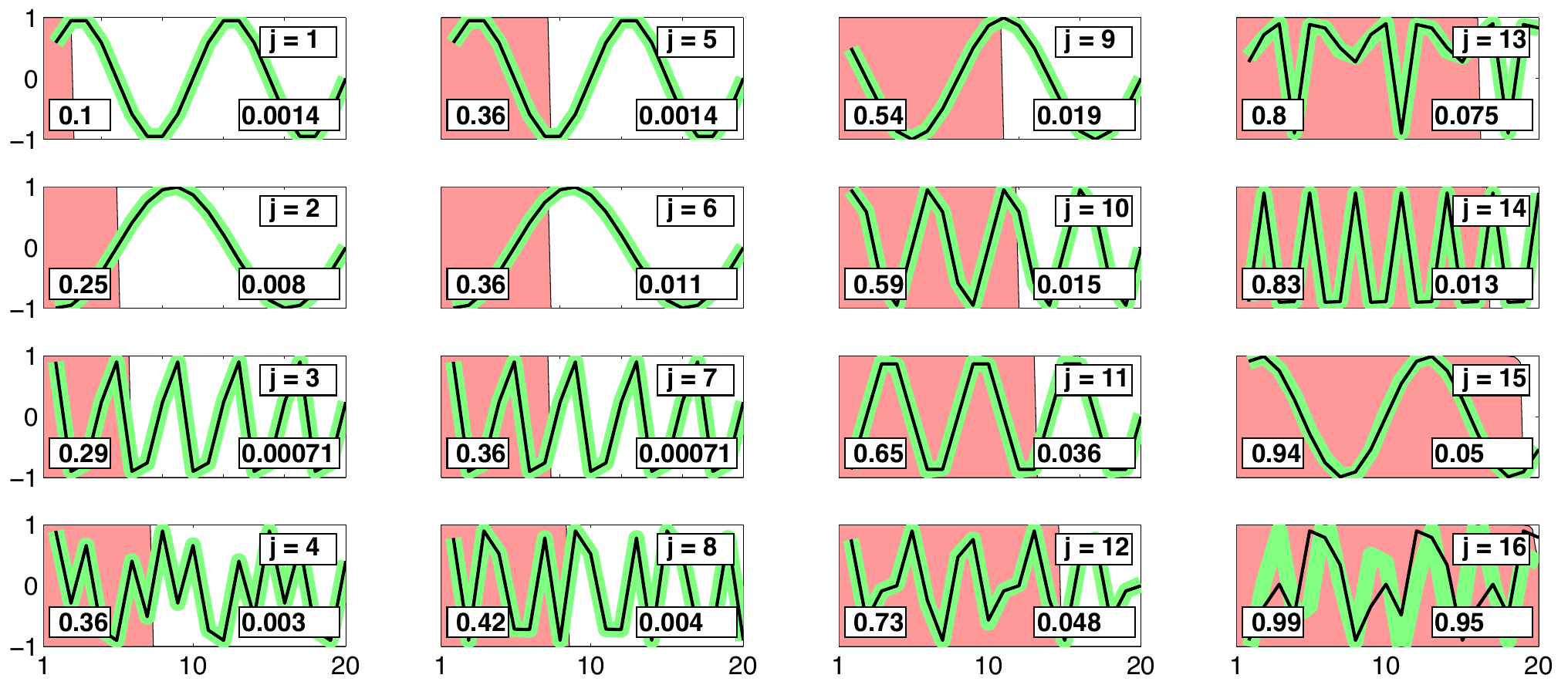}
\caption{Incremental pattern storing in a neural memory.  Each panel
  shows a 20-timestep sample of the correct training pattern $p^j$ (black
  line) overlaid on its reproduction (green line). The memory fraction
  used up until pattern $j$ is indicated by the panel fraction filled in
  red; the quota value is printed in the left bottom corner of each
  panel.   }
\label{memManMain}
\end{figure}

This ability to load patterns incrementally suggests a solution to a notorious problem
in neural network training, known as \emph{catastrophic forgetting},
which manifests itself in a disruption of previously learnt functionality
when learning new functionality. Although a number of proposals have
been made which partially alleviate the problem in special
circumstances \cite{French03,Grossberg05}, catastrophic forgetting was
still listed as an open challenge in an expert's report solicited by
the NSF in 2007 \cite{DouglasSejnowski08} which collected the main
future challenges in learning theory.

\paragraph*{Recognizing dynamical patterns.} Boolean conceptor
operations enable the combination of positive and negative evidence in
a neural architecture for dynamical pattern recognition. For a
demonstration I use a common benchmark, the \emph{Japanese vowel}
recognition task \cite{Kudoetal99}. The data of this benchmark consist
in preprocessed audiorecordings of nine male native speakers
pronouncing the Japanese di-vowel /ae/. The training data consist of
30 recordings per speaker, the test data consist of altogether 370
recordings, and the task is to train a recognizer which has to
recognize the speakers of the test recordings. This kind of data
differs from the periodic or chaotic patterns that I have been using
so far, in that the patterns are non-stationary (changing in their
structure from beginning to end), multi-dimensional (each recording
consisting of 12 frequency band signals), stochastic, and of finite
duration. This example thus also demonstrates that conceptors can be
put to work with data other than single-channel stationary patterns.

A small (10 neurons) reservoir was created. It was driven with all
training recordings from each speaker $j$ in turn ($j = 1,\ldots,9$),
collecting reservoir response signals, from which a conceptor $C^j$
characteristic of speaker $j$ was computed. In addition, for each
speaker $j$, a conceptor $N^j = \neg\,(C^1 \vee \ldots \vee C^{j-1}
\vee C^{j+1} \vee \ldots C^9)$ was computed.  $N^j$ characterizes the
condition ``this speaker is not any of the other eight speakers''.
Patterns need not to be loaded into the reservoir for
this application, because they need not be re-generated.

In testing, a recording $p$ from the test set was fed to the
reservoir, collecting a reservoir response signal $x$. For each of the
conceptors, a \emph{positive evidence} $E^+(p,j) = x' C^j x$ was
computed.  $E^+(p,j)$ is a non-negative number indicating how well the
signal $x$ fits into the ellipsoid of $C^j$. Likewise, the
\emph{negative evidence} $E^-(p,j) = x' N^j x$ that the sample $p$ was
not uttered by any of the eight speakers other than speaker $j$ was
computed. Finally, the \emph{combined evidence} $E(p,j) = E^+(p,i) +
E^-(p,i)$ was computed.  This gave nine combined evidences
$E(p,1),\ldots, E(p,9)$. The pattern $p$ was then classified as
speaker $j$ by choosing the speaker index $j$ whose combined evidence
$E(p,j)$ was the greatest among the nine collected evidences.

In order to check for the impact of the random selection of the
underlying reservoir, this whole procedure was repeated 50 times,
using a freshly created random reservoir in each trial. Averaged over
these 50 trials, the number of test misclassifications was 3.4. If the
classification would have been based solely on the positive or
negative evidences, the average test misclassification numbers would
have been 8.4 and 5.9 respectively. The combination of positive and
negative evidence, which  was enabled by Boolean operations,
 was crucial.
 
 State-of-the-art machine learning methods achieve between 4 and 10
 misclassifications on the test set (for instance
 \cite{Rodriguezetal05,Sivaramakrishnanetal07,OrsenigoVercellis10,Chatzis10}).
 The Boolean-logic-conceptor-based classifier thus compares favorably
 with existing methods in terms of classification performance.  The
 method is computationally cheap, with the entire learning procedure
 taking a fraction of a second only on a standard notebook computer.
 The most distinctive benefit however is incremental extensibility. If
 new training data become available, or if a new speaker would be
 incorporated into the recognition repertoire, the additional training
 can be done using only the new data without having to re-run previous
 training data. This feature is highly relevant in engineering
 applications and in cognitive modeling and missing from almost all
 state-of-the-art classification methods.

\paragraph*{Autoconceptors and content-addressable memories.} So far I
have been describing examples where conceptors $C^j$ associated with
patterns $p^j$ were computed at training time, to be  later
plugged in to re-generate or classify patterns.   A conceptor $C$ matrix has the
same size as the reservoir connection matrix $W$.  Storing conceptor
matrices means to store network-sized objects. This is implausible
under aspects of biological modeling. Here I
describe how conceptors can be created on the fly, without having to
store them, leading to content-addressable neural memories.

 If the system
has no pre-computed conceptors at its disposal, loaded
patterns can still be  re-generated in a two-stage process. First, the target
pattern $p$ is selected by driving the system with a brief
initial ``cueing'' presentation of the pattern (possibly in a noisy
version). During this phase, a preliminary conceptor
$C^{\mbox{\scriptsize cue}}$ is created by an online adaptation process.
This preliminary $C^{\mbox{\scriptsize cue}}$ already enables the system
to re-generate an imperfect version of the pattern $p$.  Second, after
the cueing phase has ended, the system continues to run in an
autonomous mode (no external cue signal), initially using
$C^{\mbox{\scriptsize cue}}$, to continuously generate a pattern. While
this process is running, the conceptor in the loop is continuously adapted
by a simple online adaptation rule. This rule can be described in
geometrical terms as ``adapt the current conceptor $C(n)$ such that its
ellipsoid matches better the shape of the point cloud of
the current reservoir state dynamics''.  Under this rule one
obtains a reliable convergence of the generated pattern toward a highly
accurate replica of the target pattern $p$ that was given as
a cue. 

 \begin{figure}[htbp]
\center
{\bf  A}\includegraphics[width=48mm]{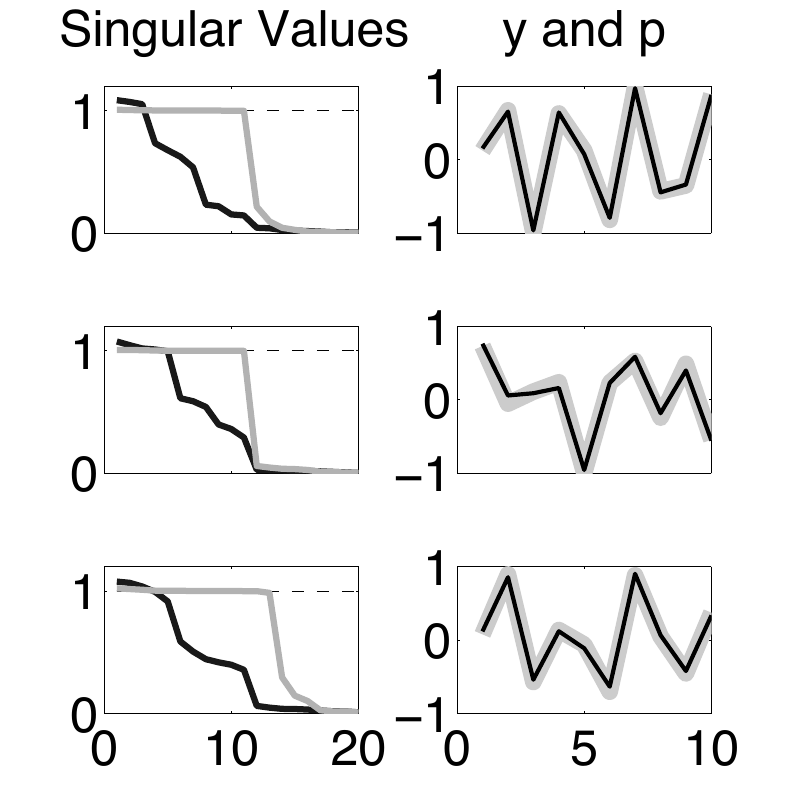}
\hspace{2mm}{\bf { B}}
\includegraphics[width=48mm]{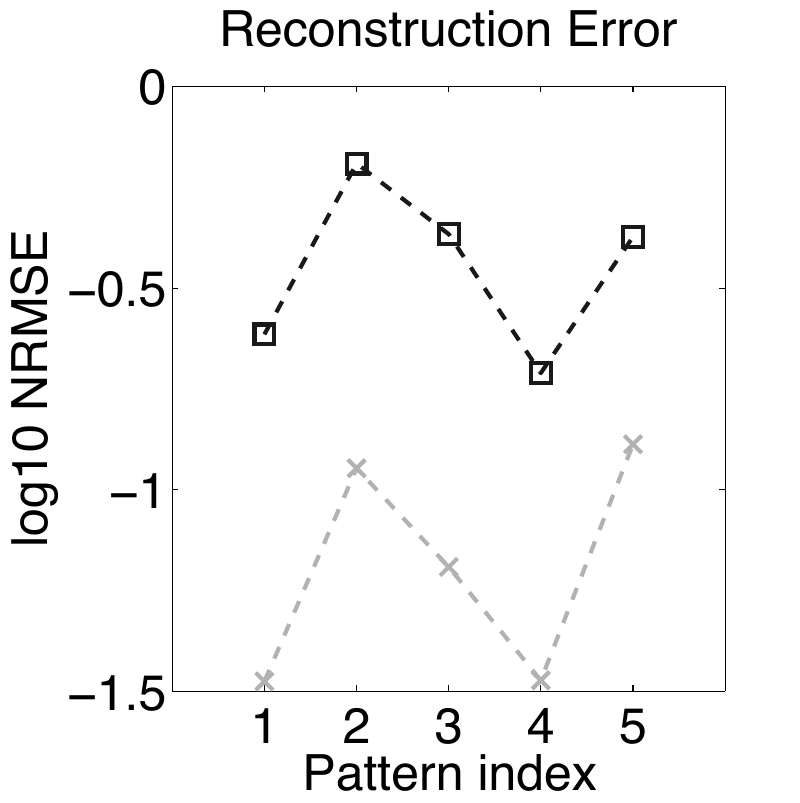}\\
{\bf  C}
\includegraphics[width=48mm]{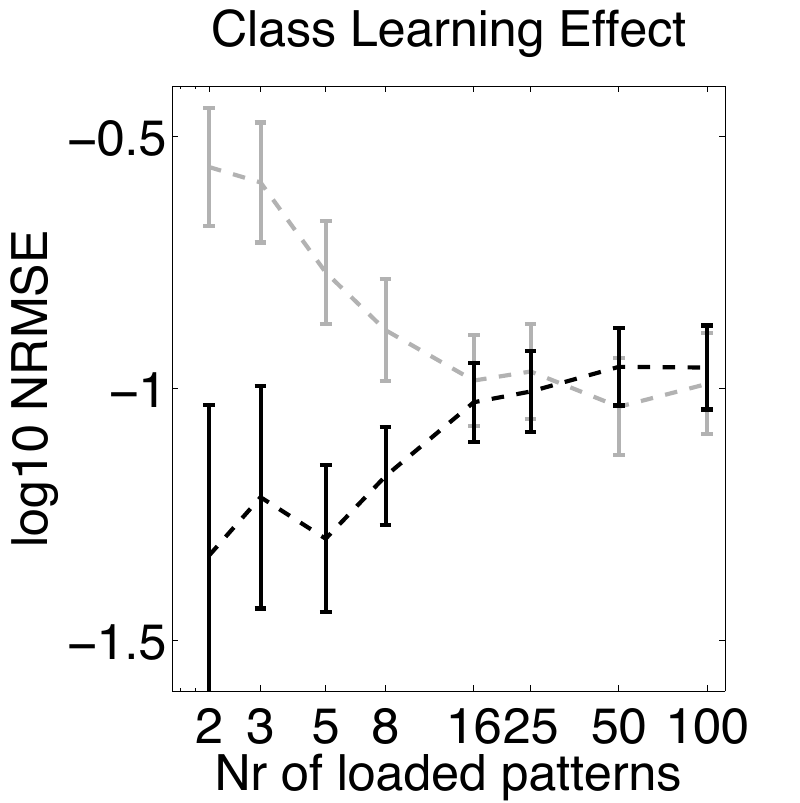}
\caption{Content-addressable memory.  {\bf { A}} First three of 
  five loaded patterns. Left panels show the leading 20 singular
  values of $C^{\mbox{\scriptsize cue}}$ (black) and
  $C^{\mbox{\scriptsize auto}}$ (gray). Right panels show an overlay
  of the original driver pattern (black, thin) and the reconstruction
  at the end of auto-adaptation (gray, thick).  {\bf { B}} Pattern
  reconstruction errors directly after cueing (black squares) and at
  end of auto-adaptation (gray crosses).  {\bf { C}} Reconstruction
  error of loaded patterns (black) and novel patterns drawn from the
  same parametric family (gray) versus the number of loaded patterns,
  averaged over 5 repetitions of the entire experiment and 10 patterns
  per plotting point. Error bars indicate standard deviations. }
\label{figContAddressMain}
\end{figure}

Results of a demonstration are illustrated in Figure
\ref{figContAddressMain}. A 200-neuron reservoir was loaded with 5
patterns consisting of a weighted sum of two irrational-period sines,
sampled at integer timesteps. The weight ratio and the phaseshift were
chosen at random; the patterns thus came from a family of patterns
parametrized by two parameters. The cueing time was 30 timesteps, the
free-running auto-adaptation time was 10,000 timesteps, leading to an
auto-adapted conceptor $C^{\mbox{\scriptsize auto}}$ at the end of
this process. On average, the reconstruction error improved from about
-0.4 (log10  NRMSE measured directly after the cueing) to -1.1 (at the end
of auto-adaptation).  It can be shown
analytically that the auto-adaptation process pulls many singular
values down to  zero. This effect renders the combined
reservoir-conceptor loop very robust against noise, because all noise
components in the directions of the nulled singular values become
completely suppressed. In fact, all
results shown in Figure \ref{figContAddressMain} were obtained with
strong state noise (signal-to-noise ratio equal to 1) inserted into
the reservoir during the post-cue auto-adaptation.

The system functions as a \emph{content-addressable memory} (CAM):
loaded items can be recalled by cueing them. The paradigmatic example
of a neural CAM are auto-associative neural networks (AANNs), pioneered
by Palm \cite{Palm80} and Hopfield \cite{Hopfield82}. In contrast to
conceptor-based CAM, which store and re-generate dynamical patterns,
AANNs store and cue-recall static patterns. Furthermore, AANNs do
not admit an incremental storing of new patterns, which is possible in
conceptor-based CAMs. The latter thus represent an advance in neural
CAMs in two fundamental aspects.

To further elucidate the properties of conceptor CAMs, I ran a suite
of simulations where the same reservoir was loaded with increasing
numbers of patterns, chosen  at random from the same 2-parametric
family (Figure \ref{figContAddressMain} {\bf C}).  After loading with
$k = 2, 3, 5, \ldots, 100$ patterns, the reconstruction accuracy was
measured at the end of the auto-adaptation. Not surprisingly, it
deteriorated with increasing memory load $k$ (black line). In
addition, I also cued the loaded reservoir with patterns that were
\emph{not} loaded, but were drawn from the same family. As one would
expect, the re-construction accuracy of these novel patterns was worse
than for the loaded patterns -- but only for small $k$. When the
number of loaded patterns exceeded a certain threshold, recall
accuracy became essentially equal for loaded and novel patterns. These
findings can be explained in intuitive terms as follows. When few
patterns are loaded, the network memorizes individual patterns by
``rote learning'', and subsequently can recall these patterns better
than other patterns from the family. When more patterns are loaded,
the network learns a representation of the entire parametric class of
patterns. I call this the \emph{class learning effect}.

%% The class learning effect can be geometrically interpreted in terms of
%% a \emph{plane attractor} \cite{Eliasmith05} arising in the space of
%% conceptor matrices $C$ (Figure \ref{figContAddressMain} {\bf 
%%   D}). The learnt parametric class of patterns is represented by a
%% $d$-dimensional manifold ${M}$ in this space, where $d$ is the number of
%% defining parameters for the pattern family (in our example, $d = 2$). The
%% cueing procedure creates an initial conceptor  $C^{\mbox{\scriptsize
%%     cue}}$ in the vicinity of ${M}$, which is then attracted
%% toward ${M}$ by the auto-adaptation dynamics. While an in-depth 
%% analysis of this situation  reveals that this picture is not
%% mathematically correct in some detail,  the plane attractor
%% metaphor yields a good phenomenal description of conceptor CAM class
%% learning. 

%% Plane attractors have been invoked as an explanation for a number of
%% biological phenomena, most prominently gaze direction control
%% \cite{Eliasmith05}. In such phenomena, points on the plane attractor
%% correspond to \emph{static} fixed points (for instance, a direction of
%% gaze). In contrast, points on $M$ correspond to conceptors which in
%% turn define \emph{temporal} patterns. Again, the conceptor framework
%% ``dynamifies'' concepts that have previously been worked out for
%% static patterns only.

\paragraph*{Toward biological feasibility: random feature conceptors.}
Several computations involved in adapting conceptor matrices  are
non-local and therefore biologically infeasible. It is however
possible to approximate matrix conceptors with another mechanism which
only requires local computations. The idea is to project (via random
projection weights $F$) the reservoir state into a \emph{random
  feature space} which is populated by a large number of neurons
$z_i$; execute the conceptor operations individually on each of these
neurons by multiplying a \emph{conception weight} $c_i$ into its
state; and finally to project back to the reservoir by another set of
random projection weights $G$ (Figure \ref{figRandFeatures}).

The original reservoir-internal random connection weigths $W$ are
replaced by a dyade of two random projections of first $F$, then $G$,
and the original reservoir state $x$ segregates into a reservoir state
$r$ and a random feature state $z$.  The conception weights $c_i$
assume the role of conceptors. They can be learnt and adapted by
procedures which are directly analog to the matrix conceptor case.
What had to be non-local matrix computations before now turns into
local, one-dimensional (scalar) operations. These operations are biologically
feasible in the modest sense that any information needed to adapt a synaptic
weight is locally available at that synapse.  All laws and
constructions concerning Boolean operations and aperture carry over.

 \begin{figure}[htb]
\center
\includegraphics[width=90mm]{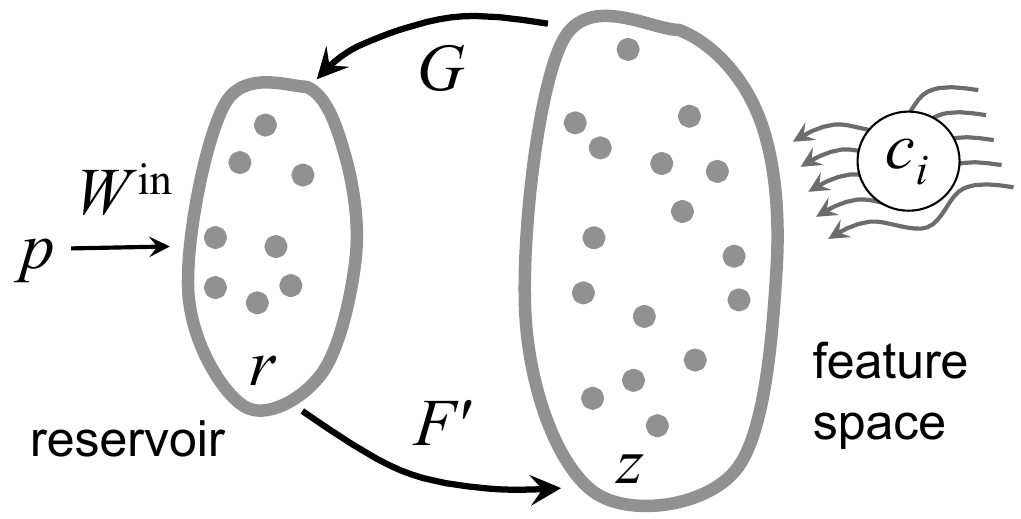}
\caption{Random feature conceptors. This neural architecture has two
  pools of neurons, the reservoir and the feature space.}
\label{figRandFeatures}
\end{figure}

A set of conception weights $c_i$ corresponding to a particular
pattern can be neurally represented and ``stored'' in the form of the
connections of a single neuron to the feature space. A dynamical
pattern thus can be represented by a single neuron. This enables a
highly compact neural representation of dynamical patterns. A machine
learning application is presented below.

 I re-ran with such random feature conceptors a choice of
the simulations that I did with matrix conceptors, using a number of
random features that was two to five times as large as the
reservoir. The outcome of these simulations: the accuracy of pattern
re-generation is essentially the same as with matrix conceptors, but
setting the aperture is more sensitive.

\paragraph*{A hierarchical  classification and de-noising architecture.} 

Here I present a system which combines in a multi-layer neural
architecture many of the items introduced
so far.  The input to this system
is a (very) noisy signal which at a given time is being generated by
one out of a number of possible candidate pattern generators. The task is to
recognize the current generator, and simultaneously to re-generate a
clean version of the noisy input pattern.

\begin{figure}[htbp]
\center
{\bf  A}\includegraphics[width=85mm]{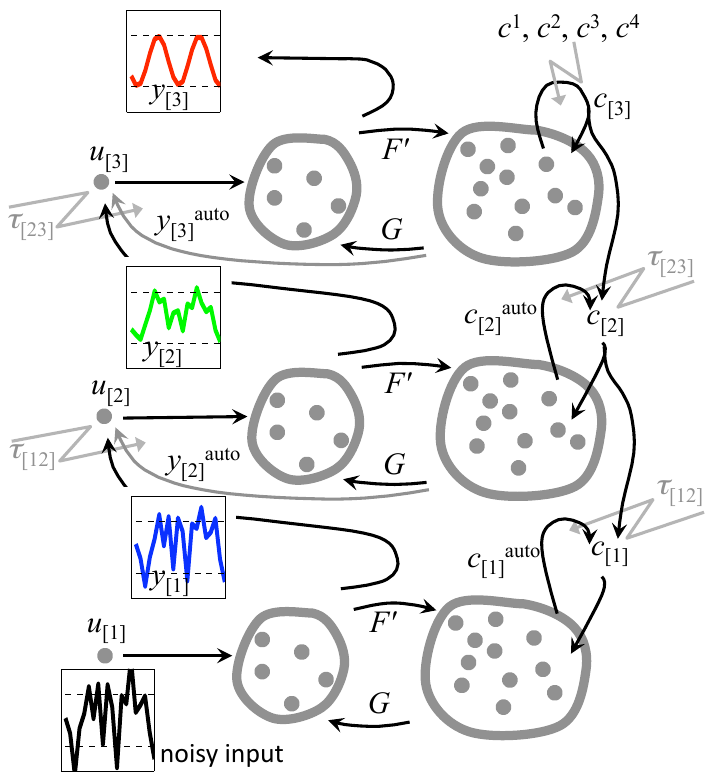}
{\bf { B}}
\includegraphics[width=50mm]{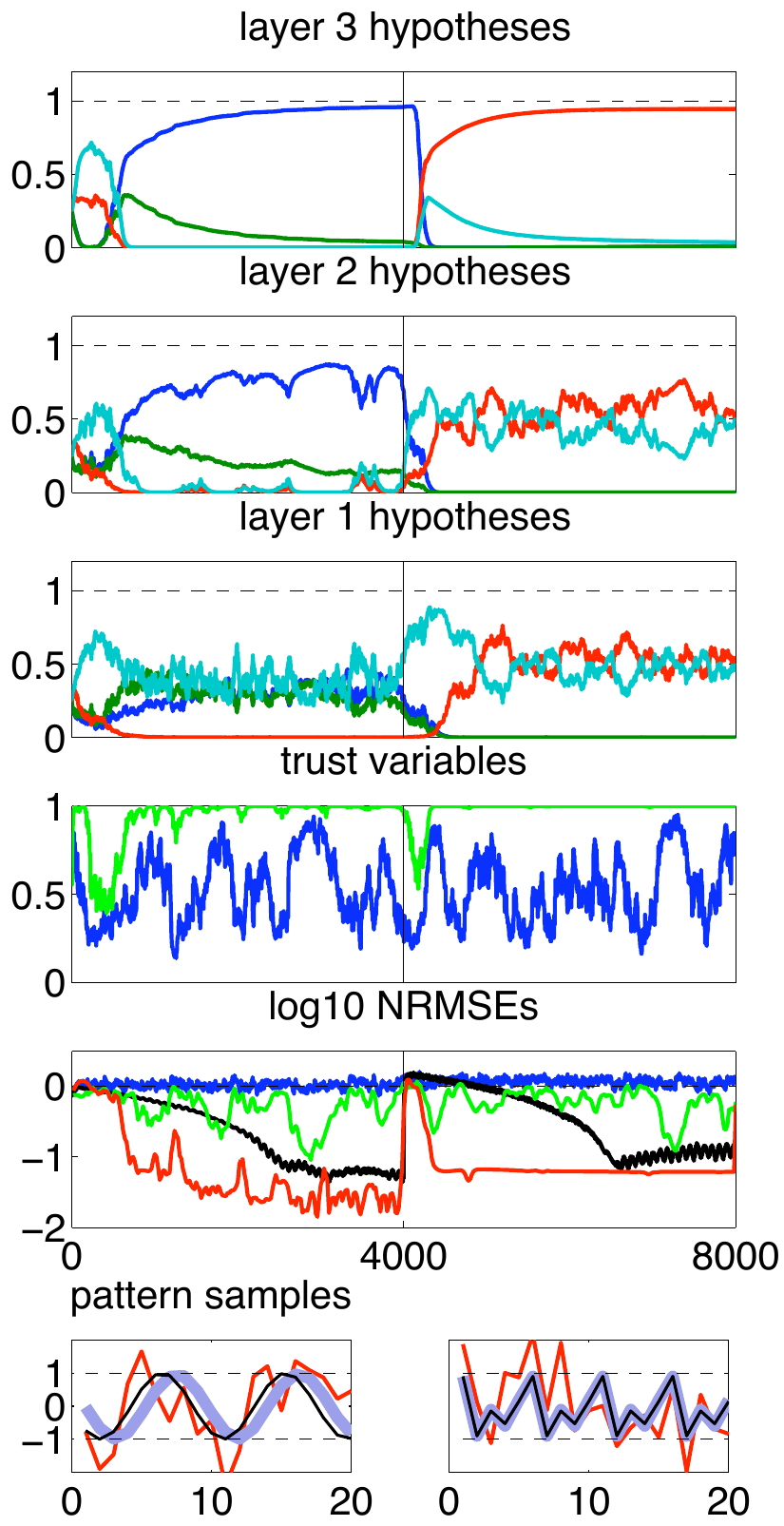}\\
\caption{Simultaneous signal de-noising and classification. {\bf 
    A.} Schema of architecture.  {\bf  B.} Simulation results.
  Panels from above: first three panels: hypothesis vectors
  $\gamma^j_{[l]}(n)$ in the three layers. Color coding: $p^1$ blue,
  $p^2$ green, $p^3$ red, $p^4$ cyan. Fourth panel: trust
  variables $\tau_{[1,2]}(n)$ (blue) and $\tau_{[2,3]}(n)$ (green).
  Fifth panel: signal reconstruction errors (log10 NRMSE) of $y_{[1]}$
  (blue),  $y_{[2]}$ (green) and $y_{[3]}$ (red)
  versus clean signal $p^j$. Black line: linear 
  baseline  filter. Bottom panels: 20-step samples from the end of the
  two presentation periods. Red: noisy input; black: clean input;
  thick gray: cleaned output signal  $y_{[3]}$. }
\label{fig3Layer}
\end{figure}

I explain the architecture with an example. It uses three processing
layers to de-noise an input signal $u_{[1]}(n) = p^j(n) + noise$, with
$p^j$ being one of the four patterns $p^1,\ldots,p^4$ used before in
this report (shown for instance in Figure \ref{fig1Main} {\bf B}).
The architecture implements the following design principles (Figure
\ref{fig3Layer} {\bf A}).  (i) Each layer is a random feature based
conceptor system (as in Figure \ref{figRandFeatures} {\bf B}). The
four patterns $p^1, \ldots, p^4$ are initially loaded into each of the
layers, and four \emph{prototype} conceptor weight vectors $c^1,
\ldots, c^4$ corresponding to the patterns are computed and stored.
(ii) In a bottom-up processing pathway, the noisy external input
signal $u_{[1]}(n) = p^j(n) + noise$ is stagewise de-noised, leading
to signals $y_{[1]}, y_{[2]},y_{[3]}$ on layers $l = 1, 2, 3$, where
$y_{[3]}$ should be a highly cleaned-up version of the input
(subscripts $[l]$ refer to layers, bottom layer is $l = 1$). (iii) The
top layer auto-adapts a conceptor $c_{[3]}$ which is constrained to be
a weighted OR combination of the four prototype conceptors. In a
suggestive notation this can be written as $c_{[3]}(n) =
\gamma^1_{[3]}(n)\, c^1 \vee \ldots \vee \gamma^4_{[3]}(n) \,c^4$. The
four weights $\gamma^j_{[3]}$ sum to one and represent a
\emph{hypothesis} vector expressing the system's current belief about
the current driver $p^j$. If one of these $\gamma^j_{[3]}$ approaches
1, the system has settled on a firm classification of the current
driving pattern. (iv) In a top-down pathway, conceptors $c_{[l]}$ from
layers $l$ are passed down to the respective layers $l-1$ below.
Because higher layers should have a clearer conception of the current
noisy driver pattern than lower layers, this passing-down of
conceptors ``primes'' the processing in layer $l-1$ with valuable
contextual information. (v) Between each pair of layers $l, l+1$, a
\emph{trust} variable $\tau_{[l,l+1]}(n)$ is adapted by an online
procedure. These trust variables range between 0 and 1. A value of
$\tau_{[l,l+1]}(n) = 1$ indicates maximal confidence that the signal
$y_{[l+1]}(n)$ comes closer to the clean driver $p^j(n)$ than the
signal $y_{[l]}(n)$ does, that is, the stage-wise denoising actually
functions well when progressing from layer $l$ to $l+1$. The trust
$\tau_{[l,l+1]}(n)$ evolves by comparing certain noise ratios that are
observable locally in layers $l$ and $l+1$. (vi) Within layer $l$, an
internal auto-adaptation process generates a candidate de-noised
signal $y_{[l]}^{\mbox{\scriptsize auto}}$ and a candidate local
autoconceptor $c_{[l]}^{\mbox{\scriptsize auto}}$. The local estimate
$y_{[l]}^{\mbox{\scriptsize auto}}$ is linearly mixed with the signal
$y_{[l-1]}$, where the trust $\tau_{[l-1,l]}$ sets the mixing rate.
The mixture $u_{[l]} = \tau_{[l-1,l]}\, y_{[l]}^{\mbox{\scriptsize
    auto}} + (1- \tau_{[l-1,l]}) \, y_{[l-1]}$ is the effective signal
input to layer $l$. If the trust $\tau_{[l-1,l]}$ reaches its maximal
value of 1, layer
$l$ will ignore the signal from below and work entirely by
self-generating a pattern. (vii) In a similar way, the effective
conceptor in layer $l$ is a trust-negotiated mixture $c_{[l]} =
(1-\tau_{[l,l+1]})\, c_{[l]}^{\mbox{\scriptsize auto}} +
\tau_{[l,l+1]} \, c_{[l+1]}$. Thus if the trust $\tau_{[l,l+1]}$ is
maximal, layer $l$ will be governed entirely by the passed-down
conceptor $c_{[l+1]}$.

Summarizing, the higher the trusts inside the hierarchy, the more will
the system be auto-generating conceptor-shaped signals, or conversely,
at low trust values the system will be strongly permeated from below
by the outside driver. If the trust variables reach their maximum
value of 1, the system will run in a pure ``confabulation'' mode and
generate an entirely noise-free signal $y_{[3]}$ -- at the risk of
doing this under an entirely misguided hypothesis $c_{[3]}$.  The key
to make this architecture work thus lies in the trust variables. It
seems to me that maintaining a measure of trust (or call it
confidence, certainty, etc.) is an intrinsically necessary component
in any signal processing architecture which hosts a top-down pathway
of guiding hypotheses (or call them context, priors, bias, etc.).

Figure \ref{fig3Layer} {\bf B} shows an excerpt from a simulation run.
The system was driven first by an initial 4000 step period of $p^1 +
noise$, followed by 4000 steps of $p^3 + noise$. The signal-to-noise
ratio was 0.5 (noise twice as strong as signal). The system
successfully settles on the right hypothesis (top panel) and generates
very clean de-noised signal versions (bottom panel). The crucial item
in this figure is the development of the trust variable
$\tau_{[2,3]}$. At the beginning of each 4000 step period it briefly
drops, allowing the external signal to permeate upwards through the
layers, thus informing the local auto-adaptation loops about ``what is
going on outside''. After these initial drops the trust rises to almost 1,
indicating that the system firmly ``believes'' to have detected the right
pattern. It then generates pattern
versions that have almost no mix-in from the noisy external driver.

As a baseline comparison I also trained a standard linear transversal
filter which computed a de-noised  input pattern point based on the
preceding $K = 2600$ input values. The filter length $K$ was set
equal to the number of trainable parameters in the neural
architecture. The performance of this linear de-noising filter (black
line in Figure  \ref{fig3Layer}) is inferior to the architecture's
performance both in terms of accuracy and response time.

It is widely believed that top-down hypothesis-passing through a
processing hierarchy plays a fundamental role in biological cognitive
systems \cite{Friston05,Clark12}. However, the current best artificial
pattern recognition systems
\cite{GravesSchmidhuber08,Krizhevskyetal12} use purely bottom-up
processing -- leaving room for further improvement by including
top-down guidance.  A few hierarchical architectures which exploit
top-down hypothesis-passing have been proposed
\cite{Friston05,HintonSalakhutdinov06,Grossberg13,GedeonArathorn07}.
All of these  are designed for recognizing static patterns,
especially images. The conceptor-based architecture presented here
appears to be the first hierarchical system which
targets dynamical patterns and uses top-down hypothesis-passing.
Furthermore, in contrast to state-of-the-art pattern recognizers, it
admits an incremental extension of the pattern repertoire.

\paragraph*{Intrinsic conceptor logic.}

In mathematical logics the \emph{semantics} (``meaning'') of a symbol
or operator is formalized as its \emph{extension}. For instance, the
symbol {\sf cow} in a logic-based knowledge representation system in
AI is semantically interpreted by the set of all (physical) cows, and
the OR-operator $\vee$ is interpreted as set union: {\sf cow} $\vee$
{\sf horse} would refer to the set comprising all cows and horses.
Similarly, in cognitive science, \emph{concepts} are semantically
referring to their extensions, usually called \emph{categories} in
this context \cite{MedinRips05}.  Both in mathematical logic and
cognitive science, extensions need not be confined to physical
objects; the modeler may also define extensions in terms of
mathematical structures, sensory perceptions, hypothetical worlds,
ideas or facts. But at any rate, there is an ontological difference
between the two ends of the semantic relationship.

This ontological gap dissolves in the case of conceptors. The 
natural account of the ``meaning'' of a matrix conceptor $C$ is the shape of the
neural state cloud it is derived from. This shape is given by the
correlation matrix $R$ of neural states. Both $C$ and $R$ have the
same mathematical format: positive semi-definite matrices of identical
dimension.  Figure \ref{figSemanticsMain} visualizes the difference
between classical extensional semantics of logics and the
system-internal conceptor semantics. The symbol $\models$ is the
standard mathematical notation for the semantical meaning relationship.

\begin{figure}[htb]
\center
\includegraphics[width=120mm]{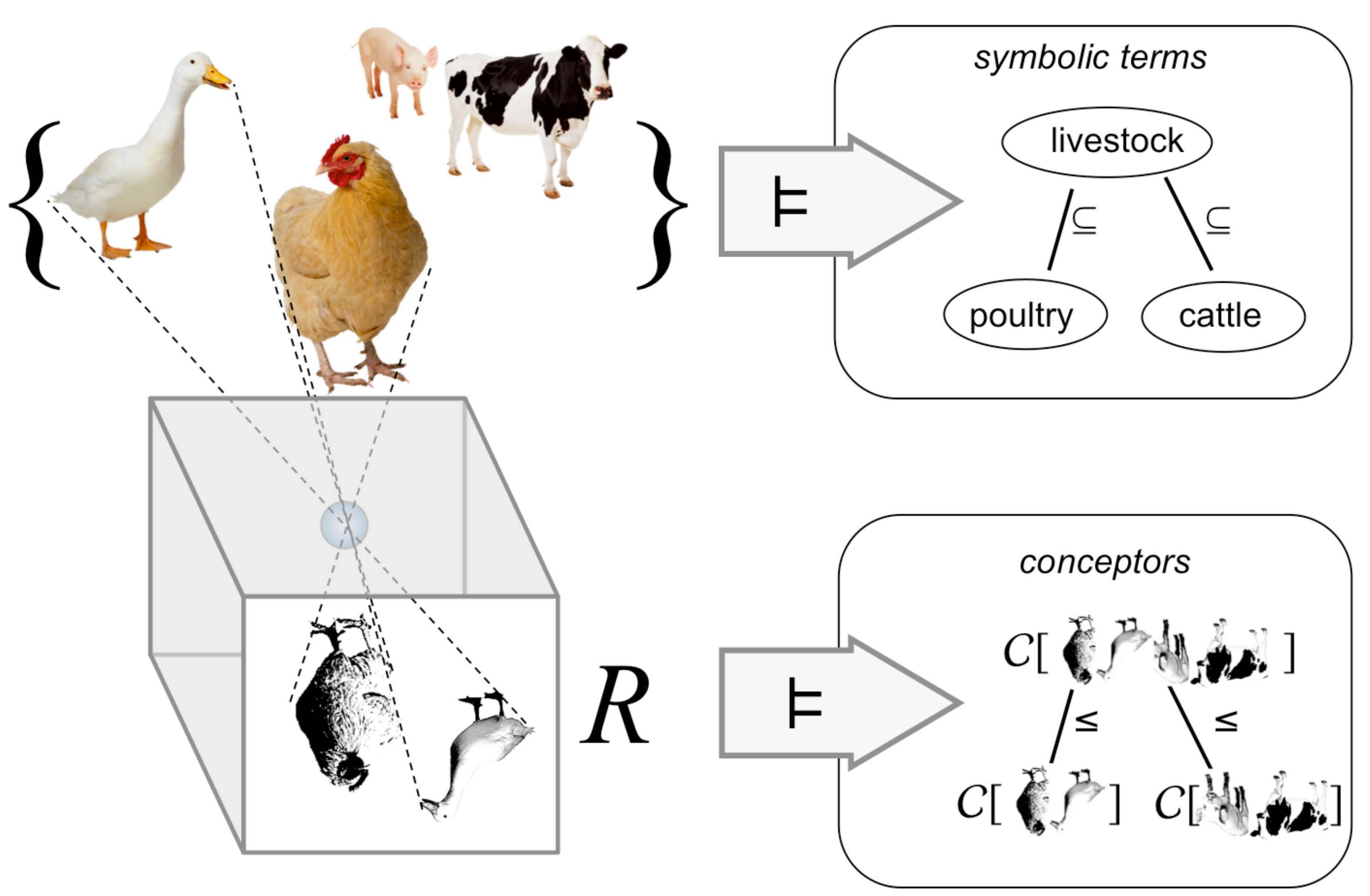}
\caption{Contrasting the extensional semantics of classical knowledge
  representation formalisms (upper half of graphics) with conceptor
  semantics (lower half).}
\label{figSemanticsMain}
\end{figure}

I have cast these intuitions into a formal specification of an
\emph{intrinsic conceptor logic} (ICL), where the semantic
relationship outlined above is formalized within the framework of
\emph{institutions} \cite{GoguenBurstall92}. This framework has been
developed in mathematics and computer science to provide a unified
view on the multitude of existing ``logics''. By formalizing ICL as an
institution, conceptor logic can be rigorously compared to other
existing logics. I highlight two findings. First, an ICL cast as an
institution is a dynamcial system in its own right: the symbols used
in this logic evolve over time. This is very much different from
traditional views on logic, where symbols are static tokens. Second,
it turns out that ICL is a logic which is \emph{decidable}. Stated in
intuitive terms, in a decidable logic it can be calculated
whether a ``concept'' $\psi$ subsumes a concept $\varphi$ (as in ``a
{\sf cow} is an {\sf animal}''). Deciding concept subsumption is a
core task in AI systems and human cognition.  In most logic-based AI
systems, deciding concept subsumption can become computationally
expensive or even impossible. In ICL it boils down to determining
whether all components of a certain conception weight vector $c_i$ are
smaller or equal to the corresponding components $c'_i$ of another
such vector, which can be done in a single processing step. This may
help explaining why humans can make classification judgements almost instantaneously.

\paragraph*{Discussion.} The human brain is a neurodynamical system
which evidently supports logico-rational reasoning
\cite{HoudeTzourioMazoyer03}. Since long this has challenged
scientists to find computational models which connect neural dynamics
with logic.  Very different solutions have been suggested.  At
the dawn of computational neuroscience, McCulloch and Pitts have already
interpreted networks of binary-state neurons as carrying out Boolean
operations \cite{McCullochPitts43}. Logical inferences of various
kinds have been realized in localist connectionist networks where
neurons are labelled by concept names \cite{Pinkas91,Shastri99a}. In
neurofuzzy modeling, feedforward neural networks are trained to carry
out operations of fuzzy logic on their inputs \cite{BrownHarris94}.
In a field known as neuro-symbolic computation, deduction rules of
certain formal logic systems
are coded or trained into neural networks 
\cite{Baderetal08,Lamb08,Borgesetal11}.  The
combinatorial/compositional structure of symbolic knowledge has been
captured by dedicated neural circuits to enable tree-structured
representations \cite{Pollack90} or variable-binding functionality
\cite{vanderVeldedeKamps06}.

All of these approaches require \emph{interface} mechanisms.  These
interface mechanisms are non-neural and code symbolic knowledge
representations into the numerical activation values of neurons and/or
the topological structure of networks.  One could say, previous
approaches \emph{code} logic \emph{into} specialized neural networks, while
conceptors \emph{instantiate} the logic \emph{of} generic recurrent neural
networks. This novel, simple, versatile, computationally efficient, neurally
not infeasible, bi-directional connection between logic and neural
dynamics opens new perspectives for computational neuroscience and
machine learning.

\newpage

\section{Introduction}\label{secIntro}

In this section I expand on the brief characterization of the
scientific context given in Section \ref{secOverview}, and introduce mathematical
notation.

\subsection{Motivation}\label{secSetting}

 Intelligent behavior is desired for robots, demonstrated by humans,
 and studied in a wide array of scientific disciplines. This research
 unfolds in two directions. In ``top-down'' oriented research, one
 starts from the ``higher'' levels of cognitive performance, like
 rational reasoning, conceptual knowledge representation, planning and
 decision-making, command of language. These phenomena are described
 in symbolic formalisms developed in mathematical logic, artificial
 intelligence (AI), computer science and linguistics. In the ``bottom-up''
 direction, one departs from ``low-level'' sensor data processing and
 motor control, using the analytical tools offered by dynamical
 systems theory, signal processing and control theory, statistics and
 information theory. For brevity I will refer to these two directions
 as the \emph{conceptual-symbolic} and the \emph{data-dynamical} sets
 of phenomena, and levels of description. The two  
 interact bi-directionally. Higher-level symbolic concepts arise from
 low-level sensorimotor data streams in short-term pattern recognition
 and long-term learning processes. Conversely,  low-level processing
 is modulated, filtered and steered by processes of attention,
 expectations, and goal-setting in a top-down fashion.

 Several schools of thought (and strands of dispute) have evolved in a
 decades-long quest for a unification of the conceptual-symbolic and
 the data-dynamical approaches to intelligent behavior. The nature of
 symbols in cognitive processes has been cast as a philosophical issue
 \cite{Searle80,FodorPylyshin88,Harnad90}. In localist connectionistic
 models, symbolically labelled abstract processing units interact by
 nonlinear \emph{spreading activation} dynamics
 \cite{Drescher91,Shastri99a}. A basic tenet of behavior-based AI is
 that higher cognitive functions \emph{emerge} from low-level
 sensori-motor processing loops which couple a behaving agent into its
 environment \cite{Brooks89,PfeiferScheier98}. Within cognitive
 science, a number of cognitive pheneomena have been described in
 terms of \emph{self-organization} in nonlinear dynamical systems
 \cite{SchoenerKelso88,SmithThelen93a,Gelder98}.  A pervasive idea in
 theoretical neuroscience is to interpret \emph{attractors} in
 nonlinear neural dynamics as the carriers of conceptual-symbolic
 representations. This idea can be traced back at least to the notion
 of cell assemblies formulated by Hebb \cite{Hebb49}, reached a first
 culmination in the formal analysis of associative memories
 \cite{Palm80,Hopfield82,Amitetal85a}, and has since then diversified
 into a range of increasingly complex models of interacting (partial)
 neural attractors
 \cite{YaoFreeman90,Tsuda00,Rabinovich08,SussilloBarak13}. Another
 pervasive idea in theoretical neuroscience and machine learning is to
 consider \emph{hierarchical} neural architectures, which are driven
 by external data at the bottom layer and transform this raw signal
 into increasingly abstract feature representations, arriving at
conceptual representations at the top layer of the
 hierarchy. Such hierarchical architectures mark the state of the art
 in pattern recognition technology \cite{LeCunetal98,Gravesetal09}.
 Many of these systems process their input data in a uni-directional,
 bottom-to-top fashion. Two notable exceptions are systems where each
 processing layer is designed according to statistical principles from
 \emph{Bayes' rule} \cite{Friston05,HintonSalakhutdinov06,Clark12},
 and models based on the iterative linear maps of \emph{map seeking
   circuits} \cite{GedeonArathorn07,Zibneretal11}, both of which
 enable top-down guidance of recognition by expectation generation.
 More generally, leading actors in theoretical neuroscience have
 characterized large parts of their field as an effort to understand
 how cognitive phenomena arise from neural dynamics
 \cite{Abbott08,Gerstneretal12}.  Finally, I point out two singular
 scientific efforts to design comprehensive cognitive brain models,
 the ACT-R architectures developed by Anderson et al.
 \cite{Andersonetal04} and the Spaun model of Eliasmith et al.
 \cite{Eliasmithetal12}. Both systems can simulate a broad selection of
 cognitive behaviors. They integrate numerous subsystems and
 processing mechanisms, where ACT-R is inspired by a top-down modeling
 approach, starting from cognitive operations, and Spaun from a
 bottom-up strategy, starting from neurodynamical processing
 principles.
 
 Despite this extensive research, the problem of integrating the
 conceptual-symbolic with the data-dynamical aspects of cognitive
 behavior cannot be considered solved. Quite to the contrary, two of
 the largest current research initiatives worldwide, the Human Brain
 Project \cite{hbp12} and the NIH BRAIN initiative \cite{Inseletal13},
 are ultimately driven by this problem. There are many reasons why
 this question is hard, ranging from experimental challenges of
 gathering relevant brain data to fundamental oppositions of
 philosophical paradigms. An obstinate stumbling block is the
 different mathematical nature of the fundamental formalisms which
 appear most natural for describing conceptual-symbolic versus
 data-dynamical phenomena: symbolic logic versus nonlinear dynamics.
 Logic-oriented formalisms can easily capture all that is
 combinatorially constructive and hierarchically organized in
 cognition: building new concepts by logical definitions, describing
 nested plans for action, organizing conceptual knowledge in large and
 easily extensible abstraction hierarchies. But logic is inherently
 non-temporal, and in order to capture cognitive \emph{processes},
 additional, heuristic ``scheduling'' routines have to be introduced
 which control the order in which logical rules are executed. This is
 how ACT-R architectures cope with the integration problem.
 Conversely, dynamical systems formalisms are predestined for modeling
 all that is continuously changing in the sensori-motor interface
 layers of a cognitive system, driven by sensor data streams. But when
 dynamical processing modules have to be combined into compounds that
 can solve complex tasks, again additional design elements have to be
 inserted, usually by manually coupling dynamical modules in ways that
 are informed by biological or engineering insight on the side of the
 researcher. This is how the Spaun model has been designed to realize
 its repertoire of cognitive functions.  Two important modeling
 approaches venture to escape from the logic-dynamics integration
 problem by taking resort to an altogether different mathematical
 framework which can accomodate both sensor data processing and
 concept-level representations: the framework of Bayesian statistics
 and the framework of iterated linear maps mentioned above. Both
 approaches lead to a unified formal description across processing and
 representation levels, but at the price of a double weakness in
 accounting for the embodiment of an agent in a dynamical environment,
 and for the combinatorial aspects of cognitive operations. It appears
 that current mathematical methods can instantiate only one of the
 three: continuous dynamics, combinatorial productivity, or a unified
 level-crossing description format.
 
 The conceptor mechanisms introduced in this report bi-directionally
 connect the data-dynamical workings of a recurrent neural network
 (RNN) with a conceptual-symbolic representation of different
 functional modes of the RNN.  Mathematically, conceptors are linear
 operators which characterize classes of signals that are being
 processed in the RNN. Conceptors can be represented as matrices
 (convenient in machine learning applications) or as neural
 subnetworks (appropriate from a computational neuroscience
 viewpoint). In a bottom-up way, starting from an operating RNN,
 conceptors can be learnt and stored, or quickly generated on-the-fly,
 by what may be considered the simplest of all adaptation rules:
 learning a regularized identity map.  Conceptors can be combined by
 elementary logical operations (AND, OR, NOT), and can be ordered by a
 natural abstraction relationship. These logical operations and
 relations are defined via a formal semantics.  Thus, an RNN engaged
 in a variety of tasks leads to a learnable representation of these
 operations in a logic formalism which can be neurally implemented.
 Conversely, in a top-down direction, conceptors can be inserted into
 the RNN's feedback loop, where they robustly steer the RNN's
 processing mode. Due to their linear algebra nature, conceptors can
 be continuously morphed and ``sharpened'' or ``defocussed'', which
 extends the discrete operations that are customary in logics into the
 domain of continuous ``mental'' transformations. I highlight the
 versatility of conceptors in a series of demonstrations: generating
 and morphing many different dynamical patterns with a single RNN;
 managing and monitoring the storing of patterns in a memory RNN;
 learning a class of dynamical patterns from presentations of a small
 number of examples (with extrapolation far beyond the training
 examples); classification of temporal patterns; de-noising of
 temporal patterns; and content-addressable memory systems. The
 logical conceptor operations enable an incremental extension of a
 trained system by incorporating new patterns without interfering with
 already learnt ones. Conceptors also suggest a novel answer to a
 perennial problem of attractor-based models of concept
 representations, namely the question of how a cognitive trajectory
 can leave an attractor (which is at odds with the very nature of an
 attractor). Finally, I outline a version of conceptors which is
 biologically plausible in the modest sense that only local
 computations and no information copying are needed.

\subsection{Mathematical Preliminaries}\label{secPreliminaries}

  I  assume that the reader is familiar with properties of
 positive semidefinite matrices, the singular value decomposition, and
 (in some of the analysis of adaptation dynamics) the usage of the
 Jacobian of a dynamical system for analysing stability properties.

$[a,b], (a,b), (a,b], [a,b)$ denote the closed (open, half-open)
interval between real numbers  $a$
and $b$.

$A'$ or $x'$ denotes the transpose of a matrix $A$ or vector $x$. $I$
is the identity matrix (the size will be clear from the context or be
expressed as $I_{n\times n}$). The $i$th unit vector is denoted by
$e_i$ (dimension will be clear in context). The trace of a square matrix $A$ is
denoted by $\mbox{tr}\,A$. The singular value decomposition of a
matrix $A$ is written as $USV' = A$, where $U,V$ are orthonormal and
$S$ is the diagonal matrix containing the singular values of $A$,
assumed to be in descending order unless stated otherwise. $A^\dagger$
is the pseudoinverse of $A$. All matrices and vectors will be real and
this will not be explicitly mentioned. 

I use the Matlab notation to
address parts of vectors and matrices, for instance $M(:,3)$ is the
third column of a matrix $M$ and $M(2:4,:)$ picks from $M$ the
submatrix consisting of rows 2 to 4. Furthermore, again like in
Matlab, I use the operator diag in a ``toggling'' mode:
$\mbox{diag}\,A$ returns the diagonal vector of a square matrix $A$,
and $\mbox{diag}\,d$ constructs a diagonal matrix from a vector $d$ of
diagonal elements. Another Matlab notation that will be used is
``$.\ast$'' for the element-wise multiplication of vectors and
matrices of the same size, and ``$^{.\wedge} $'' for element-wise
exponentation of vectors and matrices. 

$\mathcal{R}(A)$ and $\mathcal{N}(A)$ denote the
range and null space of a matrix $A$. For linear subspaces
$\mathcal{S}, \mathcal{T}$ of $\mathbb{R}^n$, $\mathcal{S}^{\perp}$ is the orthogonal
complement space of $\mathcal{S}$ and  $\mathcal{S} + \mathcal{T}$ is
the direct sum $\{x + y \mid x \in \mathcal{S}, y \in \mathcal{T} \}$
of $\mathcal{S}$ and $\mathcal{T}$. 
$\mathbf{P}_{\mathcal{S}}$ is the $n \times n$ dimensional
projection matrix on a linear subspace $\mathcal{S}$ of $\mathbb{R}^n$.  For a
$k$-dimensional linear subspace $\mathcal{S}$ of $\mathbb{R}^n$,
$\mathbf{B}_{\mathcal{S}}$ denotes any $n \times k$ dimensional matrix
whose columns form an orthonormal basis of $\mathcal{S}$. Such
matrices $\mathbf{B}_{\mathcal{S}}$ will occur only in contexts where
the choice of basis can be arbitrary. It holds that
$\mathbf{P}_{\mathcal{S}} = \mathbf{B}_{\mathcal{S}}
(\mathbf{B}_{\mathcal{S}})'$. 

$E[x(n)]$ denotes the expectation (temporal average) of a stationary
signal $x(n)$ (assuming it is well-defined, for instance, coming from
an ergodic source).

For a matrix $M$, $\| M \|_{\mbox{\scriptsize fro}}$ is the Frobenius
norm of $M$. For real $M$, it is the square root of the summed squared
elements of $M$. If $M$ is positive semidefinite with SVD $M = U S
U'$, $\| M \|_{\mbox{\scriptsize fro}}$ is the same as the 2-norm of the
diagonal vector of $S$, i.e.\ $\| M \|_{\mbox{\scriptsize fro}} =
((\mbox{diag}S)' \, (\mbox{diag}S))^{1/2}$. Since in this report I
will exclusively use the Frobenius norm for matrices, I sometimes omit
the subscript and write  $\| M \|$ for simplicity.

In a number of simulation experiments, a network-generated signal
$y(n)$ will be matched against a target pattern $p(n)$. The accuracy
of the match will be quantified by the normalized root mean square
error (NRMSE), 
$\sqrt{[(y(n) - p(n))^2] / [(p(n)^2]}$, where $[\cdot]$ is the mean
operator over data points $n$.

 The symbol $N$
is reserved for the size of a reservoir (= number of neurons)
throughout.

\section{Theory and Demonstrations}\label{secTheoryDemos}

This is the main section of this report. Here I develop in detail the concepts,
mathematical analysis, and algorithms, and I illustrate various
aspects in computer simulations.

Figure \ref{figOverviewTextSections} gives a navigation guide through
the dependency tree of the components of this section.

\begin{figure}[htb]
\center
\includegraphics[width=145mm]{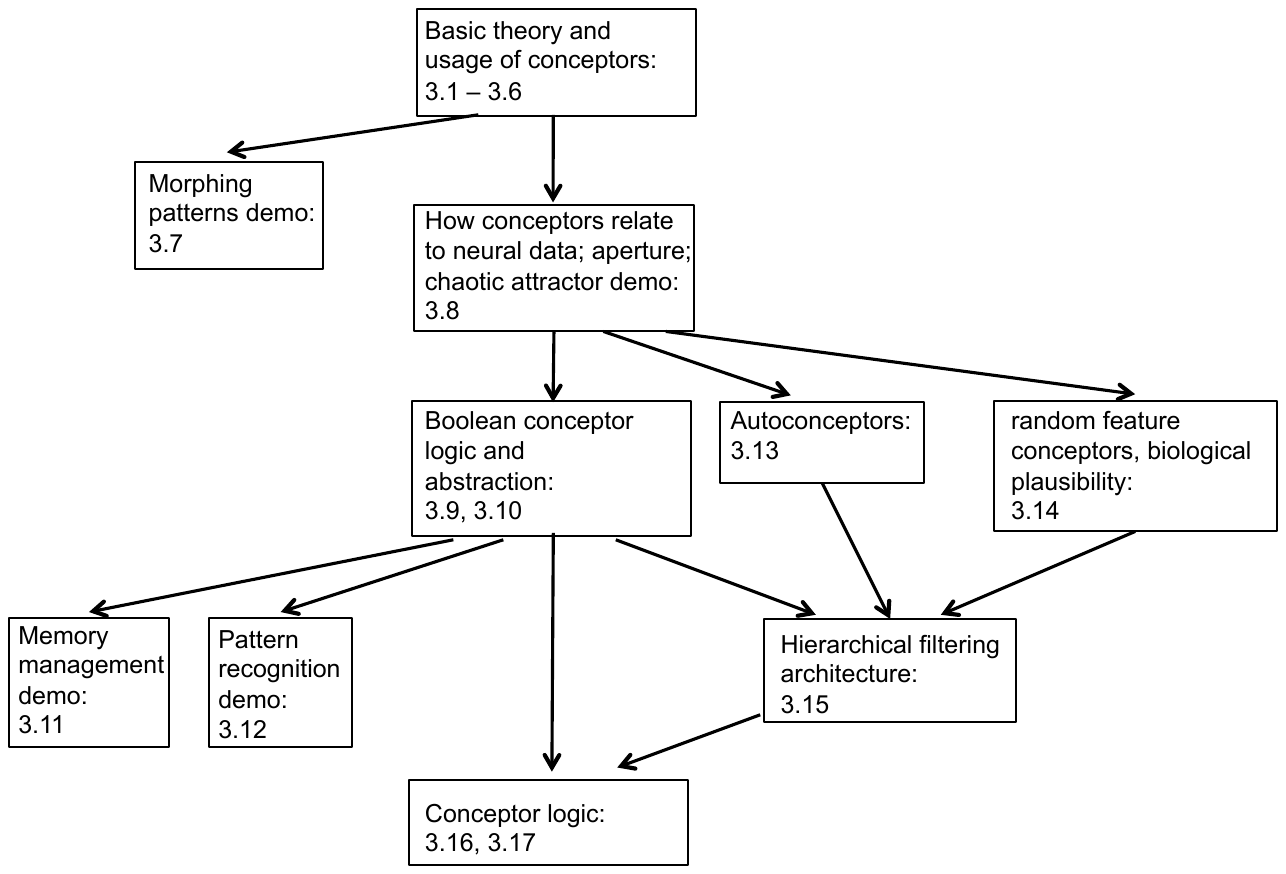}
\caption{Dependency tree of subsections in Section \ref{secTheoryDemos}.}
\label{figOverviewTextSections}
\end{figure}

 The program code (Matlab) for all simulations can be retrieved
from\\
\url{http://minds.jacobs-university.de/sites/default/files/uploads/...\\
...SW/ConceptorsTechrepR1Matlab.zip}.

\subsection{Networks and Signals}

Throughout this report, I will be using discrete-time recurrent neural
networks made of simple $\tanh$ neurons, which will be driven by an
input time series $p(n)$. In the case of 1-dimensional input, these
networks consist of (i) a ``reservoir'' of $N$ recurrently connected
neurons whose activations form a state vector $x = (x_1,\ldots,x_N)'$,
(ii) one external input neuron that serves to drive the reservoir with
training or cueing signals $p(n)$ and (iii) another external neuron
which serves to read out a scalar target signal $y(n)$ from the
reservoir (Fig.\ \ref{Fig0}). The system operates in discrete
timesteps $n = 0, 1, 2,\ldots$ according to the update equations
 \begin{eqnarray}
 x(n+1) & = & \tanh(W\,x(n) + W^{\mbox{\scriptsize
     in}}\,p(n+1) + b)\label{eq1}\\
 y(n) & = &  W^{\mbox{\scriptsize out}}\, x(n), \label{eq2}
 \end{eqnarray}
 \noindent where $W$ is the $N \times N$ matrix of reservoir-internal
 connection weights, $W^{\mbox{\scriptsize in}}$ is the $N \times 1$
 sized vector of input connection weights,  $W^{\mbox{\scriptsize
     out}}$ is the $1 \times N$ vector of readout weights, and $b$ is a
 bias. The $\tanh$ is
 a sigmoidal function that is applied to the network state $x$
 component-wise. Due to the $\tanh$, the \emph{reservoir state space}
 or simply \emph{state space} is $(-1, 1)^N$.

  The input
 weights and the bias are fixed at random values and are not subject
 to modification through training. The output weights
 $W^{\mbox{\scriptsize out}}$ are learnt. The reservoir weights $W$
 are learnt in some of the case studies below, in others they remain
 fixed at their initial random values. If they are learnt, they are
 adapted from a random initialization denoted by $W^\ast$.  Figure
 \ref{Fig0} {\bf A} illustrates the basic setup.
 
I will call the driving signals $p(n)$ \emph{patterns}. In most parts
 of this report, patterns will be periodic.  Periodicity comes in two
 variants. First, \emph{integer-periodic} patterns have the property
 that $p(n) = p(n+k)$ for some positive integer $k$. Second,
 \emph{irrational-periodic} patterns are discretely sampled from 
 continuous-time periodic signals, where the sampling interval and the
 period length of the continuous-time signal have an irrational ratio.
 An example is $p(n) = \sin(2\,\pi\, n / (10\,\sqrt{2}))$. These two
 sorts of drivers will eventually lead to different kinds of
 attractors trained into reservoirs: integer-periodic signals with
 period length $P$ yield attractors consisting of $P$ points in reservoir
 state space, while irrational-periodic signals  give rise to
 attracting sets which can be topologically characterized as
 one-dimensional cycles that are homeomorphic to the unit cycle in $\mathbb{R}^2$.

\begin{figure}[htb]
 \center
 \includegraphics[width=140 mm]{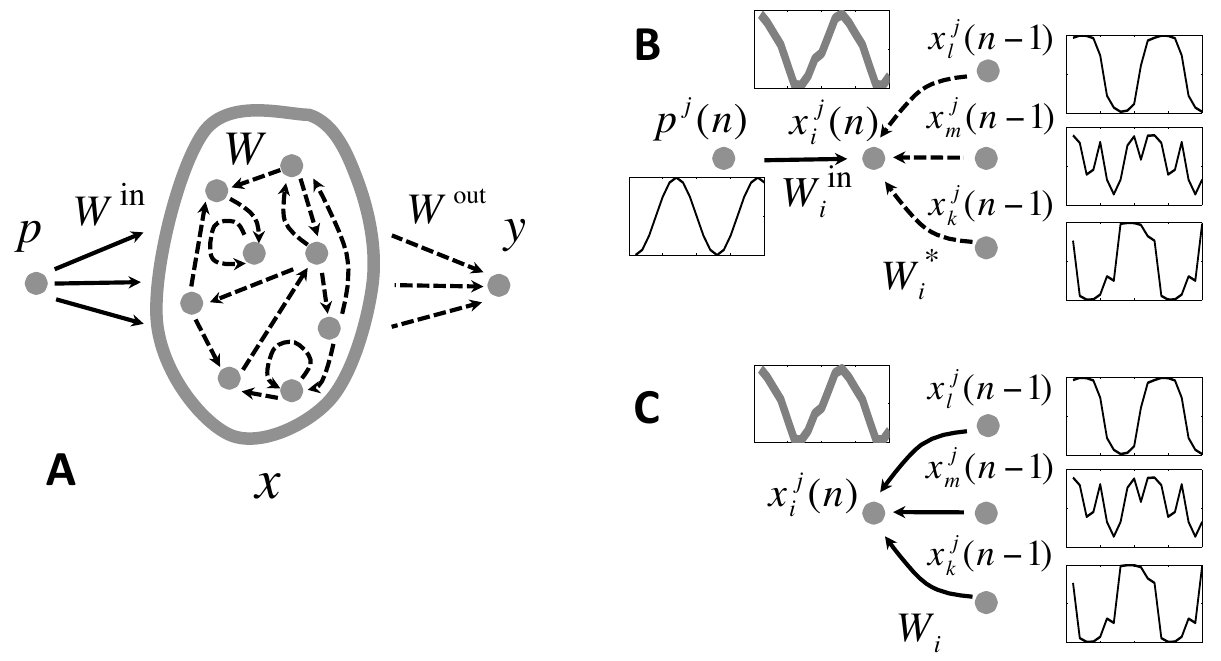}
 \caption{{\bf A.} Basic system setup. Through input connections
   $W^{\mbox{\scriptsize in}}$, an input neuron feeds a driving signal
   $p$ to a ``reservoir'' of $N = 100$ neurons which are recurrently
   connected to each other through connections $W$. From the
   $N$-dimensional neuronal activation state $x$, an output signal $y$
   is read out by connections $W^{\mbox{\scriptsize out}}$. All broken
   connections are trainable. {\bf B.} During the initial driving of the
   reservoir with driver $p^j$, using initial random weights $W^\ast$, 
   neuron $x_i$ produces its signal (thick gray line) based on external
   driving input $p$ and feeds from other neurons $x$ from within the reservoir
   (three shown).   {\bf C.} After training new reservoir weights $W$,
   the same neuron should produce the same signal based only on the
   feeds from other reservoir neurons.}
 \label{Fig0}
 \end{figure}

\subsection{Driving a Reservoir with Different Patterns}\label{sec:InitialDrivingDemo}

A basic  theme in this report is to develop methods by which a
collection of different patterns can be loaded in, and retrieved from,
a single reservoir. The key for these methods is an elementary dynamical
phenomenon: if a reservoir is driven by a pattern, the entrained
network states are confined to a linear subspace of network state
space which is characteristic of the pattern. In this subsection I
illuminate this phenomenon by a concrete example. This example
 will be re-used and extended on several occasions throughout this report. 

I use four patterns. The first two are irrational periodic and the
 last two are integer-periodic:  (1) a sinewave of period $\approx
 8.83$ sampled at integer times (pattern $p^1(n)$) (2) a
 sinewave $p^2(n)$ of period $\approx 9.83$ (period of $p^1(n)$ plus 1), (3)  a random 5-periodic
 pattern $p^3(n)$ and  (4) a slight variation $p^4(n)$ thereof (Fig.
 \ref{Fig1} left column).

 A reservoir with $N = 100$ neurons is randomly created. At creation
 time the input weights $W^{\mbox{\scriptsize in}}$ and the bias $b$
 are fixed at random values; these will never be modified thereafter.
 The reservoir weights are initialized to random values $W^\ast$; in
 this first demonstration they will not be subsequently modified either.  The
 readout weights are initially undefined (details in Section
 \ref{secGeneralSetupExpDetail}). 

In four successive and
 independent runs, the network is driven by feeding the respective
 pattern $p^j(n)$ as input ($j = 1,\ldots,4$), using the update
 rule 
\begin{displaymath}
 x^j(n+1)  =  \tanh(W^\ast\,x^j(n) + W^{\mbox{\scriptsize
     in}}\,p^j(n+1) + b).
 \end{displaymath}
 After an initial washout time, the reservoir dynamics becomes
 entrained to the driver and the reservoir state $x^j(n)$ exhibits an
 involved nonlinear response to the driver $p^j$. After this washout,
 the reservoir run is continued for $L = 1000$ steps, and the obtained
 states $x^j(n)$ are collected into  $N \times L = 100 \times 1000$ sized
 state collection matrices $X^j$ for subsequent use.

 The second column
 in Fig. \ref{Fig1} shows traces of three randomly chosen
 reservoir neurons in the four driving conditions. It is apparent that
 the reservoir has become entrained to the driving input.
 Mathematically, this  entrainment is captured by the concept of the
 \emph{echo state property}: any random initial state of a reservoir
 is ``forgotten'', such that after a washout period the current
 network state is a function of the driver. The echo state property is
 a fundamental condition for RNNs to be useful in learning tasks \cite{Jaeger01a,BuehnerYoung06,HermansSchrauwen11,Yildizetal12,Straussetal12,ManjunathJaeger13}. It
 can be ensured by an appropriate scaling of the reservoir weight
 matrix. All networks employed in this report possess the echo state
 property.

 \begin{figure}[htb]
 \center
 \includegraphics[width=145 mm]{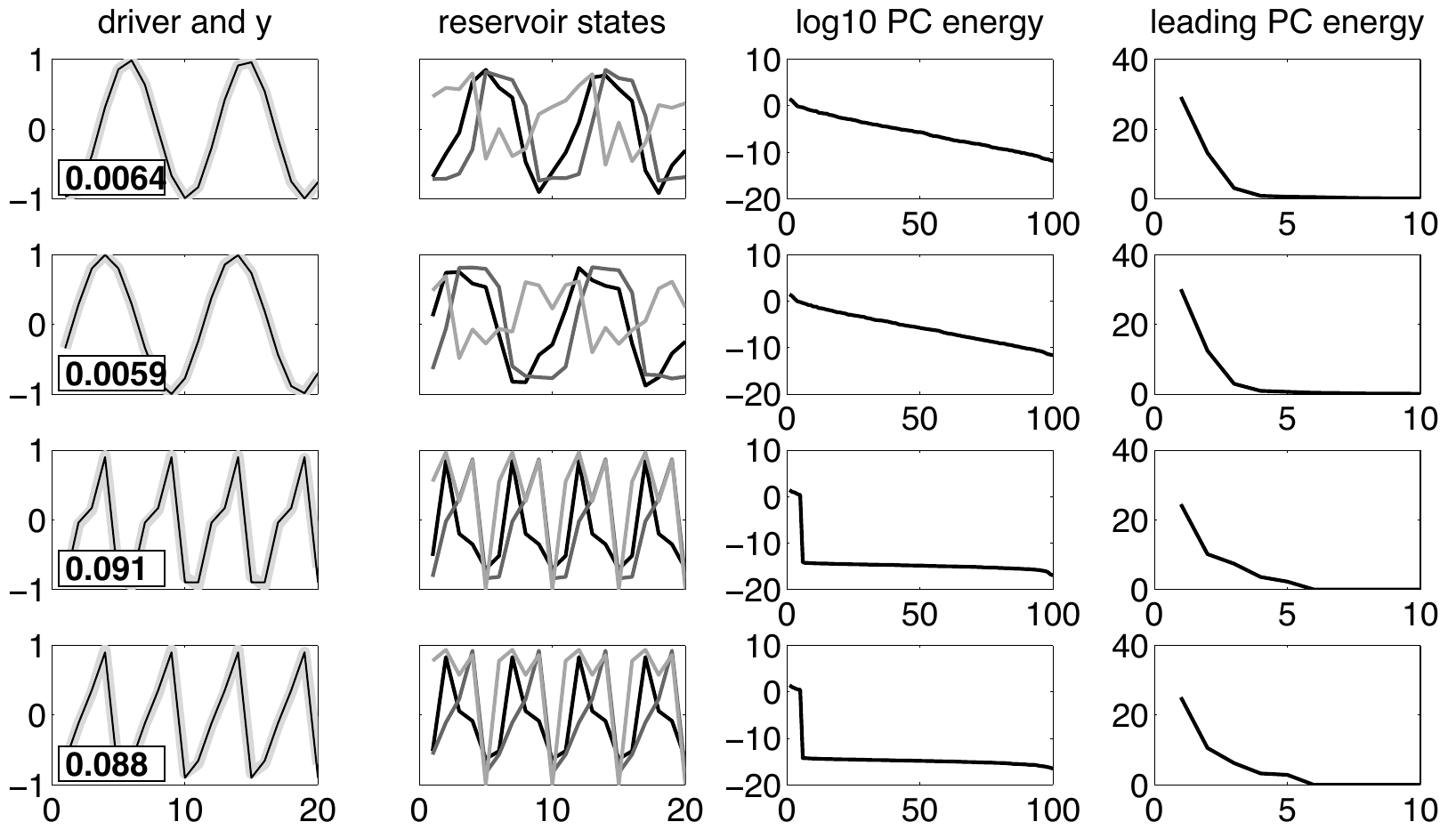}
 \caption{The subspace phenomenon.  Each row of panels documents
   situation when the reservoir is driven by a particular input
   pattern. ``Driver and y'': the driving pattern (thin black line)
   and the signals retrieved with conceptors (broad light gray line).
   Number inset is the NRMSE between original driver and retrieved
   signal. ``Reservoir states'': activations of three randomly picked
   reservoir neurons. ``Log10 PC energy'': $\log_{10}$ of reservoir
   signal energies in the principal component directions. ``Leading PC
   energy'': close-up on first ten signal energies in linear scale.
   Notice that the first two panels in each row show discrete-time
   signals; points are connected by lines only for better visual
   appearance.}
 \label{Fig1}
 \end{figure}
 
 A principal component analysis (PCA) of the 100 reservoir signals
 reveals that the driven reservoir signals are concentrated on a few
 principal directions. Concretely, for each of the four driving
 conditions, the reservoir state correlation matrix was estimated by
 $R^j = X^j\,(X^j)' / L$, and its SVD $U^j \Sigma^j (U^j)' = R^j$ was
 computed, where the columns of $U^j$ are orthonormal eigenvectors of
 $R^j$ (the principal component (PC) vectors), and the diagonal of
 $\Sigma^j$ contains the singular values of $R^j$, i.e.\ the energies (mean
 squared amplitudes) of the principal signal components. Figure
 \ref{Fig1} (third and last column) shows a plot of these principal
 component energies. The energy spectra induced by the two
 irrational-period sines look markedly different from the spectra
 obtained from the two 5-periodic signals. The latter lead to nonzero
 energies in exactly 5 principal directions because the driven
 reservoir dynamics periodically visits 5 states (the small but
 nonzero values in the $\log_{10}$ plots in Figure \ref{Fig1} are
 artefacts earned from rounding errors in the SVD computation). In
 contrast, the irrational-periodic drivers lead to reservoir states
 which linearly span all of $\mathbb{R}^N$ (Figure \ref{Fig1}, upper two
 $\log_{10}$ plots). All four drivers however share a relevant
 characteristic (Figure \ref{Fig1}, right column): the total reservoir
 energy is concentrated in a quite small number of leading principal
 directions.

When one inspects the excited reservoir dynamics in these four driving
conditions, there is little surprise that the neuronal activation
traces look similar to each other for the first two and in the second
two cases (Figure \ref{Fig1}, second column). This ``similarity'' can
be quantified in a number of ways. Noting that the geometry of the
``reservoir excitation space'' in driving condition $j$ is
characterized by a hyperellipsoid with main axes $U^j$ and axis
lengths $\mbox{diag}\,\Sigma^j$, a natural way to define a similarity
between two such ellipsoids $i, j$ is to put 
\begin{equation}\label{eqSimR}
\mbox{sim}_{i,j}^R = \frac{\| (\Sigma^i)^{1/2}\, (U^i)' U^j
  (\Sigma^j)^{1/2}\|^2}{\|\mbox{diag}\Sigma^i \| \, \|\mbox{diag}\Sigma^j \|}.
\end{equation}
 The measure $\mbox{sim}_{i,j}^R$
ranges in $[0,1]$. It is 0 if and only if the reservoir signals $x^i,
x^j$ populate orthogonal linear subspaces, and it is 1 if and only if
$R^i = a\,R^j$ for some scaling factor $a$. The measure
$\mbox{sim}_{i,j}^R$ can be understood as a generalized squared cosine between
$R^i$ and $R^j$. Figure \ref{Figcompare} {\bf  A} shows the
similarity matrix $(\mbox{sim}_{i,j}^R)_{i,j}$ obtained from
(\ref{eqSimR}). The similarity values contained in this matrix appear
somewhat counter-intuitive, inasmuch as the reservoir responses to the
sinewave patterns come out as having similarities of about 0.6 with
the 5-periodic driven reservoir signals; this does not agree with the
strong visual dissimilarity apparent in the
state plots in Figure \ref{Fig1}. In Section \ref{sec:SimMeasure} I
will introduce another similarity measure which agrees better with
intuitive judgement.

\begin{figure}[htbp]
 \center
{\bf  A}
 \includegraphics[width=40 mm]{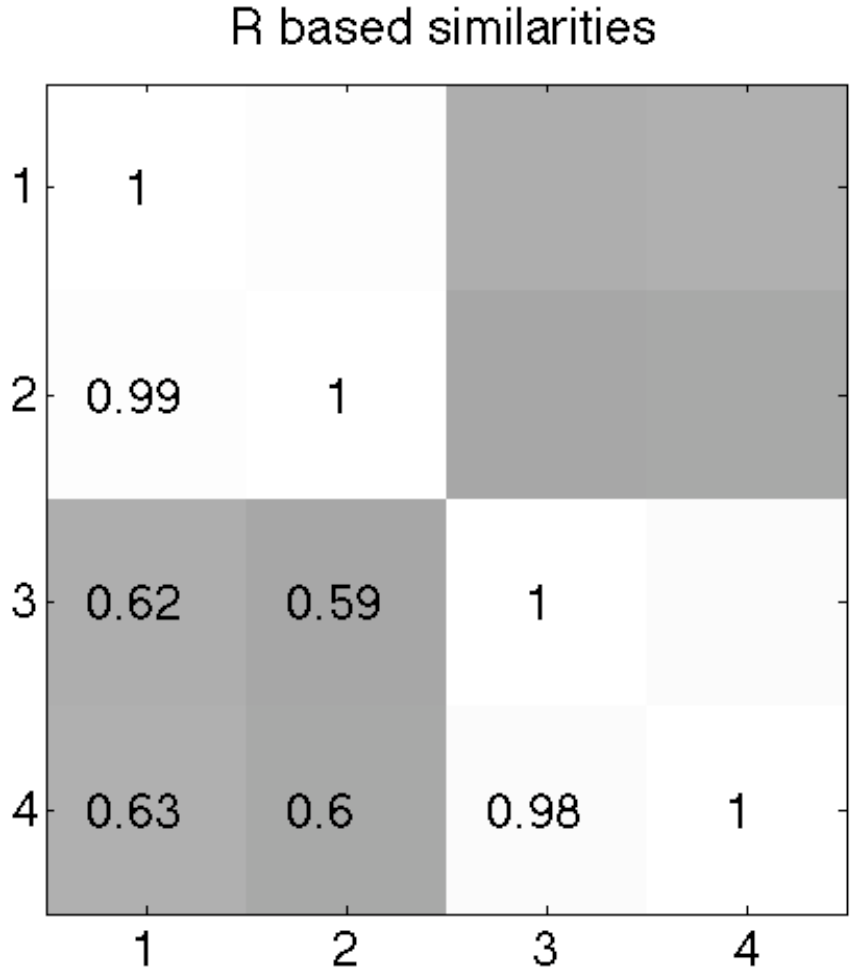}
\hspace{0.3cm}{\bf  B}
\includegraphics[width=40 mm]{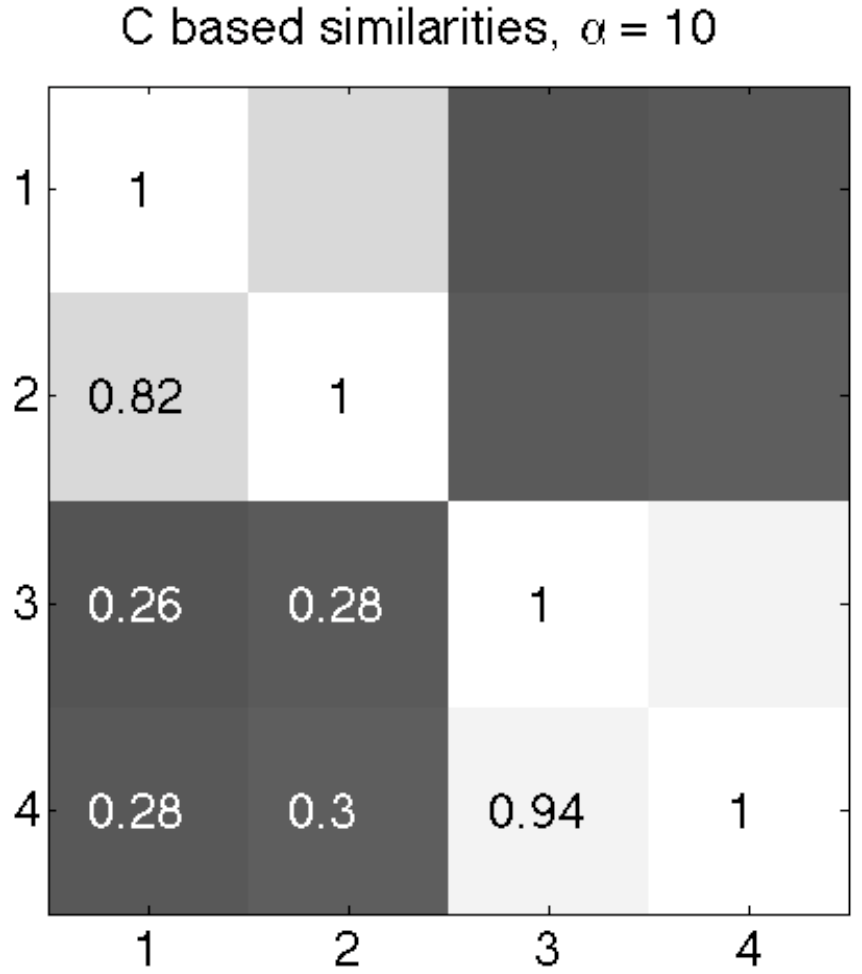}
\hspace{0.3cm}{\bf  C}
\includegraphics[width=40 mm]{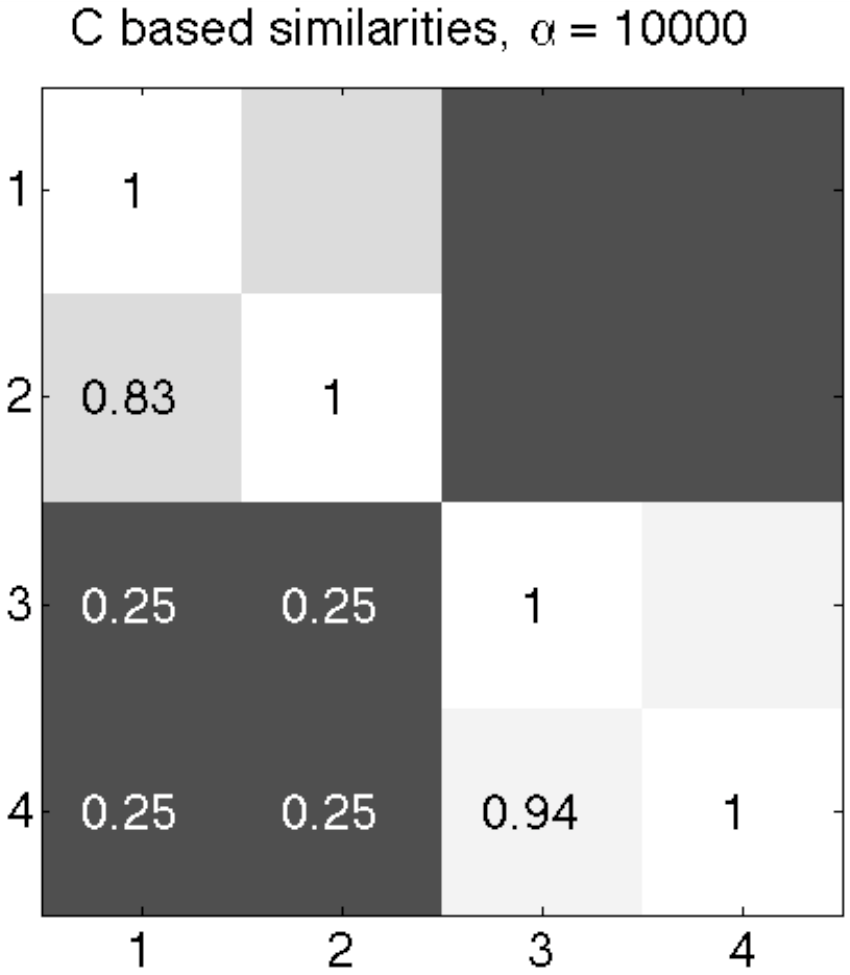}
 \caption{   Matrix plots of pairwise similarity  between the
 subspaces excited in the four driving conditions.
   Grayscale coding: 0 = black, 1 = white. {\bf  A:} similarity $\mbox{sim}_{i,j}^R$
   based on the data correlation matrices $R^i$. {\bf  B,C:}
   similarities based on conceptors $C(R^i,\alpha)$ for two different
   values of aperture $\alpha$. For explanation see text.}
 \label{Figcompare}
 \end{figure}

\subsection{Storing Patterns in a Reservoir, and Training the
  Readout}\label{sec:StoringGeneric} 

One of the objectives of this report is a method for storing several
driving patterns in a single reservoir, such that these stored
patterns can later be retrieved and otherwise be controlled or
manipulated. In this subsection I explain how the initial ``raw''
reservoir weights $W^\ast$ are adapted in order to ``store'' or
``memorize'' the drivers, leading to a new reservoir weight matrix
$W$. I continue with the four-pattern-example used above.
 
 The guiding idea
 is to enable the reservoir to re-generate the driven responses
 $x^j(n)$ \emph{in the absence of the driving input}. Consider any neuron $x_i$
 (Fig.\ \ref{Fig0}{\bf B}). During the driven runs $j = 1,\ldots,4$,
 it has been updated per 
\begin{displaymath}
x^j_i(n+1) = \tanh(W_i^\ast\,x^j(n) +
 W_i^{\mbox{\scriptsize in}}\, p^j(n+1) + b_i),
\end{displaymath}
where $W_i^\ast$ is the $i$-th row in $W^\ast$,
$W_i^{\mbox{\scriptsize in}}$ is the $i$-th element of
$W^{\mbox{\scriptsize in}}$, and $b_i$ is the $i$th bias
component. The objective for determining new reservoir weights $W$ is
that the trained reservoir should be able to oscillate in the same
four ways as in the external driving conditions, but without the
driving input --- $W$ should have 'internalized' the effects of
external input.  That is, the new input internalization weights $W_i$
leading to neuron $i$ should approximate
\begin{displaymath}
\tanh(W_i^\ast\,x^j(n) +
W_i^{\mbox{\scriptsize in}}\, p^j(n+1) + b_i) \approx \tanh(W_i\,x^j(n)
+ b_i)
\end{displaymath}
as accurately as possible, for $j = 1,\ldots,4$. Concretely, we
optimize a mean square error criterion and compute
\begin{equation}\label{eq:storeMSECrit}
W_i = \mbox{argmin}_{\tilde{W}_i} \sum_{j = 1,\ldots,K} \sum_{n =
  1,\ldots,L} (W_i^\ast\,x^j(n) +
W_i^{\mbox{\scriptsize in}}\, p^j(n+1) - \tilde{W}_i\,x^j(n))^2,
\end{equation}
where $K$ is the number of patterns to be stored (in this example $K =
4$). This is a  linear regression task, for which a number of standard
algorithms are available. I employ ridge regression (details in Section
 \ref{secGeneralSetupExpDetail}).

 The readout neuron $y$  serves as passive observer of the reservoir
 dynamics. The objective to determine its connection weights $W^{\mbox{\scriptsize
     out}}$ is simply to replicate the driving input, that is, $W^{\mbox{\scriptsize
     out}}$ is computed (again by ridge regression) such
 that it minimizes the squared error $(p^j(n) - W^{\mbox{\scriptsize
     out}} x^j(n))^2$, averaged over time and the four driving
 conditions. 

 I will refer to this preparatory training as \emph{storing} patterns
 $p^j$ in a reservoir, and call a reservoir \emph{loaded} after
 patterns have been stored. 

\subsection{Conceptors: Introduction and Basic Usage in
 Retrieval}\label{sec:RetrieveGeneric}

How can
 these stored patterns be individually \emph{retrieved} again? After all, the
 storing process has superimposed impressions of all patterns on all
 of the re-computed
 connection weights $W$ of the network -- very much like the pixel-wise
 addition of different images would yield a mixture image in which the
 individual original images are hard to discern. One would need some
 sort of filter which can disentangle again the superimposed components
 in the connection weights. In this section I explain how such 
 filters can be obtained. 
 
 The guiding idea is that for retrieving pattern $j$ from a loaded
 reservoir, the reservoir dynamics should be restricted to the linear
 subspace which is characteristic for that pattern. For didactic
 reasons I start with a simplifying assumption (to be dropped
 later).  Assume that there exists a (low-dimensional) linear
 subspace $\mathcal{S}^j \subset \mathbb{R}^N$ such that all state
 vectors contained in the driven state collection $X^j$ lie in
 $\mathcal{S}^j$. In our  example, this is actually the case
 for the two 5-periodic patterns. Let
 $\mathbf{P}_{\mathcal{S}^j}$ be the projector matrix which projects
 $\mathbb{R}^N$ on $\mathcal{S}^j$. We may then hope that if we run
 the loaded reservoir autonomously (no input), constraining its states
 to $\mathcal{S}^j$  using the update rule
\begin{equation}\label{eqProjectorConstraint}
x(n+1)  = \mathbf{P}_{\mathcal{S}^j}  \, \tanh(W\,x(n)  + b),
\end{equation}
it will oscillate in a way that is closely related to the way how it
oscillated when it was originally driven by $p^j$. 

However, it is not typically the case that the states obtained in the
original driving conditions are confined to a proper linear subspace
of the  reservoir
state space.  Consider the sine driver $p^1$ in our example. The linear
span of the reservoir response state is all of $\mathbb{R}^N$ (compare
the $\log_{10}$ PC energy plots in Figure \ref{Fig1}). The associated
projector would be the identity, which would not help to single out an
individual pattern in retrieval.  But actually we are not interested
in those principal directions of reservoir state space whose
excitation energies are negligibly small (inspect again the quick drop
of these energies in the third column, top panel in Figure \ref{Fig1}
-- it is roughly exponential over most of the spectrum, except for an
even faster decrease for the very first few singular values). Still
considering the sinewave pattern $p^1$: instead of
$\mathbf{P}_{\mathbb{R}^N}$ we would want a projector that projects on
the subspace spanned by a ``small'' number of leading principal
components of the ``excitation ellipsoid'' described by the
sine-driver-induced correlation matrix $R^1$.  What qualifies as a
``small'' number is, however, essentially arbitrary. So we want a
method to shape projector-like matrices from reservoir state
correlation matrices $R^j$ in a way that we can adjust, with a
control parameter, how many of the leading principal components should
become registered in the projector-like matrix.

At this point I  give names to the 
projector-like matrices and the adjustment parameter. I call the
latter the \emph{aperture} parameter, denoted by $\alpha$. The
projector-like matrices will be called \emph{conceptors} and generally
be denoted by the symbol $C$. Since conceptors are derived from the
ellipsoid characterized by a reservoir state corrlation matrix $R^j$,
and parametrized by the aperture parameter, I also sometimes write
$C(R^j,\alpha)$ to make this dependency transparent.  

There is a natural and convenient solution to meet all the intuitive
objectives for conceptors that I discussed up to this point. Consider
a reservoir driven by a pattern $p^j(n)$, leading to driven states
$x^j(n)$ collected (as columns) in a state collection matrix $X^j$,
which in turn yields a reservoir state correlation matrix $R^j =
X^j(X^j)' / L$. We define a conceptor $C(R^j,\alpha)$ with the aid of
a cost function $\mathcal{L}(C\, |\, R^j,\alpha)$, whose minimization yields
$C(R^j,\alpha)$. The cost function has two components. The first
component reflects the objective that $C$ should behave as a projector
matrix for the states that occur in the pattern-driven run of the
reservoir. This component is $E_n[\| x^j(n) - Cx^j(n) \|^2]$, the time-averaged
deviation of projections $Cx^j$
from the state vectors $x^j$.  The second component of $\mathcal{L}$
adjusts how many of the leading directions of $R^j$ should become
effective for the projection. This component is
$\alpha^{-2}\|C\|^2_{\mbox{\scriptsize fro}}$. This leads to the
following definition.

\begin{definition}\label{defConceptor}
Let $R = E[xx']$ be an $N \times N$ correlation matrix and $\alpha \in
(0,\infty)$. The \emph{conceptor matrix} $C = C(R,\alpha)$ \emph{associated with} $R$ \emph{and} $\alpha$
is 
\begin{equation}\label{eqDefConceptor}
C(R,\alpha) = \mbox{\emph{argmin}}_{C}\; E[\| x - Cx \|^2] +
\alpha^{-2}\,\|C\|^2_{\mbox{\scriptsize \emph{fro}}}. 
\end{equation}
\end{definition}

The minimization criterion (\ref{eqDefConceptor}) uniquely specifies
$C(R,\alpha)$. The conceptor matrix can be effectively computed from
$R$ and $\alpha$. This is spelled out in the following proposition,
which also lists elementary algebraic properties of conceptor
matrices:

\begin{proposition}\label{propCompConceptor}
Let $R = E[x\,x']$ be a correlation matrix and $\alpha \in
(0,\infty)$. Then,
\begin{enumerate} 
\item $C(R,\alpha)$ can be directly computed from $R$ and $\alpha$ by
\begin{equation}\label{eq:CompConceptor}
C(R,\alpha) = R\,(R + \alpha^{-2}\,I)^{-1} = (R + \alpha^{-2}\,I)^{-1}
\, R,
\end{equation}
\item if $R = U \Sigma U'$ is the SVD of $R$, then the SVD of
$C(R,\alpha)$ can be written as $C = U S U'$, i.e. $C$ has the same
principal component vector orientation as $R$, 
\item the singular values $s_i$ of $C$ relate to the singular values
$\sigma_i$ of $R$ by $s_i = \sigma_i / (\sigma_i + \alpha^{-2})$,
\item the singular values of $C$ range in  $[0,1)$,
\item $R$ can be recovered from $C$ and $\alpha$ by
\begin{equation}\label{eqRecoverR}
R = \alpha^{-2}\,(I - C)^{-1}\, C = \alpha^{-2}\,C \, (I - C)^{-1}.
\end{equation}
\end{enumerate}
\end{proposition} 

The proof is given in Section \ref{secProofPropcompconceptor}. Notice that
all inverses appearing in this proposition are well-defined because
$\alpha > 0$ is assumed, which implies that all singular values of
$C(R,\alpha)$ are properly smaller than 1. I will later
generalize conceptors to include the limiting cases  $\alpha = 0$ and
$\alpha = \infty$ (Section \ref{secApertureSemantics}).  

In practice, the correlation matrix $R = E[xx']$ is  estimated from a finite sample
$X$, which leads to the approximation $\hat{R} = X X' / L$, where $X =
(x(1),\ldots,x(L))$ is a matrix containing reservoir states $x(n)$
collected during a learning run. 

Figure \ref{FigSingValsFalloff} shows the singular value spectra of
$C(R,\alpha)$ for various values of $\alpha$, for our example cases of
$R = R^1$ (irrational-period sine driver) and $R = R^3$ (5-periodic
driver). We find that the nonlinearity inherent in
(\ref{eq:CompConceptor}) makes the conceptor matrices come out ``almost'' as
projector matrices: the singular values of $C$ are mostly close to 1
or close to 0. In the case of the 5-periodic driver, where the
excited network states populate a 5-dimensional subspace of
$\mathbb{R}^N$, increasing $\alpha$ lets $C(R,\alpha)$ converge to a
projector onto that subspace.

\begin{figure}[htbp]
 \center
\includegraphics[width=140 mm]{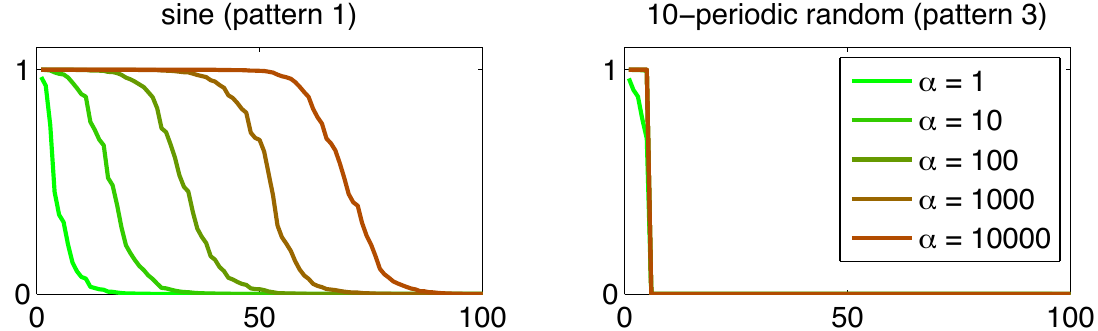}
 \caption{How the singular values of a conceptor depend on
   $\alpha$. Singular value spectra are shown for the first sinewave
   pattern and the first 5-periodic random pattern. For explanation
   see text.}
 \label{FigSingValsFalloff}
 \end{figure}

If one has a conceptor matrix $C^j = C(R^j,\alpha)$ derived from a pattern $p^j$
through the reservoir state correlation matrix $R^j$ associated with
that pattern, the conceptor matrix can be used in an autonomous run (no
external input) using the update rule
 \begin{equation}\label{eqCconstrainedRule}
x(n+1)  = C^j  \, \tanh(W\,x(n)  + b),
\end{equation}
where the weight matrix $W$ has been shaped by storing patterns among
which there was $p^j$. Returning to our example, four conceptors
$C^1,\ldots, C^4$ were computed with $\alpha = 10$ and the loaded
reservoir was run under rule (\ref{eqCconstrainedRule}) from a random
initial state $x(0)$. After a short washout period, the network
settled on stable periodic dynamics which were closely related to the
original driving patterns. The network dynamics was observed through
the previously trained output neuron.  The left column in Figure
\ref{Fig1} shows the autonomous network output as a light bold gray
line underneath the original driver. To measure the achieved accuracy,
the autonomous output signal was  phase-aligned with the
driver (details in Section \ref{secGeneralSetupExpDetail}) and then the NRMSE
was computed (insets in Figure panels). The NRMSEs indicate that the
conceptor-constrained autonomous runs could successfully separate from
each other even the closely related pattern pairs $p^1$ versus $p^2$ and $p^3$ versus 
$p^4$. 

\paragraph*{A note on terminology.} Equation (\ref{eqCconstrainedRule})
shows a main usage of conceptor matrices: they are inserted into the
reservoir state feedback loop and cancel (respectively, dampen) those
reservoir state  components which correspond to
directions in state space associated with zero (or small,
respectively) singular values in the conceptor matrix. In most of this
report, such a direction-selective damping in the reservoir feedback
loop will be effected by way of inserting matrices $C$ like in
Equation (\ref{eqCconstrainedRule}).  However, inserting a matrix is
not the only way by which such a direction-selective damping can be
achieved. In Section \ref{secBiolPlausible}, which deals with
biological plausibility issues, I will propose a neural circuit which
achieves a similar functionality of direction-specific damping of
reservoir state components by other means and with slightly
differing mathematical properties. I understand the concept of a
``conceptor'' as comprising any mechanism which effects a
pattern-specific damping of reservoir signal components.  Since in
most parts of this report this will be achieved with conceptor
matrices, as in (\ref{eqCconstrainedRule}), I will often refer to
these $C$ matrices as ``conceptors'' for simplicity.  The reader
should however bear in mind that the notion of a conceptor is more
comprehensive than the notion of a conceptor matrix. I will not spell
out a formal definition of a ``conceptor'', deliberately leaving this
concept open to become instantiated by a variety of computational
mechanisms of which only two are formally defined in this report (via
conceptor matrices, and via the neural circuit given in Section
\ref{secBiolPlausible}).

\subsection{A Similarity Measure for Excited Network
  Dynamics}\label{sec:SimMeasure}  

In Figure \ref{Figcompare} {\bf A} a similarity matrix is presented
which compares the excitation ellipsoids represented by the
correlation matrices $R^j$ by the similarity metric (\ref{eqSimR}). I
remarked at that time that this is not a fully satisfactory metric,
because it does not agree well with intuition. We obtain a more
intuitively adequate similiarity metric if conceptor matrices are used as
descriptors of ``subspace ellipsoid geometry'' instead of the raw
correlation matrices, i.e.\ if we employ the metric
\begin{equation}\label{eqSimC}
\mbox{sim}_{i,j}^\alpha = \frac{\| (S^i)^{1/2}\, (U^i)' U^j
  (S^j)^{1/2}\|^2}{\|\mbox{diag}S^i \| \, \|\mbox{diag}S^j \|},
\end{equation}
where $US^jU'$ is the SVD of $C(R^j,\alpha)$. Figure \ref{Figcompare}
{\bf B},{\bf C} shows the similarity matrices arising in our standard
example for $\alpha = 10$ and $\alpha = 10,000$. The intuitive
dissimilarity between the sinewave and the 5-periodic patterns, and
the intuitive similarity between the two sines (and the two
5-periodic pattern versions, respectively) is revealed much more
clearly than on the basis of $\mbox{sim}_{i,j}^R$. 

When interpreting similarities $\mbox{sim}_{i,j}^R$ or
$\mbox{sim}_{i,j}^\alpha$ one should bear in mind that one is not
comparing the original driving patterns but the excited reservoir
responses. 

\subsection{Online Learning of
   Conceptor Matrices}\label{secOnlineAdaptC}

The minimization criterion (\ref{eqDefConceptor}) immediately leads to
a stochastic gradient  online  method for adapting $C$:
\begin{proposition}\label{propStochAdaptC}
  Assume that a stationary source $x(n)$ of $N$-dimensional reservoir
  states is available. Let $C(1)$ be any $N \times N$ matrix, and
  $\lambda > 0$ a  learning rate. Then the stochastic gradient
  adaptation
\begin{equation}\label{eqStochAdaptC}
C(n+1) = C(n) + \lambda\, \left((x(n) - C(n)\,x(n))\,x'(n) - \alpha^{-2}\,C(n)\right)
\end{equation}
will lead to $\lim_{\lambda \downarrow 0}\, \lim_{n \to \infty}\,C(n) = C(E[x\,x'], \alpha)$.
\end{proposition}
The proof is straightforward if one employs generally known facts
about stochastic gradient descent and the fact that $E[\| x - Cx \|^2] +
\alpha^{-2}\,\|C\|^2_{\mbox{\scriptsize {fro}}}$ is positive
  definite quadratic in
  the $N^2$-dimensional space of elements of $C$ (shown in the proof
  of Proposition \ref{propCompConceptor}), and hence provides a
  Lyapunov function for the gradient descent
  (\ref{eqStochAdaptC}). The gradient  of $E[\| x - Cx \|^2] +
\alpha^{-2}\,\|C\|^2_{\mbox{\scriptsize {fro}}}$ with respect to $C$
is
\begin{equation}\label{eqGradCallo}
\frac{\partial}{\partial C} \; E[\| x - Cx \|^2] +
\alpha^{-2}\,\|C\|^2_{\mbox{\scriptsize {fro}}} = (I - C)\,E[xx'] - \alpha^{-2}\,C,
\end{equation}
which immediately yields (\ref{eqStochAdaptC}).

The stochastic update rule (\ref{eqStochAdaptC}) is very elementary.
It is driven by two components, (i) an error signal $x(n) -
C(n)\,x(n)$ which simply compares the current state with its
$C$-mapped value, and (ii) a linear decay term. We will make heavy use
of this adaptive mechanism in Sections \ref{secAutoC} \emph{ff.} This
observation is also illuminating the intuitions behind the definition
of conceptors. The two components strike a compromise (balanced by
$\alpha$) between (i) the objective that $C$ should leave reservoir
states from the target pattern unchanged, and (ii) $C$ should have
small weights. In the terminology of machine learning one could say,
``a conceptor is a regularized identity map''.

\subsection{Morphing Patterns}\label{secMorphing}

Conceptor matrices offer a way to morph RNN dynamics. Suppose that a reservoir
has been loaded with some patterns, among which there are $p^i$ and
$p^j$ with corresponding conceptors $C^i, C^j$. Patterns that are
intermediate between $p^i$ and $p^j$ can be obtained by running the
reservoir via (\ref{eqCconstrainedRule}), using a linear mixture
between $C^i$ and $C^j$:
\begin{equation}\label{eqMorph}
x(n+1) = \left((1-\mu)C^i + \mu C^j   \right)\,\tanh(W \, x(n) + b).
\end{equation}
Still using our four-pattern example, I demonstrate how this morphing
works out for morphing (i) between the two sines, (ii) between the two
5-periodic patterns, (iii) between a sine and a 5-periodic pattern.

\paragraph*{Frequency Morphing of Sines}

In this demonstration, the morphing was done for the two sinewave
conceptors $C^1 = C(R^1,10)$ and $C^2 = C(R^2,10)$. The morphing
parameter $\mu$ was allowed to range from $-2$ to $+3$ (!). The
four-pattern-loaded reservoir was run from a random initial state for
500 washout steps, using (\ref{eqMorph}) with $\mu = -2$. Then
recording was started. First, the run was
continued with the intial $\mu = -2$ for 50 steps. Then, $\mu$ was
linearly ramped up from $\mu = -2$ to $\mu = 3$ during 200 steps.
Finally, another 50 steps were run with the final setting $\mu = 3$. 

Note that morph values $\mu = 0$ and $\mu = 1$ correspond to
situations where the reservoir is constrained by the original
conceptors $C^1$ and $C^2$, respectively. Values $0 \leq \mu \leq 1$
correspond to interpolation. Values $-2 \leq \mu < 0$ and $1 < \mu
\leq 3$ correspond to extrapolation. The extrapolation range on either
side  is twice as long as the interpolation range. 

In addition, for eight equidistant values $\mu_k$ in $-2 \leq \mu < 3$, the
reservoir was run with a mixed conceptor $C = (1-\mu_k)C^1 + \mu_k
C^2$ for 500 steps, and the obtained observation signal $y(n)$ was
plotted in a delay-embedded representation, yielding ``snapshots'' of
the reservoir dynamics at these $\mu$ values (a delay-embedding plot of a
1-dimensional signal $y(n)$ creates a 2-dimensional plot by plotting
value pairs $(y(n),y(n-d))$ with a delay $d$ chosen to yield an
appealing visual appearance).

\begin{figure}[htb]
 \center
\includegraphics[width=145 mm]{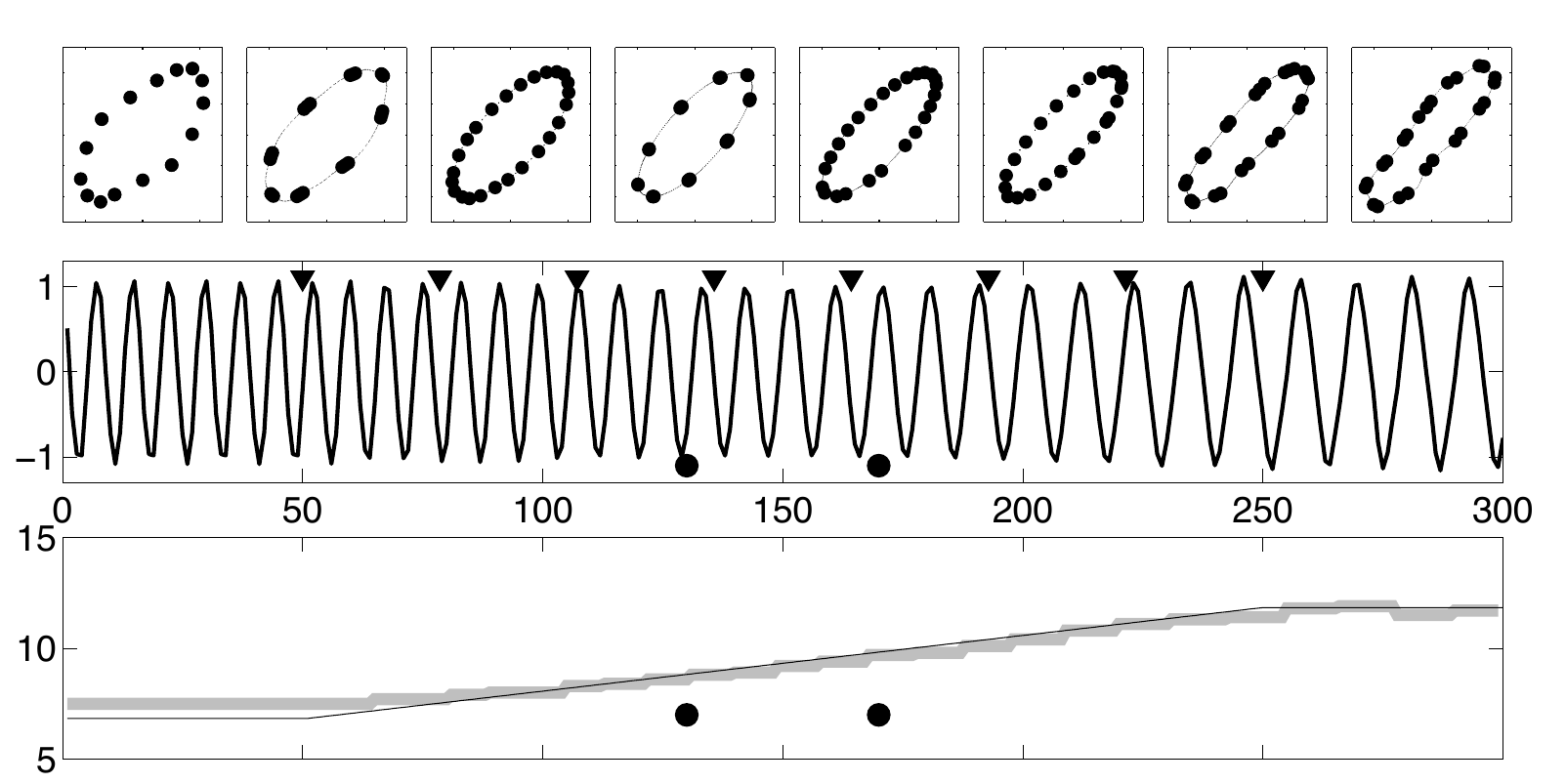}
 \caption{Morphing between (and beyond) two sines. The morphing range
   was $-2 \leq \mu \leq 3$. Black circular dots in the two bottom
   panels mark the points $\mu = 0$ and $\mu =
   1$, corresponding to situations where the two original conceptors
   $C^1, C^2$ were active in unadulterated form. Top: Delay-embedding
   plots of network observation signal $y(n)$ (delay = 1 step). Thick
   points show 25 plotted points, thin points show 500 points
   (appearing as connected line). The eight panels have a plot range
   of $[-1.4, 1.4] \times [-1.4, 1.4]$. Triangles in center panel mark the
   morph positions corresponding to the delay embedding ``snapshots''.
   Center: the network observation signal $y(n)$ of a morph run. Bottom: Thin black
   line: the period length obtained from morphing between (and
   extrapolating beyond) the original period lengths. Bold gray line:
   period lengths measured from the observation signal $y(n)$.}
 \label{FigMorphSines}
 \end{figure}
 
 Figure \ref{FigMorphSines} shows the findings. The reservoir
 oscillates over the entire inter/extrapolation range with a waveform
 that is approximately equal to a sampled sine. At the morph values $\mu = 0$
 and $\mu = 1$ (indicated by dots in the Figure), the system is in
 exactly the same modes as they were plotted earlier in the first two
 panels of the left column in Figure \ref{Fig1}. Accordingly the fit
 between the original driver's period lenghtes and the autonomously
 re-played oscillations is as good as it was reported there (i.e.
 corresponding to a steady-state NRMSE of about 0.01). In the
 extrapolation range, while the linear morphing of the mixing
 parameter $\mu$ does not lead to an exact linear morphing of the
 observed period lengths, still the obtained period lengths steadily
 continue to decrease (going left from $\mu = 0$) and to increase
 (going right from $\mu = 1$). 

In sum, it is possible to use conceptor-morphing to extend
sine-oscillatory reservoir dynamics from two learnt oscillations of
periods $\approx 8.83, 9.83$ to a range between $\approx 7.5 - 11.9$
(minimal and maximal values of period lengths shown in the
Figure). The post-training sinewave generation thus \emph{extrapolated}
 beyond the period range spanned by the two training
samples by a factor of about 4.4. From a perspective of machine
learning this extrapolation  is  remarkable.  Generally speaking, when neural pattern
generators are trained from demonstration data (often done in
robotics, e.g. \cite{ItoTani04,Reinhartetal12}), \emph{interpolation}
of recallable patterns is what one expects to achieve, while
extrapolation is deemed hard.

From a perspective of neurodynamics, it is furthermore remarkable that
the dimension of interpolation/extrapolation was the \emph{speed} of
the oscillation. Among the infinity of potential generalization
dimensions of patterns, speedup/slowdown of pattern generation has a
singular role and is particularly difficult to achieve. The reason is
that speed cannot be modulated by postprocessing of some underlying
generator's output -- the prime generator itself must be modulated
\cite{wyffelsetal13}.  Frequency adaptation of neural oscillators is an
important theme in research on biological pattern generators (CPGs)
(reviews: \cite{Grillner06, Ijspeert08}).  Frequency adaptation has
been modeled in a number of ways, among which (i) to use a highly
abstracted CPG model in the form of an ODE, and regulate speed by
changing the ODE's time constant; (ii) to use a CPG model which
includes a pacemaker neuron whose pace is adaptive; (iii) to use
complex, biologically quite detailed, modular neural architectures in
which frequency adapatation arises from interactions between modules,
sensor-motoric feedback cycles, and tonic top-down input. However, the
fact that humans can execute essentially arbitrary motor patterns at
different speeds is not explained by these models.  Presumably this
requires a generic speed control mechanism which takes effect already
at higher (cortical, planning) layers in the motor control hierarchy.
Conceptor-controlled frequency adaptation might be of interest as a
candidate mechanism for such a ``cognitive-level'' generic speed
control mechanism.

\paragraph*{Shape Morphing of an Integer-Periodic Pattern}

In this demonstration, the conceptors $C(R^3,1000)$ and $C(R^4,1000)$
from the 5-periodic patterns $p^3$ and $p^4$  were morphed, again
with $-2 \leq \mu \leq 3$.  Figure \ref{FigMorphRandPattern} depicts
the network observer $y(n)$ for a morph run of 95 steps which was
started with $\mu = -2$ and ended with $\mu = 3$, with a linear $\mu$
ramping in between. It can be seen that the differences between the
two reference patterns (located at the points marked by dots) become
increasingly magnified in both extrapolation segments. At each of the
different points in each 5-cycle, the ``sweep'' induced by the
morphing is however neither linear nor of the same type across all 5
points of the period (right panel). A simple algebraic rule that would
describe the geometric characteristics of such morphings cannot be
given. I would like to say, it is ``up to the discretion of the
network's nonlinear dynamics'' how the morphing command is
interpreted; this is especially true for the extrapolation range. If
reservoirs with a different initial random $W^\ast$ are used,
different morphing geometries arise, especially at the far ends of the
extrapolation range (not shown).

\begin{figure}[htbp]
 \center
\includegraphics[width=145 mm]{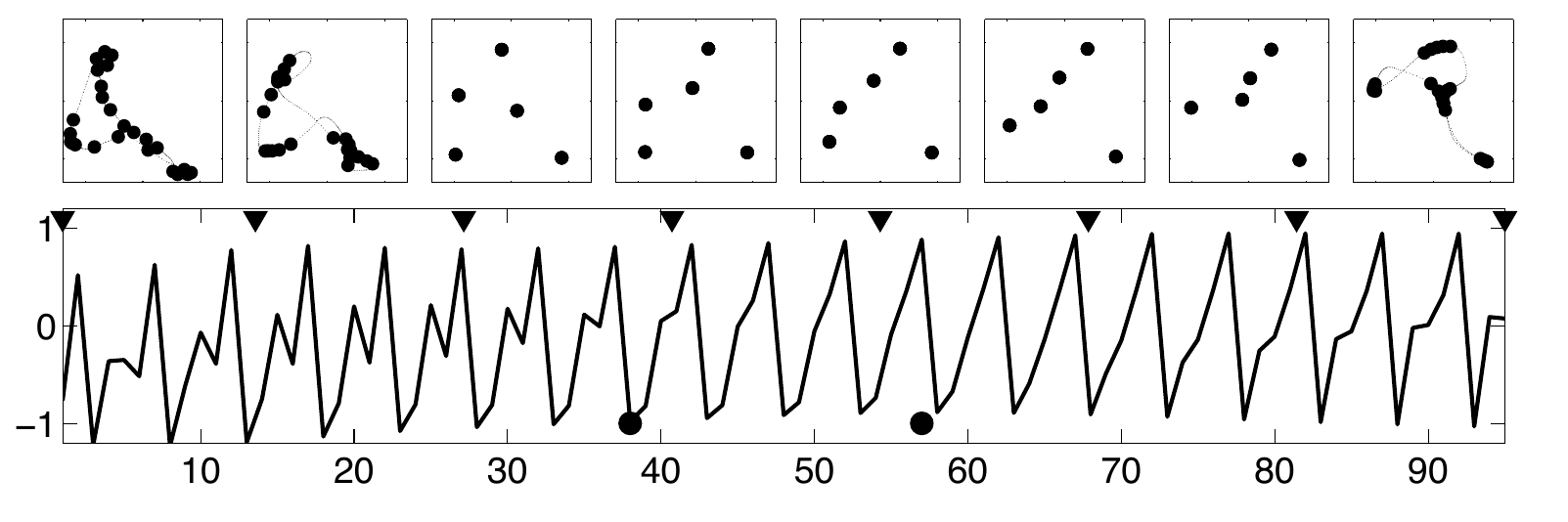}
 \caption{Morphing between, and extrapolating beyond, two versions of a 5-periodic
   random pattern. The morphing range
   was $-2 \leq \mu \leq 3$. Bottom: Network observation from a morphing
   run. Dots mark the points $\mu = 0$ and $\mu =
   1$, corresponding to situations where the two original conceptors
   $C^1, C^2$ were active in unadulterated form. The
   network observation signal $y(n)$ is shown. Top: Delay-embedding
   ``snapshots''. Figure layout similar to Figure \ref{FigMorphSines}.}
 \label{FigMorphRandPattern}
 \end{figure}

The snapshots displayed in Figure \ref{FigMorphRandPattern} reveal
that the morphing sweep takes the reservoir through two bifurcations
(apparent in the transition from snapshot 2 to 3, and from 7 to 8). In
the intermediate morphing range (snapshots 3 -- 7), we observe a
discrete periodic attractor of 5 points. In the ranges beyond, on both
sides the
attracting set becomes topologically homomorphic to a continuous
cycle. From a visual inspection, it appears that  these bifurcations
``smoothly'' 
preserve some geometrical characteristics of the observed signal
$y(n)$. A mathematical characterisation of this phenomenological
continuity across bifurcations remains for future investigations.  

\paragraph*{Heterogeneous Pattern Morphing}

Figure \ref{FigMorphRand2Sine} shows a morph from the 5-periodic
pattern $p^3$ to the irrational-periodic sine $p^2$ (period length
$\approx 9.83$). This time the morphing range was $0 \leq \mu \leq 1$,
(no extrapolation). The Figure shows a run with an initial 25
steps of $\mu = 0$, followed by a 50-step ramp to $\mu = 1$ and a tail
of 25 steps at the same $\mu$ level. One observes a gradual change of
signal shape and period along the morph. From a dynamical systems
point of view this gradual change is unexpected. The reservoir is,
mathematically speaking, an autonomous system under the influence of a
slowly changing control parameter $\mu$. On both ends of
the morph, the system is in an attractor. The topological nature of
the attractors (seen as subsets of state space) is different (5
isolated points vs.\ a homolog of a 1-dim circle), so there must be a
at least one bifurcation taking place along the morphing
route. Such a bifurcations would usually be accompanied by a sudden
change of some qualitative characteristic of the system trajectory. We
find however no trace of a dynamic rupture, at least not by visual
inspection of the output trajectory. Again, a more in-depth formal
characterization of what geometric properties are ``smoothly'' carried
through these bifurcations is left for future work. 

Figure \ref{figMorphMain} in Section \ref{secOverview} is a compound
demonstration of the three types of pattern morphing that I here
discussed individually.

A possible application for the pattern morphing
by conceptors is to effect smooth gait changes in walking robots, a
problem that is receiving some attention in that field. 

\begin{figure}[htbp]
 \center
\includegraphics[width=145 mm]{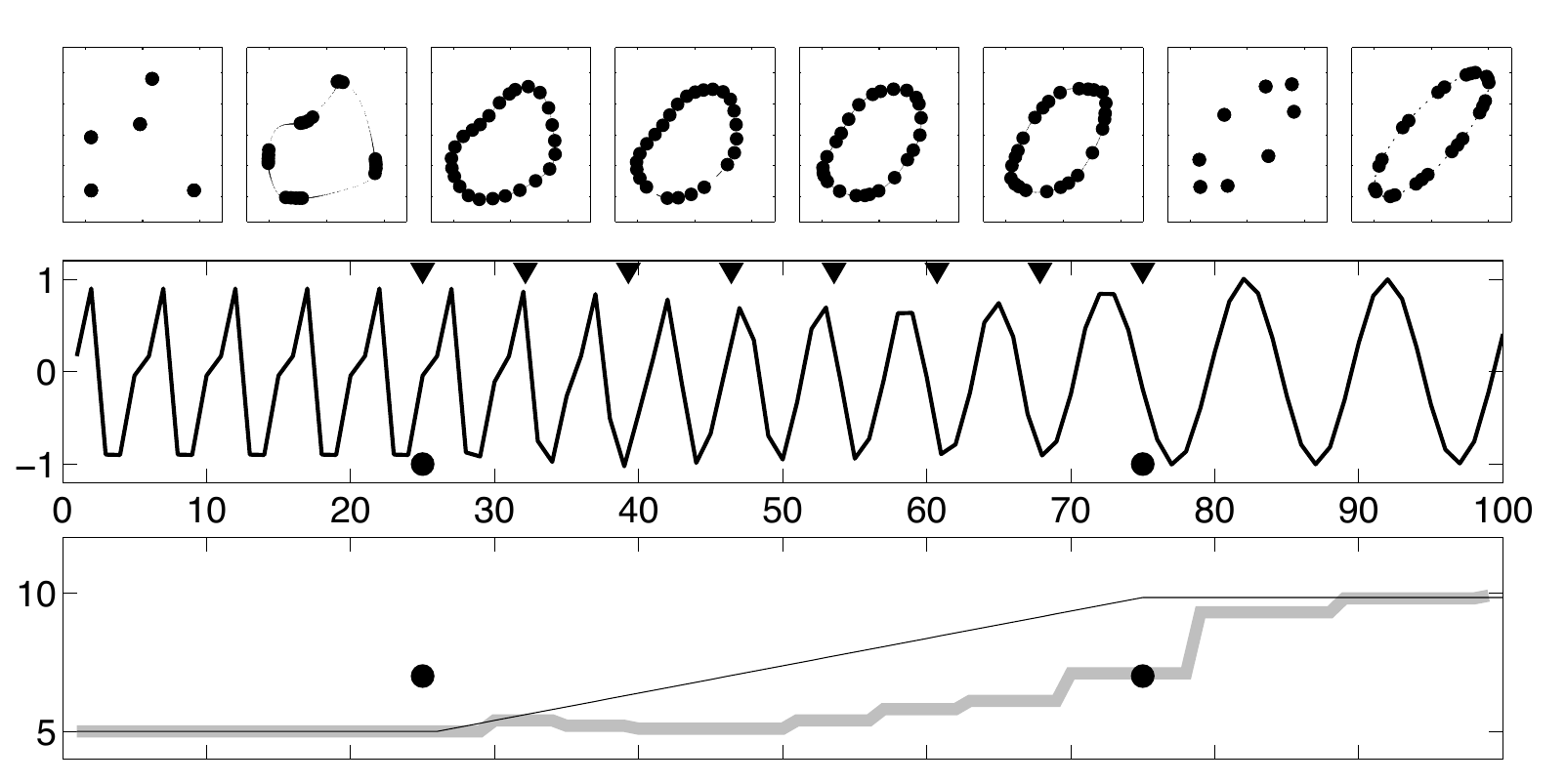}
 \caption{Morphing from  a 5-periodic
   random pattern to an irrational-periodic sine. The morphing range
   was $0 \leq \mu \leq 1$. Figure layout otherwise is as in Figure
   \ref{FigMorphSines}. }
 \label{FigMorphRand2Sine}
 \end{figure}

\subsection{Understanding Aperture}

\subsubsection{The Semantics of $\alpha$ as ``Aperture''}\label{secApertureSemantics}

Here
I  show how the parameter $\alpha$ can be interpreted as a scaling
of signal energy, and motivate why I call it ``aperture''.

We can rewrite
$C(R,\alpha) = C(E[xx'],\alpha)$ as follows:

\begin{eqnarray}\label{eqSemAlpha1}
C(E[xx'],\alpha) & = & E[xx'] (E[xx'] +\alpha^{-2}I)^{-1}  = E[(\alpha
x)(\alpha x)'] (E[(\alpha x) (\alpha x)'] + I)^{-1}\nonumber\\
& = & C(E[(\alpha x)(\alpha x)'],1) = C( \alpha^2\,E[xx'],1).
\end{eqnarray}

Thus, changing from $C(R,1)$ to $C(R,\alpha)$ 
can be interpreted as
scaling the reservoir data by a factor of $\alpha$, or
expressed in another way, as scaling the signal energy of the
reservoir signals by a factor of $\alpha^2$. This is directly analog to what
adjusting the aperture effects in an optical camera. In optics, the
term \emph{aperture} denotes the diameter of the effective lens
opening, and the amount of light energy that reaches the film is
proportional to the squared aperture. This has motivated the
naming of the parameter $\alpha$ as \emph{aperture}.

\subsubsection{Aperture Adaptation and Final Definition of Conceptor
  Matrices}\label{secApertureAdaptDefC} 

It is easy to verify that if $C_\alpha = C(R,\alpha)$ and $C_\beta =
C(R,\beta)$ are two versions of a conceptor $C$ differing in their
apertures $0 < \alpha, \beta < \infty$, they are related to
each other by
\begin{equation}\label{eqSemAlpha2}
C_\beta = C_\alpha \, \left(C_\alpha +
  \left(\frac{\alpha}{\beta}\right)^2 \, (I - C_\alpha)\right)^{-1},
\end{equation}
\noindent where we note that $C_\alpha + (\alpha/\beta)^2 (I - C_\alpha)$ is always
invertible. $C_\beta$ is thus a function of $C_\alpha$ and the ratio
$\gamma = \beta/\alpha$.  This motivates to introduce an
\emph{aperture adaptation} operation $\varphi$ on conceptors $C$, as follows:
\begin{equation}\label{eqSemAlpha3}
\varphi(C,\gamma) = C\,(C+\gamma^{-2}(I-C))^{-1},
\end{equation}
\noindent where $\varphi(C,\gamma)$ is the conceptor version obtained
from $C$ by adjusting the aperture of $C$ by a factor of
$\gamma$. Specifically, it holds  that $C(R,\alpha) =
\varphi(C(R,1),\alpha)$.  

We introduce the notation $R_C = C(I-C)^{-1}$, which leads to the
following easily verified data-based version of (\ref{eqSemAlpha3}):
\begin{equation}\label{eqSemAlpha0c1}
R_{\varphi(C,\gamma)} = \gamma^2 \, R_C.
\end{equation} 

When we treat Boolean operations further below, it will turn out that
the NOT operation will flip zero singular values of $C$ to unit
singular values. Because of this circumstance, we admit unit singular
values in conceptors and formally define

\begin{definition}\label{defconceptor}
A \emph{conceptor matrix} is a positive semidefinite matrix whose singular
values range in $[0,1]$. We denote the set of all $N \times N$
conceptor matrices by $\mathcal{C}_N$. 
\end{definition}

Note that Definition \ref{defConceptor} defined the concept of a
\emph{conceptor matrix associated with a state correlation matrix} $R$
\emph{and an aperture} $\alpha$, while  Definition \ref{defconceptor}
specifies the more general class of \emph{conceptor
  matrices}. Mathematically, conceptor matrices (as in Definition
\ref{defconceptor}) are more general than the conceptor matrices
associated with a state correlation matrix $R$, in that the former may
contain unit singular values.

Furthermore, in the context of Boolean operations it will also become
natural to admit aperture adaptations of sizes $\gamma = 0$ and
$\gamma = \infty$. The inversion in Equation (\ref{eqSemAlpha3}) is
not in general well-defined for such $\gamma$ and/or conceptors with
unit singular values, but we can generalize those relationships to the
more general versions of conceptors and aperture adaptations by a
limit construction:

\begin{definition}\label{defLimitPhi}
Let $C$ be a conceptor and $\gamma \in [0,\infty]$. Then 
\begin{equation}\label{eq:defLimitPhi}
\varphi(C,\gamma) = \left\{\begin{array}{ll}
    C\,(C+\gamma^{-2}(I-C))^{-1} & \quad \mbox{for } 0 < \gamma <
    \infty\\
\lim_{\delta \downarrow 0} C\,(C+\delta^{-2}(I-C))^{-1} & \quad
    \mbox{for } \gamma = 0\\
\lim_{\delta \uparrow \infty}  C\,(C+\delta^{-2}(I-C))^{-1} & \quad
    \mbox{for } \gamma = \infty
 \end{array}  \right.
\end{equation}
\end{definition}  

It is a mechanical exercise to show that the limits in
(\ref{eq:defLimitPhi}) exist, and to calculate the singular values for
$\varphi(C,\gamma)$. The results are collected in the following
proposition.

\begin{proposition}\label{propapadapt}
Let $C = U S U'$ be a conceptor and $(s_1,\ldots,s_N)' = \mbox{\emph{
  diag}}S$
the vector of its singular values. Let $\gamma \in [0,\infty]$. Then
$\varphi(C,\gamma) = U S_\gamma U'$ is the conceptor with singular
values  $(s_{\gamma,1},\ldots,s_{\gamma,N})'$, where 

\begin{equation}\label{eqpropadapt}
s_{\gamma,i} = \left\{\begin{array}{ll}
s_i / (s_i + \gamma^{-2}(1-s_i)) & \quad \mbox{for }\;\; 0 < s_i < 1,\; 0 < \gamma <
\infty \\
0 &  \quad \mbox{for  }\;\; 0 < s_i < 1,\; \gamma = 0\\
1 &  \quad \mbox{for  }\;\; 0 < s_i < 1,\; \gamma = \infty\\
0 & \quad \mbox{for  }\;\; s_i = 0,\; 0 \leq \gamma \leq \infty\\
1 & \quad \mbox{for  }\;\; s_i = 1,\; 0 \leq \gamma \leq \infty\\
  \end{array}    \right. 
\end{equation} 
\end{proposition} 

Since aperture adaptation of $C = USU'$ only changes the singular
values of $C$, the following fact is obvious:

\begin{proposition}\label{propTransformInvarianceApAdapt}
If $V$ is orthonormal, then $\varphi(VCV',\gamma) =
V\,\varphi(C,\gamma)\,V'$. 
\end{proposition}

Iterated application of aperture adaptation corresponds to multiplying
the adaptation factors:

\begin{proposition}\label{propIterateApAdapt}
Let $C$ be a conceptor and $\gamma, \beta \in [0, \infty]$. Then
$\varphi (\varphi(C,\gamma), \beta) = \varphi(C, \gamma \beta)$.
\end{proposition}

The proof is a straightforward algebraic verification using (\ref{eqpropadapt}).

Borrowing again terminology from photography, I call a conceptor with SVD
$C = U S U'$ \emph{hard} if all singular values in $S$ are 0 or 1 (in
photography, a  film with an
extremely ``hard'' gradation yields pure black-white images with no
gray tones.) Note that $C$ is hard if and only if it is a projector
matrix. If $C$ is hard, the following holds:
\begin{eqnarray}
C & = & C^\dagger  \; = \; C' \; = \; CC, \label{eqSemAlpha3b}\\
\varphi(C,\gamma) & = & C \quad \mbox{for }\gamma \in [0,\infty].\label{eqSemAlpha3c}
\end{eqnarray}
The first claim amounts to stating that $C$ is a projection
operator, which is obviously the case; the second claim follows
directly from (\ref{eqpropadapt}).

 Besides the aperture, another illuminating characteristic of a
 conceptor matrix is the mean value of its singular values, i.e.\ its
 normalized trace $q(C) = \mbox{trace}(C) / N$. It ranges in
 $[0,1]$. Intuitively, this quantity measures the fraction of
 dimensions from 
 the  $N$-dimensional reservoir state space that is claimed
 by $C$. I will call it the \emph{quota} of $C$. 

\subsubsection{Aperture Adaptation: Example}\label{secApertureAdapt}

In applications one will often need to adapt the aperture to optimize
the quality of $C$. What ``quality'' means depends on the task at
hand. I present an illustrative example, where the reservoir is loaded
with very fragile patterns. Retrieving them requires a prudent choice
of $\alpha$. Specifically, I loaded a reservoir of size $N = 500$ with
four chaotic patterns, derived from the well-known R\"{o}ssler,
Lorenz, Mackey-Glass, and H\'{e}non attractors (details of this
example are given in Section
\ref{secExpDetailApertureAdapt}). Four conceptors
$C_R, C_L, C_{MG}, C_H$ were computed, one for each attractor, using
$\alpha = 1$.  Then, in four retrieval
experiments, the aperture of each of these was adapted using
(\ref{eqSemAlpha3}) in a geometric succession of five different
$\gamma$, yielding five versions of each of the $C_R, C_L, C_{MG},
C_H$. Each of these was used in turn for a constrained run of the
reservoir according to the state update rule $x(n+1) = C \;
\tanh(W\,x(n) + W^{\mbox{\scriptsize in}}\,p(n+1) + b)$, and the
resulting output observation was plotted in a delay-embedding format.

\begin{figure}[htbp]
{\sf {\bf A}}$\;\;$\includegraphics[width=65 mm]{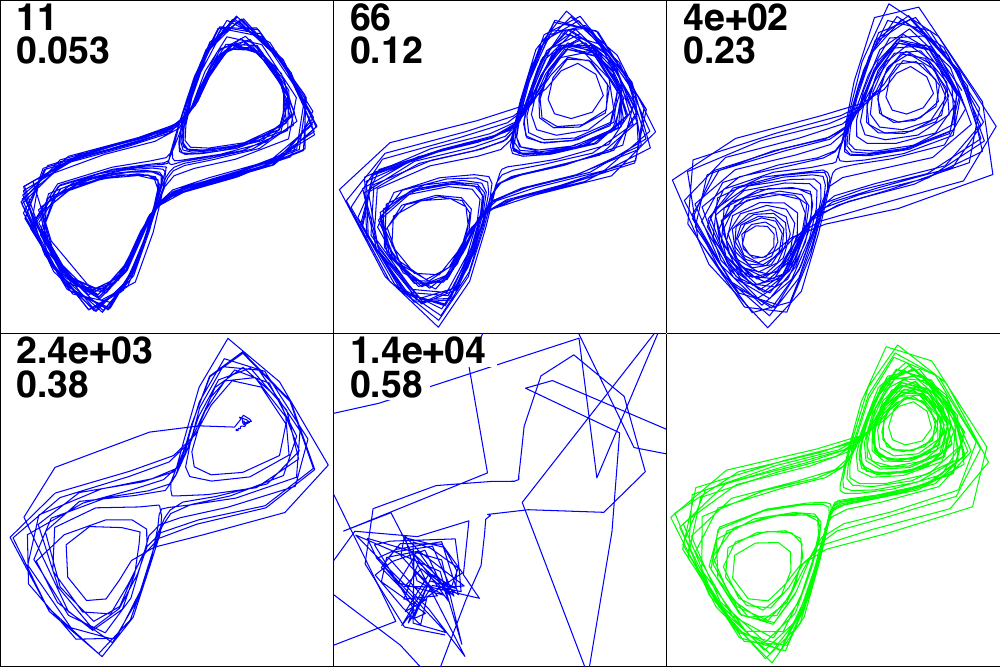}
$\;\;${\sf {\bf B}}$\;\;$\includegraphics[width=65 mm]{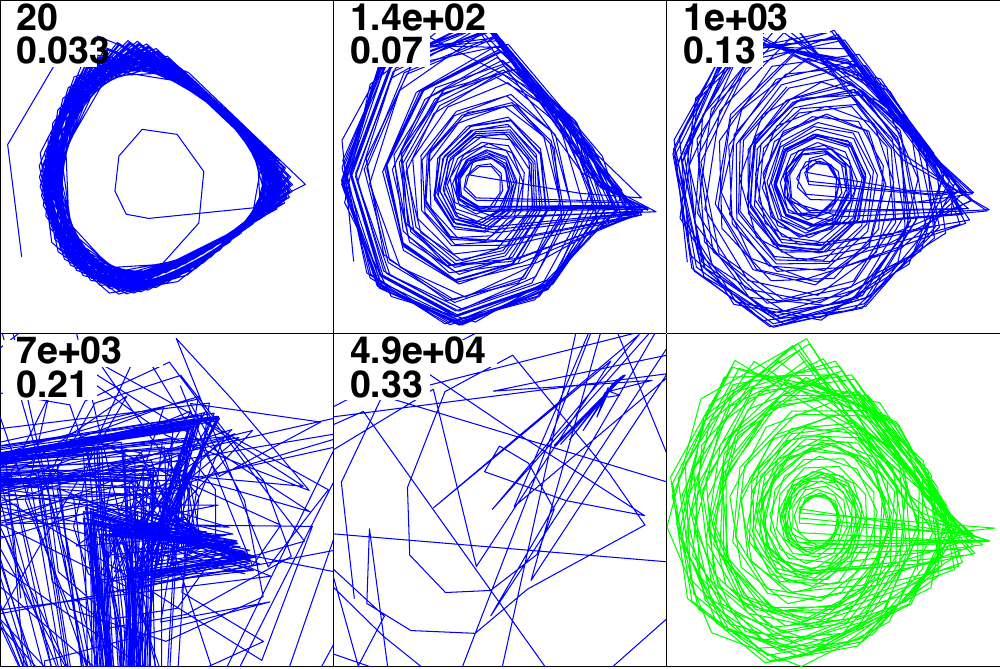}
\newline $\quad$ \newline $\quad$
{\sf {\bf C}}$\;\;$\includegraphics[width=65 mm]{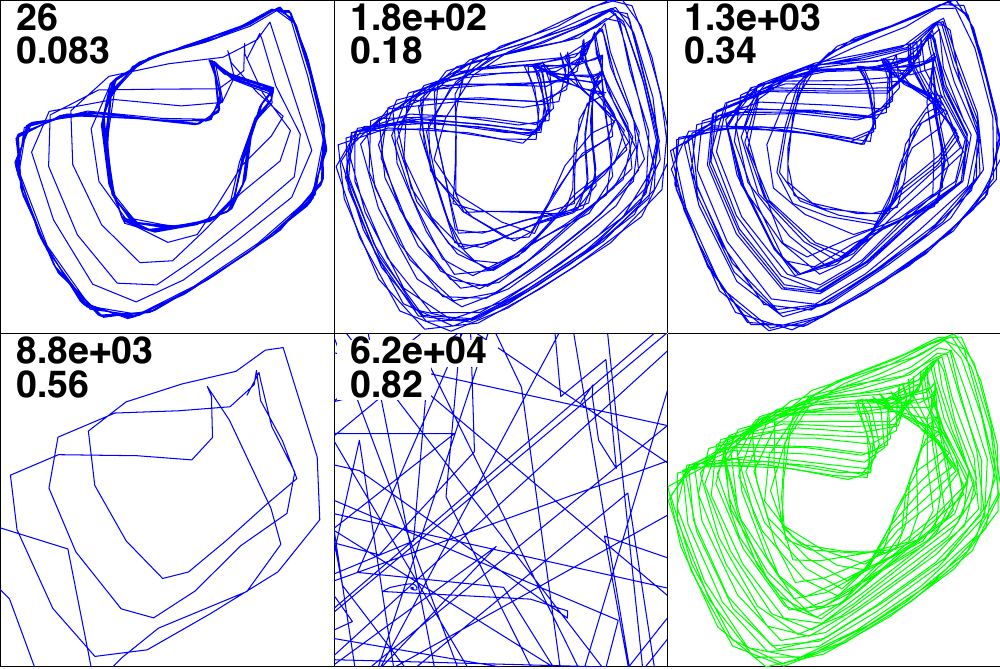}
$\;\;${\sf {\bf D}}$\;\;$\includegraphics[width=65
mm]{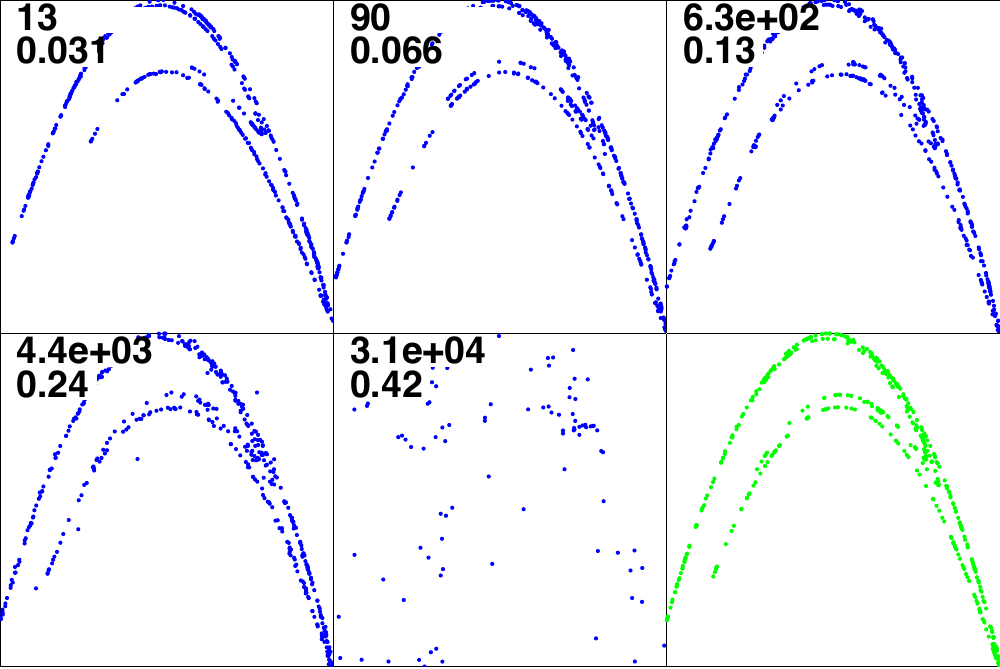}
%\vspace{-0.1cm}
\caption{Invoking conceptors to retrieve four chaotic signals from a
  reservoir. {\bf A} Lorenz, {\bf B} R\"{o}ssler, {\bf C}
  Mackey-Glass, and {\bf D} H\'{e}non attractor. All four are
  represented by delay-embedding plots of the reservoir observation
  signal  $y(n)$. The
  plot range is $[0,1]\times[0,1]$ in every panel. {\bf A} -- {\bf C}
  are attractors derived from differential equations, hence subsequent
  points are joined with lines; {\bf D} derives from an iterated map
  where joining lines has no meaning. Each 6-panel block shows five
  patterns generated by the reservoir under the control of differently
  aperture-adapted versions of
  a conceptor (blue, first five panels) and a plot of the original
  chaotic reference signal (green, last panel). Empty panels indicate
  that the $y(n)$ signal was outside the $[0,1]$ range. The right upper panel
  in each block shows a version which, judged by visual inspection,
  comes satisfactorily close to the original. First number given in a
  panel: aperture $\alpha$; second number: quota $q(C)$.}
\label{figChaosClover}
\end{figure}

Figure \ref{figChaosClover} displays the findings. Per each attractor,
the five apertures were hand-selected such that the middle one (the
third) best re-generated the original chaotic signal, while the first
failed to recover the original.  One should mention that it is not
trivial in the first place to train an RNN to stably generate any
single chaotic attractor timeseries, but here we require the loaded
network to be able to generate any one out of four such signals, only by
constraining the reservoir by a conceptor with a suitably adapted
aperture. Number insets in the panels of figure \ref{figChaosClover}
indicate the apertures and quotas used per run.

\subsubsection{Guides for Aperture Adjustment}\label{secApAdjustGuide}

The four chaotic attractors considered in the previous subsection were
``best'' (according to visual inspection) reconstructed with apertures
between 630 and 1000. A well-chosen aperture
is clearly important for working with conceptors. In all
demonstrations reported so far I chose a ``good'' aperture based on
experimentation and human judgement. In practice one will often
need automated criteria for optimizing the aperture which do not rely
on human inspection. In this subsection I propose two measures which
can serve as such a guiding criterion.

{\bf A criterion based on reservoir-conceptor interaction.} 
Introducing an interim state variable $z(n)$ by splitting the
conceptor-constrained reservoir update equation
(\ref{eqCconstrainedRule}) into
\begin{equation}\label{eqzVar}
z(n+1) = \tanh(W\,x(n)+b),\quad x(n+1) = C(R,\alpha)\,z(n+1),
\end{equation}
I define the \emph{attenuation} measurable $a$ as
\begin{equation}\label{eqBlindout}
a_{C,\alpha } = E[\|z(n) - x(n) \|^2] / E[\|z(n) \|^2], 
\end{equation}
where the states $x(n), z(n)$ are understood to result from a
reservoir constrained by $C(R,\alpha)$. The attenuation is
the fraction of the reservoir signal energy which is suppressed by
applying the conceptor. Another useful way to conceive of this
quantity is to view it as noise-to-signal ratio, where the ``noise''
is the component $z(n) - x(n)$ which is filtered away from the
unconstrained reservoir signal $z(n)$. It turns out in simulation
experiments that when the aperture is varied, the attenuation
$a_{C,\alpha }$ passes through a minimum, and at this
minimum, the pattern reconstruction performance peaks.

In Figure \ref{figBlindout}{\bf A} the $\log_{10}$ of $a_{C,\alpha }$ is
plotted for a sweep through a range of apertures $\alpha$, for
each of the four chaotic attractor conceptors (details in
Section \ref{secExpDetailApertureAdapt}). As $\alpha$ grows, the
attenuation $a_{C,\alpha }$ first declines roughly linearly in the
log-log plots, that is, by a power law of the form $a_{C,\alpha } \sim
\alpha^{-K}$. Then it enters or 
passes through a trough. The aperture values that yielded
visually optimal reproductions of the chaotic patterns coincide with
the point where the bottom of the trough is reached.

\begin{figure}[htbp]
\center
{\sf {\bf A}}$\;$\includegraphics[width=65 mm]{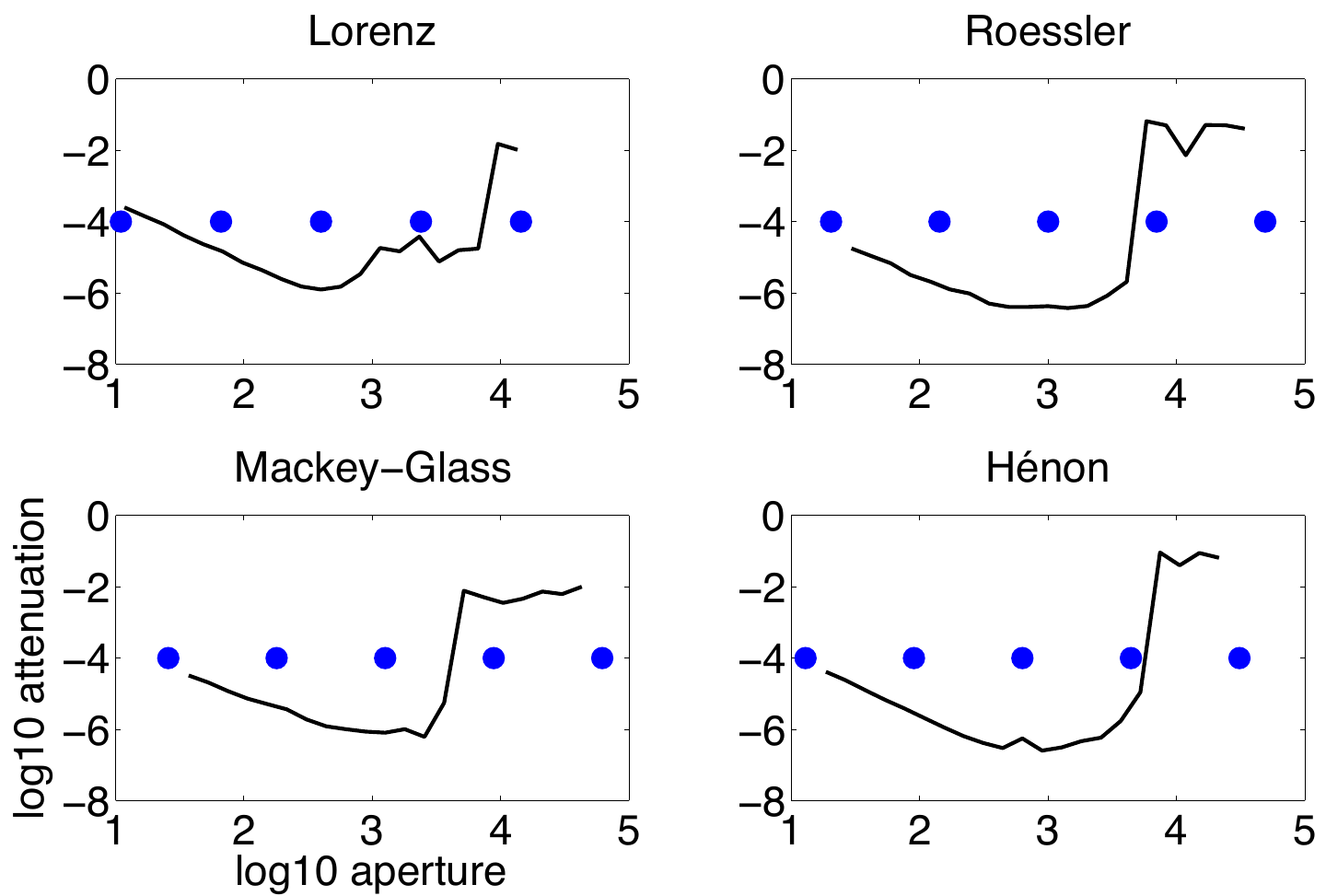}
$\;\;${\sf {\bf B}}$\;$\includegraphics[width=65 mm]{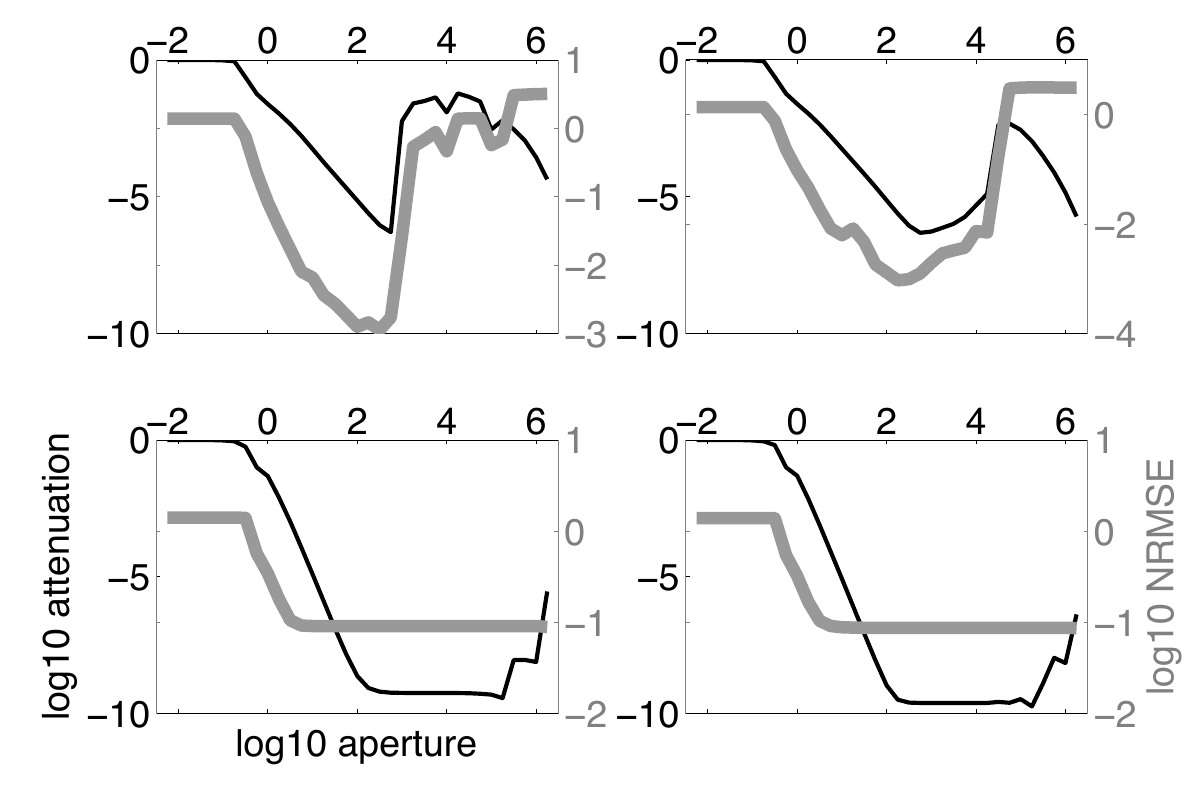}
\caption{Using attenuation to locate optimal apertures. {\bf A}
  Dependancy of attenuation on aperture for the four chaotic
  attractors. The blue dots mark the apertures used to generate the
  plots in Figure \ref{figChaosClover}. {\bf B}
  Dependancy of attenuation on aperture for the two sinewaves (top
  panels) and the two 5-point periodic patterns (bottom) used in Sections
  \ref{sec:InitialDrivingDemo}{\emph{ff.}} These plots also provide
  the NRMSEs for the accuracy of the reconstructed patterns (gray). For
  explanation see text.   }
\label{figBlindout}
\end{figure}

Figure \ref{figBlindout}{\bf B} gives similar plots for the two
irrational-period sines and the two 5-point periodic patterns treated
in earlier sections.  The same reservoir and storing procedures as
described at that place were utilized again here. The dependence of
attenuation on aperture is qualitatively the same as in the
chaotic attractor example. The attenuation plots are overlaid with the
NRMSEs of the original drivers vs.\ the conceptor-constrained
reservoir readout signals. Again, the ``best'' aperture -- here quantified
by the NRMSE -- coincides remarkably well with the trough
minimum of the attenuation.

Some peculiarities visible in the plots {\bf B} deserve a short
comment. (i) The initial constant plateaus in all four plots result
from $C(R,\alpha) \approx 0$ for the very small apertures in this
region, which leads to $x(n) \approx 0, z(n) \approx \tanh(b)$. (ii)
The jittery climb of the attenuation towards the end of the plotting
range in the two bottom panels is an artefact due to roundoff errors
in SVD computations which blows up singular values in
conceptors which in theory
should  be zero. Without rounding error involved, the attenuation plots would
remain at their bottom value once it is reached. (iii) In
the top two panels, some time after having passed through the trough
the attenuation value starts to decrease again. This is due to the
fact that for the irrational-period sinewave signals, all singular
values of the conceptors are nonzero. As a consequence, for
increasingly large apertures the conceptors will converge to the
identity matrix, which would have zero attenuation.  \newline 

{\bf A criterion based on conceptor matrix properties.} 
A very simple criterion for aperture-related
``goodness'' of a conceptor can be obtained from monitoring the
gradient of the squared Frobenius norm
\begin{equation}\label{eqNormGradCriterion}
\nabla(\gamma) = \frac{d}{d\,\log(\gamma)} \; \| \varphi(C,\gamma) \|^2
\end{equation}
with respect to the logarithm of $\gamma$. To get an intuition about
the semantics of this criterion, assume that $C$ has been obtained
from data with a correlation matrix $R$ with SVD $R = U \Sigma U'$.
Then $\varphi(C,\gamma) = R(R + \gamma^{-2}I)^{-1}$ and $\|
\varphi(C,\gamma) \|^2 = \| \Sigma(\Sigma + \gamma^{-2}I)^{-1} \|^2 = \|
\gamma^{2}\Sigma(\gamma^{2}\Sigma + I)^{-1} \|^2$. That is,
$\varphi(C,\gamma)$ can be seen as obtained from data scaled by a factor
of $\gamma$ compared to $\varphi(C,1) = C$. The criterion $\nabla(\gamma)$
therefore measures the sensitivity of (the squared norm of) $C$ on
(expontential) scalings of data. Using again photography as  a
metaphor: if the aperture of a lens is set to the value where
$\nabla(\gamma)$ is maximal, the sensitivity of the  image (= conceptor) to
changes in brightness (= data scaling) is maximal.

Figure \ref{figApViaNorm} shows the behavior of this criterion again
for the standard example of loading two irrational sines and two
integer-periodic random patterns. Its maxima coincide largely with the
minima of the attenuation criterion, and both with what was ``best''
performance of the respective pattern generator. The exception is the
two integer-periodic patterns (Figure \ref{figApViaNorm} {\bf B}
bottom panels) where the $\nabla$ criterion would suggest a slightly
too small aperture.

\begin{figure}[htb]
\center
\includegraphics[width=90 mm]{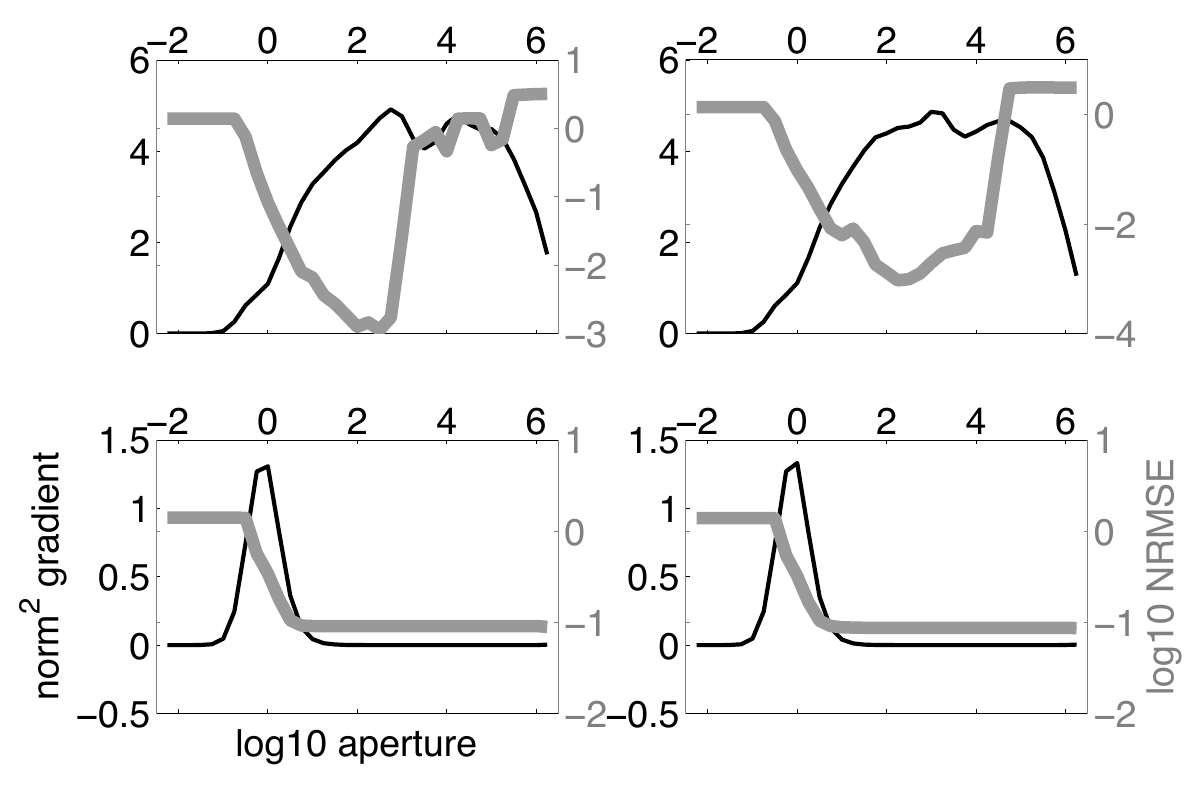}
\caption{The norm-gradient based criterion to determine ``good''
  apertures for the basic demo example from Sections
  \ref{sec:InitialDrivingDemo} and \ref{sec:RetrieveGeneric}. Plots
  show $\nabla(\gamma)$ against the $\log_{10}$ of aperture $\gamma$.
  Figure layout similar as in Figure \ref{figBlindout} {\bf B}. For
  explanation see text. }
\label{figApViaNorm}
\end{figure}

Comments on criteria for guiding aperture selection:

\begin{itemize}
\item The two presented criteria based on attenuation and norm
gradient are purely heuristic. A theoretical analysis would require
a rigorous definition of ``goodness''. Since tasks vary in their
objectives, such an analysis would have to be carried out on a
case-by-case basis for varying ``goodness'' characterizations. Other
formal criteria besides the two presented here can easily be construed (I
experimented with dozens of alternatives (not documented), some of
which performed as well as the two instances reported
here). Altogether this appears to be a wide field for
experimentation. 
\item The attenuation-based criterion needs trial runs with the
reservoir to be calculated, while the norm-gradient criterion can be
computed offline. The former seems to be particularly suited for
pattern-generation tasks where the conceptor-reservoir feedback loop
is critical (for instance, with respect to stability). The latter may
be more appropriate in machine learning tasks where conceptors are
used for classifying reservoir dynamics in a ``passive'' way without
coupling the conceptors into the network updates. I will give an
example in Section \ref{subsec:JapVow}.
\end{itemize}

\subsection{Boolean Operations on Conceptors}\label{secBoolean}

\subsubsection{Motivation}

Conceptor matrices can be submitted to operations that can be meaningfully
called AND, OR, and NOT. There are two justifications for using these
classical logical terms:

\begin{description}
\item[Syntactical / algebraic:] Many  algebraic laws
governing Boolean algebras are preserved; for hard
conceptor matrices the preservation is exact. 
\item[Semantical:] These operations on conceptor matrices correspond
dually to operations on the data that give rise to the conceptors via
(\ref{eq:CompConceptor}). Specifically, the OR operation can be
semantically interpreted on the data level by merging two datasets,
and the NOT operation by inverting the principal component weights of
a dataset. The AND operation can be interpreted on the data level by
combining de Morgan's rule (which states that $x \wedge y = \neg(\neg
x \vee \neg y)$) with the semantic interpretations of OR and NOT.
\end{description}

The mathematical structures over conceptors that arise from the
Boolean operations are richer than  standard Boolean logic, in that
aperture adaptation operations can be included in the picture. One
obtains a formal framework which one might call ``adaptive Boolean
logic''. 

There are two major ways how such a theory of conceptor logic may be
useful:

\begin{description}
    \item[A logic for information processing in RNNs (cognitive and
  neuroscience):] The dynamics of any $N$-dimensional RNN (of any
  kind, autonomously active or driven by external input), when monitored
  for some time period $L$, yields an $N \times L$ sized state
  collection matrix $X$ and its corresponding $N \times N$ correlation
  matrix $R$, from which a conceptor matrix $C = R(R+I)^{-1}$ can be obtained
  which is a ``fingerprint'' of the activity of the network for this
  period. The Boolean theory of conceptors can be employed to analyse
  the relationships between such ``activity fingerprints'' obtained at
  different intervals, different durations, or from different driving
  input. An interesting long-term research goal for cognitive
  neuroscience would be to map the logical structuring described on
  the network data level, to Boolean operations carried out by
  task-performing subjects.
\item[An algorithmical tool for RNN control (machine learning):] By
controlling the ongoing activity of an RNN in a task through
conceptors which are derived from logical operations, one can
implement ``logic control'' strategies for RNNs. Examples will be
given  in Section \ref{subsec:memmanage}, where Boolean operations on
conceptors will be key for an efficient memory management in
RNNs;  in Section \ref{subsec:JapVow}, where Boolean operations
will enable to combine positive and negative evidences for
finite-duration pattern
recognition; and in Section \ref{secHierarchicalArchitecture},
where Boolean operations will help to simultaneously de-noise and
classify signals. 
\end{description}

\subsubsection{Preliminary Definition of Boolean Operations}

Defining Boolean operators through their data semantics
is transparent and simple when the concerned data correlation matrices
are nonsingular. In this case, the resulting conceptor matrices are
nonsingular too and have singular values ranging in the open interval
$(0,1)$. I treat this situation in this subsection.  However,
conceptor matrices with a singular value range of $[0,1]$ frequently
arise in practice. This leads to technical complications which will be
treated in the next subsection. The definitions given in the present
subsection  are preliminary and  serve expository purposes.

In the remainder of this subsection, conceptor matrices $C,B$ are assumed to
derive from nonsingular correlation matrices.

I begin with OR. Recall that a conceptor matrix $C$ (with aperture 1) derives
from a data source (network states) $x$ through $R = E[xx'], C =
C(R,1) = R(R+I)^{-1}$. Now consider a second conceptor $B$ of the same
dimension $N$ as $C$, derived from another data source $y$ by $Q = E[yy'],
B = B(Q,1) = Q(Q+I)^{-1}$.    I define

\begin{equation}\label{eqDefOR}
C \vee B := (R + Q) (R + Q + I)^{-1}, 
\end{equation}

and name this the OR operation. Observe that $R + Q =
E[[x,y][x,y]']$, where $[x,y]$ is the $N \times 2$ matrix made of
vectors $x, y$.  $C \vee B$ is thus obtained by a merge of the two 
data sources which previously went into $C$ and $B$, respectively.  This
provides a  semantic interpretation of the OR operation. 

Using (\ref{eq:CompConceptor}), it is
straightforward to verify that $C \vee B$ can be directly computed
from $C$ and $B$ by
 \begin{equation}\label{eqCompOR}
C \vee B = \left(I + \left(C(I-C)^{-1} + B(I-B)^{-1}\right)^{-1} \right)^{-1}, 
\end{equation}
where the assumption of nonsingular $R, Q$ warrants that all
inverses in this equation are well-defined.

I now turn to the NOT operation. For $C = C(R,1) = R(R+I)^{-1}$ with
nonsingular $R$ I define it by
\begin{equation}\label{eqDefNOT}
\neg C :=  R^{-1}(R^{-1}+I)^{-1}.
\end{equation}

Again this can be semantically interpreted on the data level. Consider
the SVDs $R = U\Sigma U'$ and $R^{-1} = U\Sigma^{-1}U'$. $R$ and $R^{-1}$ have
the same principal components $U$, but the variances $\Sigma, \Sigma^{-1}$ of
data that would give rise to $R$ and $R^{-1}$ are inverse to each
other. In informal terms, $\neg C$ can be seen as arising from data
which co-vary inversely compared to data giving rise to $C$.

Like in the case of OR, the negation of $C$ can be computed directly
from $C$. It is easy to see that 
\begin{equation}\label{eqCompNOT}
\neg C = I - C.
\end{equation}

Finally, I consider AND. Again, we introduce it on the  data
level. Let again $C = R(R+I)^{-1}, B = Q(Q+I)^{-1}$. The OR
operation was introduced as addition on data correlation matrices, and
the NOT operation as inversion. Guided by de Morgan's law $a \wedge b
= \neg(\neg a \vee \neg b)$ from Boolean
logic, we obtain a correlation matrix  $(R^{-1} + Q^{-1})^{-1}$ for $C
\wedge B$. Via (\ref{eq:CompConceptor}), from this correlation matrix  we
are led to

\begin{equation}\label{eqDefAND}
C \wedge B := (R^{-1} + Q^{-1})^{-1} \left((R^{-1} + Q^{-1})^{-1} + I\right)^{-1}.
\end{equation}

Re-expressing $R,Q$ in terms of $C,B$ in this equation, elementary
transformations (using (\ref{eq:CompConceptor})) again allow us to compute AND
directly:

\begin{equation}\label{eqCompAND}
C \wedge B = (C^{-1} + B^{-1} - I)^{-1}.
\end{equation}

By a routine transformation of equations, it can be verified that the de
Morgan's laws $C \vee B = \neg(\neg C \wedge \neg B)$ and $C \wedge B
= \neg(\neg C \vee \neg B)$ hold for the direct computation expressions
(\ref{eqCompOR}), (\ref{eqCompNOT}) and (\ref{eqCompAND}). 

\subsubsection{Final Definition of Boolean
  Operations}\label{secFormalBooleanDef}

We notice that the direct computations (\ref{eqCompOR}) and
(\ref{eqCompAND}) for OR and AND are only well-defined for conceptor matrices
whose singular values range in $(0,1)$.  
I now  generalize the definitions for AND and OR to cases where
 the concerned
conceptors may contain singular values 0 or 1. Since the direct
computation (\ref{eqCompAND}) of AND is simpler than the direct
computation (\ref{eqCompOR}) of OR, I carry out the generalization
for AND and then transfer it to OR through de Morgan's rule. 

Assume
that $C = USU', B = VTV'$ are the SVDs of conceptors $C, B$, where $S$
and/or $T$ may contain zero singular values. The direct computation
(\ref{eqCompAND}) is then not well-defined. 

Specifically, assume that  $\mbox{diag}(S)$ contains $l
\leq N$ nonzero singular values and that $\mbox{diag}(T)$ contains $m
\leq N$ nonzero singular values, i.e.\ $\mbox{diag}(S) =
(s_1,\ldots, s_l,0,\ldots,0)'$ and $\mbox{diag}(T) =
(t_1,\ldots, t_m,0,\ldots,0)'$. Let $\delta$ be a positive real
number. Define $S_\delta$ to be the diagonal matrix which has a
diagonal $(s_1,\ldots, s_l,\delta,\ldots,\delta)'$, and
similarly $T_\delta$ to have diagonal $(t_1,\ldots,
t_m,\delta,\ldots,\delta)'$. Put $C_\delta = US_\delta U', B_\delta
= V T_\delta V'$. Then $C_\delta \wedge B_\delta = (C_\delta^{-1} +
B_\delta^{-1} - I)^{-1}$ is well-defined. We now define

\begin{equation}\label{eqCompANDgenDef}
C \wedge B = \lim_{\delta \to 0}(C_\delta^{-1} + B_\delta^{-1} - I)^{-1}.
\end{equation}

The limit  in this equation is well-defined and can be resolved into an
efficient algebraic computation:

\begin{proposition}\label{propANDDef}
  Let $\mathbf{B}_{\mathcal{R}(C) \cap \mathcal{R}(B)}$ be a matrix
  whose columns form an arbitrary orthonormal basis of $\mathcal{R}(C)
  \cap \mathcal{R}(B)$. Then, the matrix $\mathbf{B}'_{\mathcal{R}(C)
    \cap \mathcal{R}(B)} (C^\dagger + B^\dagger - I)
  \mathbf{B}_{\mathcal{R}(C) \cap \mathcal{R}(B)}$ is invertible, and
  the limit (\ref{eqCompANDgenDef}) exists and is equal to
\begin{eqnarray}
\lefteqn{C \wedge B = \lim_{\delta \to 0}(C_\delta^{-1} + B_\delta^{-1} -
I)^{-1} = }\nonumber\\
& = &  \mathbf{B}_{\mathcal{R}(C) \cap \mathcal{R}(B)} \, \left(
  \mathbf{B}'_{\mathcal{R}(C) \cap \mathcal{R}(B)} \,(C^\dagger +
  B^\dagger - I) \,\mathbf{B}_{\mathcal{R}(C) \cap \mathcal{R}(B)}
\right)^{-1} \, \mathbf{B}'_{\mathcal{R}(C) \cap \mathcal{R}(B)}. \label{eqCompANDgenComp}
\end{eqnarray}
Equivalently, let $\mathbf{P}_{\mathcal{R}(C) \cap \mathcal{R}(B)} =
\mathbf{B}_{\mathcal{R}(C) \cap \mathcal{R}(B)} \,
\mathbf{B}_{\mathcal{R}(C) \cap \mathcal{R}(B)}'$ be the projector
matrix on $\mathcal{R}(C) \cap \mathcal{R}(B)$. Then $C \wedge B$ can
also be written as
\begin{equation}\label{eqCompANDgenComp_1}
C \wedge B =  \left(
  \mathbf{P}_{\mathcal{R}(C) \cap \mathcal{R}(B)} \,(C^\dagger +
  B^\dagger - I) \,\mathbf{P}_{\mathcal{R}(C) \cap \mathcal{R}(B)}
\right)^{\dagger}.
\end{equation}
\end{proposition}

The proof and an algorithm to compute a basis matrix
$\mathbf{B}_{\mathcal{R}(C) \cap \mathcal{R}(B)}$ are given in Section
\ref{secProofAndDef}.

The formulas (\ref{eqCompANDgenComp}) and (\ref{eqCompANDgenComp_1})
not only extend the formula (\ref{eqCompAND}) to cases where $C$ or
$B$ are non-invertible, but also ensures numerical stability in cases
where $C$ or $B$ are ill-conditioned. In that situation, the
pseudoinverses appearing in (\ref{eqCompANDgenComp}),
(\ref{eqCompANDgenComp_1}) should be computed with appropriate
settings of the numerical tolerance which one can specify in common
implementations (for instance in Matlab) of the SVD. One should generally
favor (\ref{eqCompANDgenComp}) over (\ref{eqCompAND}) unless one can
be sure that $C$ and $B$ are well-conditioned.

The direct computation (\ref{eqCompNOT}) of
NOT is well-defined for $C$ with a singular value range $[0,1]$, thus
nothing remains to be done here.

Having available the general and numerically robust computations of AND
via (\ref{eqCompANDgenComp}) or (\ref{eqCompANDgenComp_1}) and of NOT  via (\ref{eqCompNOT}), we
invoke  de Morgan's rule $C \vee B = \neg(\neg C
\wedge \neg B)$ to obtain a general and robust computation for OR on
the basis of (\ref{eqCompANDgenComp}) resp.\ (\ref{eqCompANDgenComp_1})  and
(\ref{eqCompNOT}). Summarizing, we obtain the final definitions for
 Boolean operations on conceptors:

\begin{definition}\label{def:finalBoolean}
\begin{eqnarray*}
\neg\,C & := & I - C,\\
C \wedge B & := & \left(
  \mathbf{P}_{\mathcal{R}(C) \cap \mathcal{R}(B)}\, (C^\dagger +
  B^\dagger - I) \,\mathbf{P}_{\mathcal{R}(C) \cap \mathcal{R}(B)}
\right)^{\dagger},\\
C \vee B & := & \neg\,(\neg \, C \, \wedge \, \neg\, B),
\end{eqnarray*}
where  $\mathbf{P}_{\mathcal{R}(C) \cap \mathcal{R}(B)}$ is the projector matrix
  on $\mathcal{R}(C)
  \cap \mathcal{R}(B)$. 
\end{definition}

This definition is consistent with the
preliminary definitions given in the previous subsection. For AND this
is clear: if $C$ and $B$ are nonsingular, $\mathbf{P}_{\mathcal{R}(C)
  \cap \mathcal{R}(B)}$ is the identity and the pseudoinverse is the
inverse, hence (\ref{eqCompAND}) is recovered (fact 1). We noted in the
previous subsection  that de Morgan's laws hold for conceptors derived
from nonsingular correlation matrices (fact 2). Furthermore, if $C$ is derived from a
nonsingular correlation matrix, then $\neg C$ also corresponds to a
nonsingular correlation matrix (fact 3). Combining facts 1 -- 3 yields
that the way of defining OR via de Morgan's rule from AND and NOT in
Definition \ref{def:finalBoolean} generalises
(\ref{eqDefOR})/(\ref{eqCompOR}).

For later use I state a technical result which gives a
characterization of OR in terms of a limit over correlation matrices:

\begin{proposition}\label{propLimitOr}
For a conceptor matrix $C$ with SVD $C = USU'$ let $S^{(\delta)}$ be a
version of $S$ where all unit singular values (if any) have been
replaced by $1 - \delta$, and let $C^{(\delta)} =
US^{(\delta)} U'$. Let $R_C^{(\delta)} = C^{(\delta)} (I -
C^{(\delta)})^{-1}$. Similarly, for another conceptor $B$ let $R_B^{(\delta)} = B^{(\delta)} (I -
B^{(\delta)})^{-1}$. Then
\begin{equation}\label{eqLimitOr}
C \vee B = I - \lim_{\delta \downarrow 0}\, (R_C^{(\delta)} +
R_B^{(\delta)} + I)^{-1}  = \lim_{\delta \downarrow  0}\, (R_C^{(\delta)}
+ R_B^{(\delta)})\,(R_C^{(\delta)} + R_B^{(\delta)} + I)^{-1}.    
\end{equation}
\end{proposition}

The proof is given in Section \ref{secProofpropLimitOr}. Finally I
note that de Morgan's rule also holds for AND (proof in Section
\ref{secProofpropdeMorganAND}): 

\begin{proposition}\label{propdeMorganAND}
\begin{displaymath}
C \wedge B = \neg\,(\neg \, C \, \vee \, \neg\, B).
\end{displaymath}
\end{proposition}

\subsubsection{Facts Concerning Subspaces}\label{secFactsSubspaces}

For an $N \times N$ matrix $M$, let $\mathcal{I}(M) = \{x \in
\mathbb{R}^N \mid Mx = x\}$ be the \emph{identity space} of $M$. This is
the eigenspace of $M$ to the eigenvalue 1, 
a linear subspace of $\mathbb{R}^N$. The identity spaces, null spaces,
and ranges of conceptors are related to Boolean operations in various
ways. The facts collected here are technical, but will be useful in deriving
further results later.

\begin{proposition}\label{propSpaces} Let $C, B$ be any conceptor matrices,
  and $H, G$ hard conceptor matrices of the
  same dimension. Then the following facts hold:
\begin{enumerate}
\item $\mathcal{I}(C) \subseteq \mathcal{R}(C)$.
\item $\mathcal{I}(C^\dagger) = \mathcal{I}(C)$ and
$\mathcal{R}(C^\dagger) = \mathcal{R}(C)$ and $\mathcal{N}(C^\dagger) = \mathcal{N}(C)$. 
\item $\mathcal{R}(\neg C) = \mathcal{I}(C)^\perp$ and
$\mathcal{I}(\neg C) = \mathcal{N}(C)$ and
$\mathcal{N}(\neg C) = \mathcal{I}( C)$.
\item $\mathcal{R}(C \wedge B) = \mathcal{R}(C) \cap \mathcal{R}(B)$ and $\mathcal{R}(C \vee B) = \mathcal{R}(C) + \mathcal{R}(B)$.
\item  $\mathcal{I}(C \wedge B) = \mathcal{I}(C) \cap \mathcal{I}(B)$ and $\mathcal{I}(C \vee B) = \mathcal{I}(C) + \mathcal{I}(B)$.
\item $\mathcal{N}(C \wedge B) = \mathcal{N}(C) + \mathcal{N}(B)$ and  $\mathcal{N}(C \vee B) = \mathcal{N}(C) \cap \mathcal{N}(B)$. 
\item $\mathcal{I}(\varphi(C,\gamma)) = \mathcal{I}(C)$ for $\gamma
\in [0,\infty)$ and $\mathcal{R}(\varphi(C,\gamma)) = \mathcal{R}(C)$
for  $\gamma \in (0,\infty]$ and $\mathcal{N}(\varphi(C,\gamma)) =
\mathcal{N}(C)$ for  $\gamma \in (0,\infty]$. 
\item $A = A \wedge C \; \Longleftrightarrow \; \mathcal{R}(A)
\subseteq \mathcal{I}(C)$ and $A = A \vee C \; \Longleftrightarrow \;
\mathcal{I}(A)^\perp \subseteq \mathcal{N}(C)$.
\item $\varphi(C, 0)$ and $\varphi(C, \infty)$ are hard.
\item $\varphi(C, 0) = \mathbf{P}_{\mathcal{I}(C) }$ and $\varphi(C,
\infty) = \mathbf{P}_{\mathcal{R}(C) }$. 
\item $H = H^\dagger = \mathbf{P}_{\mathcal{I}(H) }$.
\item $\mathcal{I}(H) = \mathcal{R}(H) =  \mathcal{N}(H)^\perp$.
\item $\neg H = \mathbf{P}_{\mathcal{N}(H) } = \mathbf{P}_{\mathcal{I}(H)^\perp } $.
\item $H \wedge G = \mathbf{P}_{\mathcal{I}(H)\, \cap \, \mathcal{I}(G)}$.
\item  $H \vee G = \mathbf{P}_{\mathcal{I}(H) \,+ \,\mathcal{I}(G)}$.
\end{enumerate}
\end{proposition} 

The proof is given in Section \ref{secProofpropSpaces}.

\subsubsection{Boolean Operators and Aperture Adaptation}\label{secBooleanAperture}

The Boolean operations are related to aperture adaptation in a
number of ways: 

\begin{proposition}\label{propBooleanAperture}
Let $C, B$ be $N \times N$ sized conceptor matrices and $\gamma, \beta \in
[0,\infty]$. We declare $\infty^{-1} = \infty^{-2} = 0$ and $0^{-1} = 0^{-2} = \infty$. Then, 
\begin{enumerate}
\item $\neg \varphi(C,\gamma) = \varphi(\neg C, \gamma^{-1})$, 
\item $\varphi(C,\gamma) \vee \varphi(B, \gamma) = \varphi(C \vee B,
\gamma)$,
\item $\varphi(C,\gamma) \wedge \varphi(B, \gamma) = \varphi(C \wedge B, \gamma)$,
\item $\varphi(C,\gamma) \vee \varphi(C, \beta) = \varphi(C,
\sqrt{\gamma^{2}+\beta^{2}})$,
\item $\varphi(C,\gamma) \wedge \varphi(C, \beta) = \varphi(C,
(\gamma^{-2}+\beta^{-2})^{-1/2})$.
\end{enumerate}
\end{proposition}

The proof can be found in Section \ref{secProofpropBooleanAperture}. Furthermore, with the aid of aperture adaptation and OR it is
possible to implement an incremental model extension, as
follows. Assume that  conceptor $C$ has been obtained from a
dataset $X$ comprised of $m$ data vectors $x$, via $R = XX'/m, \; \; C = R(R
+ \alpha^{-2}I)^{-1}$. Then, $n$ new data vectors $y$ become
available, collected as columns in a data matrix $Y$. One wishes to
update the original conceptor $C$ such that it also incorporates the
information from $Y$, that is, one wishes to obtain 
\begin{equation}\label{eqTargetUpdateC}
\tilde{C} = \tilde{R}(\tilde{R} + \alpha^{-2} I)^{-1},
\end{equation}
 where $\tilde{R}$ is the
updated correlation matrix obtained by $Z = [X Y], \tilde{R} =
ZZ'/(m+n)$. But now furthermore assume that the original training data
$X$ are no longer available. This situation will not be uncommon in
applications. The way to a direct computation of
(\ref{eqTargetUpdateC}) is barred. In this situation, the extended
model $\tilde{C}$ can be computed from $C, Y, m, n$ as follows. Let $C_Y =
YY'(YY' + I)^{-1}$. Then,
\begin{eqnarray}
\tilde{C} & = & \varphi\left(\varphi(C, \;m^{1/2}\alpha^{-1}) \vee C_Y,\;
(m+n)^{1/2}\alpha\right)\label{eqUpdateC1}\\
& = & I - \left(\frac{m}{m+n}(I - C)^{-1}C +
\frac{n}{m+n}\alpha^2 YY' + I\right)^{-1}.\label{eqUpdateC2}
\end{eqnarray}

These formulas can be verified by elementary transformations using
(\ref{eq:CompConceptor}), (\ref{eqRecoverR}), (\ref{eqDefOR}) and
(\ref{eqSemAlpha0c1}), noting that $C$ cannot have unit singular
values because it is obtained from a bounded correlation matrix $R$,
thus $(I - C)$ is invertible.

\subsubsection{Logic Laws}\label{secLogicLaws}

Many laws from Boolean logic carry over to the operations AND, OR, NOT
defined for conceptors, sometimes with modifications. 

\begin{proposition}\label{propBooleanElementaryLaws}
Let $I$ be the $N \times N$ identity matrix, $0$ the zero matrix, and
$B, C, D$ any conceptor matrices of size $N \times N$ (including $I$
or $0$). Then the following
laws hold:
\begin{enumerate}
\item De Morgan's rules: $C \vee B = \neg\,(\neg \, C \, \wedge \,
\neg\, B)$ and $C \wedge B = \neg\,(\neg \, C \, \vee \,
\neg\, B)$. 
\item Associativity: $(B \wedge C) \wedge D = B \wedge (C \wedge D)$
and 
$(B \vee C) \vee D = B \vee (C \vee D)$. 
\item Commutativity: $B  \wedge C = C \wedge B$ and  $B  \vee C = C
\vee B$.
\item Double negation: $\neg (\neg C) = C$.
\item Neutrality of $0$ and $I$: $C \vee 0 = C$ and $C \wedge I = C$.
\item Globality of $0$ and $I$: $C \vee I = I$ and $C \wedge 0 = 0$.
\item Weighted self-absorption for OR: $C \vee C = \varphi(C,\sqrt{2})$ and
$\varphi(C,\sqrt{1/2}) \vee \varphi(C,\sqrt{1/2}) = C$.
\item Weighted self-absorption for AND: $C \wedge C = \varphi(C,1/\sqrt{2})$ and
$\varphi(C,\sqrt{2}) \wedge \varphi(C,\sqrt{2}) = C$.
\end{enumerate}
\end{proposition}

The proofs are given in Section \ref{secProofpropBooleanElementaryLaws}. From among
the classical laws of Boolean logic, the general absorption rules $A =
A \wedge (A \vee B) = A \vee (A \wedge B)$ and the laws of
distributivity do \emph{not} hold  for conceptors.

While the absorption rules $A = A \wedge (A \vee B) = A \vee (A \wedge
B)$ are not valid for conceptor matrices, it is  possible to
``invert'' $\vee$ by $\wedge$ and vice versa in a way that is
reminiscent of absorption rules:

\begin{proposition}\label{propPseudoabsorbtion}
  Let $A, B$ be conceptor matrices of size $N \times N$. Then,
\begin{enumerate}
\item $C = \left(\mathbf{P}_{\mathcal{R}(A)} \left(I + A^\dagger - (A
    \vee B)^\dagger  \right)  \mathbf{P}_{\mathcal{R}(A) }
    \right)^{\dagger}$ is a conceptor matrix and 
\begin{equation}\label{eqAbsorb1}
A = (A \vee B) \wedge C.
\end{equation}
\item  $C = I - \left(\mathbf{P}_{\mathcal{I}(A)^\perp} \left(I +
  (I-A)^\dagger - (I - (A \wedge B))^\dagger  \right) \mathbf{P}_{\mathcal{I}(A)^\perp} \right)^{\dagger}$ is a conceptor matrix and
\begin{equation}\label{eqAbsorb2}
A = (A \wedge B) \vee C.
\end{equation}
\end{enumerate}   
\end{proposition}
The proof is given in Section \ref{secProofpropPseudoabsorbtion}.

\subsection{An Abstraction Relationship between Conceptors}\label{secAbstraction}

The existence of (almost) Boolean operations between conceptors
suggests that conceptors may be useful as models of \emph{concepts}
(extensive discussion in Section \ref{secLogic}).  In this subsection
I add substance to this interpretation by introducing an abstraction
relationship between conceptors, which allows one to organize a set of
conceptors in an abstraction hierarchy.

In order to equip the set $\mathcal{C}_N$ of $N \times N$ conceptors
with an ``abstraction'' relationship, we need to identify a partial
ordering on $\mathcal{C}_N$ which meets our intuitive expectations
concerning the structure of ``abstraction''. A natural candidate is
the partial order $\leq$ defined on the set of $N \times N$ real
matrices by $X \leq Y$ if $Y - X$ is positive semidefinite. This
ordering is often called the \emph{L\"{o}wner ordering}. I will
interpret and employ the L\"{o}wner ordering as an abstraction
relation. The key facts which connect this ordering to Boolean
operations, and which justify to interpret $\leq$ as a form of logical
abstraction, are collected in the following

 \begin{proposition}\label{propBasicAbstraction}
Let $\mathcal{C}_N$ be the set of
conceptor matrices of size $N$. Then the following facts hold.  
\begin{enumerate}
\item An $N \times N$ matrix $A$ is a conceptor matrix if and only if
$0 \leq A \leq I_{N \times N}$.
\item $0_{N \times N}$ is the global minimal element and $I_{N\times
N}$ the global maximal element of $(\mathcal{C}_N, \leq)$.
\item $A \leq B$ if and only if $\neg A \geq \neg B$.
\item Let $A, B \in \mathcal{C}_N$ and $B \leq A$. Then 
\begin{displaymath}
C = \mathbf{P}_{\mathcal{R}(B)}\, (B^\dagger -
\mathbf{P}_{\mathcal{R}(B)}\, A^\dagger\, \mathbf{P}_{\mathcal{R}(B)}
+ I)^{-1}\,\mathbf{P}_{\mathcal{R}(B)} 
\end{displaymath}
is a conceptor matrix and
\begin{displaymath}
B = A \wedge C.
\end{displaymath}
\item Let again $A, B \in \mathcal{C}_N$ and $A \leq B$. Then 
\begin{displaymath}
C = I  - \mathbf{P}_{\mathcal{I}(B)^\perp} \; \left( (I-B)^\dagger -
  \mathbf{P}_{\mathcal{I}(B)^\perp} \; (I-A)^\dagger \;
  \mathbf{P}_{\mathcal{I}(B)^\perp} + I \right)^{-1}
  \;\mathbf{P}_{\mathcal{I}(B)^\perp} 
\end{displaymath}
is a conceptor matrix and
\begin{displaymath}
B = A \vee C.
\end{displaymath}
\item If for $A, B, C \in \mathcal{C}_N$ it holds that $A \wedge C =
B$, then $B \leq A$.
\item If for $A, B, C \in \mathcal{C}_N$ it holds that $A \vee C =
B$, then $A \leq B$.
\item For $A \in \mathcal{C}_N$ and $\gamma \geq 1$ it holds that $A
\leq \varphi(A, \gamma)$; for  $\gamma \leq 1$ it holds that
$\varphi(A, \gamma) \leq A$.
\item If $A \leq B$, then $\varphi(A,\gamma) \leq \varphi(B,\gamma)$
for $\gamma \in [0,\infty]$.
\end{enumerate}
\end{proposition}

The proof is given in Section \ref{secProofpropBasicAbstraction}. The
essence of this proposition can be re-expressed succinctly as follows:

\begin{proposition}\label{propLoewnerBoolean}
For conceptors $A, B$ the following conditions are equivalent:
\begin{enumerate}
\item $A \leq B$.
\item There exists a conceptor $C$ such that $A \vee C = B$.
\item There exists a conceptor $C$ such that $A = B \wedge C$.
\end{enumerate}
\end{proposition}

Thus, there is an equivalence between ``going upwards'' in the $\leq$
ordering on the one hand, and merging conceptors by OR on the other
hand. In standard logic-based knowledge representation formalisms, a
\emph{concept} (or \emph{class}) $B$ is defined to be more
\emph{abstract} than some other concept/class $A$ exactly if there is
some concept/class $C$ such that $A \vee C = B$. This motivates me to
interpret $\leq$ as an \emph{abstraction} ordering on $\mathcal{C}_N$.

\subsection{Example: Memory Management in
  RNNs}\label{subsec:memmanage}

In this subsection I demonstrate the usefulness of 
Boolean operations by introducing a memory management scheme for
RNNs.  I will show how it is possible

\begin{enumerate}
\item to store patterns  in an RNN \emph{incrementally}: if
patterns $p^1,\ldots, p^m$ have already been stored, a new pattern
$p^{m+1}$ can be stored in addition without interfering with the
previously stored patterns, and without having to know them;
\item to maintain a measure of the \emph{remaining memory capacity} of
the RNN which indicates how many more patterns can
still be stored;
  \item to \emph{exploit redundancies}: if the new pattern is similar
in a certain sense to already stored ones, loading it consumes less
memory capacity than when the new pattern is dissimilar to the already
stored ones.
\end{enumerate}

 Biological brains can learn new patterns during their lifetime. For
 artificial neural networks (ANNs) ``lifelong learning'' presents a
 notorious difficulty. Training some task into an ANN typically means
 to adapt connection weights, often by gradient descent optimization.
 When an ANN has been trained on some task in the past and subsequently
 is trained on a new task, the new weight adaptations are prone to
 destroy the previously learnt competences. This \emph{catastrophic
   forgetting} (or \emph{catastrophic interference}) phenomenon has
 been recognized since long. Although a number of proposals have been
 made which partially alleviate the problem in special circumstances
 (\cite{French03,Grossberg05,McCallum07}, skeptical overview:
 \cite{MoeHelgesenStranden05}), catastrophic forgetting was still
 considered a main challenge for neural learning theory in an expert's
 hearing solicited by the NSF in 2007 (\cite{DouglasSejnowski08}).

\subsubsection{General Principle and Algorithm}\label{subsubsecGenPrinAlg}

Recall that in the original pattern storing procedure, the initial
random weight matrix $W^\ast$ is recomputed to obtain the input
internalization weight
matrix $W$ of the loaded reservoir, such that
\begin{displaymath}
x^j(n+1) = \tanh(W^\ast\,x^j(n) + W^{\mbox{\scriptsize in}}\,
p^j(n+1) + b) \approx \tanh(W\,x^j(n) + b), 
\end{displaymath} 
where $p^j(n)$ is the $j$-th pattern signal and $x^j(n)$ is the
reservoir state signal obtained when the reservoir is driven by the
$j$-th pattern. For a transparent memory management, it is more
convenient to keep the original $W^\ast$ and record the weight changes
into an \emph{input simulation matrix} $D$, such that
\begin{equation}\label{eqDMatUpdate}
x^j(n+1) = \tanh(W^\ast\,x^j(n) + W^{\mbox{\scriptsize in}}\,
p^j(n+1) + b) \approx \tanh(W^\ast\,x^j(n) + D \,x^j(n) + b).
\end{equation}

$D$ simulates the additive impact $W^{\mbox{\scriptsize in}}\,
p^j(n)$ of the input $p^j(n)$ on the network 'potential' before the
wrapping with the $\tanh$.

In a non-incremental batch training mode, $D$ would be computed by
regularized linear regression to minimize the following squared error:
\begin{equation}\label{eqDMatError}
D = \mbox{argmin}_{\tilde{D}} \sum_{j = 1,\ldots,K} \sum_{n = n_0+1,
  \ldots, L} \| W^{\mbox{\scriptsize in}}\,p^j(n) - \tilde{D} \,x^j(n-1)\|^2,
\end{equation}
where $K$ is the number of patterns and $L$ is the length of the
training sequence (subtracting an initial washout period of length
$n_0$). Trained in this way, the sum $W^\ast + D$ would be essentially
identical (up to differences due to using another regularization
scheme) to the input internalization weights $W$ obtained in the
original pattern storing procedure. In fact, the performance of loading a
reservoir with patterns via an input simulation matrix $D$ as in
(\ref{eqDMatUpdate}) is indistinguishable from what is obtained in the
original procedure (not reported).

At recall time, the conceptor $C^j$ is inserted and the pattern $p^j$
is re-generated by running
\begin{equation}\label{e19} 
{x}(n+1) = C^j\, \tanh\left( W^\ast \, {x}(n) + D\, {x}(n) + b\right), \quad
y(n) = W^{\mbox{\scriptsize out}}\, {x}(n).
\end{equation}

 Another variant of the loading procedure is even more
  minimalistic and aims at replacing only the very input $p^j(n)$ by
  minimizing 
\begin{equation}\label{e18a}
\sum_{j=1,\ldots,K} \sum_{n=n_0+1,\ldots,L} \| p^j(n) - R\,{x}^j(n)  \|^2.
\end{equation}

For $d$-dimensional input this yields a $d \times N$ matrix $H$ of
what I call \emph{input recreation weights} which at recall time is
utilized by
\begin{equation}\label{e19a} 
{x}(n+1) = C^j\, \tanh\left( W^\ast \, {x}(n) + \, W^{\mbox{\scriptsize
    in}} \, H \, {x}(n) + b\right), \quad
y(n) = W^{\mbox{\scriptsize out}}\, {x}(n).
\end{equation}

On the grounds of mathematical intuition, loading patterns by input
internalization weights $W$ should be superior to the variant with input
simulation weights $D$ or input recreation weights $R$, because of the
greater number of re-computed parameters (leading to greater accuracy)
and the more comprehensive impact of regularization (leading to
greater dynamical stability, among other). 
However, in all the various simulation experiments that I carried out
so far I found only small to negligable deteriorations in performance
when using $D$ or $R$ instead of $W$, both with respect to accuracy
and with respect to dynamical stability. In the various simulation demos
covered in this report,

With the aid of Boolean operations and input simulation weights it
becomes possible to incrementally load a collection of $d$-dimensional
patterns $p^1,
p^2,\ldots$ into a reservoir, such that (i) loading $p^{m+1}$ does not
interfere with previously loaded $p^1,\ldots, p^m$; (ii) similarities
between patterns are exploited to save memory space; (iii) the amount
of still free memory space can be monitored.

Here when I speak of ``memory
space'', I am referring to the $N$-dimensional vector space $\mathcal{M} =
\mathbb{R}^N$ spanned by reservoir states ${x}$. Furthermore, when I
will be speaking of ``components'' of a reservoir state ${x}$, I refer to
projections of ${x}$ on some linear subspace of $\mathcal{M}$, where the subspace
will be clear from context.

The incremental loading procedure unfolds in loading cycles $m = 1, 2,
\ldots$. After completion of the $m$-th cycle, input simulation
weights $D^m$, output weights $W^{\mbox{\scriptsize out}, m}$, and
conceptors $C^1, \ldots, C^m$ are obtained, such that when \eqref{e19}
is run with $D^m$, $W^{\mbox{\scriptsize out}, m}$ and one of the
$C^j$ ($1 \leq j \leq m$), the pattern $p^j$ is re-generated. $D^{m+1}$
and $W^{\mbox{\scriptsize out}, m+1}$ are computed from the previously
obtained $D^{m}$ and $W^{\mbox{\scriptsize out}, m}$ by adding
increments
\begin{equation}\label{e20}
D^{m+1} = D^{m} + D_{\mbox{\scriptsize inc}}^{m+1}, \quad
W^{\mbox{\scriptsize out}, m+1} = W^{\mbox{\scriptsize out}, m} +
W^{\mbox{\scriptsize out}, m+1}_{\mbox{\scriptsize inc}},
\end{equation}
with zero-weight initialization  in the first loading cycle
\begin{equation}\label{e21}
D^{1} = \mathbf{0} + D_{\mbox{\scriptsize inc}}^{1}, \quad
W^{\mbox{\scriptsize out}, 1} = \mathbf{0} +
W^{\mbox{\scriptsize out}, 1}_{\mbox{\scriptsize inc}}.
\end{equation}

Conceptors $C^1, C^2, \ldots$ are computed as usual from reservoir
states collected in runs of the native network ${x}(n+1) =
\tanh(W^\ast \, {x}(n) + W^{\mbox{\scriptsize in}}\, p^j(n) + {b})$. The key for incremental loading is a conceptor-based
characterization of the memory space claimed by the $m$ previously
loaded patterns.  Concretely, in loading cycle $m+1$ we make use of
the conceptor
\begin{equation}\label{e22}
A^m = C^1 \vee \ldots \vee C^m,
\end{equation}
which due to the associativity of conceptor-OR can be incrementally
computed by $A^{m} = A^{m-1} \vee C^{m}$ with initialization by the
zero conceptor. In intuitive terms,  $A^{m}$ characterizes the
geometry of the reservoir
state cloud induced by all patterns loaded up to cycle $m$. The
complement $F^m = \neg A^m$ in turn characterizes the memory space
that is still ``free'' for use in loading cycle $m+1$.  

The new input simulation weights $D^{m+1}$ and output weights
$W^{\mbox{\scriptsize out}, m+1}$ are computed on the basis of $F^m$ and $D^m$
as follows:

\begin{enumerate}
    \item Drive the native reservoir with pattern $p^{m+1}$ via
  ${x}^{m+1}(n+1) = \tanh(W^\ast \, {x}(n)^{m+1} +
  W^{\mbox{\scriptsize in}}\, p^{m+1}(n) + {b})$ for $L$ steps,
  collect reservoir states ${x}^{m+1}(n_0), \ldots, {x}^{m+1}(L-1)$ 
  column-wise into a state collection matrix $X^{m+1}$ of size $N
  \times (L - n_0)$, furthermore collect time-shifted reservoir states
  ${x}^{m+1}(n_0+1), \ldots, {x}^{m+1}(L)$ into 
   $X^{m+1}_+$, collect pattern samples $p^{m+1}(n_0+1), \ldots,
  p^{m+1}(L)$ into a $d \times (L- n_0)$ matrix $P^{m+1}$, and 
compute $C^{m+1}$ by \eqref{eq:CompConceptor}.
    \item Obtain the increment $D_{\mbox{\scriptsize inc}}^{m+1}$ by
  regularized linear regression from these states ${x}^{m+1}$ as the
  minimizer of the loss
\begin{equation}\label{e23Rev1}
\sum_{n=n_0, \ldots, L} \|W^{\mbox{\scriptsize in}}\,p^{m+1}(n) - D^m
\, {x}^{m+1}(n-1) - D_{\mbox{\scriptsize inc}}^{m+1} \, F^m \, {x}^{m+1}(n-1)  \|^2.
\end{equation}
Concretely, this means to compute
\begin{equation}\label{e23aRev1}
D^{m+1}_{\mbox{\scriptsize inc}} = \left((SS'/(L-n_0) + \alpha^{-2}
I)^{-1} \, S {T^D}' / (L-n_0)\right)', 
\end{equation}
where $S = F^{m} \, X^{m+1}$, $T^D =  W^{\mbox{\scriptsize in}} P^{m+1}
  - D^{m} \, X^{m+1}$ contains the targets for the linear regression
  leading to $D^{m+1}_{\mbox{\scriptsize inc}}$, and $\alpha^{-2}$ is the
Tychonov regularizer. 
\item  Obtain  the increment $W^{\mbox{\scriptsize out}, m+1}_{\mbox{\scriptsize
    inc}}$ by regularized linear regression as the minimizer of the
loss
\begin{equation}\label{e24}
\sum_{n=n_0, \ldots, L} \|p^{m+1}(n) - W^{\mbox{\scriptsize out},
  m}\,{x}^{m+1}(n) - W^{\mbox{\scriptsize out}, m+1}_{\mbox{\scriptsize
    inc}}\,F^m \, {x}^{m+1}(n)  \|^2.
\end{equation}
Concretely this means to compute
\begin{equation}\label{e24a}
W^{\mbox{\scriptsize  out}\, m+1}_{\mbox{\scriptsize inc}} =
\left((S_+ S'_+ /(L-n_0) + a^2_{\mbox{\scriptsize out}}\,I)^{-1} 
\, S_+ {T^{\mbox{\scriptsize out}}}'/(L-n_0)\right)',
\end{equation}
where $S_+ = F^{m} \, X^{m+1}_+$,  $T^{\mbox{\scriptsize out}} =
P^{m+1} - W^{\mbox{\scriptsize out}, m} X^{m+1}_+$, and
$a^2_{\mbox{\scriptsize out}}$ is a regularization coefficient.
  \item Update $A^{m+1} = A^m \vee C^{m+1}, \; F^{m+1} = \neg A^{m+1},
\; D^{m+1} = D^m + D_{\mbox{\scriptsize inc}}^{m+1}, \;
W^{\mbox{\scriptsize out}, m+1} = W^{\mbox{\scriptsize out}, m} +
W^{\mbox{\scriptsize out}, m+1}_{\mbox{\scriptsize inc}}$.
\end{enumerate}

It is interesting to note that it is intrinsically impossible to
\emph{unlearn} patterns selectively and ``decrementally''. Assume that patterns
$p^1, \ldots, p^m$ have been trained, resulting in $D^m$. Assume that
one wishes to unlearn again $p^m$. As a result of this unlearning one
would want to obtain $D^{m-1}$. Thus one would have to compute
$D^m_{\mbox{\scriptsize inc}}$ from $D^m$, $A^m$ and $p^m$ (that is, from
$D^m$ and $C^m$), in order to recover $D^{m-1} = D^m -
D^m_{\mbox{\scriptsize inc}}$. However, the way to identify
$D^m_{\mbox{\scriptsize inc}}$ from $D^m$, $A^m$ and $C^m$ is barred because
of the redundancy exploitation inherent in step 2. Given only
$D^m$, and not knowing the patterns $p^1,\ldots,p^{m-1}$ which must be
preserved, there is no way to identify which directions of reservoir
space must be retained to preserve those other patterns. The best one
can do is to put $\tilde{A}^{m-1} = A^m - C^m = A^m \wedge \neg C^m$
and re-run step 3 using $\tilde{A}^{m-1}$ instead of $A^{m-1}$ and
putting $T = W^{\mbox{\scriptsize in}}\, P^j$ in step 2. This leads
to a version $\tilde{D}^m_{\mbox{\scriptsize inc}}$ which coincides
with the true $D^m_{\mbox{\scriptsize inc}}$ only if there was no
directional overlap between $C^m$ and the earlier $C^{m'}$, i.e. if
$D^{m-1}X^m = 0$ in the original incremental learning procedure. To
the extent that $p^m$ shared state directions with the other patterns,
i.e. to the extent that there was redundancy, unlearning $p^m$ will
degrade or destroy patterns that  share state directions with
$p^m$. 

The incremental pattern learning method offers the commodity to
measure how much ``memory space'' has already been used after the
first $j$ patterns have been stored. This quantity
is the quota  $q(A^j)$. When it approaches 1, the reservoir is
``full'' and an attempt to store another pattern will fail because the
$F^m$ matrix  will be close to zero.

Two demonstrations, detailed below, illustrate various aspects of the
incremental storing procedure.  In the first demonstration, the
patterns were periodic with integer period lengths. In the second
demonstration, the patterns came from a 2-parametric family of created
from weighted and phase-shifted sums of sinewaves with irrational
period lengths. These two kinds of patterns display interestingly
different characteristics in incremental storage.

\subsubsection{Demonstration 1: Incremental Loading of
  Integer-Periodic Patterns} \label{subsec:memManageDemo1}

In the first demo I loaded $K = 16$ integer-periodic patterns into an
$N = 100$ sized reservoir,  using the
incremental loading procedure detailed above.  The patterns were sines
of integer period length or random periodic signals. Period lengths
ranged between 3 and 15 (details documented in Section
\ref{secDetailmemmanageDemo1}).  Figure \ref{memManFig1} displays 
characteristic impressions.

\begin{figure}[htbp]
\center
\includegraphics[width=145 mm]{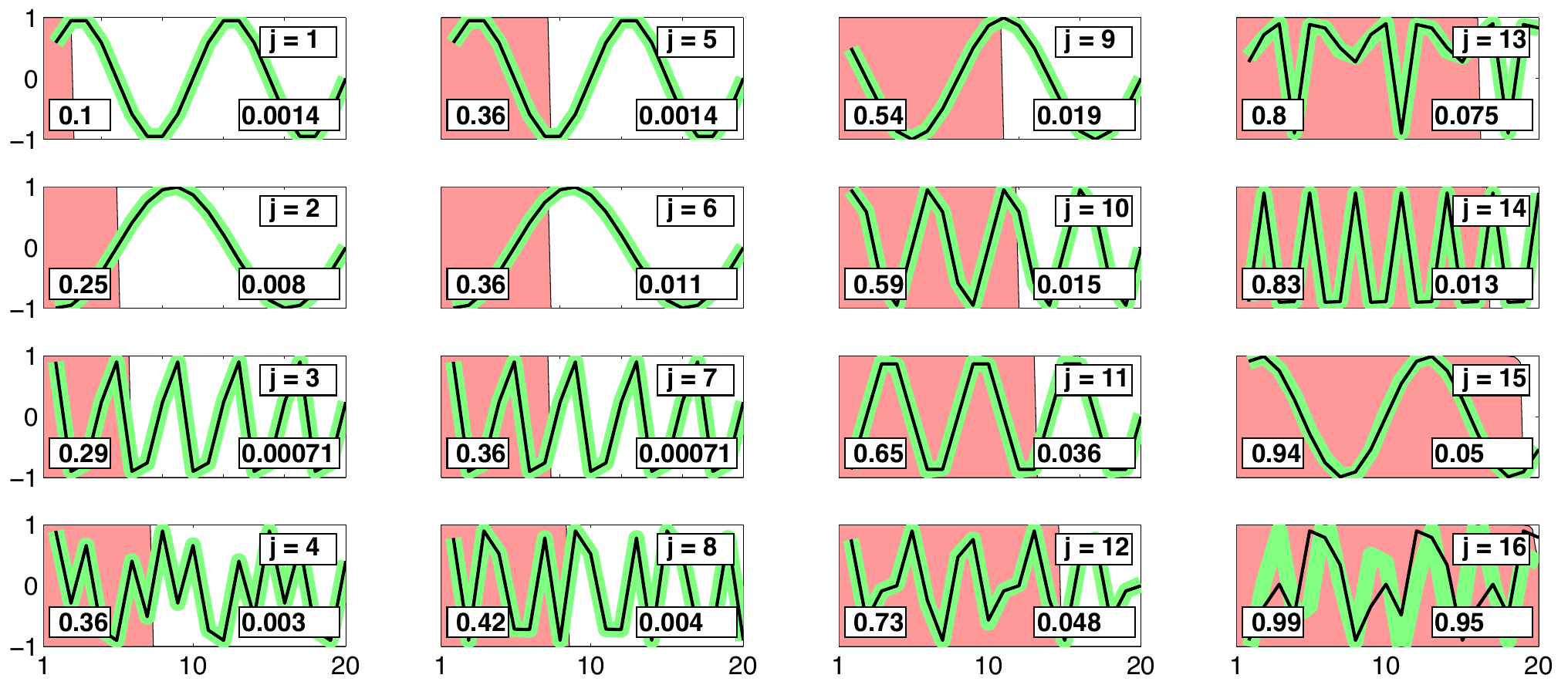}
\caption{Incremental storing, first demonstration (figure repeated
  from Section \ref{secOverview} for convenience). 13 patterns with
  integer period lengths ranging between 3 and 15 were stored.
  Patterns were sinewaves with integer periods or random. Patterns $j
  = 5,6,7$ are identical to $j = 1, 2, 3$.  Each panel shows a
  20-timestep sample of the correct training pattern $p^j$ (black
  line) overlaid on its reproduction (green line). The memory fraction
  used up until pattern $j$ is indicated by the panel fraction filled
  in red; this quota value is printed in the left bottom corner of each
  panel. The red areas in each panel in fact show the singular value
  spectrum of $A^j$ (100 values, $x$ scale not shown). The NRMSE is
  inserted in the bottom right corners of the panels. Note that the
  conceptor-controlled reproduction of all patterns was carried out
  \emph{after} the last pattern had been loaded.}
\label{memManFig1}
\end{figure}

\noindent \textbf{Comments.} When a reservoir is driven with a signal that
has an integer period, the reservoir states (after an initial washout
time) will entrain to this period, i.e.\ every neuron likewise will
exhibit an integer-periodic activation signal. Thus, if the period
length of driver $p^j$ is $L^j$, the state correlation matrix as well as
the conceptor $C^j$ will be matrices of rank $L^j$. An aperture $\alpha
= 1000$ was used in this demonstration. The large size of this
aperture and the fact that the state correlation matrix has rank $L^j$ leads to a conceptor
$C^j$ which comes close to a projector matrix, i.e.\ it has $L^j$
singular values that are close to one and $N-L^j$ zero singular
values. Furthermore, if a new pattern $p^{j+1}$ is presented, the
periodic reservoir state vectors arising from it will generically be
linearly independent of all state vectors that arose from earlier
drivers. Both effects together (almost projector $C^j$ and linear
independence of nonzero principal directions of these $C^j$) imply
that the sequence $A^1,\ldots, A^K$ will essentially be a sequence of
projectors, where $\mathcal{R}(A^{j+1})$  will comprise $L^{j+1}$ more
dimensions than  $\mathcal{R}(A^{j})$. This becomes clearly apparent in Figure
\ref{memManFig1}: the area under the singular value plot of $A^j$ has
an almost rectangular shape, and the increments from one plot to the
next match the periods of the respective drivers, except for the last
pattern, where the network capacity is almost exhausted. 

Patterns $j = 5,6,7$ were identical to $j = 1, 2, 3$. As a
consequence, when the storage procedure is run for $j = 5,6,7$, $A^j$
remains essentially unchanged -- no further memory space is
allocated.

When the network's capacity is almost exhausted in the sense that the
quota $q(A^j)$ approaches 1, storing another pattern becomes
inaccurate. In this demo, this happens for that last pattern $j =
16$ (see Figure \ref{memManFig1}).

For a comparison I also loaded all 16 patterns simultaneously in a
separate simulation experiment,
computing only a single input simulation weight matrix $D$ by
minimizing \eqref{eqDMatError}. The dots in the left panel in Figure
\ref{figMemManOverview} indicate the recall accuracies based on this
$D$. The mean of NRMSEs of this comparison was 0.063, while the mean
NRMSE of recalled patterns in the incremental loading condition after
the final loading cycle was 0.078. These two mean values do not differ
much, but the recall NRMSEs scatter differently in the two conditions.
In the non-incremental loading condition, the recall accuracies are
much closer to each other than in the incremental condition, where we
find many highly precise accuracies and one poor outlier, namely the
last pattern which spilled over the remaining memory space. Apparently
in simultaneous loading the common $D$ strikes a compromise for all
loaded patterns on equal terms, whereas in incremental loading
earlier-loaded patterns receive the benefit of competing only with the
patterns that had been loaded before.

\begin{figure}[tb]
 \center
   \includegraphics[width=150mm]{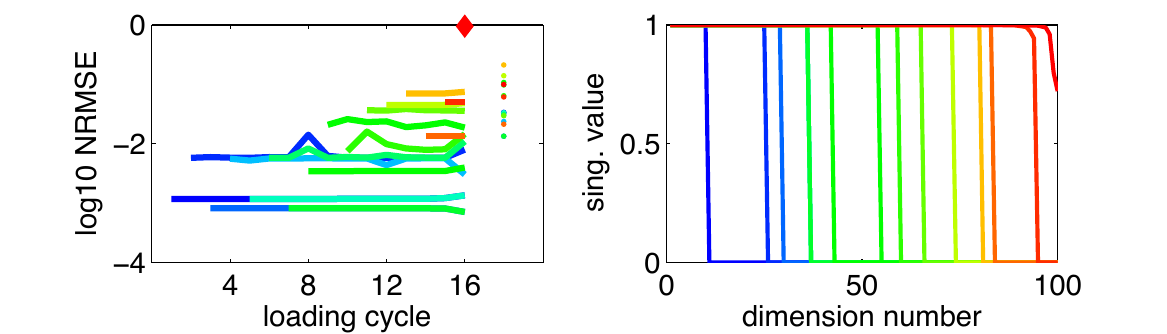}
 \caption{Incremental loading of integer-periodic patterns:
   detail. \emph{Left:}  Development of re-generation error (log10 of 
   NRMSE) for the 16 patterns from Figure \ref{memManFig1}  during the
   incremental loading procedure. Each line corresponds to one
   pattern. The red diamond marks the NRMSE of the last-loaded
   pattern. The 16 dots at the right mark the recall errors obtained
   from non-incremental loading.  \emph{Right:} The
   essentially rectangular singular value spectra of the conceptors
   $A^m$, same color code as in left panel. }
 \label{figMemManOverview}
 \end{figure}

\subsection{Incremental Loading of Integer-Periodic Patterns: Detailed
  Analysis}

Here I provide a more detailed analysis of the incremental loading of
integer-periodic patterns (demo in Section \ref{subsec:memmanage}). I
consider the case of incrementally loading patterns $p^1, p^2,
\ldots$, which have integer period lengths $L^1, L^2, \ldots$, into an
$N$-dimensional reservoir. I refer to $\mathcal{M} =
\mathbb{R}^N$ as the \emph{memory space} provided by the reservoir.

The periodic patterns $p^1, p^2, \ldots$ give rise to a sequence of
mutually orthogonal linear ``excitation'' subspaces $(\mathcal{E}^j)_{j = 1, 2,
  \ldots}$ of $\mathcal{M}$, with dimensions $L^j$, as follows.

When the reservoir is driven by pattern $p^1$, after a washout period
it will engage in an $L^1$-periodic sequence of states, that is,
${x}^1(n) = {x}^1(n+L^1 -1)$. Generically (i.e.\ with probability 1 for
random draws of network weights), these $L^1$ states ${x}^1(n), \ldots,
{x}^1(n+L^1 - 1)$ are linearly independent and span an $L^1$-dimensional
linear subspace $\mathcal{E}^1$ of $\mathcal{M}$.   Define
$\mathcal{M}^1 = \mathcal{E}^1$.

Now assume, by induction, that for the first $m$ patterns $p^1,\ldots,
p^m$ we have found $m$ pairwise orthogonal linear subspaces
$\mathcal{E}^1, \ldots, \mathcal{E}^m$ of dimensions $L^1, \ldots,
L^m$, with $\mathcal{M}^m = \mathcal{E}^1 \oplus \ldots \oplus
\mathcal{E}^m$ being the $(L^1 + \ldots L^m)$-dimensional subspace
given by the direct sum of these. Now let the native reservoir be
driven with the next pattern $p^{m+1}$. Provided that this pattern is
not a duplicate of any of the previous ones, and provided that $L^1 +
\ldots + L^m + L^{m+1} \leq N$, the collection of the $L^1 + \ldots +
L^{m+1}$ many induced reservoir states ${x}^1(n),\ldots, {x}^1(n+L^1-1),
\ldots, {x}^{m+1}(n), \ldots, {x}^{m+1}(n+L^{m+1} - 1)$ will again
generically be linearly independent, spanning an $(L^1 + \ldots +
L^{m+1})$-dimensional subspace $\mathcal{M}^{m+1}$ of $\mathcal{M}$.
Define $\mathcal{E}^{m+1}$ to be the orthogonal complement to
$\mathcal{M}^m$ within $\mathcal{M}^{m+1}$. Then $\mathcal{E}^{m+1}$
has dimension $L^{m+1}$. This concludes the induction step in the
definition of the sequence $(\mathcal{E}^j)_{j = 1, 2,
  \ldots}$. Note that this construction only works as long as $L^1 +
\ldots +  L^{m+1} \leq N$, and assumes that the
patterns $p^j$ are pairwise different. 

The spaces $\mathcal{M}^j$ and $\mathcal{E}^j$ can be characterized through the
conceptors $C^j$ associated with the patterns $p^j$. To see this,
consider first the
conceptor matrix $C^1$. It has exactly $L^1$ nonzero eigenvalues
corresponding to eigenvectors which  span $\mathcal{M}^1$. For
large aperture (I used $\alpha = 1000$ in the demo), the eigenvalue
spectrum of $C^1$ will be approximately rectangular, with $L^1$
eigenvalues close to 1 and $N-L^1$ eigenvalues exactly zero (right
panel in  Figure \ref{figMemManOverview}). In
linear algebra terms, $C^1$  approximately is the
projector matrix which projects $\mathcal{M}$ on $\mathcal{E}^1$, and
$\neg C^1$ as the projector matrix that maps $\mathcal{M}$ on
$(\mathcal{E}^1)^\bot$. Define $A^1 = C^1, F^1 = \neg C^1$, as in the
algorithm given in the previous subsection.

Again by induction, assume that $A^m = C^1 \vee \ldots \vee C^m$ is
approximately the projector of $\mathcal{M}$ on $\mathcal{M}^m$ and
$F^m = \neg{A^m}$ the complement projector of $\mathcal{M}$ on
$(\mathcal{M}^m)^\bot$. It holds that $A^{m+1} = A^m \vee C^{m+1}$ is
(approximately) the projector from $\mathcal{M}$ on $M^{m+1}$ (follows
from Proposition \ref{propSpaces}), and $F^{m+1} = \neg
A^{m+1}$ the projector on $(M^{m+1})^\bot$.

Equipped with these interpretations of $A^m$ and $F^m$ we turn to the
incremental loading procedure.  

According to the loss \eqref{e23Rev1}, $D_{\mbox{\scriptsize inc}}^{m+1}$
is optimized to map $ F^m \, {x}^{m+1}(n-1)$ to $W^{\mbox{\scriptsize
    in}}\,p^{m+1}(n) - D^m \, {x}^{m+1}(n-1)$, that is, to the
pattern-input term $W^{\mbox{\scriptsize in}}\,p^{m+1}(n)$ minus those
reservoir state components which are already produced by the input
simulation weights $D^m$. As we have just seen, $F^m$  is
approximately a
projection of reservoir states on the linear subspace of $\mathcal{M}$
which was not excited by any of the previous driving patterns, that
is, a projection on $(\mathcal{M}^m)^\bot$. Basing
the regularized linear regression on arguments $F^m \, {x}^{m+1}(n-1)$
which lie in $(\mathcal{M}^m)^\bot$
leads to a solution for $D_{\mbox{\scriptsize inc}}^{m+1}$ such
that $D_{\mbox{\scriptsize inc}}^{m+1}$ nulls all reservoir state
components that fall in the linear subspace $\mathcal{M}^m$ excited by
previous patterns. 

By an easy inductive argument it can be seen that $D^m$, in turn,
nulls all vectors in $(\mathcal{M}^m)^\bot$.

Decomposing states excited by $p^{m+1}$ as 
 ${x}^{m+1}(n) = {u} + {v}$, with ${u} \in
\mathcal{M}^m$, ${v} \in (\mathcal{M}^m)^\bot$,  one obtains
 (approximately, to the extent that the $C^j$ are projectors) 
\begin{eqnarray}
D^m \, {u} & \approx & D^m \,{x}^{m+1}(n), \label{e25}\\
 D^m \, {v} & \approx & 0 \label{e26}\\
D_{\mbox{\scriptsize inc}}^{m+1} \, {u} & \approx & 0, \label{e27}\\
D_{\mbox{\scriptsize inc}}^{m+1} \, {v} & \approx & D_{\mbox{\scriptsize
 inc}}^{m+1}\, {x}^{m+1}(n) \; \approx \; D_{\mbox{\scriptsize
    inc}}^{m+1}\, F^m \, {x}^{m+1}(n) \; \nonumber\\
 & \approx & W^{\mbox{\scriptsize
    in}}\,p^{m+1}(n+1) - D^m \, {x}^{m+1}(n), \label{e28}
\end{eqnarray}
where the  approximation \eqref{e28} is accurate to the extent that the
linear regression solution to \eqref{e23Rev1} is accurate. Hence, 
\begin{eqnarray}
D^{m+1}\,{x}^{m+1}(n) & = & (D^m + D_{\mbox{\scriptsize
    inc}}^{m+1})\,({u} +{v}) \nonumber\\
& = & D^m \,{x}^{m+1}(n) + W^{\mbox{\scriptsize in}}\,p^{m+1}(n+1) - D^m \,
{x}^{m+1}(n)\nonumber\\
& = & W^{\mbox{\scriptsize in}}\,p^{m+1}(n+1),\label{e29}
\end{eqnarray}
as desired for input simulation weights. The reasoning for output
weights computed on the basis of \eqref{e24} is analog.

\eqref{e29} explains why the input simulation weights $D^{m+1}$
can recover the pattern $p^{m+1}$ at recall time. To see that
furthermore $D^{m+1}$ also still recovers earlier loaded patterns
(avoiding catastrophic forgetting), observe that reservoir states
associated with earlier patterns fall in $\mathcal{M}^m$ and that by
\eqref{e25}--\eqref{e27}, $D^{m+1}$ restricted on $\mathcal{M}^m$
operates identically as $D^{m}$, and $D^{m+1}$ restricted on
$\mathcal{M}^{m-1}$ as $D^{m-1}$, etc.

I want to point out a possible misunderstanding of the working
principle of this incremental loading procedure. It does \emph{not}
function by splitting the memory space $\mathcal{M}$ into orthogonal
components $\mathcal{E}^j$, and loading the patterns $p^j$ separately
into these in some way. The states ${x}^{m+1}(n)$ obtained by exciting
the reservoir with $p^{m+1}$ are generically non-orthogonal to the
excitation space $\mathcal{M}^m$ of previously loaded patterns.  The
re-generation of $p^m$ ``uses'' reservoir state components from all
the subspaces $\mathcal{E}^1, \ldots, \mathcal{E}^m$. In this respect,
the incremental loading procedure  fundamentally differs from a
common strategy adopted to counter catastrophic forgetting by
orthogonalizing internal representations of patterns in some way
(survey in \cite{French03}).

\subsubsection{Demonstration 2: Incremental Loading of
   Irrational-Period Patterns}\label{subsec:memManageDemo2}
 
 In the second demo I loaded 16 patterns that were randomly taken from
 the the 2-parametric family of patterns governed by
\begin{equation}\label{e30}
p(n) = a \, \sin(2 \, \pi \, n / P) + (1-a) \, \sin(4 \, \pi \, (b + n / P) ). 
\end{equation}

These signals are weighted sums of two sines, the first with period
length $P$ and the second with period length $P/2$. The weights of
these two components are $a$ and $(1-a)$, and the second component is
phase-shifted relative to the first by a fraction $b$ of its period
length $P/2$. The reference period length $P$ was fixed to $P =
\sqrt{30}$. The parameters $a, b$ were freshly sampled from the
uniform distribution on $[0, 1]$ for each $p^j$, where 16 patterns
were used. Because $P$ is irrational, these patterns (when sampled at
integer steps) are not integer-periodic but quasi-periodic. As a
consequence, the reservoir states ${x}^j(n)$ excited by a driver
$p^j(n)$ span all of $\mathcal{M}$, and all singular values of $C^j$
are nonzero, with a reverse sigmoid-shaped spectrum like we saw
earlier for other irrational-period patterns (Figure
\ref{FigSingValsFalloff} [left]). Network parameters are reported in
Section \ref{secDetailmemmanageDemo2}. Figures \ref{memManFig2} and
\ref{figMemManOverview_irrational} display the findings.

\begin{figure}[htbp]
\center
\includegraphics[width=145 mm]{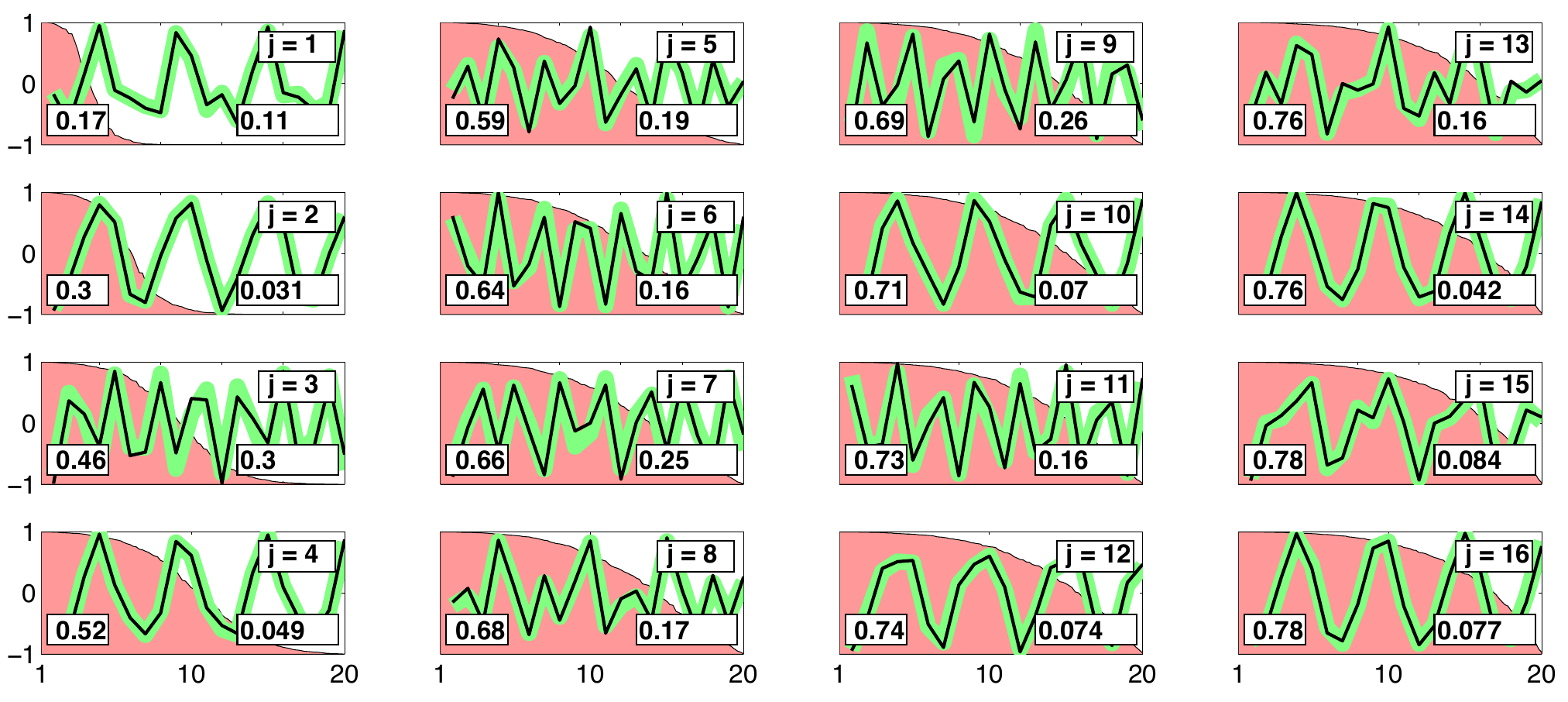}
\caption{Incremental storing, second demonstration. 16 sinewave patterns with
  irrational periods ranging between 4 and 20 were used. Plot layout
  is the same as in Figure \ref{memManFig1}.}
\label{memManFig2}
\end{figure}

\begin{figure}[tb]
 \center
   \includegraphics[width=150mm]{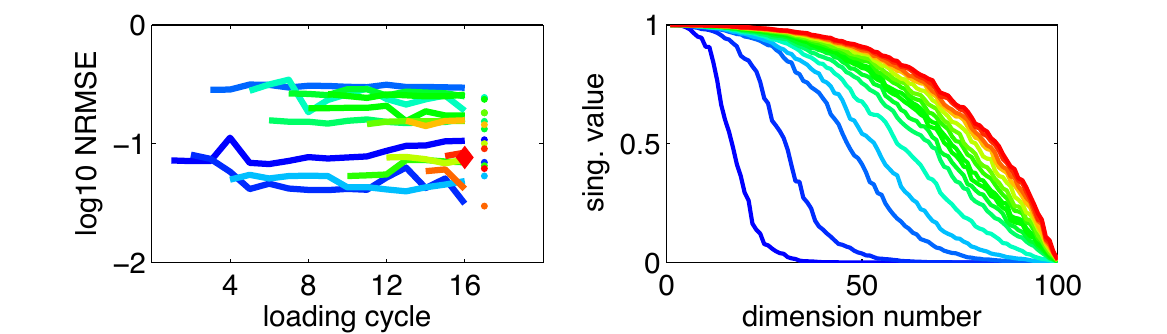}
 \caption{Detail of incremental loading of irrational-periodic
   patterns. Figure layout as in Fig.\ \ref{figMemManOverview}.}
 \label{figMemManOverview_irrational}
 \end{figure}

\noindent {\textbf Comments.}  When the driver has an irrational period
length, the excited reservoir states will span the available
reservoir space $\mathbb{R}^N$. Each reservoir state correlation matrix will have only nonzero
singular values, albeit of rapidly decreasing magnitude (these tails
are so small in magnitude that they are not visible in the first few
plots of $A^j$ in Figure \ref{memManFig2}). The fact that each driving
pattern excites the reservoir in all directions leads to the
``reverse sigmoid'' kind of  shapes of the singular values of
the $A^j$ visible in  Figure \ref{memManFig2}. 

As the iterated storing progresses,  a redundancy exploitation effect becomes
apparent: while for the first 4 patterns altogether a quota $q(A^4) =
0.52$ was allocated, the remaining 12 patterns only needed an additional
quota of $q(A^{16}) - q(A^4) = 0.26$. Stated in suggestive terms, at later stages
of the storing sequence the network had already learnt how to
oscillate in sinewave mixes in general, and only needed to learn in
addition how to oscillate at the particular newly presented
version taken from the parametrized pattern family. 
An aperture of size $\alpha = 1.5$ was used in the second
demonstration.

The
mean recall NRMSE for the 16 patterns (testing after loading the last
one) was 0.136. The mean NRMSE for these patterns when loaded
non-incrementally was 0.131. Like for integer-periodic patterns, these
values are not substantially different from each other.

\subsubsection{Integer-Periodic Versus Parametrized Patterns: Close-up
 Inspection}\label{subsecIncLoadSurvey} 

Notice the very different shape of the singular value spectra of the
$A^j$ in the integer-periodic versus the irrational-periodic patterns
(right panels in Figures \ref{figMemManOverview} and
\ref{figMemManOverview_irrational}). This suggests substantial
differences in the underlying mechanisms, and it is also obvious that
the analysis offered in the previous subsection for integer-periodic
patterns does not transfer to the irrational-periodic patterns from a
parametrized family which we considered here.

To understand better the differences between the incremental loading
of integer-periodic versus parametrized patterns, I ran a separate
suite of simulations with additional diagnostics, as follows (detail
in Section \ref{secDetailMemManageCloseup}).  The
simulation with integer-periodic patterns was repeated 10 times, using
the same global network scalings and regularization coefficients as
before, but using exclusively 6-periodic patterns throughout. Loading
sixteen such patterns into a 100-neuron reservoir should claim a
memory quota of about $6 \cdot 16 / 100 = 0.96$, just short of
over-loading. Network weights and patterns were randomly created for
each of the 10 simulations. Similarly, the parametric-family
simulation was repeated 10 times, using the same pattern family
\eqref{e30} with freshly sampled pattern parameters and network
weights in each simulation. Finally, I loaded 100 integer-periodic
patterns of period length 3 into the same reservoir that was used for
the 6-periodic patterns, again repeating the simulation 10 times. For
each individual simulation, the following diagnostic quantities were
computed (indicated colors refer to color code in Figure \ref{fig15_Rev1}):

\begin{enumerate}
    \item The mean absolute values $\overline{a^m}$ of arguments $a^m(n) = F^m \,
  {x}^m(n-1)$ that enter the regression for $D^m_{\mbox{\scriptsize
      inc}}$ (mean taken over $n$ and vector components,
  \emph{\color{orange} orange}).
    \item The mean absolute values $\overline{t^m}$ of targets $t^m(n) =
  W^{\mbox{\scriptsize in}}\, p(n) - D^{m-1}\, {x}^m(n-1)$ that enter
  the regression for $D^m_{\mbox{\scriptsize inc}}$ (\emph{\color{red}
    red}).
    \item The mean absolute size of matrix entries in the
  $D^m_{\mbox{\scriptsize inc}}$ matrices,  normalized by the inverse of
the scaling factor that $D^m_{\mbox{\scriptsize inc}}$ has to
realize.  Concretely, define  $\overline{D^m_{\mbox{\scriptsize inc}}}$ to
be the average absolute size of matrix elements in
  ${D^m_{\mbox{\scriptsize inc}}} \, \overline{a^m} /
\overline{t^m}$ (\emph{\color{blue} blue}).
\item The mean absolute size of matrix entries in $D^m$ (\emph{\color{gray} gray}).
\item The memory quota used up to the current loading cycle
(\emph{black}).
\item The condition number of the matrix inversion that has to be
carried out in the (regularized) linear regression when computing
$D^m_{\mbox{\scriptsize inc}}$. This is a number $\geq 1$. Values
close to 1 indicate that the linear regression  faces a simple
task (arguments component signals are orthogonal and have same signal
power; regression amounts to linearly combine orthogonal projections)
(\emph{\color{green} green}).
\item The recall NRMSEs for the 16 patterns $p^1, \ldots, p^{16}$
after the final loading cycle. Gives 16 NRMSE values per simulation
(\emph{\color{cyan} cyan}).
\item The recall NRMSEs computed for pattern $p^j$ after the $j$-th
loading cycle. Gives 16 NRMSE values per simulation
(\emph{\color{magenta} magenta}).
\end{enumerate} 

\begin{figure}[tb]
 \center
   \includegraphics[width=150mm]{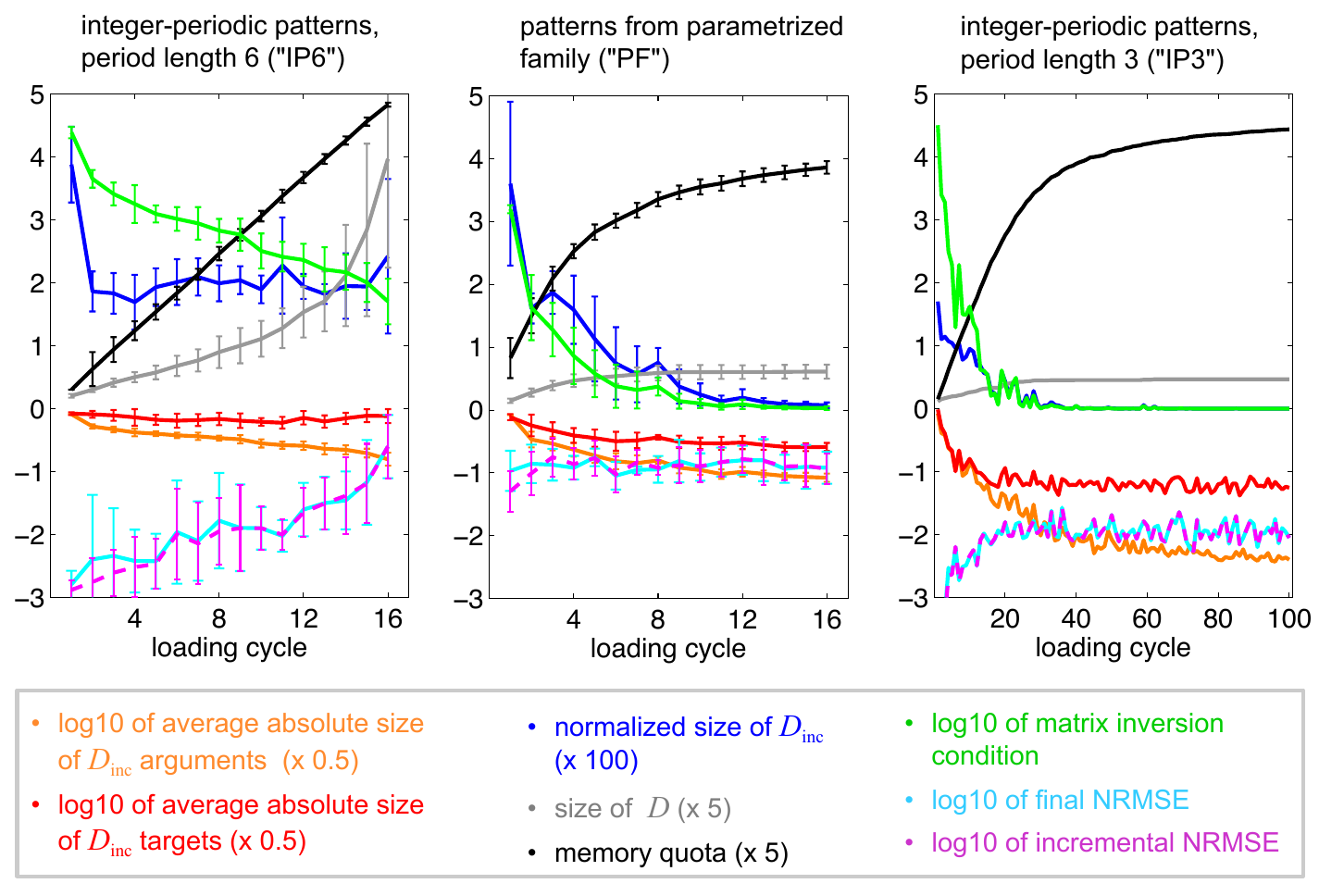}
 \caption{Detailed diagnostics of the incremental loading
   procedure. Plots show means of 10 simulations, error bars indicate
   standard deviations (omitted in third panel for clarity). Some
   curves were scaled to fit the panel, legend indicates scalings. The
   horizontal axis marks the loading cycles except for the final NRMSE
   ({\color{cyan} cyan}) where it marks the pattern number. For
   explanation see text.}
 \label{fig15_Rev1}
 \end{figure}

For brevity I refer to the integer-periodic pattern simulations as the
``IP6'' and ``IP3'' conditions and to the parametric family patterns as the ``PF''
condition, and  to the diagnostic quantities by their colors in
Figure \ref{fig15_Rev1}. Here are some observations that help to gain 
intuitive insight---future mathematical analyses notwithstanding:

\begin{itemize}
\item In both IP6 and PF, the reconstruction accuracy for a pattern
$p^j$ directly after its loading is, by and large, about the same as 
the reconstruction accuracy after the last loading cycle
(\emph{\color{cyan} cyan},
\emph{\color{magenta} magenta}). That is, subsequent loading events do not degrade
the representation of previously loaded patterns---this is also
manifest in the left panels of Figures \ref{figMemManOverview} and \ref{figMemManOverview_irrational}.
  \item In IP6 the reconstruction accuracy (\emph{\color{cyan} cyan},
\emph{\color{magenta} magenta}) deteriorates roughly linearly in terms
of the log-NRMSE as loading cycles unroll. This might be connected to
the circumstance that on the one hand, each new pattern is unrelated
to the previous ones and should require the same amount of ``coding
information'' to be stored (apparent in the linear increase of the
quota, \emph{black}), but on the other hand, increasingly little
``free memory space'' $F^m$ can be recruited for the coding and hence,
increasingly less ``degrees of freedom for coding'' are available. 
  \item In PF the final reconstruction quality (\emph{\color{cyan}
  cyan}) is about the same for all patterns, and it is about the same
as the incrementally assessed reconstruction NRMSEs
(\emph{\color{magenta} magenta}). In fact, if more patterns from this
family were to be added in further loading cycles, the reconstruction
NRMSEs still would stay on the same level (not shown).  The memory
  system can learn the entire pattern family. 
    \item The condition number (\emph{\color{green} green}) of the
  matrix that has to be inverted in the regression improves in all of
  IP6, FP and IP3, but to different degrees. Concretely, this matrix
  is $\mbox{mean}_n \{F^m \, {x}^m(n)\, (F^m \, {x}^m(n))'\} + \varrho \,
  I$, where $\varrho$ is the ridge regression regularizing
  coefficient (set to 0.001, 0.02, 0.001 in IP6, FP and IP3). In PF,
  the condition number swiftly approaches 1. This can be understood as
  an effect of dominance of regularization: given that the average
  absolute value of arguments $F^m \, {x}^m(n)$ decreases to an order of
  magnitude of 0.01 (\emph{\color{orange} orange}), the spectral
  radius of $\mbox{mean}_n \{F^m \, {x}^m(n)\, (F^m \, {x}^m(n))'\} $ will
  drop to an order of 0.0001, hence the linear regression becomes dominated
  by the  regularization part with coefficient $\varrho =  0.02 \gg 0.0001$. In
  intuitive terms, in PF the iterated re-computations of $D^m$ soon
  cease to ``learn anything substantially new''. Because the recall
  accuracies do not degrade, this means that nothing new \emph{has} to
  be learned---after a few loading cycles, the memory system has
  extracted from the examples stored so far almost all that is
  necessary to represent the entire family. 
\item  The situation is
  markedly different in IP6. Here the condition number never falls
  below about 100, the spectral radius of  $\mbox{mean}_n \{F^m \,
  {x}^m(n)\, (F^m \, {x}^m(n))'\} $ is about the same size as $\varrho =
  0.001$, thus substantial novel information becomes coded
  in each $D^m$ until the end. 
\item The interpretation that in IP6 novel information becomes coded
throughout all loading cycles whereas in PF the rate of coding new
information falls to zero is further substantiated by the development
of the (normalized) sizes $\overline{D^m_{\mbox{\scriptsize inc}}}$
of the increments that are added to $D^m$ (\emph{\color{blue}
  blue}). In IP6 these increments retain approximately the same size
through all loading cycles whereas in PF they appear to decay toward
zero (in simulations with much larger numbers of loading cycles [not
shown] this impression was numerically  verified). In the same vein,
the mean absolute sizes of elements in the $D^m$ matrices (\emph{\color{gray}
  gray}) grows superlinearly in IP6 and levels out in PF. 
  \item The two different modes of operation that we find contrasted
in the IP6 and PF conditions can occur both together in a pattern
loading sequence. In the IP3 condition, the 3-periodic patterns can be
viewed as being drawn from a 2-parametric family (the random length-3
patterns were normalized to all have the same minimal and maximal
values, thus can be described up to period shifts by the binary
parameter ``the maximal value point follows the minimal value point
directly yes/no'' and the continuous parameter that gives the
non-extremal point value). When the first few (about 20) patterns from
this collection are loaded, they lead to a similar phenemenology as in
the IP6 condition: each additional pattern is ``novel'' to the memory
system. When more loading cycles are executed, the phenomenology
changes to the type seen in the PF condition: the network (more
precisely, the input simulation matrix $D^m$) has captured the ``law''
of the entire family and not much further information has to be
coded. 
\end{itemize}

While these simulations help us to get basic intuitions about the
mechanisms that enable incremental pattern loading in different
conditions, a formal analysis would be quite complex and is left for
future work. Questions that would need to be addressed include the
effects of regularization (which are strong), aperture settings, and
asymptotic behavior when the number of loaded patterns grows to
infinity.

\subsubsection{Incremental Loading of Arbitrary
  Patterns}\label{subsecIncLoadMixed} 

When one tries to store patterns (i) whose conceptor singular value
spectra are not close to rectangular and (ii) which do not come from a
parametric family, the incremental loading procedure breaks down (not
shown). The reason for this failure is clear on intuitive grounds.
When the singular value spectra of conceptors $C^m$ are
``sigmoid-shaped'' as e.g.\ in Figure
\ref{figMemManOverview_irrational}, the memory space claimed by
patterns loaded up to loading cycle $m$ will have nonzero components
in all directions of $\mathcal{M}$---technically, the singular value
spectrum of $A^m$ will nowhere be zero. The argument vectors $F^{m}\,
{x}^{m+1}(n) = \neg\, A^{m}\, {x}^{m+1}(n)$ which enter the regression
for $D^m_{\mbox{\scriptsize inc}}$ likewise will have signal
components in all directions of $\mathcal{M}$, and executing the
procedure for incremental loading will lead to a confounding of
already stored with newcoming patterns.

This diagnosis directly hints at a solution: make the singular
value spectra of the conceptors $C^m$ associated with (arbitrary)
patterns $p^j$ rectangular or approximately rectangular. One possible
procedure to achieve this runs like follows:

\begin{enumerate}
\item When loading pattern $p^m$, compute the associated conceptor
$C^m$ as before. 
\item Compute its SVD $U\,S\,U' = C^m$. $S$ is a diagonal matrix which
contains  the singular values of $C^m$ on its diagonal. Denote the
singular values by $\sigma_i$. 
\item Change these $\sigma_i$ in a way that yields new
$\tilde{\sigma}_i$ which are either close to 0 or close to 1. A
drastic procedure is thresholding: $\tilde{\sigma}_i = 1$ if $\sigma_i
> 0.5$ else  $\tilde{\sigma}_i = 0$. A softer procedure is to pass
$\sigma_i$ through some sufficiently steep sigmoid. 
\item Use the ``rectangularized'' $\tilde{C}^m = U \tilde{S} U'$
instead of the original $C^m$ in the incremental loading procedure
which otherwise remains unchanged (where $\tilde{S}$ is the diagonal
matrix with the $\tilde{\sigma}_i$ on its diagonal). 
\end{enumerate}

\begin{figure}[tb]
 \center
   \includegraphics[width=150mm]{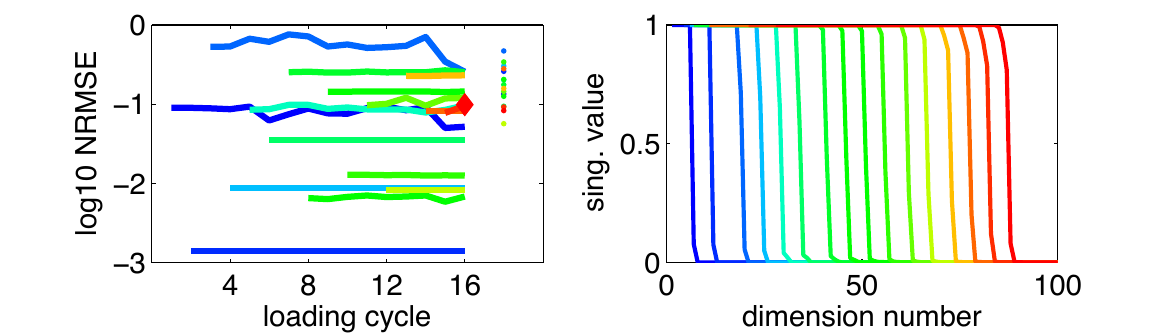}
 \caption{Incremental loading of diverse patterns, using conceptors
   with singular value spectra transformed to  approximately
   rectangular shape. Figure layout as in Figure \ref{figMemManOverview}.} 
 \label{fig16_Rev1}
 \end{figure}
 
 For a demonstration, I loaded a sequence of patterns where $p^m$ was
 integer-periodic with period 5 when $m$ was uneven and where $p^m$
 was one of the patterns from the 2-parametric family used above for
 even $m$ (simulation detail in Section \ref{secDetailMemManageArbitrary}). The original incremental loading procedure would fail, but
 when the singular values of conceptors $C^m$ were passed through a
 steep sigmoid by $\tilde{\sigma}_i = \left(\tanh(50 \, (2\,\sigma_i -
   1)) \right) / 2$, incremental loading functioned well, as revealed
 in Figure \ref{fig16_Rev1}. Interestingly, the incremental loading gave
 better recall accuracies than simultaneous loading: the mean NRMSE
 for incrementally loaded patterns, assessed after the final loading
 cycle, was 0.094, whereas the mean NRMSE for the same patterns loaded
 simultaneously was 0.19 (spending a fair effort on optimizing
 scaling and regularization parameters in the latter case). This is
 surprising and further investigations are needed to understand why/when
 incremental loading may give better results than simultaneous
 loading.

I conclude this section by remarking that when conceptors are computed
by the auto-adaptation rule introduced below in Section
\ref{secAutoCAll}, approximately rectangular singular value spectra
conducive for incremental loading
are automatically obtained.

\textbf{Note added in revision 4, November 2024:} This conceptor-based
approach to incremental learning without catastrophic forgetting has
in the meantime been adapted to deep (forward) neural networks in the
works of Xu 'Owen' He \cite{HeJaeger18,He23}.

\subsection{Example: Dynamical Pattern
  Recognition}\label{subsec:JapVow}

In this subsection I present another demonstration of the usefulness
of Boolean operations on conceptor matrices. I describe a training scheme for
a pattern recognition system which reaches (or surpasses) the
classification test performance of state-of-the-art recognizers on a
widely used benchmark task. Most high-performing existing classifiers
are trained in discriminative training schemes.  Discriminative
classifier training exploits the contrasting differences between the
pattern classes. This implies that if the repertoire of
to-be-distinguished patterns becomes extended by a new pattern, the
classifier has to be re-trained on the entire dataset, re-visiting
training data from the previous repertoire. In contrast, the system
that I present is trained in a ``pattern-local'' scheme which admits
an incremental extension of the recognizer if new pattern types were
to be included in its repertoire. Furthermore, the classifier can be
improved in its exploitation phase by incrementally incorporating
novel information contained in a newly incoming test pattern. The
key to this local-incremental classifier training is agin Boolean
operations on conceptors.

Unlike in the rest of this report, where I restrict the presentation
to stationary and potentially infinite-duration signals, the
patterns here are nonstationary and of short duration. This subsection
thus also serves as a demonstration how conceptors function with short
nonstationary patterns.

I use is the \emph{Japanese Vowels} benchmark dataset. It has been
donated by \cite{Kudoetal99} and is publicly available at the UCI
Knowledge Discovery in Databases Archive
(\url{http://kdd.ics.uci.edu/}). This dataset has been used in dozens
of articles in machine learning as a reference demonstration and thus
provides a quick first orientation about the positioning of a new
classification learning method. The dataset consists of 640 recordings
of utterances of two successive Japanese vowels /ae/ from nine male
speakers. It is grouped in a training set (30 recordings from each of
the speakers = 270 samples) and a test set (370 further recordings,
with different numbers of recordings per speaker). Each sample
utterance is given in the form of a 12-dimensional timeseries made
from the 12 LPC cepstrum coefficients. The durations of these
recordings range between 7 and 29 sampling timesteps. The task is to
classify the speakers in the test data, using the training data to
learn a classifier. Figure \ref{figJapData} (top row) gives an
impression of the original data.

\begin{figure}[htb]
\center
\includegraphics[width=140 mm]{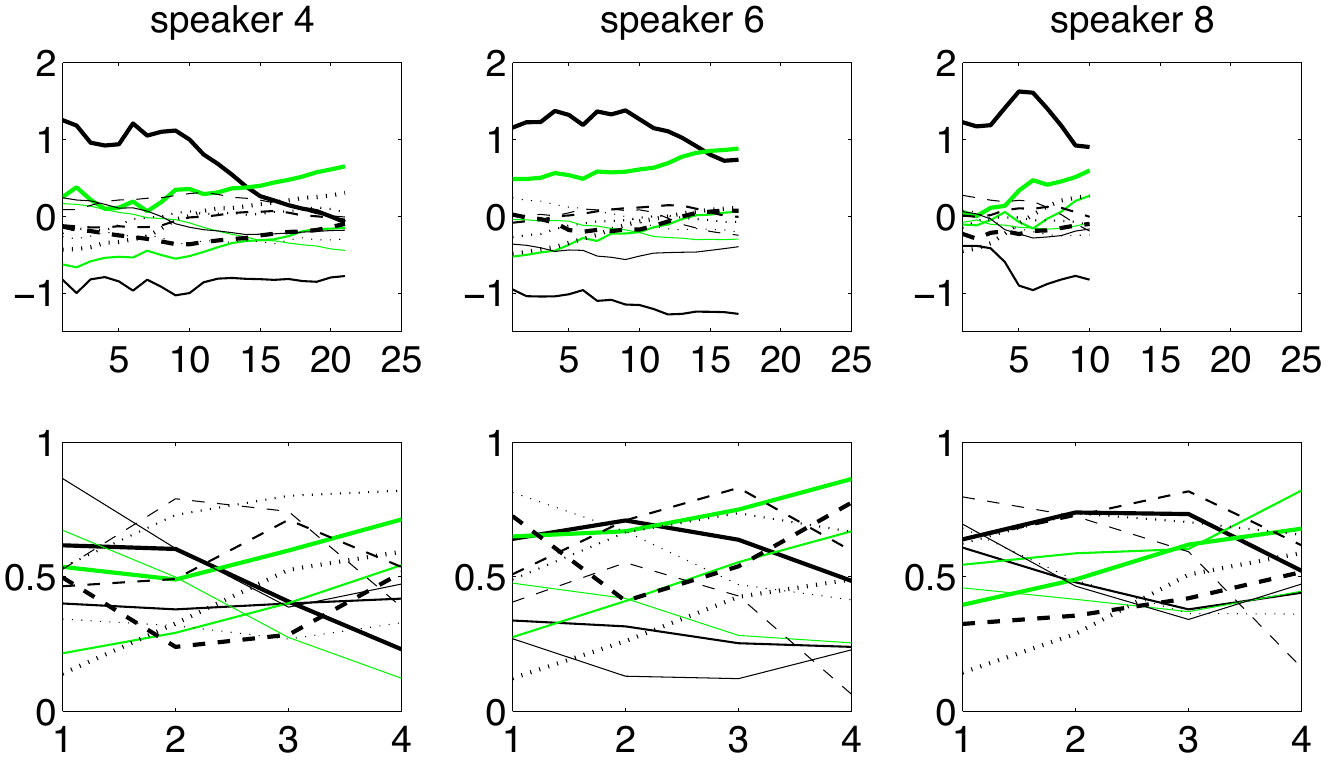}
\caption{Three exemplary utterance samples from Japanese Vowels
  dataset. Plots show values of twelve signal channels against
  discrete timesteps. Top row: raw data as provided in benchmark repository,
  bottom row: standardized format after preprocessing.}
\label{figJapData}
\end{figure}

I preprocessed the raw data  into a standardized format by (1)
shift-scaling each of the twelve channels such that per channel, the
minimum/maximum value across all training samples was 0/1; (2)
interpolating each channel trace in each sample by a cubic polynomial;
(3) subsampling these on four equidistant support points. The same
transformations were applied to the test data. Figure \ref{figJapData}
(bottom row) illustrates the normalized data format. 

The results reported in the literature for this benchmark typically
reach an error rate (percentage of misclassifications on the test set)
of about 5 -- 10 test errors (for instance,
\cite{Rodriguezetal05,Sivaramakrishnanetal07,OrsenigoVercellis10}
report from 5 -- 12 test misclassifications, all using specialized
versions of temporal support vector machines). The best result that I
am aware of outside my own earlier attempts \cite{Jaegeretal07} is
reported by \cite{Chatzis10} who reaches about 4 errors, using refined
hidden Markov models in a non-discriminative training scheme. It is
however possible to reach zero errors, albeit with an extraordinary
effort: in own work \cite{Jaegeretal07} this was robustly achieved by
combining the votes of 1,000 RNNs which were each independently
trained in a discriminative scheme.

Here I present a ``pocket-size'' conceptor-based classification
learning scheme which can be outlined as follows:

\begin{enumerate}
\item A single, small ($N = 10$ units) random reservoir network is initially created. 
\item This reservoir is driven, in nine independent sessions, with the 30
preprocessed training samples of each speaker $j$ ($j = 1,\ldots,9)$, and a
conceptor $C^+_j$ is created from the network response (no ``loading''
of patterns; the reservoir remains unchanged throughout). 
  \item In exploitation, a preprocessed sample $s$ from the test set
is fed to the reservoir and the induced reservoir states $x(n)$ are
recorded and transformed into a single vector $z$. For each conceptor
then the \emph{positive evidence} quantity $z'\,C^+_j\,z$ is computed.
This leads to a classification by deciding for $j = \mbox{argmax}_i \,
z'\,C^+_i\,z$ as the speaker of $s$. The idea behind this procedure is
that if the reservoir is driven by a signal from speaker $j$, the
resulting response $z$ signal will be located in a linear subspace of
the (transformed, see below) reservoir state space
whose overlap with the ellipsoids given by the $C^+_i$ is largest for
$i = j$.  
\item In order to further improve the classification quality, for each
speaker $j$ also a conceptor $C^{-}_j = \neg \bigvee\{C^+_1, \ldots,
  C^+_{j-1},C^+_{j+1}, \ldots, C^+_9\}$ is computed. This conceptor
can be understood as representing the event ``not any of the other
speakers''. This leads to a \emph{negative evidence} quantity
$z'\,C^-_j\,z$ which can likewise be used as a basis for
classification. 
  \item By adding the positive and negative evidences, a
\emph{combined evidence} is obtained which can be paraphrased as
``this test sample seems to be from speaker $j$ and seems not to be
from any of the others''.
\end{enumerate} 

In more detail, the procedure was implemented as
follows. A 10-unit reservoir system with 12 input units and a constant bias
term with the update equation 
\begin{equation} \label{eqJapVRNNupdate}
x(n+1) = \tanh(W\,x(n) + W^{\mbox{\scriptsize     in}}s(n) + b)
\end{equation}
was created by randomly creating the $10 \times 10$ reservoir weight
matrix $W$, the $10 \times 12$ input weight matrix
$W^{\mbox{\scriptsize in}}$ and the bias vector $b$ (full
specification in Section \ref{subsecJapVowExp}). Furthermore, a random
starting state $x_{\mbox{\scriptsize start}}$, to be used in every run
in training and testing, was created. Then, for each speaker $j$, the
conceptor $C^+_j$ was learnt from the 30 preprocessed training samples
$s_j^k(n)$ (where $j = 1,\ldots,9; \; k = 1,\ldots,30; n =
1,\ldots,4$) of this speaker, as follows:

\begin{enumerate}
\item For each training sample $s_j^k$ ($k = 1, \ldots, 30$) of
this speaker, the system (\ref{eqJapVRNNupdate}) was run with this
input, starting from $x(0) = x_{\mbox{\scriptsize  start}}$, yielding
four network states $x(1), \ldots, x(4)$. These states were
concatenated with each other and with the driver input into a $4 \cdot
(10 + 12) = 88$ dimensional vector $z_j^k = [x(1); s_j^k(1); \ldots; x(4);
s_j^k(4)]$. This vector contains the entire network response to the
input  $s_j^k$ and the input itself. 
  \item The 30 $z_j^k$ were assembled as columns into a $88 \times 30$
matrix $Z$ from which a correlation matrix $R_j = ZZ' / 30$ was
obtained. A preliminary conceptor $\tilde{C}^+_j = R_j(R_j + I)^{-1}$
was computed from $R_j$ (preliminary because in a later step the
aperture is optimized). Note that $\tilde{C}^+_j$ has size $88 \times
88$.
\end{enumerate}

After all ``positive evidence'' conceptors $\tilde{C}^+_j$ had been created,
preliminary ``negative evidence'' conceptors $\tilde{C}^-_j$ were
computed as
\begin{equation}\label{eqNegConceptor}
\tilde{C}^-_j = \neg \bigvee  \{\tilde{C}^+_1, \ldots,
  \tilde{C}^+_{j-1},\tilde{C}^+_{j+1}, \ldots, \tilde{C}^+_9\}.
\end{equation} 

An important factor for good classification performance is to find
optimal apertures for the conceptors, that is, to find aperture
adaptation factors $\gamma^+, \gamma^-$ such that final conceptors
$C^+_j = \varphi(\tilde{C}^+_j, \gamma^+), C^-_j =
\varphi(\tilde{C}^-_j, \gamma^-)$ function well for classification. A
common practice in machine learning would be to optimize $\gamma$ by
cross-validation on the training data. This, however, is expensive,
and more crucially, it would defy the purpose to design a learning
procedure which can be incrementally extended by novel pattern classes
without having to re-inspect all training data. Instead of
cross-validation I used the $\nabla$ criterion described in Section
\ref{secApAdjustGuide} to find a good aperture. Figure
\ref{figGammaOptJapVow} shows how this criterion varies with $\gamma$
for an exemplary case of a $\varphi(\tilde{C}^+_j, \gamma^+)$ sweep.
For each of the nine $\tilde{C}^+_j$, the value $\tilde{\gamma}^+_j$
which maximized $\nabla$ was numerically computed, and the mean of
these nine values was taken as the common $\gamma^+$ to get the nine
$C^+_j = \varphi(\tilde{C}^+_j, \gamma^+)$. A similar procedure was
carried out to arrive at $C^-_j = \varphi(\tilde{C}^-_j, \gamma^-)$.

\begin{figure}[htb]
\center
\includegraphics[width=80 mm]{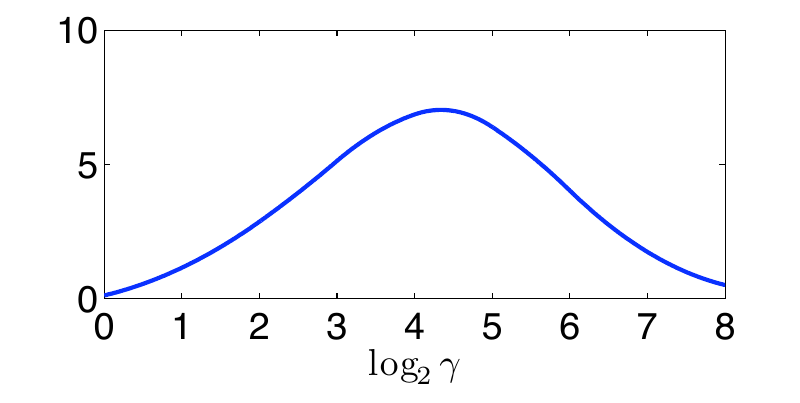}
\caption{The criterion $\nabla$ from an
  exemplary conceptor plotted
  against the log 2 of candidate aperture adaptations $\gamma$.}  
\label{figGammaOptJapVow}
\end{figure}

The conceptors $C^+_j, C^-_j$ were then used for classification as
follows. Assume $z$ is an 88-dimensional combined states-and-input
vector as described above, obtained from driving the reservoir with a
preprocessed test  sample. Three kinds of classification
hypotheses were computed, the first only based on $C^+_j$, the second
based on  $C^-_j$, and one based on a combination of both. Each
classification hypothesis is a 9-dimensional vector with ``evidences''
for the nine speakers. Call these evidence vectors $h^+, h^-, h^{+-}$
for the three kinds of classifications. The first of these was
computed by setting $\tilde{h}^+(j) = z'\, C^+_j\,z$, then normalizing
$\tilde{h}^+$ to a range of $[0,1]$ to obtain $h^+$. Similarly $h^-$
was obtained from using $z'\, C^-_j\,z$, and $h^{+-} = (h^+ + h^-) / 2$ was
simply the mean of the two former. Each hypothesis vector leads to a
classification decision by opting for the speaker $j$ corresponding to
the largest component in the hypothesis vector. 

This classification procedure was carried out for all of the 370 test
cases, giving 370 hypothesis vectors of each of the three
kinds. Figure \ref{figJapVowEvidences} gives an impression.

\begin{figure}[htb]
\center
\includegraphics[width=150 mm]{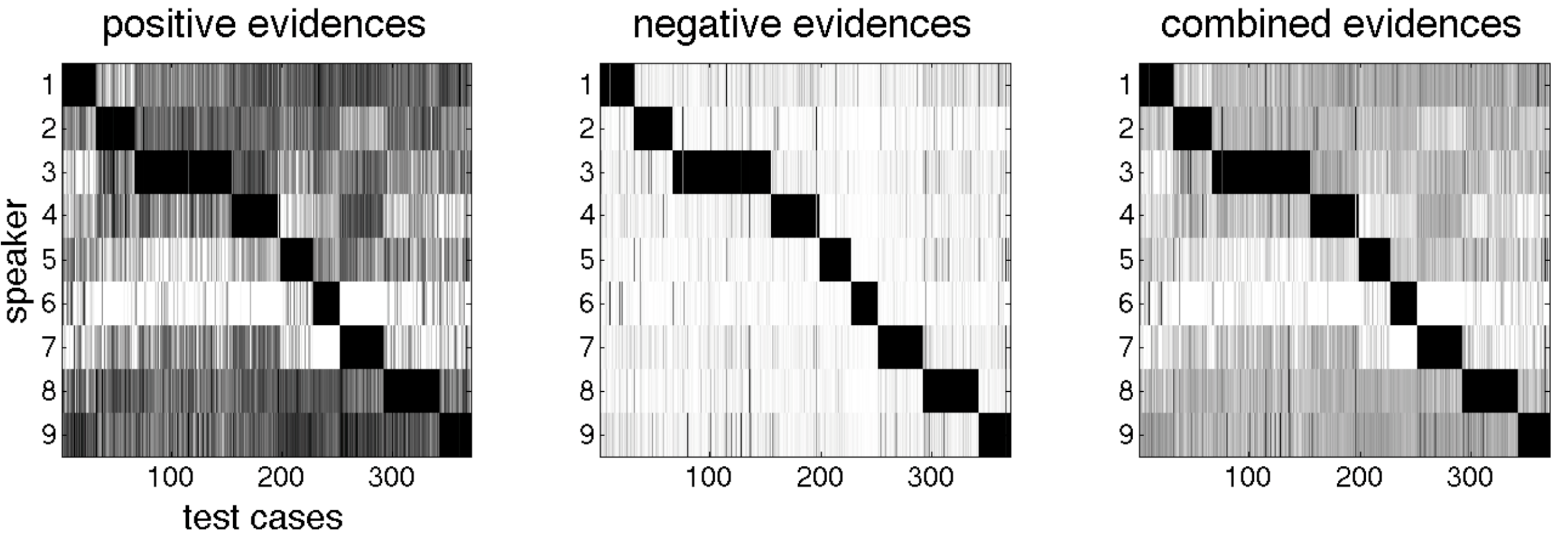}
\caption{Collected evidence vectors $h^+, h^-, h^{+-}$ obtained in a
  classification learning experiment. Grayscale coding: white = 0,
  black = 1. Each panel shows  370 evidence vectors. The (mostly)
  black segments along the diagonal correspond to the correct
  classifications (test samples were sorted by speaker). For explanation see text.}  
\label{figJapVowEvidences}
\end{figure}

\paragraph*{Results:} The outlined classification experiment was repeated 50
times with random new reservoirs. On average across the 50 trials, the
optimal apertures $\gamma^+ \; / \; \gamma^-$ were found as 25.0 /
27.0 (standard deviations 0.48 / 0.75). The
number of misclassifications for the three types of classification
(positive, negative, combined evidence) were 8.5 / 5.9 / 4.9 (standard
deviations 1.0 / 0.91 / 0.85). The training errors for the combined
classification (obtained from applying the classification procedure on
the training samples) was zero in all 50 trials. For comparison, a
carefully regularized linear classifier based on the same $z$ vectors
(detail in Section \ref{subsecJapVowExp}) reached 5.1 misclassifications across the 50 trials.

While these results are at the level of state-of-the-art classifiers
on this benchmark, this basic procedure can be refined, yielding a
significant improvement. The idea is to compute the evidence for
speaker $j$ based on a conceptor $\bar{C}^+_j$ which itself is based
on the assumption that the test sample $s$ belongs to the class $j$, that
is, the computed evidence should reflect a quantity ``if $s$ belonged
to class $j$, what evidence can we collect under this
assumption?''. Recall that $C^+_j$ is obtained from the 30
training samples through $C^+_j = R \, (R +
(\gamma^+)^{-2}I)^{-1}$, where $R = ZZ'/30$ is the correlation matrix
of the 30 training coding vectors belonging to speaker $j$. Now
add the test vector $z$ to $Z$, obtaining $\bar{Z} = [Z z], \bar{R} =
\bar{Z}\bar{Z}'/31, \bar{C}^+_j = \bar{R} (\bar{R} +
(\gamma^+)^{-2}I)^{-1}$, and use $\bar{C}^+_j$ in the procedure
outlined above instead of $C^+_j$. Note that, in application scenarios
where the original training data $Z$ are no longer available at test
time, $\bar{C}^+_j$ can be directly computed
from $C^+_j$ and $z$ through the model update formulas
(\ref{eqUpdateC1}) or (\ref{eqUpdateC2}). The negative evidence
conceptor is accordingly obtained by $\bar{C}^-_j = \neg\,\bigvee
\{\bar{C}^+_1,\ldots,\bar{C}^+_{j-1}, \bar{C}^+_{j+1}, \ldots,
\bar{C}^+_9  \}$. 

\paragraph*{Results of refined classification procedure:} Averaged over 50
learn-test trials with independently sampled reservoir weights, the
number of misclassifications for the three types of classification
(positive, negative, combined evidence) were 8.4 / 5.9 / 3.4 (standard
deviations 0.99 / 0.93 / 0.61). The training misclassification errors for the combined
classification was zero in all 50 trials.

The detection of good apertures through the $\nabla$ criterion worked
well. A manual grid search through candidate apertures found that a
minimum test misclassification rate of 3.0  (average
over the 50 trials) from the combined classificator was obtained with
an aperture $\alpha^+ = 20, \alpha^- = 24$ for both the positive and negative
conceptors. The automated aperture detection yielded apertures
$\alpha^+ = 25, \alpha^- = 27$ and a (combined classificator)
misclassification rate of 3.4, close to the optimum.

{\bf Discussion.} The following observations are worth noting.  
\begin{description}
\item[Method also applies to static pattern classification.] In
  the presented classification method, temporal input samples $s$
  (short preprocessed nonstationary timeseries) were transformed into
  static coding vectors $z$ as a basis for constructing conceptors.
  These $z$ contained the original input signal $s$ plus the state
  response from a small reservoir driven by $s$. The reservoir was
  only used to augment $s$ by some random nonlinear interaction terms
  between the entries in $s$. Conceptors were created and used in
  classification without referring back to the reservoir dynamics.
  This shows that conceptors can also be useful in \emph{static}
  pattern classification.
    \item[Extensibility.] A classification model consisting of learnt
  conceptors $C^+_j, C^-_j$ for $k$ classes can be easily
  \emph{extended by new classes}, because the computations needed for
  new $C^+_{k+1}, C^-_{k+1}$ only require positive training samples of
  the new class. Similarly, an existing model $C^+_j, C^-_j$ can be
  \emph{extended by new training samples} without re-visiting original
  training data by an application of the model extension formulae
  (\ref{eqUpdateC1}) or (\ref{eqUpdateC2}). In fact, the refined
  classification procedure given above can be seen as an ad-hoc
  conditional model extension by the test sample. 
\item[Including an ``other'' class.] Given a learnt classification
model  $C^+_j, C^-_j$ for $k$ classes it appears straightforward to
include an ``other'' class by including $C^+_{\mbox{\scriptsize other}}
= \neg \bigvee \{C^+_1,\ldots, C^+_k \}$ and recomputing the negative
evidence conceptors from the set $\{C^+_1,\ldots, C^+_k,
C^+_{\mbox{\scriptsize other}} \}$ via (\ref{eqNegConceptor}). I have
not tried this out yet. 
\item[Discriminative nature of combined classification.] The
classification of the combined type, paraphrased above as ``sample seems to
be  from class $j$ and seems not to be from any of the
others'', combines information from all classes into an evidence vote
for a candidate class $j$. Generally, in discriminative learning
schemes for classifiers, too, contrasting information \emph{between}
the classes is exploited. The difference is that in those schemes,
these differences are worked in at learning time, whereas in the
presented conceptor-based scheme they are evaluated at test time. 
\item[Benefits of Boolean operations.] The three aforementioned points
-- extensibility, ``other'' class, discriminative classification --
all hinge on the availability of the NOT and OR operations, in
particular, on the associativity of the latter. 
\item[Computational efficiency.] The computational steps involved in
learning and applying conceptors are constructive. No iterative
optimization steps are involved (except that standard implementations
of matrix inversion are iterative). This leads to short computation
times. Learning conceptors from the 270 preprocessed data samples,
including determining good apertures, took 650 ms and classifying a
test sample took 0.7 ms for the basic and 64 ms for the refined
procedure (on a dual-core 2GHz Macintosh notebook computer, using
Matlab).
\item[Competitiveness.] The test misclassification rate of 3.4 is
slightly better than the best rate of about 4 that I am aware of in
the literature outside own work \cite{Jaegeretal07}.  Given that the
zero error  performance in  \cite{Jaegeretal07} was achieved with an
exceptionally expensive model (combining 1,000 independently sampled
classifiers), which furthermore is trained in a discriminative setup
and thus is not extensible, the attained performance level, the
computational efficiency, and the extensibility of the  conceptor-base
model render it a competitive alternative to existing classification
learning methods. It remains to be seen though how it performs on
other datasets. 
  \item[Regularization by aperture adaptation?] In supervised
classification learning tasks, it is generally important to
regularize models to find the best balance between overfitting
and under-exploiting training data. It appears that the role of
regularization is here played by the aperture adaptation, though a
theoretical analysis remains to be done. 
\item[Early stage of research.] The proposed classifier learning
scheme was based on numerous ad-hoc design decisions, and quite
different ways to exploit conceptors for classification are easily
envisioned. Thus, in sum, the presented study should  be regarded as
no more than a first demonstration of the basic usefulness of conceptors for
classification tasks.
\end{description}

\subsection{Autoconceptors}\label{secAutoCAll}

\subsubsection{Motivation and Overview}\label{secAutoC}

In the preceding sections I have defined conceptors as transforms $C =
R(R + \alpha^{-2}I)^{-1}$ of reservoir state correlation matrices $R$.
In order to obtain some conceptor $C^j$ which captures a driving
pattern $p^j$, the network was driven by $p^j$ via $x(n+1) =
\tanh(W^\ast\,x(n) + W^{\mbox{\scriptsize in}}p^j(n+1) + b)$, the
obtained reservoir states were used to compute $R^j$, from which $C^j$
was computed. The conceptor $C^j$ could then later be
exploited via the conceptor-constrained update rule $x(n+1) = C^j\,
\tanh(Wx(n) + b)$ or its variant $x(n+1) = C^j\, \tanh(W^\ast\, x(n) +
D\,x(n) + b)$.

This way of using conceptors, however, requires that the conceptor
matrices $C^j$ are computed at learning time (when the original
drivers are active), and they have to be \emph{stored} for later
usage. Such a procedure  is useful and feasible in engineering or
machine learning applications, where the conceptors $C^j$ may be
written to file for later use. It is also adequate for
theoretical investigations of reservoir dynamics, and logical analyses
of relationships between reservoir dynamics induced by different
drivers, or constrained by different conceptors. 

However, storing conceptor matrices is entirely implausible from a
perspective of  neuroscience. A conceptor matrix has
the same size as the original reservoir weight matrix, that is, it is
as large  an entire network (up to
a saving factor of one half due to the symmetry of conceptor
matrices). It is hard to envision plausible models for computational
neuroscience where learning a new pattern  by some RNN
essentially would amount to creating an entire new network.  

This motivates to look for ways of how conceptors can be used for
constraining reservoir dynamics without the necessity to store
conceptors in the first place. The network would have to create
conceptors ``on the fly'' while it is performing some relevant task.
Specifically, we are interested in tasks or functionalities which are
relevant from a computational neuroscience point of view. This objective
also motivates to focus on algorithms which are not immediately
biologically implausible. In my opinion, this largely excludes
computations which explicitly exploit the SVD of a matrix (although it
has been tentatively argued that neural networks can perform principal component
analysis \cite{Oja82} using biologically observed mechanisms).

In the next subsections I  investigate a version
of conceptors with associated modes of usage where there is no
need to store conceptors and where  computations are 
online adaptive and local in the sense that the information necessary
for adapting a synaptic weight is available at the concerned
neuron. I will demonstrate the workings of these conceptors and
algorithms in two functionalities, (i) content-addressable memory
(Section \ref{secAutoCMemExample}) and
(ii) simultaneous de-noising and classification of a signal (Section \ref{secHierarchicalArchitecture}). 

In this line of modeling, the conceptors are created by the reservoir
itself at the time of usage. There is no role for an external engineer
or superordinate control algorithm to ``plug in'' a conceptor.
I will speak of \emph{autoconceptors} to distinguish these
autonomously network-generated conceptors from the conceptors that are
externally stored and externally inserted into the reservoir dynamics.
In discussions I will sometimes refer to those
``externalistic'' conceptors as \emph{alloconceptors}.

Autoconceptors, like alloconceptors, are positive semidefinite
matrices with singular values in the unit interval. The semantic
relationship to data, aperture operations, and Boolean operations are
identical for allo- and autoconceptors. However, the way how
autoconceptors are generated is different from alloconceptors, which
leads to additional constraints on their algebraic characteristics.
The set of autoconceptor matrices is a proper subset of the conceptor
matrices in general, as they were defined in Definition
\ref{defconceptor}, i.e. the class of positive semidefinite matrices
with singular values ranging in the unit interval. The additional
constraints arise from the circumstance that the reservoir states
$x(n)$ which shape an autoconceptor $C$ are themselves depending on
$C$.

The treatment of autoconceptors will be structured as follows. I will
first introduce the basic equations of autoconceptors and their
adaptation dynamics (Section \ref{secAutoBasicEqs}), demonstrate their
working in a of content-addressable memory task (Section
\ref{secAutoCMemExample}) and mathematically analyse central
properties of the adaptation dynamics (Section \ref{subsecCAD}). The
adaptation dynamics however has non-local aspects which render it
biologically implausible. In order to progress toward biologically
feasible autoconceptor mechanisms, I will propose 
neural circuits which implement autoconceptor dynamics in ways that
require only local information for synaptic weight changes (Section
\ref{secBiolPlausible}).

\subsubsection{Basic Equations}\label{secAutoBasicEqs}

 The basic system equation for autoconceptor systems is
\begin{equation}\label{eqBasicAutoC1}
x(n+1) = C(n)\,\tanh(W^\ast\,x(n) + W^{\mbox{\scriptsize in}}p(n+1) + b)
\end{equation} 
or variants thereof, like 
\begin{equation}\label{eqBasicAutoC2}
x(n+1) = C(n)\,\tanh(W\,x(n)  + b)
\end{equation}
or
\begin{equation}\label{eqBasicAutoC3}
x(n+1) = C(n)\,\tanh(W^\ast x(n) + Dx(n)  + b),
\end{equation}
the latter two for the situation after having patterns stored. The
important novel element in these equations is that $C(n)$ is time-dependent.
Its evolution will be governed by adaptation rules that I will
describe presently. $C(n)$
need not be positive semidefinite at all times; only when the
adaptation of $C(n)$ converges, the resulting $C$ matrices will have
the algebraic properties of conceptors.

One can conceive of the system (\ref{eqBasicAutoC1}) as a two-layered
neural network, where the two layers have the same number of neurons,
and where the layers are reciprocally connected by the connection
matrices $C$ and $W$ (Figure \ref{fig:autoCArchitecture}). The two
layers have states
\begin{eqnarray}
r(n+1) & = & \tanh(W^\ast z(n) + W^{\mbox{\scriptsize
    in}}p(n+1) + b) \label{eqBasicAutoC4}\\
 z(n+1) & = & C\,r(n+1). \label{eqBasicAutoC5}
\end{eqnarray} 
The $r$ layer has sigmoidal (here: $\tanh$) units
and the $z$ layer has linear ones. The customary reservoir state $x$
becomes split into two states $r$ and $z$, which can be conceived of
as states of two pools of neurons.

\begin{figure}[htbp]
\center
\includegraphics[width=70 mm]{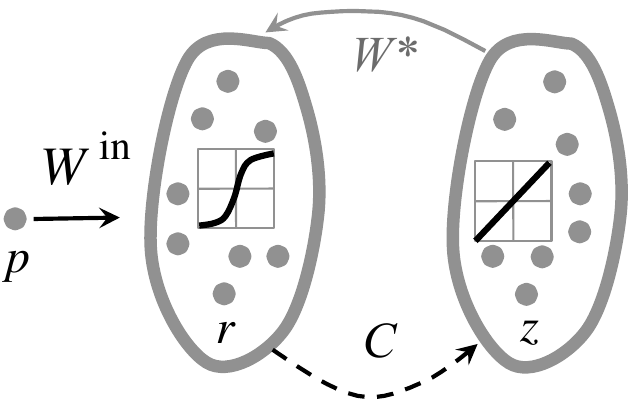}
\caption{Network representation of a basic autoconceptor system. Bias $b$
  and optional readout mechanisms are omitted. The broken arrow
  indicates that $C$
  connections are online adaptive. For explanation see text. }
\label{fig:autoCArchitecture}
\end{figure}

In order to determine an adaptation law for $C(n)$, I replicate the
line of reasoning that was employed to motivate the design of
alloconceptors in Section \ref{sec:RetrieveGeneric}. Alloconceptors
were designed to act as ``regularized identity maps'', which led to
the defining criterion (\ref{eqDefConceptor}) in Definition
\ref{defConceptor}:
\begin{displaymath}
C(R,\alpha) = \mbox{argmin}_{C}\; E[\| x - Cx \|^2] +
\alpha^{-2}\,\|C\|^2_{\mbox{\scriptsize fro}}. 
\end{displaymath}
The reservoir states $x$ that appear in this criterion resulted from
the update equation $x(n+1) = \tanh(W^\ast x(n) + W^{\mbox{\scriptsize
  in}}p(n+1) + b)$. This led to the explicit solution
(\ref{eq:CompConceptor}) stated in Proposition
\ref{propCompConceptor}:
\begin{displaymath}
C(R,\alpha) = R\,(R + \alpha^{-2}\,I)^{-1},
\end{displaymath}
where $R$ was the reservoir state correlation matrix $E[xx']$.  I 
 re-use this criterion (\ref{eqDefConceptor}), which leads
to an identical formula $C = R\,(R + \alpha^{-2}\,I)^{-1}$
for autoconceptors. The crucial difference is that now the state
correlation matrix $R$ depends on $C$:
\begin{equation}\label{eqRAuto}
R = E[zz'] = E[Cr(Cr)'] = C\,E[rr']\,C =: CQC, 
\end{equation}
where we introduce $Q = E[rr']$. This transforms the direct computation formula
(\ref{eq:CompConceptor}) to a fixed-point equation:
\begin{displaymath}
C =  CQC\,(CQC + \alpha^{-2}\,I)^{-1},
\end{displaymath}
which is equivalent to
\begin{equation}\label{eqCfixedpoint}
(C-I)CQC - \alpha^{-2}C = 0.
\end{equation}
Since $Q$ depends on $r$ states, which in turn depend on $z$ states,
which in turn depend on $C$ again, $Q$ depends on $C$ and should be
more appropriately be written as $Q_C$. Analysing the
fixed-point equation $(C-I)CQ_C C - \alpha^{-2}C = 0$ is a little
inconvenient, and I defer this to Section \ref{subsecCAD}. When one uses
autoconceptors, however, one does not need to explicitly solve
(\ref{eqCfixedpoint}). Instead, one can resort to a version of the incremental
adaptation rule (\ref{eqStochAdaptC}) from Proposition
\ref{propStochAdaptC}:
\begin{displaymath}
C(n+1) = C(n) + \lambda\, \left((z(n) - C(n)\,z(n))\,z'(n) -
  \alpha^{-2}\,C(n)\right),
\end{displaymath}
which implements a stochastic gradient descent with respect to the
cost function $E[\| z - Cz \|^2] +
\alpha^{-2}\,\|C\|^2_{\mbox{\scriptsize fro}}$. In the new sitation
given by (\ref{eqBasicAutoC1}), the state
$z$ here depends on $C$. This is, however, of no concern for
using (\ref{eqStochAdaptC}) in practice. We thus complement the
reservoir state update
rule (\ref{eqBasicAutoC1}) with the conceptor update rule
(\ref{eqStochAdaptC}) and comprise this in a definition:

\begin{definition}\label{defAutoCRNN}
An \emph{autoconceptive reservoir network} is a two-layered RNN with fixed
weights $W^\ast, W^{\mbox{\scriptsize \emph{in}}}$ and online adaptive
weights $C$, whose dynamics are given by
\begin{eqnarray}
z(n+1) &  = & C(n)\,\tanh(W^\ast\,z(n) + W^{\mbox{\scriptsize
    in}}p(n+1) + b)\label{eqAutoCRNN1}\\ 
C(n+1) & = & C(n) + \lambda\, \left((z(n) - C(n)\,z(n))\,z'(n) -
  \alpha^{-2}\,C(n)\right),\label{eqAutoCRNN2}
\end{eqnarray}
where $\lambda$ is a learning rate and $p(n)$ an input signal. Likewise,
when the update equation (\ref{eqAutoCRNN1}) is replaced by
variants of the kind (\ref{eqBasicAutoC2}) or (\ref{eqBasicAutoC3}), we will
speak of autoconceptive networks.
\end{definition}

I will  derive in Section \ref{subsecCAD} that if the driver
$p$ is stationary and if $C(n)$ converges under these rules, then the
limit $C$ is positive semidefinite with singular values in the set
$(1/2,1) \cup \{0\}$. Singular values asymptotically obtained under the evolution
(\ref{eqAutoCRNN2}) are either ``large'' (that is, greater than 1/2)
or they are zero, but they cannot be ``small'' but nonzero.  If the aperture
$\alpha$ is fixed at increasingly smaller values, increasingly many
singular values will be forced to zero. Furthermore, the analysis in
Section \ref{subsecCAD} will also reveal that among the nonzero
singular values, the majority will be close to 1. Both effects
together mean that autoconceptors are typically approximately hard
conceptors, which can be regarded as an intrinsic
mechanism of contrast enhancement, or noise suppression.

\subsubsection{Example: Autoconceptive Reservoirs as Content-Addressable
 Memories}\label{secAutoCMemExample}

 In previous sections I
demonstrated how loaded patterns can be retrieved if the associated
conceptors are plugged into the network dynamics. These conceptors
must have been stored beforehand. The actual memory functionality thus
resides in whatever mechanism is used to store the conceptors;
furthermore, a conceptor is a heavyweight object with the size of the
reservoir itself. It is biologically implausible to create and
``store'' such a network-like object for every pattern that is to be
recalled. 

In this section I describe how autoconceptor dynamics can be used to
create content-address\-able memory systems. In such systems, recall is
triggered by a cue presentation of the item that is to be recalled.
The memory system then should in some way autonomously ``lock into'' a
state or dynamics which autonomously re-creates the cue item. In the
model that will be described below, this ``locking into''   spells
out as running the reservoir in autoconceptive mode (using equations
(\ref{eqBasicAutoC3}) and (\ref{eqAutoCRNN2})),
by which process a conceptor corresponding to the cue pattern 
shapes itself and enables the reservoir to autonomously re-generate the
cue. 

The archetype of  content-addressable neural memories is the Hopfield
network \cite{Hopfield82}. In these networks, the cue is a static
pattern (technically a vector, in demonstrations often an image),
which  typically is corrupted by noise or incomplete. If the  Hopfield
network has been previously trained on the uncorrupted complete
pattern, its recurrent dynamics will converge to an attractor state which
re-creates the trained original from the corrupted cue. This
\emph{pattern completion} characteristic is the essence of the memory
functionality in Hopfield networks. In the autoconceptive model, the
aspect of completion manifests itself in that the cue is a \emph{brief}
presentation of a dynamic pattern, too short for a conceptor to be
properly adapted. After the cue is switched off, the autoconceptive
dynamics continues to shape the conceptor in an entirely autonomous
way, until it is properly developed and the reservoir re-creates the
cue. 

This autoconceptive adaptation is superficially analog to the
convergence to an attractor point in Hopfield networks. However, there
are important conceptual and mathematical differences between the two
models. I will discuss them at the end of this section.

\paragraph*{Demonstration of basic architecture.} To display the core idea
of a content-addressable memory, I ran simulations according to the
following scheme:

\begin{enumerate}
    \item {\bf Loading.} A collection of $k$ patterns $p^j$ ($j =
  1,\ldots,k$) was loaded in an $N$-dimensional reservoir, yielding an
  input simulation matrix $D$ as described in Equation
  (\ref{eqDMatUpdate}), and readout weights $W^{\mbox{\scriptsize
      out}}$, as described in Section \ref{sec:StoringGeneric}. No
  conceptors are stored.
\item {\bf Recall.} For each pattern $p^j$, a recall run was executed
which consisted of three stages:
\begin{enumerate}
\item {\bf Initial washout.} Starting from a zero network state, the reservoir
was driven with $p^j$ for $n_{\mbox{\scriptsize washout}}$ steps, in
order to obtain a task-related reservoir state.
\item {\bf Cueing.} The reservoir was continued to be driven with
$p^j$ for another $n_{\mbox{\scriptsize cue}}$ steps. During this
cueing period,  $C^j$ was adapted by using $r(n+1) =
\tanh(W r(n) + W^{\mbox{\scriptsize in}} p^j(n) + b)$, $C^j(n+1) = C^j(n) +
\lambda^{\mbox{\scriptsize cue}}\, ((r(n) - C^j(n)\,r(n))\,r'(n) - \alpha^{-2}\,C^j(n))$. At
the beginning of this period, $C^j$ was initialized to the zero
matrix. At the end of this period, a conceptor $C^{j\;\mbox{\scriptsize
  cue}}$ was obtained.
\item {\bf Autonomous recall.} The network run was continued for
another $n_{\mbox{\scriptsize recall}}$ steps in a mode
where the input was switched off and replaced by the input simulation
matrix $D$, and where the conceptor $C^{j\;\mbox{\scriptsize
  cue}}$ was further adapted 
autonomously by the autoconceptive update mechanism, via $z(n+1) =
C^j(n) \, \tanh(W z(n) + D z(n) + b)$, $C^j(n+1) = C^j(n) +
\lambda^{\mbox{\scriptsize recall}}\, ((z(n) - C^j(n)\,z(n))\,z'(n) - \alpha^{-2}\,C^j(n))$. At
the end of this period, a conceptor $C^{j\;\mbox{\scriptsize recall}}$
was available.
\end{enumerate}
  \item {\bf Measuring quality of conceptors.} The quality of the
conceptors $C^{j\;\mbox{\scriptsize cue}}$ and $C^{j\;\mbox{\scriptsize
    recall}}$ was measured by separate offline runs without conceptor
adaptation using $r(n) = \tanh(W z(n) + D z(n) + b)$; $z(n+1) =
C^{j\;\mbox{\scriptsize cue}} \, r(n)$ (or $z(n+1) =
C^{j\;\mbox{\scriptsize recall}} \, r(n)$, respectively). A
reconstructed pattern $y(n) = W^{\mbox{\scriptsize out}}r(n)$ was
obtained and its similarity with the original pattern $p^j$ was
quantified in terms of a NRMSE.
\end{enumerate}

I carried out two instances of this experiment, using two different
kinds of patterns and parametrizations:

\begin{description}
    \item[4-periodic pattern.] The patterns were random
  integer-periodic patterns of period 4, where per pattern the four
  pattern points were sampled from a uniform distribution and then
  shift-scaled such that the range became $[-1\;\; 1]$. This
  normalization implies that the patterns are drawn from an
  essentially 3-parametric family (2 real-valued parameters for fixing
  the two pattern values not equal to $-1$ or $1$; one integer
  parameter for fixing the relative temporal positioning of the $-1$
  and $1$ values). Experiment parameters: $k = 10, N = 100, \alpha =
  100, n_{\mbox{\scriptsize washout}} = 100, n_{\mbox{\scriptsize
      cue}} = 15, n_{\mbox{\scriptsize recall}} = 300,
  \gamma^{\mbox{\scriptsize cue}} = 0.02, \gamma^{\mbox{\scriptsize
      recall}} = 0.01$ (full detail in Section
  \ref{secContentAddressableExperiment}).
  \item[Mix of 2 irrational-period sines.] Two sines of period
lengths $\sqrt{30}$ and $\sqrt{30}/2$ were added with random phase
angles and random amplitudes, where however the two amplitudes were
constrained to sum to 1. This means that patterns were drawn
from a 2-parametric family. Parameters: $k = 10, N = 200, \alpha =
100, n_{\mbox{\scriptsize washout}} = 100, n_{\mbox{\scriptsize cue}}
= 30, n_{\mbox{\scriptsize recall}} = 10000, \gamma^{\mbox{\scriptsize
    cue}} = \gamma^{\mbox{\scriptsize recall}} =  0.01$. Furthermore,
during the auto-adaptation period, strong Gaussian iid noise was added to the
reservoir state before applying the $\tanh$, with a signal-to-noise
rate of 1. 
\end{description}
\begin{figure}[htbp]
\center
{\bf A.}\includegraphics[width=65mm]{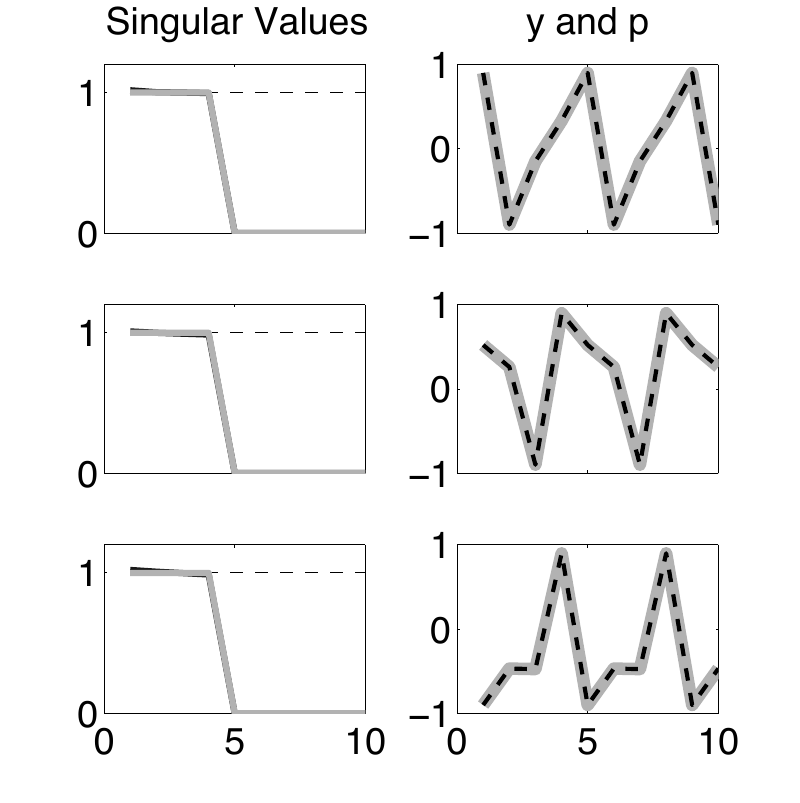}
{\bf B.}\includegraphics[width=65mm]{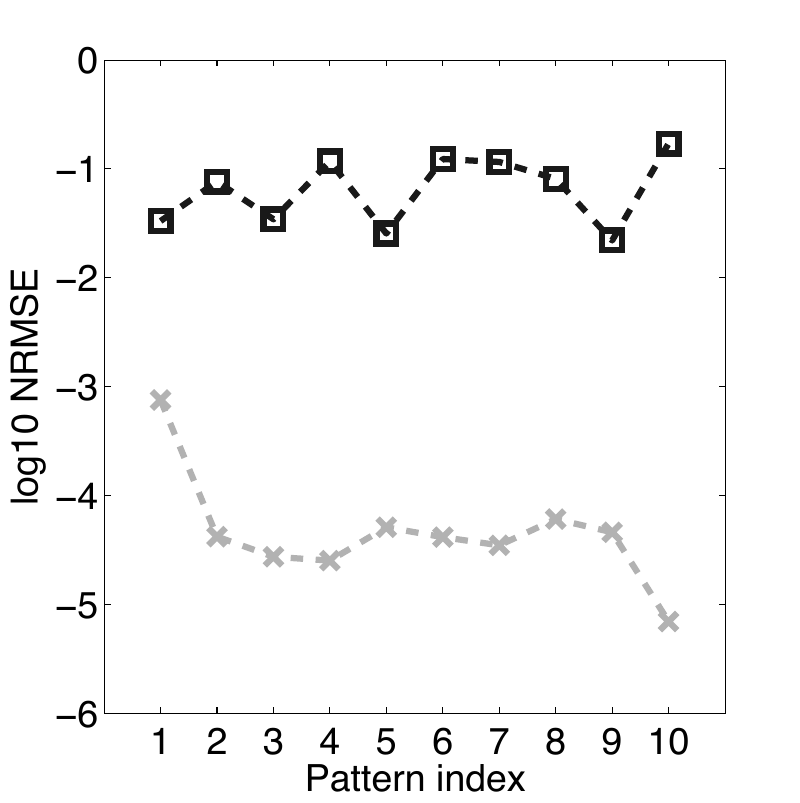}\\
{\bf C.}\includegraphics[width=65mm]{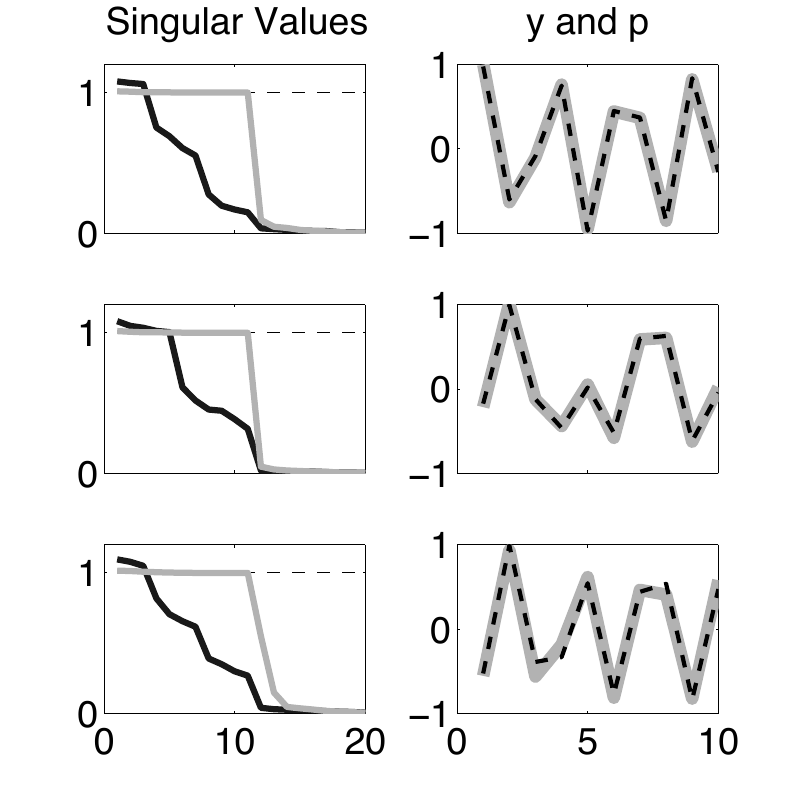}
{\bf D.}\includegraphics[width=65mm]{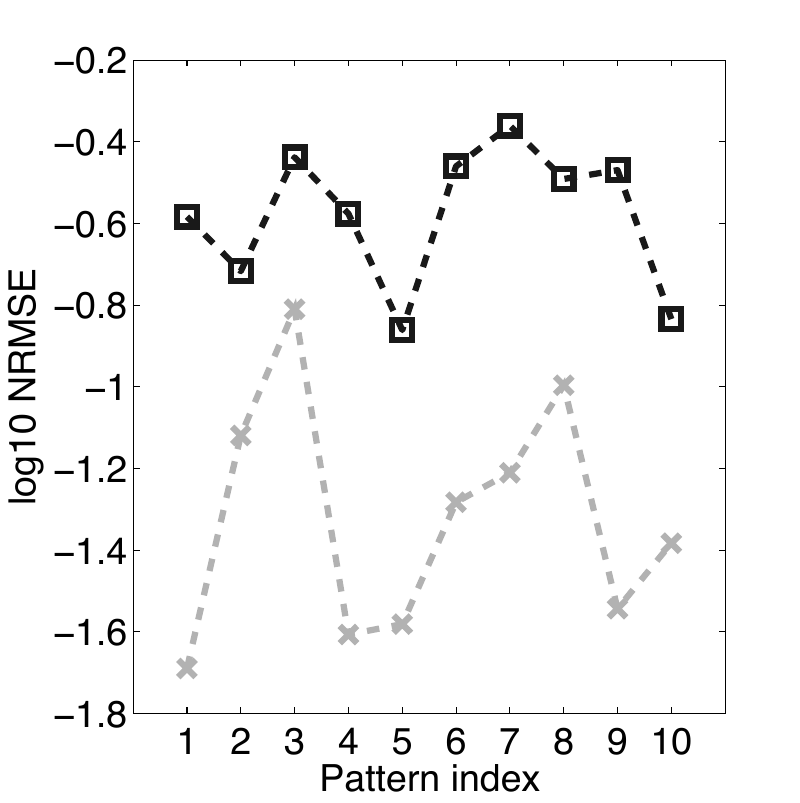}
\caption{Basic content-addressable memory demos. {\bf A}, {\bf B}:
  4-periodic pattern, {\bf C}, {\bf D}: mix of sines
  pattern.   Panels
  {\bf A}, {\bf C} show the first three of the 10 patterns. The
  singular value plots show the first 10 (20, respectively) singular values of
  $C^{j\;\mbox{\scriptsize cue}}$ (black) and $C^{j\;\mbox{\scriptsize
    recall}}$ (light gray). The ``y and p'' panels show the reconstructed
  pattern $y$ obtained with $C^{j\;\mbox{\scriptsize
    recall}}$ (bold light gray) and the original training pattern
  $p^j$ (broken black), after optimal phase-alignment. {\bf B}, {\bf
  D} plot the pattern reconstruction NRMSEs in log10 scale for the
  reconstructions obtained from $C^{j\;\mbox{\scriptsize cue}}$ (black
  squares) and from $C^{j\;\mbox{\scriptsize
    recall}}$ (gray crosses).  For
  explanation see text. }
\label{figContAdress}
\end{figure}

Figure \ref{figContAdress} illustrates the outcomes of these two
experiments. Main observations:

\begin{enumerate}
\item In all cases, the quality of the preliminary conceptor
$C^{j\;\mbox{\scriptsize cue}}$ was very much improved by the
subsequent auto-adaptation (Panels {\bf B}, {\bf
  D}), leading to an ultimate pattern reconstruction whose quality is
similar to the one that would be obtained from precomputed/stored
conceptors.  
  \item The effects of the autoconceptive adaptation are reflected in
the singular value profiles of $C^{\mbox{j\;\scriptsize cue}}$ versus
$C^{j\;\mbox{\scriptsize recall}}$ (Panels {\bf A}, {\bf C}). This is
especially well visible in the case of the sine mix patterns (for the
period-4 patterns the effect is too small to show up in the plotting
resolution). During the short cueing time, the online adaptation of
the conceptor from a zero matrix to $C^{j\;\mbox{\scriptsize cue}}$
only manages to build up a preliminary profile that could be
intuitively called ``nascent'', which  then ``matures'' in the
ensuing network-conceptor interaction during the autoconceptive recall
period. 
\item The conceptors $C^{j\;\mbox{\scriptsize recall}}$ have an almost
rectangular singular value profile. In the next section I will show
that if autoconceptive adaptation converges, singular values are
either exactly zero or greater than 0.5 (in fact, typically close to
1), in agreement with what can be seen here. Autoconceptive adaptation
has a strong tendency to lead to almost hard conceptors. 
  \item The fact that adapted autoconceptors typically have a close to
rectangular singular value spectrum renders the auto-adaptation
process quite immune against even strong state noise.  Reservoir state
noise components in directions of the nulled eigenvectors are entirely
suppressed in the conceptor-reservoir loop, and state noise components
within the nonzero conceptor eigenspace do not impede the development
of a ``clean'' rectangular profile. In fact, state  noise is even
beneficial:
it speeds up the auto-adaptation process without a noticeable loss in
final pattern reconstruction accuracy (comparative simulations not
documented here).

This noise robustness however depends on the existence of zero
singular values in the adapting autoconceptor $C$. In the simulations
reported above, such zeros were present from the start because the
conceptor was initialized as the zero matrix. If it had been
initialized differently (for instance, as identity matrix), the
auto-adaptation would only asymptotically pull (the majority of)
singular values to zero, with noise robustness only gradually
increasing to the degree that the singular value spectrum of $C$
becomes increasingly rectangular. If noise robustness is desired, it
can be reached by additional adaptation mechanisms for $C$. In
particular, it is helpful to include a thresholding mechanism: all
singular values of $C(n)$ exceeding a suitable threshold are set to 1,
all singular values dropping below a certain cutoff are zeroed (not shown).
\end{enumerate}

\paragraph*{Exploring the effects of increasing memory load -- patterns from
a parametrized family.} A central
theme in neural memory research is the capacity of a neural storage
system. In order to explore how recall accuracy depends on the memory
load, I carried out two further experiments, one for each pattern
type. Each of these experiments went along the following scheme:

\begin{enumerate}
\item Create a reservoir. 
\item In separate trials, load this reservoir with an increasing
number $k$ of patterns (ranging from $k = 2$ to $k = 200$ for the 4-period
and from $k = 2$ to $k = 100$ for the mixed sines). 
  \item After loading, repeat the recall scheme described above, with
the same parameters. Monitor the recall accuracy obtained from
$C^{j\;\mbox{\scriptsize recall}}$ for the first 10 of the loaded
patterns (if less than 10 were loaded, do it only for these). 
  \item In addition, per trial, also try to cue and ``recall'' 10 novel
patterns that were drawn randomly from the 4-periodic and mixed-sine
family, respectively, and which were not part of the collection loaded
into the reservoir. Monitor the ``recall'' accuracy of these novel
patterns as well.
\item Repeat this entire scheme 5 times, with freshly created
 patterns, but re-using always the same reservoir.
\end{enumerate}

\begin{figure}[htb]
  \center 
{\bf  A.}\includegraphics[width=65mm]{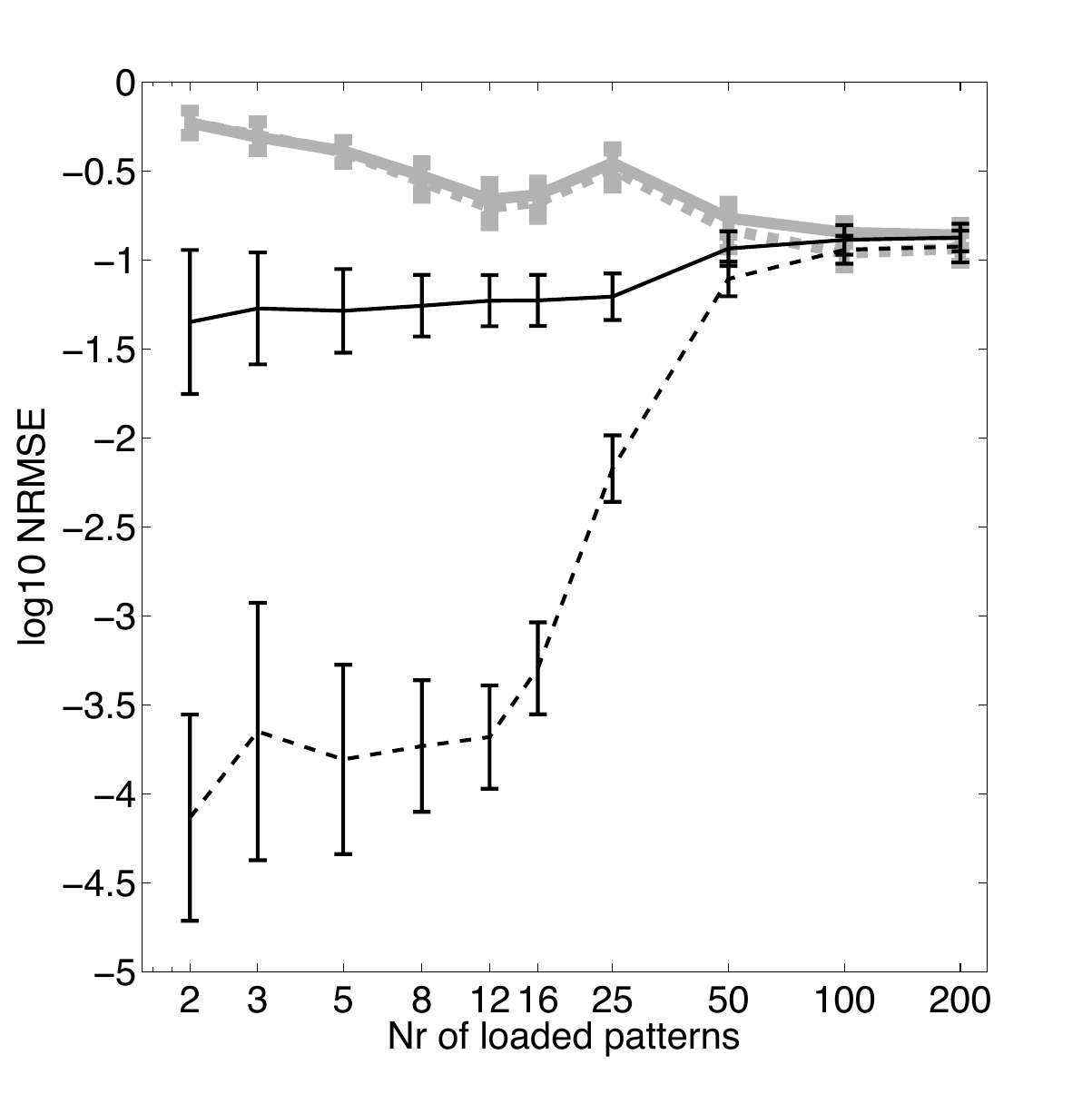}
  {\bf     B.}\includegraphics[width=65mm]{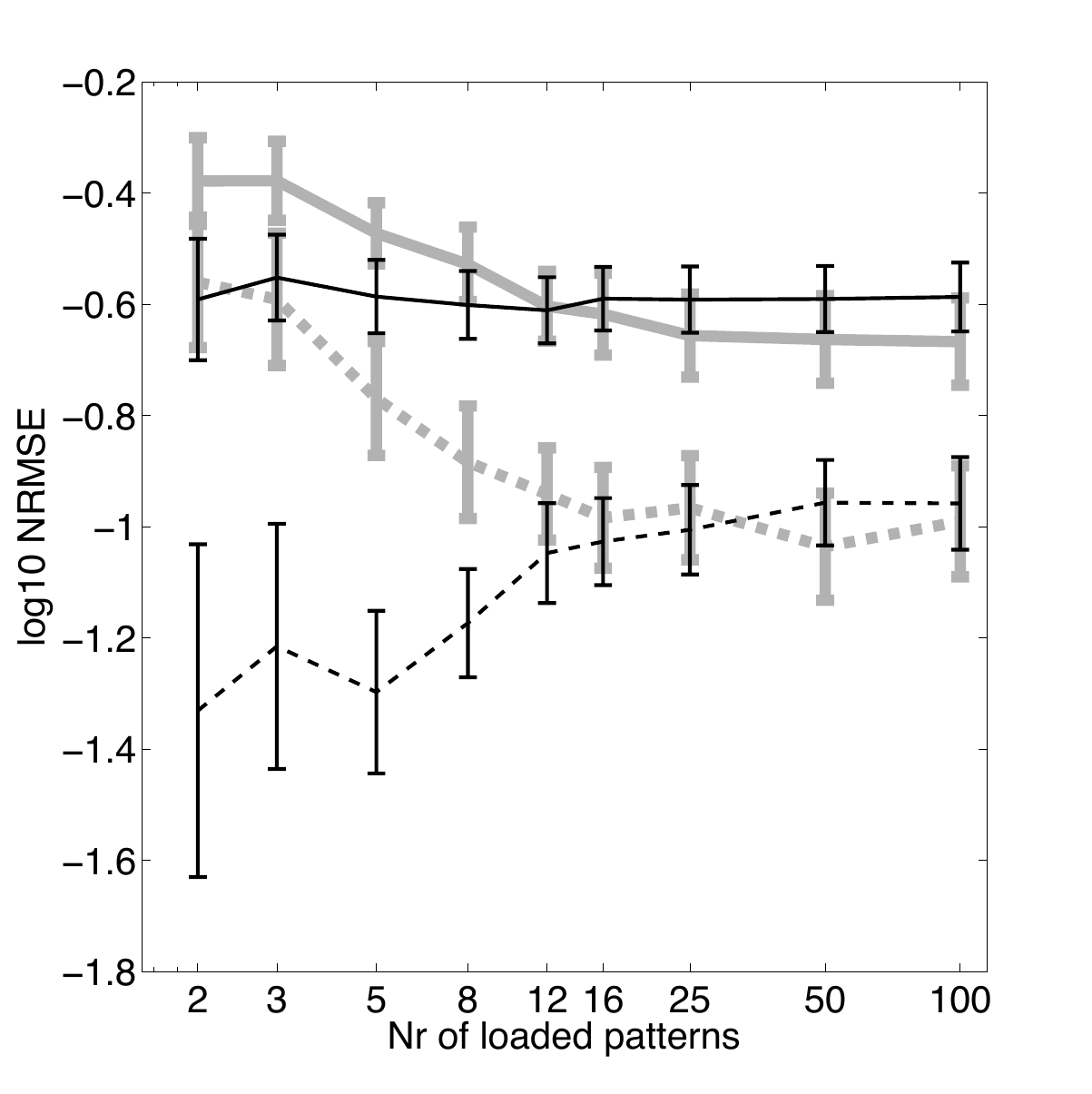}
\caption{Exploring the effects of memory load. {\bf A:} 4-periodic
    patterns, {\bf B:} mix-of-sines patterns. Each diagram shows the
    log10 NRMSE of recalling loaded patterns with
    $C^{\mbox{\scriptsize cue}}$ (black solid line) and with
    $C^{\mbox{\scriptsize recall}}$ (black broken line), as well as
    of ``recalling'' patterns not contained in the loaded set, again
    obtained 
    from $C^{\mbox{\scriptsize cue}}$ (gray solid line) and with
    $C^{\mbox{\scriptsize recall}}$ (gray broken line). Error bars
    indicate 95 \% confidence intervals. Both axes are in logarithmic
    scaling. For explanation see text. }
\label{figContAdressLoadSweep}
\end{figure}

Figure \ref{figContAdressLoadSweep} shows the results of these
experiments. The plotted curves are the summary averages
over the 10 recall targets and the 5 experiment repetitions. Each
plot point in the diagrams thus reflects an average over 50 NRMSE
values (except in cases where $k < 10$ patterns were stored; then
plotted values correspond to averages over $5k$ NRMSE values for
recalling of loaded patterns). I  list the main findings:

\begin{enumerate}
\item For all numbers of stored patterns, and for both the \emph{recall
loaded patterns} and \emph{recall novel patterns} conditions, the
autoconceptive ``maturation'' from  $C^{\mbox{\scriptsize cue}}$ to
$C^{\mbox{\scriptsize recall}}$ with an improvement of recall accuracy
is found again.
  \item The final $C^{\mbox{\scriptsize recall}}$-based recall
accuracy in the \emph{recall loaded pattern} condition has a sigmoid
shape for both pattern types. The steepest ascent of the sigmoid
(fastest deterioration of recall accuracy with increase of memory
load) occurs at about the point where the summed quota of all
$C^{\mbox{\scriptsize cue}}$ reaches the reservoir size $N$ -- the
point where the network is ``full'' according to this criterion (a
related effect was encountered in the incremental
loading study reported in Section \ref{subsec:memmanage}). When the
memory load is further increased beyond this point (one might say the
network becomes ``overloaded''), the recall accuracy does not break down
but levels out on a plateau which still translates into a recall
performance where there is a strong similarity between the target
signal and the reconstructed one.
\item In the \emph{recall novel patterns} conditions, one finds a
steady improvement of recall accuracy with increasing memory load. For
large memory loads, the accuracy in the \emph{recall novel patterns}
condition is virtually the same as in the \emph{recall loaded
  patterns} conditions. 
\end{enumerate}

Similar findings were obtained in other simulation studies (not
documented here) with other types of patterns, where in each study the
patterns were drawn from a parametrized family. 

A crucial characteristic of these experiments is that the patterns
were samples from a paramet\-rized family. They shared a family
resemblance. This mutual relatedness of patterns is exploited by the
network: for large numbers $k$ of stored patterns, the storing/recall
mechanism effectively acquires a model of the entire parametric
family, a circumstance revealed by the essentially equal recall
accuracy in the \emph{recall loaded patterns} and \emph{recall novel
  patterns} conditions. In contrast, for small $k$, the \emph{recall
  loaded patterns} condition enables a recall accuracy which is 
superior to the  \emph{recall
 novel patterns} condition: the memory system stores/recalls
individual patterns. I find this worth a special emphasis:

\begin{itemize}
    \item For small numbers of loaded patterns (before the point of
  network overloading) the system stores and recalls individual
  patterns. \emph{The input simulation matrix $D$ represents {\bf
      individual} patterns.}
\item For large numbers of loaded patterns (overloading the network),
the system learns a  representation of the parametrized pattern
family and can be cued with, and will ``recall'', any pattern from that
family.  \emph{The input simulation matrix $D$ represents the {\bf
     class} of patterns.}  
\end{itemize}

At around the point of overloading, the system, in a sense, changes
its nature from a mere storing-of-individuals device to a
learning-of-class mechanism. I call this the \emph{class
learning effect}.

\paragraph*{Effects of increasing memory load -- mutually unrelated
  patterns.}  A precondition
for the class learning effect  is that the parametric pattern family
is simple enough to become represented by the network. If the pattern
  family is too richly structured to be captured by a given network
  size, or if patterns do not have a family resemblance at all, the
  effect cannot arise. If a network is loaded with such patterns, and
  then cued with novel patterns, the ``recall'' accuracy will be on
  chance level; furthermore, as $k$ increases beyond the overloading
  region, the recall accuracy of patterns contained in the loaded
  collection will decline to chance level too. 

In order to demonstrate this, I loaded the same 100-unit reservoir
that was used in the 4-periodic pattern experiments with random
periodic patterns whose periods ranged from 3 through 9. While
technically this is still a parametric family, the number of
parameters needed to characterize a sample pattern is 8, which renders
this family far too complex for a 100-unit reservoir. Figure
\ref{figContAdressLoadSweepUnrelated} illustrates what, expectedly,
happens when one loads increasingly large numbers of such effectively
unrelated patterns.  The NRMSE for  the \emph{recall novel  patterns}
condition is about 1 throughout, which corresponds to entirely
uncorrelated pattern versus reconstruction pairs; and this NRMSE is
also approached for large $k$ in the \emph{recall loaded
  patterns} condition.

\begin{figure}[htb]
  \center 
\includegraphics[width=70mm]{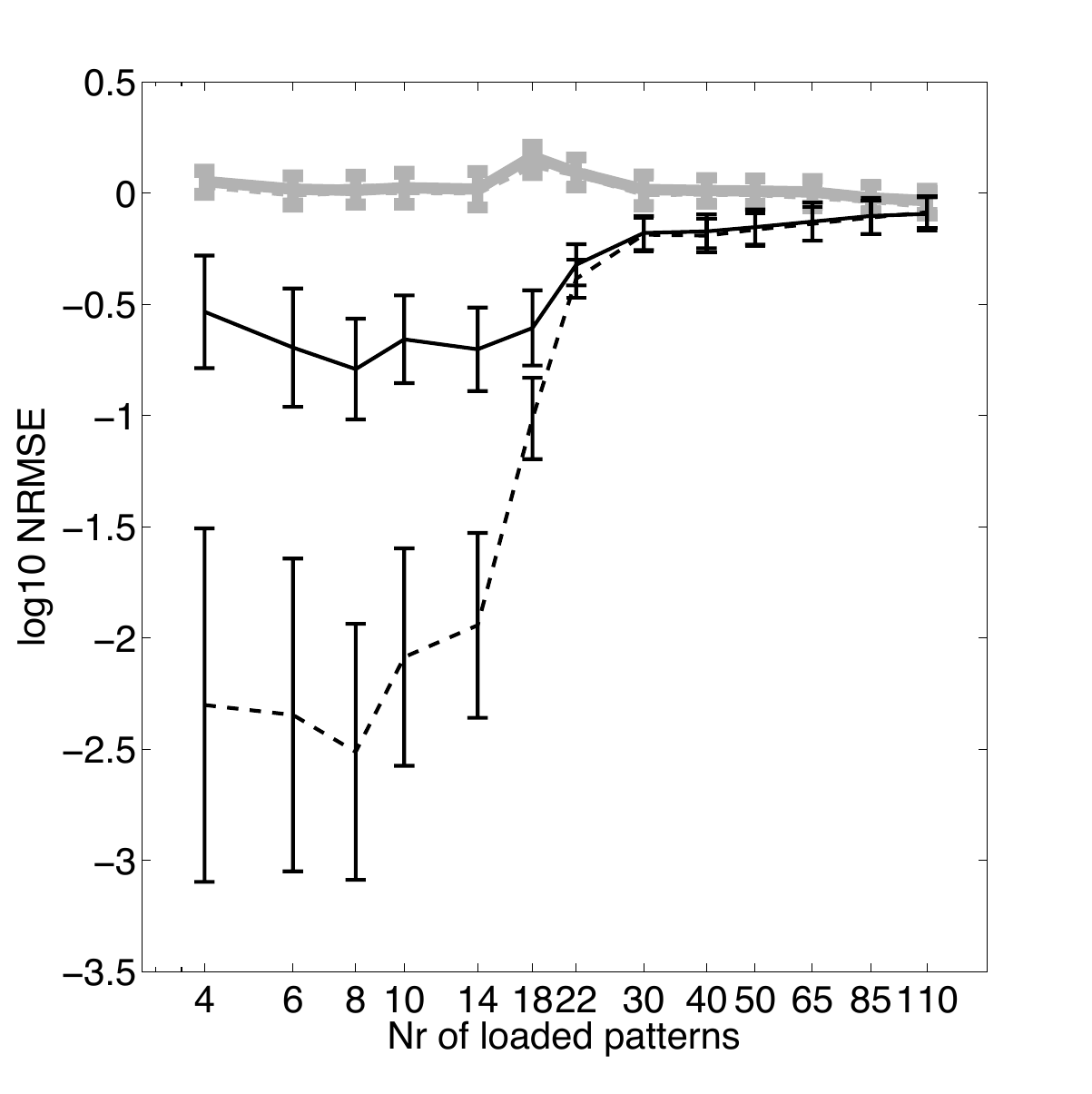}
 \caption{Effects of memory load on recall accuracy for unrelated
   patterns. Figure layout as in Figure \ref{figContAdressLoadSweep}.
   For explanation see text. }
\label{figContAdressLoadSweepUnrelated}
\end{figure}

\paragraph*{Discussion.} Neural memory mechanisms -- how to store patterns
in, and retrieve from, neural networks -- is obviously an important
topic of research. Conceptor-based mechanisms bring novel aspects to
this widely studied field. 

The paradigmatic model for content-addressable
storage of patterns in a neural network is undoubtedly the family of
auto-associative neural networks (AANNs) whose analysis and design was
pioneered by Palm \cite{Palm80} and Hopfield \cite{Hopfield82} (with a
rich history  in theoretical neuroscience, referenced
in \cite{Palm80}). Most of these models are characterized by the
following properties:

\begin{itemize}
\item AANNs with $N$ units are used to store static
patterns which are themselves $N$-dimensional vectors. The activity
profile of the entire network coincides with the very patterns. In
many demonstrations, these  patterns are rendered as 2-dimensional images.  
\item The networks are typically employed, after training, in
pattern completion or restauration tasks, where an incomplete or
distorted $N$-dimensional pattern is set as the initial
$N$-dimensional network state. The network then should evolve toward a
completed or restored pattern state. 
  \item AANNs have symmetric connections and (typically) binary
neurons. Their recurrent dynamics can be formalized as a descent along
an energy function, which leads to convergence to fixed points which
are determined by the  input pattern. 
\item An auto-associative network is trained from a set of $k$
reference patterns, where the network weights are adapted such that
the network state energy associated with each training pattern is
minimized. If successful, this leads to an energy landscape over state space
which assumes local minima at the network states that are identical to the
reference patterns.
\end{itemize}

The comprehensive and transparent mathematical theory available 
for AANNs has left a strong imprint on our preconceptions of what are
essential features of a content-addressable neural memory.
Specifically, AANN research has settled the way how the task of
storing items in an associative memory is framed in the first place:
``given $k$ reference patterns, train a network such that in
exploitation, these patterns can be reconstructed from incomplete
cues''. This leads naturally to identifying stored memory items with
attractors in the network dynamics. Importantly, memory
items are seen as \emph{discrete}, individual entities. For
convenience I will call this the ``discrete items stored as
attractors'' (DISA) paradigm. 

Beyond modeling memory functionality proper, the DISA paradigm is
historically and conceptually connected to a wide range of models
of neural representations of conceptual knowledge, where attractors
are taken as the neural representatives of discrete concepts. To name
only three kinds of such models: point attractors (cell assemblies and
bistable neurons) in the working memory literature
\cite{Durstewitzetal00}; spatiotemporal attractors in neural field
theories of cortical representation \cite{Schoeneretal95, Freeman07,
  Freeman07a}; (lobes of) chaotic attractors as richly structured
object and percept representations
\cite{YaoFreeman90,BabloyantzLourenco94}.
        
Attractors, by definition, keep the system trajectory confined within
them. Since clearly cognitive processes do not become ultimately
trapped in attractors, it has been a long-standing modeling challenge
to account for ``attractors that can be left again'' -- that is, to
partly disengage from a strict DISA paradigm. Many answers have been
proposed. Neural noise is a plausible agent to ``kick'' a trajectory
out of an attractor, but a problem with noise is its unspecificity
which is not easily reconciled with systematic information processing.
A number of alternative ``attractor-like'' phenomena have been
considered that may arise in high-dimensional nonlinear dynamics and
offer escapes from the trapping problem: \emph{saddle point dynamics}
or \emph{homoclinic cycles} \cite{Rabinovich08,GrosKaczor10};
\emph{chaotic itinerancy} \cite{Tsuda00}; \emph{attractor
  relics}, \emph{attractor ruins}, or \emph{attractor ghosts}
\cite{SussilloBarak13}; \emph{transient attractors}
\cite{Jaeger95c}; \emph{unstable attractors} \cite{Timmeetal02};
\emph{high-dimensional attractors} (initially named \emph{partial
  attractors}) \cite{MaassJoshiSontag06}; \emph{attractor landscapes}
\cite{NegrelloPasemann08}.

All of these lines of work revolve around a fundamental conundrum: on
the one hand, neural representations of conceptual entities need to
have some kind of stability -- this renders them identifiable,
noise-robust, and temporally persistent when needed. On the other
hand, there must be cognitively meaningful mechanisms for a fast
switching between neural representational states or modes. This riddle
is not yet solved in a widely accepted way. Autoconceptive plane
attractor dynamics may lead to yet another answer. This kind of
dynamics intrinsically combines dynamical stability (in directions
complementary to the plane of attraction) with dynamical neutrality
(within the plane attractor). However, in the next section we will see
that this picture, while giving a good approximation, is  too simple.

\subsubsection{Analysis of Autoconceptor Adaptation Dynamics}\label{subsecCAD}

Here I present a formal analysis of some asymptotic properties of the  conceptor
adaptation dynamics.

\subsubsection*{Problem Statement}

We consider the system of the coupled fast network state updates and slow
conceptor adaptation given by

\begin{equation}\label{som1Eq1}
z(n+1) = C(n)\; \tanh(W \, z(n))
\end{equation}

\noindent and

\begin{eqnarray}
C(n+1) & = & C(n) + \lambda \left((z(n+1) - C(n)\,z(n+1))\,z'(n+1) -
  \alpha^{-2}\,C(n)  \right)\nonumber\\
& = & C(n) + \lambda \left((I - C(n))\, z(n+1) z'(n+1) -
  \alpha^{-2}\,C(n)   \right), \label{som1Eq2}
\end{eqnarray}

\noindent where $\lambda$ is a learning rate. When $\lambda$ is small
enough, the instantaneous state correlation $z(n)z'(n)$ in
(\ref{som1Eq2}) can be replaced by its expectation under $C$ fixed at
$C(n)$, that is, we consider the dynamical system in time $k$
$$z(k+1) = C(n) \tanh(W z(k))$$ and take the expectation of $zz'$
under this dynamics, 
\begin{eqnarray*}
E_n[z z'] & := & E_k[C(n) \tanh(W\,z(k)) \tanh(W\,z(k))' C(n)'] \\
& = & C(n) \, E_k[\tanh(W\,z(k)) \tanh(W\,z(k))'] C(n)' \;\;=: \;\;C(n)Q(n)C(n)', 
\end{eqnarray*}
where $Q(n)$ is a positive semi-definite
correlation matrix. Note that  $Q(n)$ is a function
of $C$ and itself changes on the slow timescale of the $C$ adaptation.
For  further analysis it is convenient to change to continuous time
and instead of (\ref{som1Eq2}) consider
\begin{equation}\label{som1Eq3}
\dot{C}(t) = (I - C(t))\,C(t) Q(t) C'(t)  - \alpha^{-2} C(t).
\end{equation}

I now investigate the nature of potential fixed point solutions under
this dynamics. If $C$ is a fixed point of this dynamics, $Q(t)$ is
constant. In order to investigate the nature of such fixed point
solutions, we analyse solutions in $C$ for the general fixed point
equation associated with (\ref{som1Eq3}), i.e.\ solutions in $C$ of 
\begin{equation}\label{som1Eq4}
0 = (I - C)\,C Q C'  - \alpha^{-2} C,
\end{equation}
\noindent where $Q$ is some positive semidefinite matrix. We will
denote the dimension of $C$ by $N$ throughout the remainder of this
section. Let $V
D V' = Q$ be the SVD of $Q$, where $D$
is a diagonal matrix containing  the singular values of $Q$ on its
diagonal, without loss of generality in descending order. Then
(\ref{som1Eq4}) is equivalent to 
 \begin{eqnarray}
0 & = & V' ( (I - C)\,C Q C'  - \alpha^{-2} C )V \nonumber\\
& = & (I - V'CV) \, V'CV \, D (V'CV)' - \alpha^{-2} \, V'CV. \label{som1Eq5}
\end{eqnarray}

We may therefore assume that $Q$ is in descending diagonal form $D$,
analyse solutions of 
\begin{equation}\label{som1Eq6}
0 = (I - C)\,C D C'  - \alpha^{-2} C,
\end{equation}
and then transform these solutions $C$ of (\ref{som1Eq6})
back to solutions of (\ref{som1Eq4}) by $C \to VCV'$. In the
remainder we will only consider solutions of (\ref{som1Eq6}). I will
characterize the fixed points of this system and analyse their
stability properties. 

\subsubsection*{Characterizing the Fixed-point Solutions}

\paragraph*{The case $\alpha = 0$.}

In this degenerate case, neither the discrete-time update rule (\ref{som1Eq2}) nor
the  dynamical equation (\ref{som1Eq3}) is well-defined. The aperture
cannot be set to zero in practical applications where (\ref{som1Eq2})
is used for conceptor adaptation.

However, it is clear that  (i) for any
$\alpha > 0$, $C = 0$ is a fixed point solution of (\ref{som1Eq3}),
and that (ii) if we define $B(\alpha) = \sup\{\| C \| \; | \; C \mbox{ is
a fixed-point solution of (\ref{som1Eq3})} \}$, then $\lim_{\alpha \to 0}
B(\alpha) = 0$. This justifies to set, by convention, $C = 0$ as the
unique fixed point of (\ref{som1Eq3}) in the case $\alpha = 0$. In
practical applications this could be implemented by a reset mechanism:
whenever $\alpha = 0$ is set by some superordinate control mechanism,
the online adaptation (\ref{som1Eq2}) is over-ruled and $C(n)$ is
immediately set to $0$. 

\paragraph*{The case $\alpha = \infty$.}

In the case $\alpha = \infty$ (i.e., $\alpha^{-2} = 0$) our task is to characterize the solutions $C$
of 
\begin{equation}\label{som1Eq12}
0 = (I-C)CDC'.
\end{equation}

We first assume that $D$ has full rank. Fix some $k \leq N$. We proceed to
characterize rank-$k$ solutions $C$ of (\ref{som1Eq12}). $CDC'$ is
positive semidefinite and has rank $k$, thus it has an SVD $CDC' = U
\Sigma U'$ where $\Sigma$ is diagonal nonnegative and can be assumed
to be in descending order, i.e.\ its diagonal is $(\sigma_1, \ldots,
\sigma_k, 0, \ldots, 0)'$ with $\sigma_i > 0$. Any solution $C$ of
(\ref{som1Eq12}) must satisfy $U \Sigma U' = C U \Sigma U'$, or
equivalently, $\Sigma = U'CU \Sigma$. It is
easy to see that this entails that $U' C U$ is of the form
\begin{displaymath}
U' C U = \left( \begin{array}{cc} I_{k \times k} & 0 \\
0 & A \end{array}         \right)
\end{displaymath}
for some arbitrary $(n-k) \times (n-k)$ submatrix
$A$. Requesting $\mbox{rank}(C) = k$  implies $A = 0$ and hence
\begin{equation}\label{som1Eq13}
C = U \left( \begin{array}{cc} I_{k \times k} & 0 \\
0 & 0 \end{array}         \right) U'.
\end{equation}

Since conversely, if $U$ is any orthonormal matrix, a matrix $C$ of
the form given in (\ref{som1Eq13}) satisfies $C^2 = C$, any such $C$
solves (\ref{som1Eq12}). Therefore, the rank-$k$ solutions of
(\ref{som1Eq12}) are exactly the matrices of type (\ref{som1Eq13}). 

If $D$ has rank $l < N$, again we fix a desired rank $k$ for solutions
$C$. Again let $CDC' = U \Sigma U'$, with $\Sigma$ in descending
order. $\Sigma$ has a rank $m$
which satisfies $m \leq k, l$. From considering $\Sigma = U'CU \Sigma$
it follows that $U'CU$ has the form 
\begin{displaymath}
U' C U = \left( \begin{array}{cc} I_{m \times m} & 0 \\
0 & A \end{array}         \right)
\end{displaymath}
for some  $(N-m) \times (N-m)$ submatrix
$A$. Since we prescribed $C$ to have rank $k$, the rank of $A$ is
$k-m$. Let $U_{>m}$ be the $n \times (N-m)$ submatrix of $U$ made from
the columns with indices greater than $m$. We rearrange $ U
\Sigma U' = C D C'$ to 
\begin{displaymath}
\Sigma = \left( \begin{array}{cc} I_{m \times m} & 0 \\0 & A
  \end{array} \right) \, U'\, D\, U\,   \left( \begin{array}{cc} I_{m
      \times m} & 0 \\ 0 & A \end{array} \right),    
\end{displaymath} 
from which it follows (since the diagonal of $\Sigma$ is
zero at positions greater than $m$) that $A U'_{>m} D U_{>m} A' =
0$. Since the diagonal of $D$ is zero exactly on positions $> l$, this 
is equivalent to 
\begin{equation}\label{som1Eq14}
A\, (U(1:l, m+1:N))' = A\, U(m+1:N, 1:l) = 0.
\end{equation}

 We
now find that (\ref{som1Eq14}) is already sufficient to make $C =
U\, (I_{m\times m} | 0 \,/ \,0 | A)\, U'$ solve (\ref{som1Eq12}), because
a simple algebraic calculation yields
\begin{displaymath}
C D = C C D = U \,  \left(
      \begin{array}{c} (U(1:N, 1:m))' \\ 0  \end{array} \right) \, D.
\end{displaymath}

We thus have determined the rank-$k$ solutions of (\ref{som1Eq12}) to
be all matrices of form $C = U\, (I_{m\times m} | 0 \,/\, 0 | A)\,
U'$, subject to (i) $m \leq l,k$, (ii) $\mbox{rank}(A) = k - m$, (iii)
$A\, (U(1:l, m+1:N))' = 0$. Elementary considerations (omitted here)
lead to the following generative procedure to obtain all of these
matrices: 

\begin{enumerate}
\item Choose $m$ satisfying $l - N + k \leq m \leq k$. 
\item Choose a size $N \times l$ matrix $\tilde{U}'$ made from
orthonormal columns which is zero in the last $k-m$ rows (this is
possible due to the choice of $m$). 
\item Choose an arbitrary $(N-m) \times (N-m)$ matrix $A$ of SVD form $A =
V \Delta W'$ where the diagonal matrix  $\Delta$ is in ascending order
and is zero exactly on the first $N-k$ diagonal positions (hence
$\mbox{rank}(A) = k - m$). 
\item Put
\begin{displaymath}
\tilde{\tilde{U}}' = \left( \begin{array}{cc} I_{m \times m} & 0 \\ 0 & W \end{array} \right) \, \tilde{U}'.
\end{displaymath}
This  preserves orthonormality of columns, i.e.\
$\tilde{\tilde{U}}'$ is still made of orthonormal
columns. Furthermore, it holds that $(0 | A)\, \tilde{\tilde{U}}' =
0$. 
  \item Pad $\tilde{\tilde{U}}'$ by adding arbitrary $N - l$ further
orthonormal colums to the right, obtaining an $N \times N$ orthonormal
$U'$.
\item We have now obtained a rank-$k$ solution
\begin{equation}\label{som1Eq15}
C = U \, \left( \begin{array}{cc} I_{m \times m} & 0 \\ 0 & A
  \end{array} \right) \, U',
\end{equation}
\noindent where we have put $U$ to be the transpose of the matrix $U'$
that was previously constructed. 
\end{enumerate}

\paragraph*{The case $0 < \alpha < \infty$.} 

We proceed under the assumption that $\alpha < \infty$, that is,
$\infty > \alpha^{-2} > 0$.

I first show that any solution $C$ of (\ref{som1Eq6}) is a positive
semidefinite matrix. The matrix $CDC'$ is positive semidefinite and
therefore has a SVD of the form $U \Sigma U' = CDC'$, where $U$ is
orthonormal and real and $\Sigma$ is the diagonal matrix with the
singular values of $CDC'$ on its diagonal, without loss of generality
in descending order. From $(I-C) U \Sigma U' = \alpha^{-2} \,C$ it follows
that
\begin{eqnarray*}
U \Sigma U' & = & \alpha^{-2}\,C + C\,U \Sigma U' = C\,(\alpha^{-2} I + U \Sigma
U')\\
& = & C\,(U(\alpha^{-2} I + \Sigma)U').
\end{eqnarray*}

$\alpha^{-2} I + \Sigma$ and hence $U(\alpha^{-2} I + \Sigma)U'$ are nonsingular
because $\alpha^{-2} > 0$, therefore
\begin{displaymath}
C = U \Sigma U' (U(\alpha^{-2} I + \Sigma)U')^{-1} = U \Sigma (\alpha^{-2}\,I +
\Sigma)^{-1}U' =: U S U', 
\end{displaymath}
where $S =\Sigma (\alpha^{-2}\,I + \Sigma)^{-1}$ is a diagonal
matrix, and in descending order since $\Sigma$ was in descending
order. We therefore know that any solution $C$ of (\ref{som1Eq6}) is
of the form $C = U S U'$, where $U$ is the same as in $CDC' = U \Sigma
U'$. From  $S =\Sigma (\alpha^{-2}\,I + \Sigma)^{-1}$ it furthermore
follows that $s_i < 1$ for all singular values $s_i$ of $C$, that is,
$C$ is a conceptor matrix.

We now want to obtain a complete overview of all solutions $C = USU'$
of (\ref{som1Eq6}), expressed in terms of an orthonormal real matrix $U$ and
a nonnegative real diagonal matrix $S$. This amounts  to
finding the solutions in $S$ and $U$ of
\begin{equation}\label{som1Eq7}
(S - S^2) U' D U S = \alpha^{-2}\,S,
\end{equation}
subject to $S$ being nonnegative real diagonal and $U$ being real
orthonormal. Without loss of generality we furthermore may assume that
the entries in  $S$ are in descending order. 

Some observations are
immediate. First, the rank of $S$ is bounded by the rank of $D$, that
is, the number of nonzero diagonal elements in $S$ cannot exceed the
number of nonzero elements in $D$. Second, if $U, S$ is a solution,
and $S^{\ast}$ is the same as $S$ except that some nonzero elements in
$S$ are nulled, then $U, S^{\ast}$ is also a solution (to see this,
left-right multiply both sides of (\ref{som1Eq7}) with a thinned-out
identity matrix that has zeros on the diagonal positions which one
wishes to null). 

Fix some $k \leq \mbox{rank}(D)$. We want to determine all rank-$k$
solutions $U, S$, i.e.\ where $S$ has exactly $k$ nonzero elements
that appear in descending order in the first $k$ diagonal positions.
We write $S_k$ to denote diagonal real matrices of size $k \times k$
whose diagonal entries are all positive. Furthermore, we write
$U_k$ to denote any $N \times k$ matrix whose columns are real
orthonormal.

It is clear that if $S, U$ solve (\ref{som1Eq7}) and $\mbox{rank}(S)
= k$ (and $S$ is in descending order), and if $U^{\ast}$ differs from
$U$ only in the last $N-k$ columns, then also $S, U^{\ast}$ solve
(\ref{som1Eq7}). Thus, if we have all solutions $S_k, U_k$ of
\begin{equation}\label{som1Eq8}
(S_k - S_k^2) U_k' D U_k S_k = \alpha^{-2}\,S_k,
\end{equation}
then we get all rank-$k$ solutions $S,U$ to (\ref{som1Eq6})
by padding $S_k$ with $N-k$ zero rows/columns, and extending $U_k$ to
full size $N \times n$ by appending any choice of orthonormal columns
from the orthogonal complement of $U_k$. We therefore only have to
characterize the solutions  $S_k, U_k$ of (\ref{som1Eq8}), or
equivalently, of 
\begin{equation}\label{som1Eq9}
U_k' D U_k = \alpha^{-2}\,(S_k - S_k^2)^{-1}.
\end{equation}

To find such  $S_k, U_k$, we first consider solutions  $\tilde{S}_k,
U_k$ of 
\begin{equation}\label{som1Eq10}
U_k' D U_k = \tilde{S}_k,
\end{equation}
subject to $\tilde{S}_k$ being diagonal with positive
diagonal elements. For this we employ the Cauchy interlacing theorem
and its converse. I restate, in a simple special case adapted to the
needs at hand, this result from \cite{FanPall57} where it
is presented in greater generality. 

\begin{theorem}
 (Adapted from Theorem 1 in \cite{FanPall57}, see remark
of author at the end of the proof of that theorem for a justification
of the version that I render here.) Let $A, B$ be two symmetric real
matrices with $\mbox{dim}(A) = n \geq k = \mbox{dim}(B)$, and singular values
$\sigma_1,\ldots, \sigma_n$ and $\tau_1,\ldots, \tau_k$ (in descending
order). Then there exists a real $n \times k$ matrix $U$ with $U' U = I_{k
  \times k}$ and $U' A U = B$ if and only if for $i = 1,\ldots, k$ it
holds that   $\sigma_i \geq \tau_i \geq \sigma_{n-k+i}.$
\end{theorem}

This theorem implies that if $U_k, \tilde{S}_k$ is any solution of
(\ref{som1Eq10}), with $U_k$ made of $k$ orthonormal columns and
$\tilde{S}_k$ diagonal with diagonal elements $\tilde{s}_i$ (where $j
= 1,\ldots, k$, and the enumeration is in descending order), then the
latter ``interlace'' with the diagonal entries $d_1, \ldots, d_N$ of
$D$ per $d_i \geq \tilde{s}_i \geq d_{N-k+i}.$ And conversely, any
diagonal matrix $\tilde{S}_k$, whose elements interlace with the
diagonal elements of $D$, appears in a solution $U_k, \tilde{S}_k$  of
(\ref{som1Eq10}). 

Equipped with this overview of solutions to (\ref{som1Eq10}), we revert
from (\ref{som1Eq10}) to (\ref{som1Eq9}). Solving $\tilde{S}_k =
\alpha^{-2}\,(S_k - S_k^2)^{-1}$ for $S_k$ we find that the diagonal
elements $s_i$ of $S_k$ relate to the $\tilde{s}_i$ by
\begin{equation}\label{som1Eq11}
s_i = \frac{1}{2}\left(1 \pm \sqrt{1 - \frac{4
      \alpha^{-2}}{\tilde{s}_i}}\right).
\end{equation}

Since $s_i$ must be positive real and smaller than $1$, only such
solutions $\tilde{S}_k$ to (\ref{som1Eq10}) whose entries are all
greater than $4\alpha^{-2}$ yield admissible solutions to our
original problem (\ref{som1Eq9}). The interlacing condition then
teaches us that the possible rank of solutions $C$ of (\ref{som1Eq6})
is bounded from above by the number of entries in $D$  greater than
$4\alpha^{-2}$.

For each value $\tilde{s}_i > 4\alpha^{-2}$, (\ref{som1Eq11}) gives two
solutions $s_{i,1} < 1/2 < s_{i,2}$. We will show further below that
the solutions smaller than $1/2$ are unstable while the solutions greater
than $1/2$ are stable in a certain sense. 

Summarizing and adding algorithmic detail, we obtain all rank-$k$
solutions $C = USU'$ for (\ref{som1Eq6}) as follows:

\begin{enumerate}
\item Check whether $D$ has at least $k$ entries greater than
$4\alpha^{-2}$. If not, there are no rank-$k$ solutions. If yes, proceed.
\item Find a solution in $U_k, \tilde{S}_k$ of $U'_k D U_k =
\tilde{S}_k$, with  $\tilde{S}_k$ being diagonal with 
diagonal elements greater than $4\alpha^{-2}$, and interlacing with the
elements of $D$. (Note: the proof of Theorem 1 in  \cite{FanPall57} is
constructive and could be used for finding $U_k$ given
$\tilde{S}_k$.)
\item Compute $S_k$ via (\ref{som1Eq11}), choosing between the $\pm$
options at will. 
  \item Pad $U_k$ with any orthogonal complement and $S_k$ with
further zero rows and columns to full $n \times n$ sized $U, S$, to finally
obtain a rank-$k$ solution $C = USU'$ for (\ref{som1Eq6}). 
\end{enumerate}

\subsubsection*{Stability Analysis of Fixed-point Solutions}

\paragraph*{The case $\alpha = \infty$.}

Note again that $\alpha = \infty$ is the same as $\alpha^{-2} = 0$.
 We consider the time evolution of
the quantity $\| I - C \|^2$ as $C$ evolves under the
zero-$\alpha^{-2}$ version of (\ref{som1Eq3}):
\begin{equation}\label{somEq7}
\dot{C}(t) = (I - C(t))\,C(t) Q(t) C'(t).
\end{equation}

We obtain
\begin{eqnarray}
(\| I - C \|^2)^{\displaystyle \dot{ }} & = & \mbox{trace}((I - C)(I -
C'))^{\displaystyle \dot{ }} \nonumber\\
& = &  \mbox{trace}((C-C^2)QC'C' + CCQ(C' - C'^2) - (C-C^2)QC' - CQ(C'
- C'^2) )\nonumber\\
& = & 2\; \mbox{trace}((C-C^2)Q(C'^2 -C'))\nonumber\\
& = & - 2 \; \mbox{trace}((C-C^2)Q(C' - C'^2)) \leq 0, \label{somEq8}
\end{eqnarray}
where in the last line we use that the trace of a positive
semidefinite matrix is nonnegative. This finding instructs us that no
other than the identity $C = I$ can be a stable solution of
(\ref{somEq7}), in the sense that all eigenvalues of the associated
Jacobian are negative. If $Q(t)$ has full rank for all $t$, then indeed this is the
case (it is easy to show that $\| I - C(t) \|^2$ is strictly decreasing,
hence a Lyapunov function in a neighborhood of $C = I$). 

The stability characteristics of other (not full-rank) fixed points of (\ref{somEq7})
are intricate.  If one computes the eigenvalues of the Jacobian at
rank-$k$ fixed points $C$ (i.e.\ solutions of sort $C = U(I_{k\times
  k} | 0 \; / \; 0 | 0)U'$, see (\ref{som1Eq13})), where $k < N$, one
finds negative values and zeros, but no positive values. (The
computation of the Jacobian follows the pattern of the Jacobian for
the case $\alpha < \infty$, see below, but is simpler; it is omitted here). Some
of the zeros correspond to perturbation directions of $C$ which change
only the coordinate transforming matrices $U$. These perturbations are
neutrally stable in the sense of leading from one fixed point solution
to another one, and satisfy $C + \Delta = (C + \Delta)^2$. However,
other perturbations $C + \Delta$ with the property that $C + \Delta
\neq (C + \Delta)^2$ lead to $(\| I - C \|^2)^{\textstyle \dot{ }} <
0$. After such a perturbation, the matrix $C + \Delta$ will evolve
toward $I$ in the Frobenius norm.  Since the Jacobian of $C$ has no
positive eigenvalues, this instability is non-hyperbolic. In
simulations one accordingly finds that after a small perturbation
$\Delta$ is added, the divergence away from $C$ is initially extremely
slow, and prone to be numerically misjudged to be zero.

For rank-deficient $Q$, which leads to fixed points of sort $C =
U(I_{m\times m} | 0 \; / \; 0 | A)U'$, the computation of Jacobians
becomes involved (mainly because $A$ may be non-symmetric) and
I did not construct them. In our context, where $Q$ derives from a
random RNN, $Q$ can be expected to have full rank, so a detailed
investigation of the rank-deficient case would be an academic exercise.

\paragraph*{The case $0 < \alpha < \infty$.}
This is the case of greatest practical relevance, and I spent a
considerable effort on elucidating it. 

Note that $\alpha < \infty$ is equivalent to $\alpha^{-2} > 0$.
Let $C_0 = U S U'$ be a rank-$k$ fixed point of $\dot{C} = (I - C) C D
C' - \alpha^{-2} C$, where $\alpha^{-2} > 0$, $U S U'$ is the SVD of
$C$ and without loss of generality the singular values $s_1, \ldots,
s_N$ in the diagonal matrix $S$ are in descending order, with $s_1,
\ldots, s_k > 0$ and $s_i = 0$ for $i > k$ (where $1 \leq k \leq N$).
In order to understand the stability properties of the dynamics
$\dot{C}$ in a neighborhood of $C_0$, we compute the eigenvalues of
the Jacobian $J_C = \partial \dot{C} / \partial C$ at point $C_0$.
Notice that $C$ is an $N \times N$ matrix whose entries must be
rearranged into a vector of size $N^2 \times 1$ in order to arrive at
the customary representation of a Jacobian. $J_C$ is thus an $N^2
\times N^2$ matrix which should be more correctly written as
$J_C(\mu,\nu) = \partial \,vec\,\dot{C}(\mu) / \partial
\,vec\,C(\nu)$, where $vec$ is the rearrangement operator ($1 \leq
\mu, \nu \leq N^2$ are the indices of the matrix $J_C$). Details are
given in Section \ref{secProofprop1}
within the proof of the following central proposition: 

\begin{proposition}\label{prop1}
The Jacobian $J_C(\mu,\nu) = \partial \,vec\,\dot{C}(\mu) /
\partial \,vec\,C(\nu)$ of a rank-$k$ fixed point of (\ref{som1Eq3})
has the following multiset of eigenvalues: 
\begin{enumerate}
\item $k(N-k)$ instances of 0,
\item $N(N-k)$ instances of $-\alpha^{-2}$, 
\item $k$ eigenvalues $\alpha^{-2} (1-2s_l)/(1-s_l)$, where $l = 1,\ldots,
k$,
\item $k(k-1)$ eigenvalues which come in pairs of the form 
\begin{displaymath}
\lambda_{1,2} = \frac{\alpha^{-2}}{2} \left( \frac{s_l}{s_m - 1}  +
  \frac{s_m}{s_l - 1} \pm \sqrt{ \left( \frac{s_l}{s_m - 1}  -
      \frac{s_m}{s_l - 1} \right)^2 + 4}  \right), 
\end{displaymath}
where $m < l \leq k$.
\end{enumerate}
\end{proposition}

An inspection of sort \emph{3.}\ eigenvalues reveals that whenever one of
the $s_l$ is smaller than $1/2$, this eigenvalue is positive and hence
the fixed point $C_0$ is unstable. 

If some $s_l$ is exactly equal to $1/2$, one obtains additional zero
eigenvalues by  \emph{3.} I will exclude such cases in the following
discussion, considering them to be non-generic. 

If all $s_l$ are greater than $1/2$, it is
straightforward to show that the values of sorts \emph{3.}\ and
\emph{4.}\ are negative. Altogether, $J_P$ thus has $k(N-k)$ times the
eigenvalue $0$ and otherwise negative ones.  I will call such
solutions \emph{1/2-generic}. All solutions that one will effectively
obtain when conceptor auto-adaptation converges are of this kind.

This characterization of the eigenvalue spectrum of 1/2-generic solutions
does not yet allow us to draw firm conclusions about how such a
solution  will react to perturbations. There are two reasons
why Proposition \ref{prop1}  affords but a partial insight in the stability
of 1/2-generic solutions. (A) The directions connected to zero
eigenvalues span a  $k(N-k)$-dimensional center manifold whose
dynamics remains un-analysed. It may be stable, unstable, or neutral. 
(B) When a 1/2-generic solution is perturbed, the matrix $D$ which
reflects the conceptor-reservoir interaction will change: $D$ is in
fact a function of $C$ and should be more correctly written $D = D(C)$. In our
linearization around fixed point solutions we implicitly considered
$D$ to be constant. It is unclear whether a full treatment using $D =
D(C)$ would lead to a different qualitative picture. Furthermore,  (A)
and (B) are liable to combine their effects. This is especially
relevant for the dynamics on the center manifold, because its
qualitative dynamics
is  determined by components from higher-order approximations to
(\ref{som1Eq3}) which are more susceptible to become qualitatively
changed by non-constant $D$ than the dynamical components of
(\ref{som1Eq3})  orthogonal to the center manifold. 

Taking (A) and (B) into account, I now outline a hypothetical picture
of the dynamics of (\ref{som1Eq3}) in the vicinity of 1/2-generic
fixed-point solutions. This picture is based only on plausibility
considerations, but it  is in agreement with what I observe in
simulations.

First, a dimensional argument
sheds more light on the nature of the dynamics in the
$k(N-k)$-dimensional center manifold. 
Consider a 1/2-generic rank-$k$ solution $C = USU'$ of (\ref{som1Eq7}). Recall that
the singular values $s_i$ in $S$ were derived from $\tilde{s}_i$ which interlace with the
diagonal elements $d_1, \ldots, d_N$ of $D$ by $d_i \geq \tilde{s}_i
\geq d_{N-k+i}$, and where $U'_k D U_k = \tilde{S}_k$ (Equation
(\ref{som1Eq10})). I call $C$ a \emph{1/2\&interlacing-generic} solution if the
interlacing is proper, i.e.\ if $d_i > \tilde{s}_i
> d_{N-k+i}$. Assume furthermore that $D(C)$ is constant in a
neighborhood of $C$. In this case, differential changes to $U_k$ in
(\ref{som1Eq10}) will lead to differential changes in $\tilde{S}_k$. If
these changes to $U_k$ respect the conditions (i) that $U_k$ remains
orthonormal and (ii) that $\tilde{S}_k$ remains diagonal, the changes to
$U_k$ lead to new fixed point solutions. The first constraint (i)
allows us to change $U_k$ with $(N-1)+(N-2)+\ldots+(N-k) = kN -
k(k+1)/2$ degrees of freedom. The second constraint (ii) reduces this
by $(k-1)k/2$ degrees of freedom. Altogether we have $kN -
k(k+1)/2 - (k-1)k/2 = k(N-k)$  differential directions of change of
$C$ that lead to new fixed points. This coincides with the dimension
of the center manifold associated with $C$. We can conclude that the
center manifold of a 1/2\&interlacing-generic $C$ extends exactly in the directions of
neighboring fixed point solutions. This picture is based however on the
assumption of constant $D$. If the dependency of $D$ on $C$ is
included in the picture, we would not expect to find any other fixed
point solutions at all in a small enough neighborhood of
$C$. Generically, fixed point solutions of an ODE are isolated.
Therefore, in the light of the considerations made so far, we would
expect to find isolated fixed point solutions $C$, corresponding to close
approximations of stored patterns. In a local vicinity of such
solutions, the autoconceptor adaptation would presumably progress on
two timescales: a fast convergence toward the center manifold
$\mathcal{K}$ associated with the fixed point $C$, superimposed on a
slow convergence toward $C$ within $\mathcal{K}$ (Figure
\ref{figAutoAdaptCenterManifoldsketch} {\bf A}).

\begin{figure}[htb]
\center
\includegraphics[width=130mm]{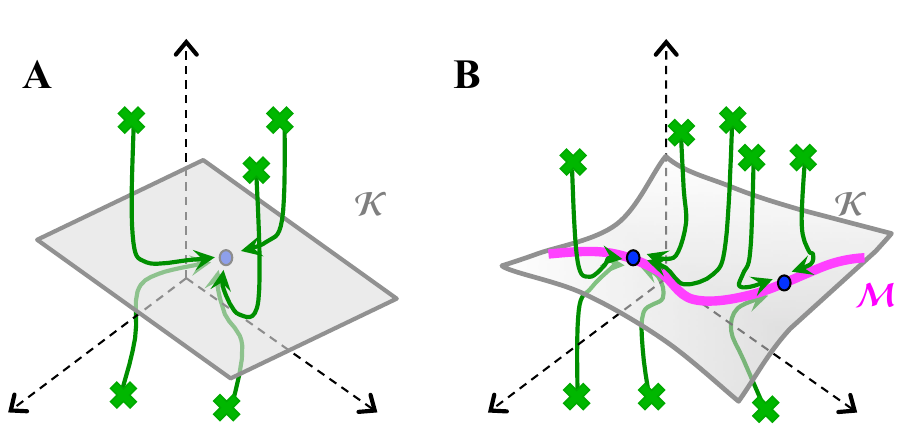}
\caption{Hypothetical phase portraits  of $C$ autoadaptation in the
  parameter space of $C$ (schematic). Blue points show stable fixed
  point solutions $C$. The gray plane in {\bf A} represents the center
  manifold $\mathcal{K}$ and in {\bf B} the merged center manifolds of
  neighboring fixed point $C$.  Green arrows represent sample
  trajectories of $C$ adaptation. Green crosses mark the starting
  points of the adaptation trajectories set by the cueing procedure.
  {\bf A.} When a small number of patterns has been loaded, individual
  stable fixed point conceptors $C$ are created.  {\bf B.} In the case
  of learning a $d$-parametric pattern class, fixed point solutions
  $C_i$ become located within a $d$-dimensional pattern manifold
  $\mathcal{M}$ (bold magenta line). For explanation see text. }
\label{figAutoAdaptCenterManifoldsketch}
\end{figure}

The situation becomes particularly interesting when many patterns from
a $d$-parametric class have been stored.  
Taking into account what the stability analysis above has revealed
about center manifolds of fixed points $C$,  I  propose the
following picture as a working hypothesis for the geometry of
conceptor adaptation dynamics that arises when a $d$-parametric
pattern class has been stored by overloading:
\begin{itemize}
\item The storing procedure leads to a number of  stable fixed point
solutions $C_i$ for the autoconceptor adaptation (blue dots in Figure
\ref{figAutoAdaptCenterManifoldsketch} {\bf B}). These $C_i$ are
associated with patterns from the pattern family, but need not
coincide with the sample patterns that were loaded.
\item The $k(N-k)$-dimensional center manifolds of the $C_i$ merge
into a comprehensive manifold $\mathcal{K}$ of the same dimension. In
the vicinity of  $\mathcal{K}$, the autoadaptive $C$ evolution leads
to a convergence toward   $\mathcal{K}$.  
\item Within $\mathcal{K}$ a $d$-dimensional submanifold $\mathcal{M}$
is embedded, representing the learnt class of patterns. Notice that we
would typically expect  $d << k(N-k)$ (examples in the previous section had $d =
2$ or $d = 3$, but $k(N-k)$ in the order of several 100). Conceptor
matrices located on $\mathcal{M}$ correspond to patterns from the
learnt class.
\item The convergence of $C$ adaptation trajectories toward
$\mathcal{K}$ is superimposed with a slower contractive dynamics
within $\mathcal{K}$ toward the class submanifold $\mathcal{M}$.  
\item The combined effects of the attraction toward $\mathcal{K}$ and
furthermore toward $\mathcal{M}$ appear in simulations as if
$\mathcal{M}$ were acting as a plane attractor. 
\item On an even slower timescale, within $\mathcal{M}$ there is an
attraction toward the isolated fixed point solutions $C_i$. This
timescale is so slow that the motion within $\mathcal{M}$  toward the
fixed points $C_i$ will be hardly observed in simulations. 
\end{itemize} 

In order to corroborate this refined picture, and especially to
confirm the last point from the list above, I carried out a long-duration
content-addressable memory simulation along the lines described in
Section \ref{secAutoCMemExample}. Ten 5-periodic patterns were loaded into
a small (50 units) reservoir. These patterns represented ten stages of
a linear morph between two similar patterns $p^1$ and $p^{10}$,
resulting in a morph sequence $p^1, p^2, \ldots, p^{10}$ where $p^i =
(1- (i-1)/9)\,p^1 + ((i-1)/9)\,p^{10}$, thus representing instances
from a 1-parametric family. Considering what was found in
Section \ref{secAutoCMemExample}, loading these ten patterns should
enable the system to re-generate by auto-adaptation any linear morph
$p_{\mbox{\scriptsize test}}$ between $p^1$ and $p^{10}$ after being
cued with $p_{\mbox{\scriptsize test}}$.

\begin{figure}[htb]
\center
\includegraphics[width=145mm]{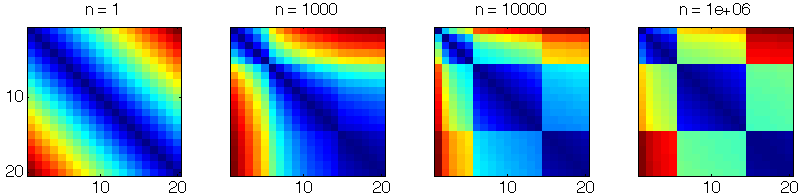}
\caption{Numerical exploration of fixed point solutions under $C$
  auto-adaptation. Each panel shows pairwise distances of 20 conceptor
  matrices obtained after $n$ auto-adaptation steps, after being cued
  along a 20-step morph sequence of cue signals. Color coding: 
  blue -- zero distance; red -- maximum distance. For explanation
  see text. }
\label{figAutoAdaptCenterManifold}
\end{figure} 

After loading, the system was cued with 20 different cues. In each of
these $j = 1,\ldots,20$ conditions, the cueing pattern
$p_{\mbox{\scriptsize test}}^j$ was the $j$-th linear interpolation
between the stored $p^1$ and $p^{10}$. The cueing was done for 20
steps, following the procedure given at the beginning of Section
\ref{secAutoCMemExample}. 
At the end of the cueing, the system will be securely driven
into a state $z$ that is very accurately connected to re-generating the
pattern $p_{\mbox{\scriptsize test}}^j$, and the conceptor matrix that
has developed by the end of the cueing would enable
the system to re-generate  a close simile of
$p_{\mbox{\scriptsize test}}^j$ (a post-cue log10 NRMSE of about $-2.7$ was
obtained in this simulation). 

After  cueing, the system was left running in conceptor
auto-adaptation mode using (\ref{eqAutoCRNN2}) for 1 Mio timesteps,
with an adaptation rate of $\lambda = 0.01$. 

At times $n = 1, 1000, 10000, 1e^6$ the situation of convergence was
assessed as follows. The pairwise distances between the current twenty
autoconceptors $C^j(n)$ were compared, resulting in a $20 \times 20$
distance matrix $D(n) = (\|C^k(n) - C^l(n) \|_{\mbox{\scriptsize
    fro}})_{k,l = 1,\ldots,20}$. Figure
\ref{figAutoAdaptCenterManifold} shows color plots of these distance
matrices. The outcome: at the beginning of autoadaptation ($n = 1$),
the 20 autoconceptors are spaced almost equally widely from each
other. In terms of the schematic in Figure
\ref{figAutoAdaptCenterManifoldsketch} {\bf B}, they would all be
almost equi-distantly lined up close to $\mathcal{M}$. Then, as the
adaptation time $n$ grows, they contract toward three point attractors
within $\mathcal{M}$ (which would correspond to a version of
\ref{figAutoAdaptCenterManifoldsketch} {\bf B} with three blue
dots). These three point attractors correspond to the three dark blue
squares on the diagonal of the last distance matrix shown in Figure
\ref{figAutoAdaptCenterManifold}.

This singular simulation cannot, of course, provide conclusive
evidence that the qualitative picture proposed in Figure
\ref{figAutoAdaptCenterManifoldsketch} is correct. A rigorous
mathematical characterization of the hypothetical manifold
$\mathcal{M}$ and its relation to the center manifolds of fixed point
solutions of the adaptation dynamics needs to be worked out. 

Plane attractors have been proposed as models for a number of
biological neural adaptation processes (summarized in
\cite{Eliasmith05}). A classical example is gaze direction control.
The fact that animals can fix their gaze in arbitrary (continuously
many) directions has been modelled by plane attractors in the
oculomotoric neural control system. Each gaze direction corresponds to
a (controlled) constant neural activation profile. In contrast to and
beyond such models, conceptor auto-adaptation organized along a
manifold $\mathcal{M}$ leads not to a continuum of \emph{constant}
neural activity profiles, but explains how a continuum of
\emph{dynamical} patterns connected by continuous morphs can be
generated and controlled.

In sum, the first steps toward an analysis of autoconceptor adaptation
have revealed that this adaptation dynamics is more involved than
either the classical fixed-point dynamics in autoassociative memories
or the plane attractor models suggested in computational
neuroscience. For small numbers of stored patterns, the picture bears
some analogies with autoassociative memories in that stable fixed
points of the autonomous adaptation correspond to stored patterns. For
larger numbers of stored patterns (class learning), the plane
attractor metaphor captures essential aspects of phenomena seen in
simulations of not too long duration.

\subsection{Toward Biologically Plausible Neural Circuits: Random
  Feature Conceptors}\label{secBiolPlausible}

The autoconceptive update equations
 \begin{eqnarray*}
 z(n+1) & = & C(n) \, \tanh(Wz(n) + Dz(n) + b)\\
 C(n+1) & = & C(n) + \lambda \, \left((z(n) - C(n) z(n)) \, z'(n) - \alpha^{-2}C(n)\right)
 \end{eqnarray*}
 could hardly be realized in biological neural systems. One problem is that
 the $C$ update needs to evaluate $C(n) z(n)$, but $z(n)$ is not an
 \emph{input} to $C(n)$ in the $z$ update but the \emph{outcome} of
 applying $C$. The input to $C$ is instead the state $r(n) =
 \tanh(Wz(n) + Dz(n) + b)$. In order to have both computations carried
 out by the same $C$, it seems that biologically hardly feasible
 schemes of 
 installing two weight-sharing copies of $C$ would be required.
 Another problem is that the update of $C$ is nonlocal: the
 information needed for updating a ``synapse'' $C_{ij}$ (that is, an
 element of $C$) is not entirely contained in the presynaptic or
 postsynaptic signals available at this synapse.

 Here I propose an architecture which solves these problems, and which
 I think has a natural biological ``feel''. The basic idea is to (i)
 randomly expand the reservoir state $r$ into a (much)
 higher-dimensional \emph{random feature space}, (ii) carry out the
 conceptor operations in that random feature space, but in a
 simplified version that only uses scalar operations on individual
 state components, and (iii) project the conceptor-modulated
 high-dimensional feature space state back to the reservoir by another
 random projection.  The reservoir-conceptor loop is replaced by a
 two-stage loop, which first leads from the reservoir to the feature
 space (through connection weight vectors $f_i$, collected column-wise
 in a random neural projection matrix $F$), and then back to the
 reservoir through a likewise random set of backprojection weights $G$
 (Figure \ref{figBiolArch} {\bf A}). The reservoir-internal connection
 weights $W$ are replaced by the combination of $F$ and $G$, and the
 original reservoir state $x$ known from the basic matrix conceptor
 framework is split into a reservoir state vector $r$ and a feature
 space state vector $z$ with components $z_i = c_i f'_i r$. The
 conception weights $c_i$ take over the role of conceptors. In full
 detail,

\begin{enumerate}
\item expand the $N$-dimensional reservoir state $r = \tanh(Wz +
W^{\mbox{\scriptsize in}}p + b)$ into the
$M$-dimensional random feature space by a random feature map $F'
= (f_1, \ldots, f_M)'$ (a synaptic connection weight matrix of size $M
\times N$) by computing the $M$-dimensional feature vector $F'\,r$,
\item  multiply each of the $M$ feature projections $f'_i\,r$ with an
adaptive \emph{conception weight} $c_i$ to get a conceptor-weighted feature state $z =
\mbox{diag}(c)\, F'\,r$, where the \emph{conception vector} $c =
(c_1,\ldots,c_M)'$ is made of the conception weights, 
\item project $z$ back to the reservoir by a random $N \times M$
backprojection matrix $\tilde{G}$, closing the loop.
\end{enumerate}
Since both $W$ and $\tilde{G}$ are random, they can be joined in a single
random map $G = W\,\tilde{G}$. This leads to the following consolidated state update
cycle of a \emph{random feature conception}  (RFC) architecture:
\begin{eqnarray}
r(n+1) & = & \tanh(G\, z(n) + W^{\mbox{\scriptsize in}}p(n) +
b),\label{eqRFC1}\\
z(n+1) & = & \mbox{diag}(c(n))\, F' \, r(n+1),\label{eqRFC2}
\end{eqnarray}
where $r(n) \in \mathbb{R}^N$ and $z(n), c(n) \in \mathbb{R}^M$.

From a biological modeling perspective there exist a number of concrete
candidate mechanisms by which the mathematical operation of
multiplying-in the conception weights could conceivably be realized. I will discuss
these later  and for the time being remain on this
abstract mathematical level of description.

\begin{figure}[htb]
\center
{\bf  A} \includegraphics[width=80mm]{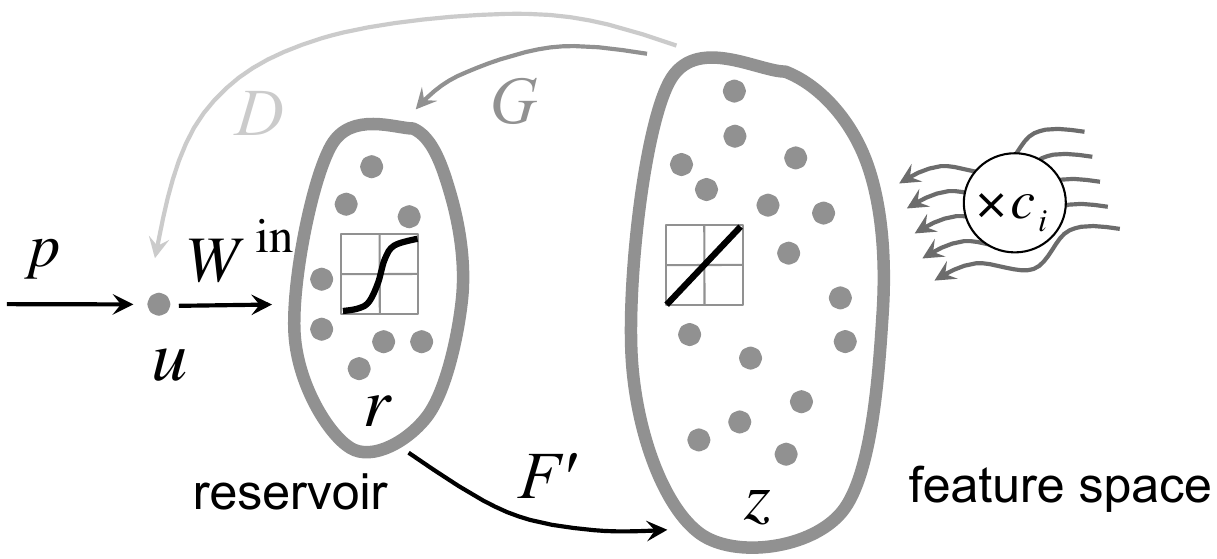}
\hspace{8mm}{\bf { B}} \includegraphics[width=40mm]{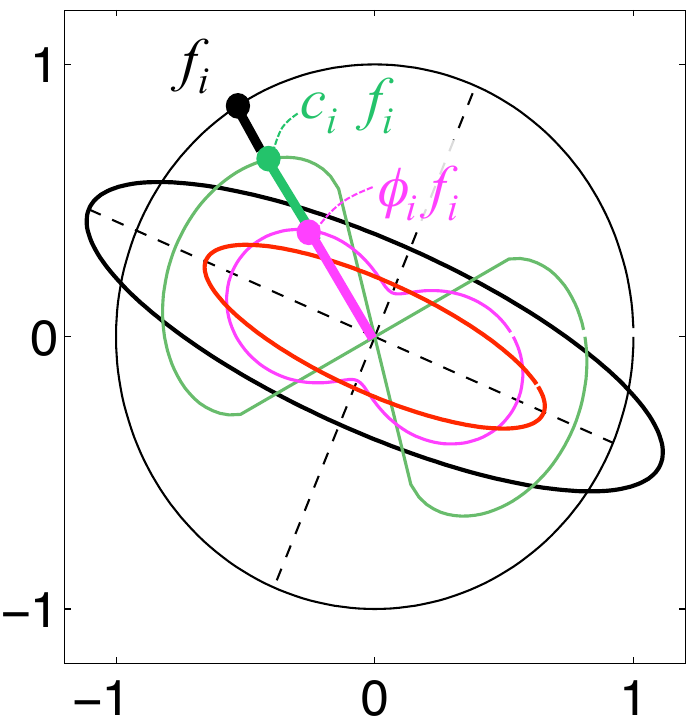}
\caption{An alternative conceptor architecture aiming at greater
  biological plausibility.  {\bf  A} Schematic
  of random feature space
  architecture. The reservoir state $r$ is projected by a
  feature map $F'$ into a higher-dimensional feature space with states
  $z$, from
  where it is back-projected by $G$ into the reservoir. The conceptor
  dynamics is realized by unit-wise multiplying conception weights
  $c_i$ into $f'_i r$ to obtain 
  the $z$ state.  The input unit $u$ is
  fed by external input $p$ or by learnt input simulation weights $D$.
  {\bf  B} Basic idea (schematic). Black ellipse:
  reservoir state $r$ correlation matrix $R$. Magenta dumbbell:
  scaling sample points $f_i$ from the unit sphere by their mean
  squared projection on reservoir states. Green dumbbell: feature
  vectors $f_i$ scaled by auto-adapted conception weights $c_i$. Red
  ellipse: the resulting virtual conceptor $C_F$. For detail see text.}
\label{figBiolArch}
\end{figure}

The conception vector $c(n)$ is adapted online and element-wise in a way that
is analog to the adaptation of matrix autoconceptors given in Definition
\ref{defAutoCRNN}. Per each element $c_i$ of $c$, the adaptation aims
at minimizing the objective function
\begin{equation}\label{eqcAdaptObjective}
E[(z_i - c_i z_i)^2] + \alpha^{-2}c_i^2,
\end{equation}
which leads to fixed point solutions satisfying 
\begin{equation}\label{eqcAdaptFPsolutions}
c_i = E[z_i^2] (E[z_i^2] + \alpha^{-2})^{-1}
\end{equation}
and a stochastic gradient descent online adaptation rule 
\begin{equation}\label{eqcAdapt}
c_i(n+1) = c_i(n) + \lambda_i \left(z_i^2(n) - c_i(n)\,z^2_i(n) -
  \alpha^{-2}\,c_i(n)\right), 
\end{equation}
where $i = 1,\ldots,M$, $\lambda_i$ is an adaptation rate, and $z_i$
is the $i$-the component of $z$. In computer simulations one will
implement this adaptation not element-wise but in an obvious
vectorized fashion.

If (\ref{eqcAdapt}) converges, the converged fixed point is either
$c_i = 0$, which always is a possible and stable solution, or it is of
the form 
\begin{equation}\label{eqSolRFCc}
c_i = 1/2 + \sqrt{(\alpha^2 \phi_i - 4) /4 \alpha^2
  \phi_i},
\end{equation}
 which is another possible stable solution provided that
$\alpha^2 \phi_i - 4 > 0$. In this formula, $\phi_i$ denotes the
expectation $\phi_i = E_r[(f'_i \,r)^2]$, the mean energy of the feature
signal $f'_i\,r$. These possible values of stable solutions can be
derived in a similar way as was done for the singular values of
autoconceptive matrix $C$ in Section \ref{subsecCAD}, but the
derivation is by far simpler (because it can be done element-wise for
each $c_i$ and thus entails only scalars, not matrices) and is left
as an exercise. Like the singular values of stable autoconceptors $C$,
the possible stable value range for conception weights obtainable
through (\ref{eqcAdapt}) is thus $\{0\} \cup (1/2, \,1)$.

Some geometric properties of random feature conceptors are illustrated
in Figure \ref{figBiolArch} {\bf B}. The black ellipse represents the
state correlation matrix $R = E[rr']$ of a hypothetical 2-dimensional
reservoir. The random feature vectors $f_i$ are assumed to have unit norm in
this schematic and therefore sample from the surface of the unit
sphere.  The magenta-colored dumbbell-shaped surface represents the weigthing of the
random feature vectors $f_i$ by the mean energies $\phi_i =
E[(f'_i\,r)^2]$ of the feature signals $f'_i\,r$. Under the
autoconception adaptation they give rise to conception weights $c_i$
according to (\ref{eqSolRFCc}) (green dumbbell surface). For values
$\alpha^2 \, \phi_i - 4 < 0$ one obtains $c_i = 0$, which shows up in the
illustration as the wedge-shaped indentation in the green curve. The
red ellipse renders the virtual conceptor $C_F$ (see below) which
results from the random feature conception weights.  

Two properties of this RFC architecture are worth
pointing out.  

First,  conceptor matrices $C$ for an $N$-dimensional
reservoir have $N(N+1)/2$ degrees of freedom. If, using conception
vectors  $c$ instead, one wishes to attain a performance level of pattern
reconstruction accuracy that is comparable to what can be achieved with
conceptor matrices $C$, one would expect that $M$ should be in
the order of $N(N+1)/2$. At any rate, this is an indication that $M$
should be significantly larger than $N$. In the simulations
below I used $N = 100, M = 500$, which worked robustly well. In
contrast, trying $M = 100$ (not documented), while likewise yielding
good accuracies, resulted in systems that were rather sensitive to
parameter settings. 

Second, the individual adaptation rates $\lambda_i$ can be chosen much
larger than the global adaptation rate $\lambda$ used for matrix
conceptors, without putting stability at risk. The reason is that the
original adpatation rate $\lambda$ in the stochastic gradient descent
formula for matrix conceptors given in Definition \ref{defAutoCRNN} is
constrained by the highest local curvature in the gradient landscape,
which leads to slow convergence in the directions of lower curvature.
This is a notorious general characteristic of multidimensional
gradient descent optimization, see for instance
\cite{FarhangBoroujeny98}.  This problem becomes irrelevant for the
individual $c_i$ updates in (\ref{eqcAdapt}).  In the simulations
presented below, I could safely select the $\lambda_i$ as large as
$0.5$, whereas when I was using the original conceptor matrix autoadaption
rules, $\lambda = 0.01$ was often the fastest rate possible. If
adaptive individual adaptation rates $\lambda_i$ would be implemented
(not explored), very fast convergence of (\ref{eqcAdapt}) should
become feasible.

\paragraph*{Geometry of feature-based conceptors.} Before I report on
simulation experiments, it may be helpful to contrast geometrical
properties of the RFC architecture and with the geometry of matrix
autoconceptors.

For the sake of discussion, I split the backprojection $N \times M$
matrix $G$ in (\ref{eqRFC1}) into a product $G = W \, F$ where the
``virtual'' reservoir weight matrix $W := GF^\dagger$ has size $N
\times N$. That is, I consider a system $z(n+1) = F\,\mbox{diag}(c(n))
\, F'\,\tanh(Wz(n) )$ equivalent to (\ref{eqRFC1}) and (\ref{eqRFC2}),
where $c$ is updated according to (\ref{eqcAdapt}). For the sake of
simplicity I omit input terms and bias in this discussion. The map $F
\circ \mbox{diag}(c) \circ F': \mathbb{R}^N \to \mathbb{R}^N$ then
plugs into the place that the conceptor matrix $C$ held in the
conceptor systems $z(n+1) = C \,\tanh(Wz(n))$ discussed in previous
sections.  The question I want to explore is how $F \circ
\mbox{diag}(c) \circ F'$ compares to $C$ in geometrical terms. A
conceptor matrix $C$ has an SVD $C = U S U'$, where $U$ is
orthonormal. In order to make the two systems directly comparable, I
assume that all feature vectors $f_i$ in $F$ have unit norm. Then $C_F
= \|F\|^{-2}_2 \,F \circ \mbox{diag}(c) \circ F' $ is positive
semidefinite with 2-norm less or equal to 1, in other words it is an
$N\times N$ conceptor matrix.

Now furthermore assume that the adaptation (\ref{eqcAdapt}) has
converged. The adaptation loop (\ref{eqRFC1}, \ref{eqRFC2},
\ref{eqcAdapt} ) is then a stationary process and the expectations
$\phi_i = E_r[(f'_i \,r)^2]$ are well-defined. Note that these expectations can
equivalently be written as $\phi_i = f'_i\, R \, f_i$, where $R =
E[rr']$.  According to what I remarked earlier, after convergence to
a
stable fixed point solution we have, for all $1 \leq i \leq M$, 
\begin{equation}\label{eqcConvergedValue}
c_i \in \left\{\begin{array}{ll} \{0\}, & \mbox{if }\alpha^2 \phi_i - 4
      \leq 0,\\ \{0, 1/2 +
\sqrt{(\alpha^2 \phi_i - 4) /4 \alpha^2 \phi_i}\}, & \mbox{if } \alpha^2 \phi_i - 4
      > 0.\end{array}   \right.
\end{equation}

Again for the sake of discussion I restrict my considerations to
converged solutions where all $c_i$ that \emph{can} be nonzero (that
is, $\alpha^2 \phi_i - 4 > 0$) are indeed nonzero.

It would be desirable to have an analytical result which gives the SVD
 of the $N\times N$ conceptor
$C_F = \|F\|^{-2}_2 \,F \circ \mbox{diag}(c) \circ F' $ under these
assumptions. Unfortunately this analysis appears to be involved and at
this point I cannot deliver it. In order to still obtain some insight
 into the geometry of $C_F$, I computed a number of such matrices numerically
 and  compared them to matrix-based autoconceptors $C$ that were
 derived from the same assumed stationary reservoir state process. The
 outcome is displayed in Figure \ref{figCversusc}.

\begin{figure}[htb]
\center
\includegraphics[width=110mm]{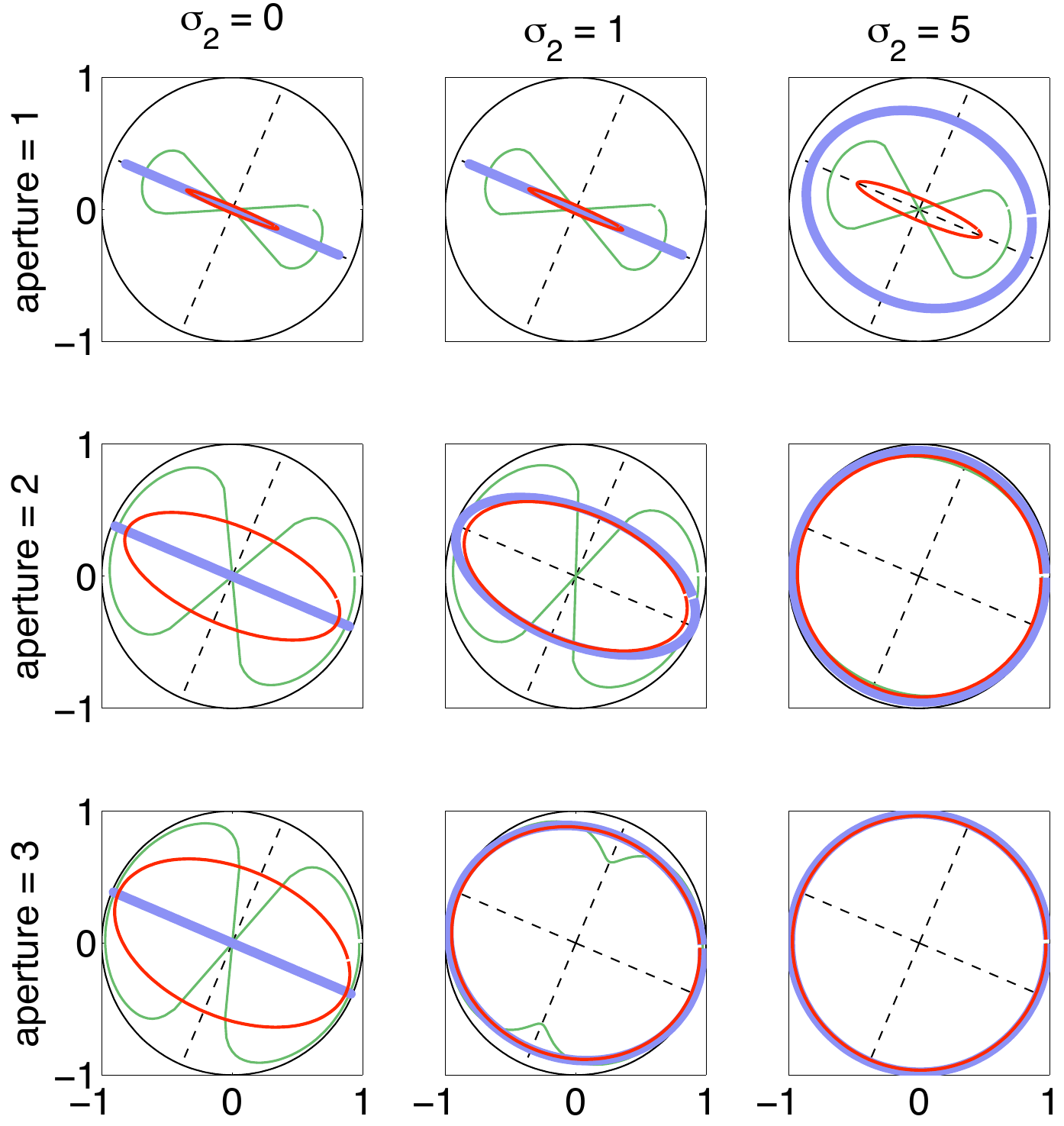}
\caption{Comparing matrix-based autoconceptors (bold blue
  ellipses) with feature-based autoconceptors $C_F$ (red
  ellipses). Broken lines mark principal directions of the reservoir
  state correlation matrix $R$. Each panel corresponds to a particular
  combination of aperture and the second singular value $\sigma_2$ of
  $R$. The dumbbell-shaped surfaces (green line) represent the
  values of the conception weights $c_i$. For explanation see text.}
\label{figCversusc}
\end{figure}

Concretely, these numerical investigations were set up as follows. The
reservoir dimension was chosen as $N = 2$ to admit plotting. The
number of features was $M = 200$. The feature vectors $f_i$ were
chosen as $(\cos(i\, 2\, \pi / M), \sin(i\, 2\, \pi / M))'$ (where $i =
1,\ldots,M$), that is, the unit vector $(1 \;0)'$ rotated in
increments of $(i/M)\, 2\, \pi$. This choice mirrors a situation where a
very large number of $f_i$ would be randomly sampled; this would
likewise result in an essentially uniform coverage of the unit circle.
The conception weights $c_i$ (and hence $C_F$) are determined by the
reservoir state correlation matrix $R = E[rr']$. The same holds for
 autoconceptor matrices $C$. For an exploration of the $C_F$ versus $C$
geometries, I thus systematically varied $R = U \Sigma U'$. The
principal directions $U$ were randomly chosen and remained the same
through all variations. The singular values $\Sigma =
\mbox{diag}(\sigma_1\; \sigma_2)$ were chosen as $\sigma_1 \equiv 10,
\sigma_2 \in\{ 0, 1, 5\}$, which gave three versions of $R$. The aperture
$\alpha$ was selected in three variants as $\alpha \in \{ 1, 2, 3\}$, which
altogether resulted in nine $(R,\alpha)$ combinations.

For each of these combinations, conception weights $c_i$ were computed via
(\ref{eqcConvergedValue}), from which $C_F$ were obtained. Each of
these maps the unit circle on an ellipse,  plotted in Figure
\ref{figCversusc} in red. The values of the $c_i$ are
represented in the figure as the dumbbell-shaped curve (green) connecting the
vectors $c_i \,f_i$. The wedge-shaped constriction to zero in some of these
curves corresponds to angular values of $f_i$ where $c_i = 0$. 

For comparison, for each of the same $(R,\alpha)$ combinations also an
autoconceptor matrix $C$ was computed using the results from Section
\ref{subsecCAD}. We saw on that occasion that nonzero singular values of $C$
are not uniquely determined by $R$; they are merely constrained by
certain interlacing bounds. To break this indeterminacy, I selected
those $C$ that had the maximal admissible singular values. According
to (\ref{som1Eq11}), this means that the singular values of $C$ were
set to $s_i = 1/2 + \sqrt{(\alpha^2 \sigma_i - 4) /4 \alpha^2
  \sigma_i}$ (where $i = 1,2$) provided the root argument was
positive, else $s_i = 0$.

Here are the main findings that can be collected from Figure
\ref{figCversusc}:

\begin{enumerate}
\item The principal directions of $C_F, C$ and $R$ coincide. The fact
that $C_F$ and $R$ have the same orientation can also be shown
analytically, but the argument that I have found is (too) involved and not
given here. This orientation of $C_F$ hinges on the circumstance that
the $f_i$ were chosen to uniformly sample the unit sphere.
  \item The $C_F$ ellipses are all non-degenerate, that is, $C_F$ has
no zero singular values (although many of the $c_i$ may be zero as
becomes apparent in the wedge constrictions in the dumbbell-shaped
representation of these values). In particular, the $C_F$ are also
non-degenerate in cases where the matrix autoconceptors $C$ are
(panels in left column and center top panel). The finding that $C_F$
has no zero singular values can be regarded as a disadvantage compared
to matrix autoconceptors, because it implies that no signal direction
in the $N$-dimensional reservoir signal space can be completely
suppressed by $C_F$. However, in the $M$-dimensional feature signal
space, we do have nulled directions. Since the experiments reported
below exhibit good stability properties in pattern reconstruction, it
appears that this ``purging'' of signals in the feature space segment
of 
the complete reservoir-feature loop is effective enough.
\item Call the ratio of the largest over the smallest singular value
of $C_F$ or $C$ the \emph{sharpness} of a conceptor (also known as
eigenvalue spread in the signal processing literature). Then sometimes
$C_F$ is sharper than $C$, and sometimes the reverse is true. If
sharpness is considered a desirable feature of concepors (which I
think it often is), then there is no universal advantage of $C$ over $C_F$
or vice versa.
\end{enumerate}

\paragraph*{System initialization and loading patterns: generic
  description.} Returning from this inspection of geometrical
properties to the system (\ref{eqRFC1}) -- (\ref{eqcAdapt}), I proceed
to describe the initial network creation and pattern loading procedure
in generic terms.  Like with the matrix conceptor systems considered
earlier in this report, there are two variants which are the analogs
of (i) recomputing the reservoir weight matrix $W^\ast$ into an input
internalization matrix $W$, as in Section
\ref{sec:StoringGeneric}, as opposed to (ii) computing an additional input
simulation matrix $D$, as in Section \ref{subsec:memmanage}. In the
basic experiments reported below I found that both work equally well.
Here I  document  the second option.  A readout weight vector
$W^{\mbox{\scriptsize out}}$ is likewise computed during loading. Let
$K$ target patterns $p^j$ be given ($j = 1,\ldots,K)$, which are to be
loaded. Here is an outline:

\begin{description}
    \item[Network creation.] A random feature map $F$, random
  input weights $W^{\mbox{\scriptsize in}}$, a random bias vector $b$,
  and a random backprojection matrix $G^\ast$ are generated. $F$ and
  $G^\ast$ are suitably scaled such that the combined $N \times N$ map
  $G^\ast \, F'$ attains a prescribed spectral radius. This spectral
  radius is a crucial system parameter and plays the same role as the
  spectral radius in reservoir computing in general (see for instance
  \cite{Verstraeten09,Yildizetal12}). All conception weights are
  initialized to $c^j_i = 1$, that is, $\mbox{diag}(c^j) = I_{M \times
    M}$.
\item[Conception weight adaptation.] The system is driven with  each
pattern $p^j$ in turn for $n_{\mbox{\scriptsize adapt}}$ steps
(discarding an initial washout), while $c^j$ is being adapted per
\begin{eqnarray*}
z^j(n+1) & = & \mbox{diag}(c^j(n))\, F' \, \tanh(G^\ast\, z^j(n) +
W^{\mbox{\scriptsize in}}p^j(n) + b),\\
c^j_i(n+1) & = & c^j_i(n) + \lambda_i \left( z^j_i(n)^2 -
  c^j_i(n)\,z^j_i(n)^2 - \alpha^{-2}\, c^j_i(n)  \right),
\end{eqnarray*}
leading to conception vectors $c^j$ at the end of this period. 
  \item[State harvesting for computing $D$ and $W^{\mbox{\scriptsize \rm 
    out}}$, and for recomputing $G$.] The conception vectors $c^j$ obtained from the previous
step are kept fixed, and for each pattern $p^j$ the input-driven system $r^j(n) =
\tanh(G^\ast\, z^j(n) + W^{\mbox{\scriptsize in}}p^j(n) + b)$; $z^j(n+1) =
\mbox{diag}(c^j)\, F' \,r^j(n)$ is run for $n_{\mbox{\scriptsize
    harvest}}$ time steps, collecting states $r^j(n)$ and $z^j(n)$. 
  \item[Computing weights.] The $N \times M$ input simulation matrix
  $D$ is computed by solving the regularized linear regression 
\begin{equation}\label{eqAdaptDCriterion}
D = \mbox{argmin}_{\tilde{D}}\; \sum_{n,j}\|
W^{\mbox{\scriptsize in}}p^j(n) - \tilde{D} z^j(n-1) \|^2 + \beta_D^2\,\|
\tilde{D} \|^2_{\mbox{\scriptsize fro}} 
\end{equation}
where $\beta_D$ is a suitably chosen Tychonov regularizer. This means
that the autonomous system update
$z^j(n+1) = \mbox{diag}(c^j)\,F'\,\tanh(G^\ast z^j(n) +
D\, z^j(n) + b)$ should be able to
simulate input-driven updates $z^j(n) = $
\\$\mbox{diag}(c^j)\,F'\,\tanh(G^\ast\, z^j(n) + W^{\mbox{\scriptsize
    in}}p^j(n) + b)$.  $W^{\mbox{\scriptsize
    out}}$ is similarly computed by solving
\begin{displaymath}
W^{\mbox{\scriptsize out}} = \mbox{argmin}_{\tilde{W}}\;
\sum_{n,j}\| p^j(n) - \tilde{W}r^j(n) \|^2 + \beta_{W^{\mbox{\scriptsize out}}}^2\,\|
\tilde{W} \|^2.
\end{displaymath}

Optionally one may furthermore
recompute $G^\ast$ by solving the trivial regularized linear regression
\begin{displaymath}
G = \mbox{argmin}_{\tilde{G}}\; \sum_{n,j}\|G^\ast z^j(n) - \tilde{G}
z^j(n) \|^2 + \beta_G^2\,\| \tilde{G} \|^2_{\mbox{\scriptsize fro}}. 
\end{displaymath}
for a suitably chosen Tychonov regularizer $\beta_G$. While $G^\ast$
and $G$ should behave almost identically on the training inputs,
the average absolute size of entries in $G$ will  be (typically much)
smaller than the original weights in $G^\ast$ as a result of the
regularization. Such regularized auto-adaptations have been found to
be beneficial in pattern-generating recurrent neural networks
\cite{ReinhartSteil10}, and in the experiments to be reported presently
I took advantage of this scheme.
\end{description} 

The feature vectors $f_i$ that make up $F$ can optionally be
normalized such that they all have the same norm. In my experiments
this was not found to have a noticeable effect.

If a stored pattern $p^j$ is 
to be retrieved, the only item that needs to be changed is the
conception vector $c^j$. This vector can either be obtained by
re-activating that $c^j$ which was adapted during the loading (which
implies that it needs to be stored in some way). Alternatively, it can
be obtained by autoadaptation without being previously stored, as in
Sections \ref{secAutoC} -- \ref{subsecCAD}. I now describe two
simulation studies which demonstrate how this scheme functions
(simulation detail documented in Section \ref{secExpBiolPlausible}). The
first study uses stored conception vectors, the second
demonstrates autoconceptive adaptation.

\paragraph*{Example 1: pattern retrieval with stored conception vectors
  $c^j$.} This simulation re-used the $N = 100$ reservoir from
Sections \ref{sec:InitialDrivingDemo} \emph{ff.} and the four driver
patterns (two  irrational-period sines, two very similar
5-periodic random patterns). The results are displayed in Figure
  \ref{figcStored}.

\begin{figure}[htb]
  \center \includegraphics[width=130mm]{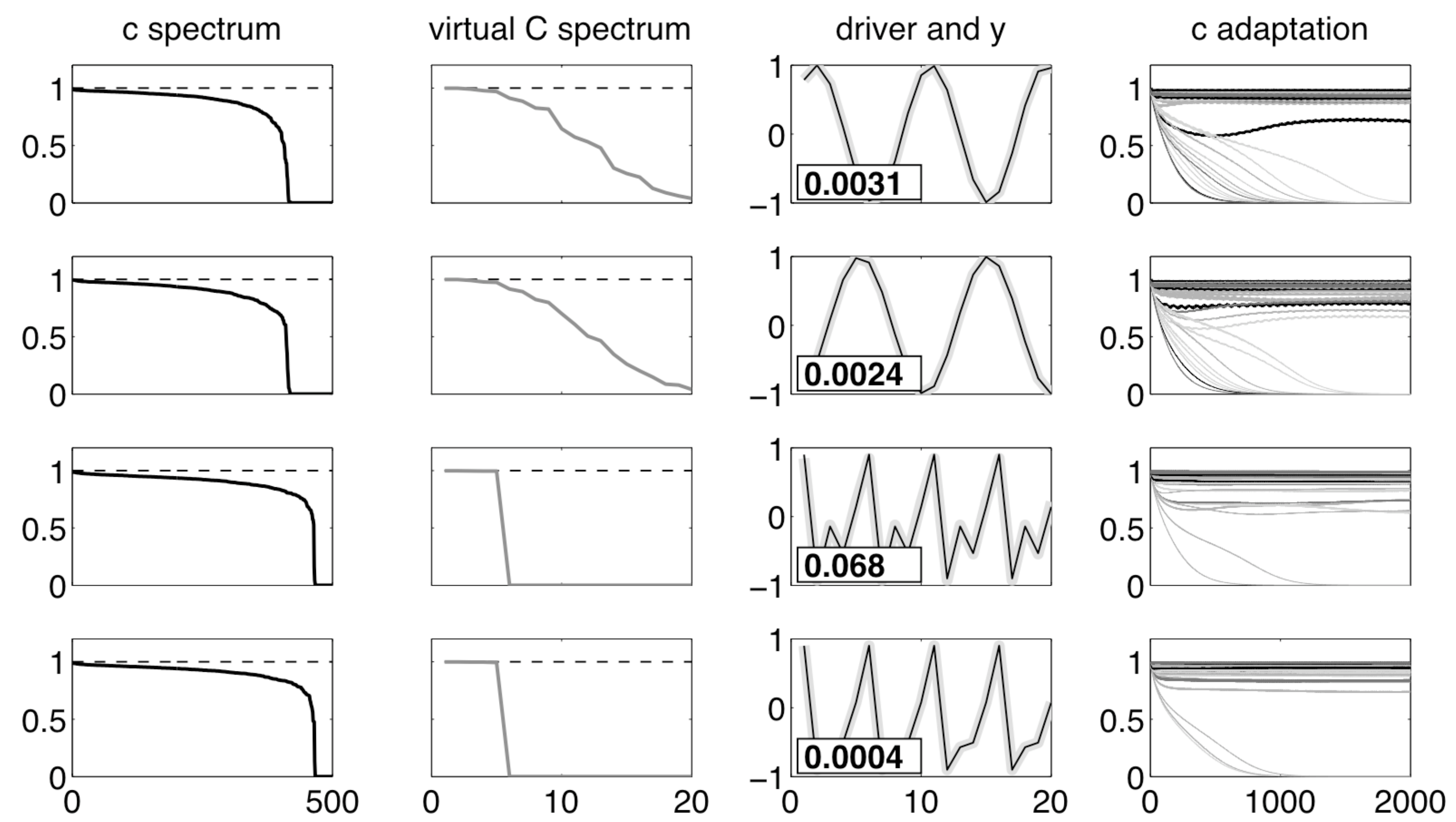}
\caption{Using stored random feature coded conceptors in a replication of the basic
  pattern retrieval experiment from Section \ref{sec:RetrieveGeneric},
  with $M = 500$ random feature vectors $f_i$.  First column: sorted
  conception vectors $c^j$. Second column: spectra of virtual
  conceptors $C_F$. Third column: reconstructed patterns (bold light
  gray) and original patterns (thin black) after phase alignment.
  NRMSEs are given in insets. Last column: The adaptation of $c^j$
  during the 2000 step runs carried out in parallel to the loading
  process. 50 of 500 traces are shown. For explanation see text.}
\label{figcStored}
\end{figure}

The loading procedure followed the generic scheme described above
(details in Section \ref{secDocExp}), with $n_{\mbox{\scriptsize
    adapt}} = 2000, n_{\mbox{\scriptsize harvest}} = 400$, $\lambda_i
= 0.5$, $\beta_G^2 = \beta_D^2 = 0.01$ and
$\beta_{W^{\mbox{\scriptsize out}}}^2 = 1$. The aperture was set to
$\alpha = 8$. The left column in Figure \ref{figcStored} shows the
resulting $c^j$ spectra, and the right column shows the evolution of
$c^j$ during this adaptation. Notice that a considerable portion of
the conception weights evolved toward zero, and that none ended in the
range $(0\;1/2)$, in agreement with theory. 

For additional insight into the dynamics of this system I also
computed  ``virtual'' matrix conceptors $C^j_F$ by $R^j = E[(r^j)'
r^j], \; C_F^j = R^j \, (R^j + \alpha^{-2})^{-1}$ (second column). The
singular value spectrum of $C^j_F$ reveals that the autocorrelation
spectra of $r^j$ signals in RFC systems is  almost
identical to the singular value spectra obtained with matrix
conceptors on earlier occasions (compare Figure \ref{FigSingValsFalloff}).

The settings of matrix scalings and aperture were quite robust;
variations in a range of about $\pm$50\% about the chosen
values preserved stability and accuracy of pattern recall (detail
in Section \ref{secExpBiolPlausible}).

For testing the recall of pattern $p^j$, the loaded system was run
using the update routine
\begin{eqnarray*}
r^j(n) & = & \tanh(G\,z^j(n) + D\, z^j(n) + b),\\ 
y^j(n) & = & W^{\mbox{\scriptsize out}}\, r^j(n),\\
z^j(n+1) & = & \mbox{diag}(c^j)\,F'\,r^j(n),
\end{eqnarray*}
starting from a random starting state $z^j(0)$ which was sampled from
the normal distribution, scaled by $1/2$. After a washout of 200
steps, the reconstructed pattern $y^j$ was recorded for 500 steps and
compared to a 20-step segment of the target pattern $p^j$. The second
column in Figure \ref{figcStored} shows an overlay of $y^j$ with $p^j$
and gives the NRMSEs. The reconstruction is of a similar quality as
was found in Section \ref{sec:RetrieveGeneric} where full conceptor
matrices $C$ were used.

\paragraph*{Example 2: content-addressed pattern retrieval.} For a
demonstration of content-addressed recall similar to the studies
reported in Section \ref{secAutoCMemExample}, I re-used the $M = 500$
system described above. Reservoir scaling parameters and the loading
procedure were identical except that conception vectors $c^j$ were not
stored. Results are collected in Figure
\ref{figcContAddress}. The cue and recall procedure for a pattern $p^j$ was
carried out as follows:

\begin{enumerate}
    \item Starting from a random reservoir state, the loaded reservoir
  was driven with the cue pattern for a washout time of 200 steps by
  $$z^j(n+1) = F' \, \tanh(G \, z^j(n) + W^{\mbox{\scriptsize
      in}}\, p^j(n) + b).$$
\item Then, for a cue period of 800 steps, the system was updated with
$c^j$ adaptation by 
\begin{eqnarray*}
z^j(n+1) & = & \mbox{diag}(c^j(n))\,F'\,\tanh(G\,z^j(n) +
W^{\mbox{\scriptsize in}}\, p^j(n) + b),\\
c^j_i(n+1) & = & c^j_i(n) + \lambda_i \left( z^j_i(n)^2 -
  c^j_i(n)\,z^j_i(n)^2 - \alpha^{-2}\, c^j_i(n)  \right),\;\;(1\leq
i\leq M)
\end{eqnarray*}
starting from an all-ones $c^j$, with an adaptation rate $\lambda_i =
0.5$ for all $i$. At the end of this period, a conception vector
$c^{j,{\mbox{\scriptsize cue}}}$ was obtained. 
\item To measure the quality of $c^{j,{\mbox{\scriptsize cue}}}$, a
separate run of 500 steps without $c$ adapation was done using
\begin{eqnarray*}
r^j(n) & = & \tanh(G\,z^j(n) +  D\, z^j(n) + b),\\  
y^j(n) & = & W^{\mbox{\scriptsize out}}\, r^j(n),\\
z^j(n+1) & = & \mbox{diag}(c^{j,{\mbox{\scriptsize cue}}})\,F'\, r^j(n)
\end{eqnarray*}
obtaining a pattern reconstruction $y^j(n)$. This was phase-aligned
with the original pattern $p^j$ and an NRMSE was computed (Figure
\ref{figcContAddress}, third column). 
\item The recall run was resumed after the cueing period and continued
for another 10,000 steps in auto-adaptation mode, using
\begin{eqnarray*}
z^j(n+1) & = & \mbox{diag}(c^j(n))\,F'\,\tanh(G\,z^j(n) + D\, z^j(n) + b),\\
c^j_i(n+1) & = & c^j_i(n) + \lambda_i \left( z^j_i(n)^2 -
  c^j_i(n)\,z^j_i(n)^2 - \alpha^{-2}\, c^j_i(n)  \right),\;\;(1\leq
i\leq K)
\end{eqnarray*}
leading to a final $c^{j,{\mbox{\scriptsize adapted}}}$ at the end of this
period. 
  \item Another quality measurement run was done identical to the
post-cue measurement run, using $c^{j,{\mbox{\scriptsize adapted}}}$
(NRMSE results in Figure \ref{figcContAddress}, third column).
\end{enumerate}

\begin{figure}[htb]
\center
\includegraphics[width=145mm]{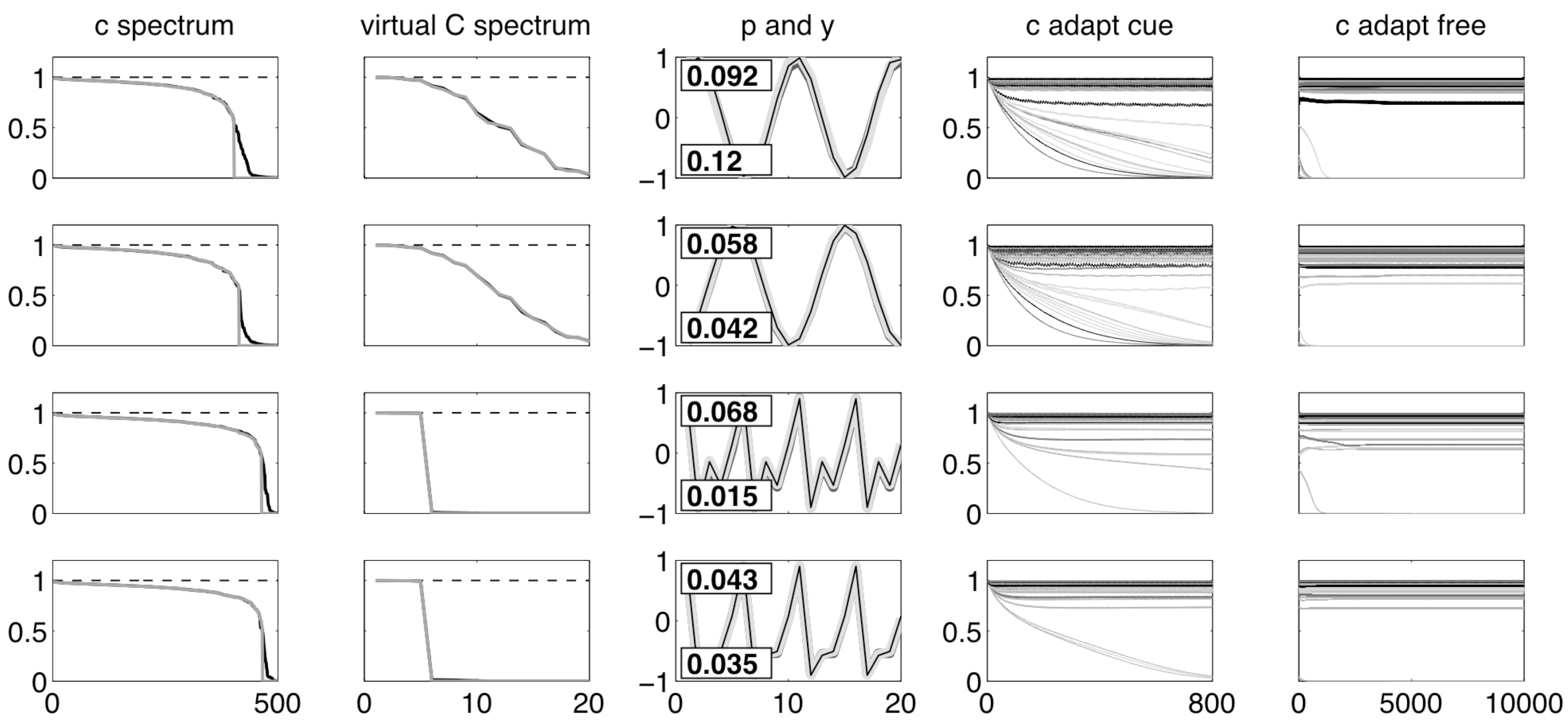}
\caption{Content-addressed recall using RFC
  conceptors with $M = 500$ feature vectors $f_i$.  First column:
  sorted feature projection weight vectors $c^j$ after the cue phase
  (black) and after 10,000 steps of autoadaptation (gray). Second
  column: spectra of virtual conceptors $C_F$ after cue (black) and at
  the end of autonomous adaptation (gray). Both spectra are almost
  identical. Third column: reconstructed patterns (bold light gray:
  after cue, bold dark gray: after autoadaptation; the latter are
  mostly covered by the former) and original patterns (thin black).
  NRMSEs are given in insets (top: after cue, bottom: after
  autoadaptation). Fourth column: The adaptation of $c^j$ during the
  cueing period. Last column: same, during the 10000 autoadaptation
  steps. 50 of 500 traces are shown. Note the different timescales in
  column 4 versus column 5.  For explanation see text.}
\label{figcContAddress}
\end{figure}

Like in the matrix-$C$-based content-addressing experiments from
Section \ref{secAutoCMemExample}, the recall quality directly after
the cue further improved during the autoconceptive adaption
afterwards, except for the first pattern. Pending a more detailed
investigation, this may be attributed to the ``maturation'' of the
$c^j$ during autoadaption which reveals itself in the convergence of a
number of $c^j_i$ to zero during autoadaptation (first and last column
in Figure \ref{figcContAddress}). We have seen similar effects in
Section \ref{secAutoCMemExample}.

An obvious difference to those earlier experiments is that the cueing
period is much longer now (800 versus 15 -- 30 steps). This is owed to
the circumstance that now the conceptor adaptation during cueing
started from an all-ones $c^j$, whereas in  Section
\ref{secAutoCMemExample} it was started from a zero $C$. In the latter
case, singular values of $C$ had to grow away from zero toward one
during cueing, whereas here they had to sink away from one toward
zero. The effects of this mirror situation are not symmetrical. In an
``immature'' post-cue conceptor matrix $C$ started from a zero $C$,
all the singular values which eventually should converge to zero are
already at zero at start time and remain there. Conversely, the
post-cue feature projection weights $c^{j,{\mbox{\scriptsize cue}}}$,
which \emph{should} eventually become zero, have not come close to
this destination even after the 800 cue steps that were allotted here
(left panels in Figure \ref{figcContAddress}). This tail of
``immature'' nonzero elements in $c^{j,{\mbox{\scriptsize cue}}}$
leads to an insufficient filtering-out of reservoir state components
which do not belong to the target pattern dynamics.

The development of the $c^j_i$ during the autonomous post-cue
adapation is not monotonous (right panels). Some of these weights
meander for a while before they settle to what appear stable final
values. This is due to the transient nonlinear reservoir--$c^j$ interactions
which remain to be mathematically analyzed.

A potentially important advantage of using random feature conceptors
$c$ rather than matrix conceptors $C$ in machine learning applications
is the faster convergence of the former in online adaptation
scenarios. While an dedicated comparison of convergence properties
between $c$ and $C$ conceptors remains to be done, one may naturally
expect that stochastic gradient descent works more efficiently for
random feature conceptors than for matrix conceptors, because the
gradient can be followed individually for each coordinate $c_i$,
unencumbered by the second-order curvature interactions which
notoriously slow down simple gradient descent in multidimensional
systems. This is one of the reasons why in the complex hierarchical
signal filtering architecture to be presented below in Section
\ref{secHierarchicalArchitecture} I opted for random feature
conceptors.

\paragraph*{Algebraic and logical rules for  conception weights.} The
various definitions and rules for aperture adaptation, Boolean
operations, and abstraction introduced previously for matrix
conceptors directly carry over to random feature conceptor. The new
definitions and rules are simpler than for conceptor matrices because
they all apply to the individual, scalar conception weights. I present
these items without detailed derivations (easy exercises). In the
following, let $c = (c_1,\ldots,c_M)', b = (b_1,\ldots,b_M)'$ be two
conception weight vectors.

Aperture adaptation (compare Definition \ref{defLimitPhi}) becomes

\begin{definition}\label{def:Apadaptcvecs}
\begin{eqnarray*}
\varphi(c_i, \gamma) & := & c_i / (c_i + \gamma^{-2}(1-c_i)) \quad
\mbox{for } \; 0 < \gamma < \infty,\\
\varphi(c_i, 0) & := & \left\{\begin{array}{lll}0 & \mbox{if} & c_i < 1, \\  1
    & \mbox{if} & c_i = 1, \end{array}\right.\\ 
\varphi(c_i, \infty) & := & \left\{\begin{array}{lll}1 & \mbox{if} & c_i > 0, \\  0
    & \mbox{if} & c_i = 0. \end{array}\right.
\end{eqnarray*}
\end{definition}

 Transferring the
matrix-based definition of Boolean operations (Definition
\ref{def:finalBoolean}) to conception weight vectors leads to the following laws:

\begin{definition}\label{def:Booleancvecs}
\begin{eqnarray*}
\neg\,c_i & := & 1 - c_i,\\
c_i \wedge b_i & := & \left\{ \begin{array}{lll} c_i b_i / (c_i + b_i -
  c_i b_i) & \mbox{ if not } & c_i = b_i = 0,\\
0 & \mbox{ if } & c_i = b_i = 0,  \end{array}\right.\\
c_i \vee b_i & := & \left\{ \begin{array}{lll} (c_i + b_i - 2c_i b_i) / (1 -
  c_i b_i) & \mbox{ if not } & c_i = b_i = 1,\\
1 & \mbox{ if } & c_i = b_i = 1.  \end{array}\right.
\end{eqnarray*}
\end{definition}

The matrix-conceptor properties connecting aperture adaptation with
Boolen operations (Proposition \ref{propBooleanAperture}) and the
logic laws (Proposition \ref{propBooleanElementaryLaws}) remain valid after the obvious modifications of
notation.

We define $c \leq b$ if for all $i = 1,\ldots,M$ it holds that $c_i
\leq b_i$. The main elements of Proposition \ref{propBasicAbstraction}
turn into 

\begin{proposition}\label{propcAbstraction}
Let $a = (a_1,\ldots,a_M)', b = (b_1,\ldots, b_M)'$ be conception
weight vectors.  Then the following facts hold. 
\begin{enumerate}
\item If $b \leq a$, then $b = a \wedge c$, where $c$ is the conception
weight vector with entries
\begin{displaymath}
c_i = \left\{\begin{array}{lll} 0 & \mbox{ if } &  b_i = 0,\\ (b_i^{-1}
    - a_i^{-1} + 1)^{-1}& \mbox{ if } &  b_i > 0.\end{array}  \right.
\end{displaymath} 
\item If $a \leq b$, then $b = a \vee c$, where $c$ is the conception
weight vector with entries
\begin{displaymath}
c_i = \left\{\begin{array}{lll} 1 & \mbox{ if } &  b_i = 1,\\ 1 - \left((1-b_i)^{-1}
    - (1-a_i)^{-1} + 1\right)^{-1}& \mbox{ if } &  b_i < 1.\end{array}  \right.
\end{displaymath}
\item If $a \wedge c = b$, then $b \leq a$.
\item If $a \vee c = b$, then $a \leq b$.  
\end{enumerate}
\end{proposition}

Note that all of these definitions and rules can be considered as
restrictions of the matrix conceptor items on the special case of
diagonal conceptor matrices. The diagonal elements of such diagonal
conceptor matrices can be identified with conception weights.
 
\paragraph*{Aspects of biological plausibility.} ``Biological
plausibility'' is a vague term inviting abuse. Theoretical
neuroscientists develop mathematical or computational models of neural
systems which range from fine-grained compartment models of single
neurons to abstract flowchart models of cognitive processes. Assessing
the methodological role of formal models in neuroscience is a complex
and sometimes controversial issue \cite{Abbott08,Gerstneretal12}.
When I speak of biological plausibility in connection with conceptor
models, I do not claim to offer a blueprint that can be directly
mapped to biological systems.  All that I want to achieve in this
section is to show that conceptor systems may be conceived which do
\emph{not} have characteristics that are decidedly \emph{not}
biologically feasible. In particular, (all) I wanted is a conceptor
system variant which can be implemented without state memorizing or
weight copying, and which only needs locally available information for
its computational operations. In the remainder of this section I
explain how these design goals may be satisfied by RFC conceptor
architectures.

My proposal for a biologically 'plausible' model will use input
recreation weights $H$ (see Section \ref{subsubsecGenPrinAlg}) instead
of input simulation weights $D$, which were used in most other demos
reported above.   

I will discuss only the adaptation of conception weights and the
learning of the input recreation weights $H$, leaving the cueing
mechanism aside.  The latter was implemented in Example 2 in the
previous subsection in an ad-hoc way just to set the stage and would
require a separate treatment under the premises of biological
plausibility.

In my discussion I will continue to use the discrete-time update
dynamics that was used throughout this report. Biological systems are
not updated according to a globally clocked cycle, so this clearly
departs from biology. Yet, even a synchronous-update discrete-time
model can offer relevant insight. The critical issues that I want to
illuminate -- namely, no state/weight copying and locality of
computations -- are independent of choosing a discrete or continuous
time setting.

I first consider the adaptation of the input recreation weights $H$.
In situated, life-long learning systems one may assume that at given
point in (life-)time, some version of $H$ is already present and
active, reflecting the system's learning history up to that point. In
the content-addressable memory example above, at the end of the cueing
period there was an abrupt
switch from driving the system with external input to an autonomous
dynamics using the system's own input simulation via $D$. Such binary
instantaneous switching can hardly be expected from biological
systems. It seems more adequate to consider a gradual blending between
the input-driven and the autonomous input simulation mode, as per
\begin{eqnarray} 
\lefteqn{z^j(n+1)  =  }\label{eqEtaMediation}\\
& = &\mbox{diag}(c^j(n))\,F'\,\tanh \left(G\,z^j(n) +
  W^{\mbox{\scriptsize in}} \,\left( \tau(n) \,H\, z^j(n) +
  (1-\tau(n))\,p(n)\right) + b\right),  \nonumber
\end{eqnarray}
where a mixing between the two modes is mediated by a ``slide ruler''
parameter $\tau$ which may range between 0 and 1 (a blending of this
kind will be used in the architecture in Section \ref{secHierarchicalArchitecture}).  

As a side remark, I mention that when one considers comprehensive
neural architectures, the question of negotiating between an
input-driven and an autonomous processing mode arises quite
generically. A point in case are ``Bayesian brain'' models of pattern
recognition and control, which currently receive much attention
\cite{Friston05,Clark12}. In those models, a neural processing layer
is driven both from ``lower'' (input-related) layers and from
``higher'' layers which autonomously generate predictions. Both
influences are merged in the target layer by some neural
implementation of Bayes' rule. Other approaches that I would like to
point out in this context are layered restricted Boltzmann machines
\cite{HintonSalakhutdinov06}, which likewise can be regarded as a
neural implementation of Bayes' rule; hierarchical neural field models
of object recognition \cite{Zibneretal11} which are based on
Arathorn's ``map seeking circuit''  model of combining
bottom-up and top-down inputs to a neural processing layer
\cite{GedeonArathorn07}; and mixture of experts models for motor
control (for example \cite{WolpertKawato98}) where a
``responsibility'' signal comparable in its function to the $\tau$
parameter negotiates a blending of different control signals.

\begin{figure}[htb]
\center
\includegraphics[width=110mm]{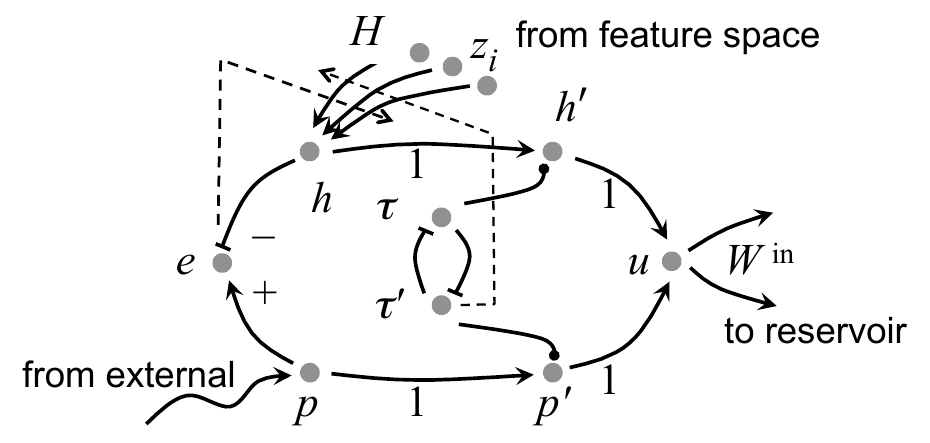}
\caption{An abstract circuit which would implement the $\tau$
  negotiation between driving a reservoir with external input $p$
  versus with recreated input $h = H z$. Abstract neurons are marked by filled
  gray circles. Connections that solely copy a
  neural state forward are marked with ``1''. Connections marked
  $-\!\bullet$ refer to multiplicative modulation. Connections that
  are inhibitory by their nature are represented by $\dashv$. Broken
  arrows indicate a controlling influence on the weight adaptation of $D$. For
  explanation see text.}
\label{figMicroCircuitH}
\end{figure}

Returning to (\ref{eqEtaMediation}), the mathematical formula could be
implemented in an abstract neural circuit as drawn in Figure
\ref{figMicroCircuitH}. Explanation of this diagram: gray dots
represent abstract firing-rate neurons (biologically realized by
individual neurons or collectives). All neurons are linear. Activation
of neuron $h$: simulated input $h(n) = Hz(n)$; of $p$: external driver
$p(n)$. Neuron $e$ maintains the value of the ``error'' $p(n) -
d(n)$. $h$ and $p$ project their activation values to $h'$ and $p'$, whose
activations are multiplicatively modulated by the activations of
neurons $\tau$ and $\tau'$. The latter maintain the values of $\tau$
and $\tau' = 1-\tau$ from (\ref{eqEtaMediation}). The activations $h'(n) =
\tau(n)\,H\,z(n)$ and $p'(n) = (1-\tau(n))\,p(n)$ are additively
combined in $u$, which finally feeds to the reservoir through
$W^{\mbox{\scriptsize in}}$. 

For a multiplicative modulation of neuronal activity a number of
biological mechanisms have been proposed, for example
\cite{Rothmanetal09,ChanceAbbott00}. The abstract model given here is
not committed to a specific such mechanism. Likewise I do not further
specify the biological mechanism which balances between $\tau$ and
$\tau'$, maintaining a relationship $\tau' = 1 - \tau$; it seems natural to see this as a suitable version of mutual
inhibition.

An in-depth discussion  by which mechanisms and for which purposes $\tau$ is
administered is beyond the scope of this report. Many scenarios are
conceivable. For the specific purpose of content-addressable memory
recall, the setting considered in this section, a natural option to
regulate $\tau$ would be to identify it with the (0-1-normalized and
time-averaged) error signal $e$. In the architecture presented in
Section \ref{secHierarchicalArchitecture} below, regulating $\tau$
 assumes a  key role and will be guided by  novel principles.   

The sole point that I want to make is that this abstract architecture
(or similar ones) requires only local information for the
adaptation/learning of $H$. Consider a synaptic connection $H_i$ from
a feature neuron $z_i$ to $h$. The learning objective
(\ref{eqAdaptDCriterion}) can be achieved, for instance, by the
stochastic gradient descent mechanism
\begin{equation}\label{eqAdaptDStochGrad}
H_i(n+1) = H_i(n) + \lambda\,\left((p(n) - H_i(n)z_i(n))\,z_i(n) -
  \alpha^{-2}\,H_i(n)  \right), 
\end{equation}
where the error $p(n) - H_i(n)z_i(n)$ is available in the activity
of the $e$ neuron. The learning rate $\lambda$ could be fixed, but a
more suggestive option would be to scale it by $\tau'(n) = 1 -
\tau(n)$, as indicated in the diagram. That is, $H_i$ would be adapted
with an efficacy proportional to the degree that the system is
currently being externally driven.

I now turn to the action and the adaptation of the conception weights,
stated in mathematical terms in equations (\ref{eqRFC2}) and
(\ref{eqcAdapt}). There are a number of possibilities to implement
these formulae in a model expressed on the level of abstract
firing-rate neurons. I inspect three of them. They are  sketched in
Figure \ref{figMicroCircuitsc}.

\begin{figure}[htb]
\center
\includegraphics[width=140mm]{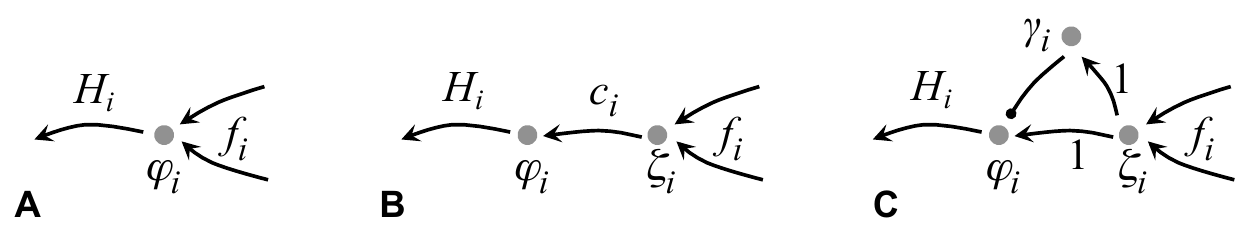}
\caption{Three candidate neural implementations of conception weight
  mechanisms. In each diagram, $\varphi_i$ is an abstract neuron whose
  activation is $\varphi(n) = z_i(n) = c_i(n) \, f'_i\,r(n)$. In {\bf  B}
  and {\bf C}, $\zeta_i$ has activation $f'_i\,r(n)$. In {\bf 
    C}, $\gamma_i$ has activation $c_i$.  For explanation see text.}
\label{figMicroCircuitsc}
\end{figure}

The simplest model (Figure \ref{figMicroCircuitsc} {\bf A}) represents
the quantity $z_i(n) = c_i(n) \, f'_i\,r(n)$ by the activation of a single
neuron $\varphi_i$. It receives synaptic input $f'_i\,r(n)$ through
connections $f_i$ and feeds to the reservoir (or to an input gating
circuit as discussed above) through the single synaptic connection
$H_i$. The weighting of $f'_i\,r(n)$ with the factor $c_i$ is effected
by some self-regulated modulation of synaptic gain. Taking into
account that $c_i$ changes on a slower timescale than $f'_i\,r(n)$,
the information needed to adapt the strength $c_i$ of this modulation
(\ref{eqcAdapt}) is a moving average of the neuron's own activation
energy $z_i^2(n)$ and the current synaptic gain $c_i(n)$, which are
characteristics of the neuron $\varphi_i$ itself and thus are
trivially locally available.

In the next model (Figure \ref{figMicroCircuitsc} {\bf B}), there is a
division of labor between a neuron $\varphi_i$ which again represents
$c_i(n) \, f'_i\,r(n)$ and a preceding neuron $\zeta_i$ which
represents $f'_i\,r(n)$. The latter feeds into the former through a
single synaptic connection weighted by $c_i$. The adaptation of the
synaptic strength $c_i$ here is based on the (locally time-averaged) squared
activity of the postsynaptic neuron $\varphi_i$, which again is
information locally available at the synaptic link $c_i$. 

Finally, the most involved circuit offered in Figure
\ref{figMicroCircuitsc} {\bf C} delegates the representation of $c_i$
to a separate neuron $\gamma_i$. Like in the second model, a neuron
$\zeta_i$ which represents $f'_i\,r(n)$ feeds to $\varphi_i$, this
time copying its own activation through a unit connection. The
$\gamma_i$ neuron multiplicatively modulates $\varphi_i$ by its
activation $c_i$. Like in the architecture described in
Figure \ref{figMicroCircuitH}, I do not commit to a specific
biological mechanism for such a multiplicative modulation. The
information needed to adapt the activation $c_i$ of neuron $\gamma_i$
according to (\ref{eqcAdapt}) is, besides $c_i$ itself, the quantity
$z_i = c_i(n) \, f'_i\,r(n)$. The latter is represented in $\varphi_i$
which is postsynaptic from the perspective of $\gamma_i$ and therefore
not directly accessible. However, the input $f'_i\,r(n)$ from neuron
$\zeta_i$ is available at $\gamma_i$, from which the quantity $z_i =
c_i \, f'_i\,r(n)$ can be inferred by neuron $\gamma_i$. The neuron
$\gamma_i$ thus needs to instantiate an intricate activation dynamics
which combines local temporal averaging of $(f'_i\,r(n))^2$ with an
execution of (\ref{eqcAdapt}). A potential benefit of this third
neural circuit over the preceding two is that a representation of
$c_i$ by a neural activation can presumably be biologically adapted on a faster
timescale than the neuron auto-modulation in system {\bf A} or the
synaptic strength adaptation in {\bf B}. 

When I first considered content-addressable memories in this report
(Section \ref{secAutoCMemExample}), an important motivation for doing
so was that storing entire conceptor matrices $C$ for later use in
retrieval is hardly an option for biological systems. This may be
different for conception vectors: it indeed becomes possible to
``store'' conceptors without having to store network-sized objects.
Staying with the notation used in Figure \ref{figMicroCircuitsc}: a
single neuron $\gamma^j$ might suffice to represent and ``store'' a
conception vector $c^j$ associated with a pattern $p^j$. The neuron
$\gamma^j$ would project to all $\varphi_i$ neurons whose states
correspond to the signals $z_i$, with synaptic connection weights
$c^j_i$, and effecting a multiplicative modulation of the activation
of the $\varphi_i$ neurons proportional to these connection weights.
 I am not in a position to judge
whether this is really an option in natural brains. For applications
in machine learning however, using stored conception vectors $c^j$ in
conjunction with RFC systems may be a relevant alternative to using
stored matrix conceptors, because vectors $c^j$ can be stored much
more cheaply in computer systems than matrices.

\emph{A speculative outlook.} I allow myself to indulge in a brief speculation of how
RFC conceptor systems might come to the surface -- literally -- in
mammalian brains. The idea is to interpret the activations of (groups
of) neurons in the neocortical sheet as representing conception
factors $c_i$ or $z_i$ values, in one of the versions shown in Figure
\ref{figMicroCircuitsc} or some other concrete realization of RFC
conceptors. The ``reservoir'' part of RFC systems might be found in
deeper brain structures. When some
patches of the neocortical sheet are activated and others not
(revealed for instance through fMRI imaging or electro-sensitive
dyes), this may then be interpreted as a specific $c^j$ vector being
active. In geometrical terms, the surface of the hyperellipsoid of the
``virtual'' conceptor would be mapped to the neocortical sheet. Since
this implies a reduction of dimension from a hypothetical reservoir
dimension $N$ to the 2-dimensional cortical surface, a dimension
folding as in self-organizing feature maps
\cite{KohonenHonkela07,Obermayeretal90,Galtieretal12} would be necessary. What a
cognitive scientist would call an ``activation of a concept'' would
find its neural expression in such an activation of a dimensionally
folded ellipsoid pertaining to a ``virtual'' conceptor $C_F$
in the cortical sheet. An intriguing further step down speculation
road is to think about Boolean operations on concepts as being
neurally realized through the conceptor operations described in
Sections \ref{secBoolean} -- \ref{subsec:JapVow}. All of this is
still too vague.
Still, some aspects of this picture have already been explored in some
detail in other contexts. Specifically, the series of neural models
for processing serial cognitive tasks in primate brains developed by Dominey
et al.\ \cite{Domineyetal95,Dominey05} combine a reservoir dynamics
located in striatal nuclei with cortical context-providing activation
patterns which shares some  characteristics with the speculations offered
here.

\subsection{A Hierarchical Filtering and Classification
  Architecture}\label{secHierarchicalArchitecture}

A reservoir equipped with some conceptor mechanism does not by
itself serve a purpose. If this computational-dynamical principle is
to be made useful for practical purposes like prediction, classification,
control or others, or if it is to be used in cognitive systems
modeling, conceptor-reservoir modules need to be integrated into
more comprehensive architectures. These  architectures take care of
which data are fed to the conceptor modules, where their output is
channelled, how apertures are adapted, and everything else that is
needed to manage a conceptor module for the overall system purpose. In
this section I present a particular architecture for the purpose of
combined signal denoising and classification as an example. This
(still simple) example introduces a number of features which may be of
more general use when conceptor modules are integrated into
architectures:

\begin{description}
\item[Arranging conceptor systems in bidirectional hierarchies:] 
a higher conceptor module is fed from a lower one by the
output of the latter (bottom-up data flow), while at the same
time the higher module co-determines the  conceptors associated
with the lower one (top-down ``conceptional bias'' control).  
\item[Neural instantiations of  individual conceptors:] Using random
feature conceptors, it becomes possible to economically
store and address individual conceptors. 
  \item[Self-regulating balance between perception and action modes:]
a conceptor-reservoir module is made to run in any mixture of two
fundamental processing modes, (i) being passively driven by external
input and (ii) actively generating an output pattern. The balance
between these modes is autonomously steered by a criterion that arises naturally
in hierarchical conceptor systems.  
\end{description}

A personal remark: the first and last of these three items constituted the
original research questions 
which ultimately guided me to  conceptors. 

\emph{The task.} The input to the system is a timeseries made of
alternating sections of the four patterns $p^1,\ldots,p^4$ used variously before in
this report: two sines of irrational period lengths, and two slightly
differing 5-periodic patterns. This signal is corrupted by strong
Gaussian i.i.d.\ noise (signal-to-noise ratio = 0.5) -- see bottom panels in Fig.\ 
\ref{figcArch}. The task is to classify which of the four patterns is
currently active in the input stream, and generate a clean version of
it. This task is a simple instance of the generic task ``classify and
clean a signal that intermittently comes from different, but familiar,
sources''.

\begin{figure}[htb]
\center
\includegraphics[width=130mm]{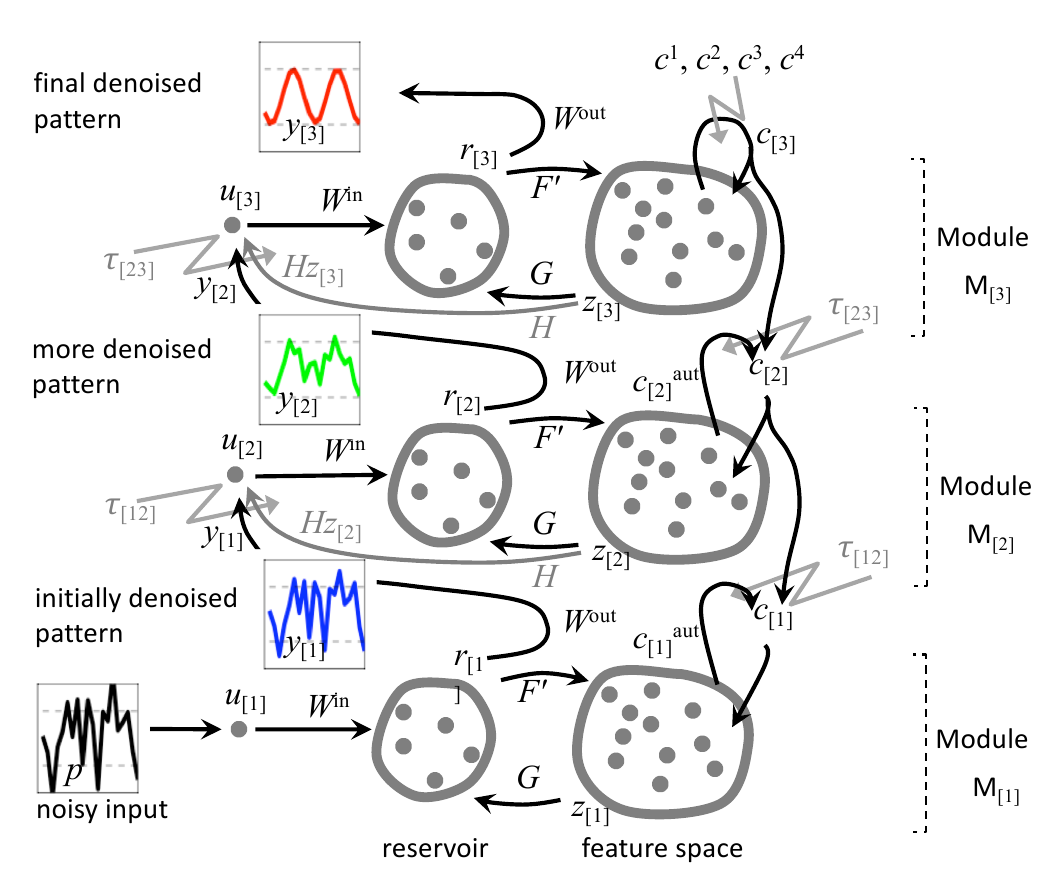}
\caption{Schematic of 3-layer architecture for signal filtering and
  classification. For explanation see text.}
\label{fig3LayerArchitecture}
\end{figure}

\emph{Architecture.} The basic idea is to
stack copies of a reservoir-conceptor loop, giving a hierarchy of such
modules (compare Figure \ref{fig3LayerArchitecture}). Here I present
an example with three layers, having essentially identical modules
$\mathcal{M}_{[1]}$ on the lowest, $\mathcal{M}_{[2]}$ on the middle, and
$\mathcal{M}_{[3]}$ on the highest layer (I  use subscript square brackets
$[l]$ to denote levels in the hierarchy).

Each module is a reservoir-conceptor loop. The conceptor is
implemented here through the $M$-dimensional feature space expansion described in the
previous section, where a high-dimensional conception weight vector
$c$ is multiplied into the feature state (as in Figure
\ref{figBiolArch}). At  exploitation time the state
update equations are 
\begin{eqnarray*}
u_{[l]}(n+1) & = &   (1-\tau_{[l-1,l]}(n))\,y_{[l-1]}(n+1) + \tau_{[l-1,l]}(n)
\,H\,z_{[l]}(n),\\ 
r_{[l]}(n+1) & = & \tanh(G \, z_{[l]}(n) + W^{\mbox{\scriptsize in}}\,u_{[l]}(n+1)  + b
  ), \\
z_{[l]}(n+1) & = &  c_{[l]}(n) \,.\!\ast \, F' r_{[l]}(n+1),\\
y_{[l]}(n+1) & = &  W^{\mbox{\scriptsize out}}\,  r_{[l]}(n+1),
\end{eqnarray*}
where $u_{[l]}$ is the effective signal input to module
$\mathcal{M}_{[l]}$, $y_{[l]}$ is the output signal from that module,
and the $\tau$ are mixing parameters which play a crucial role here
and will be detailed later.  In addition to these fast timescale state
updates there are several online adaptation processes, to be described
later, which adapt $\tau$'s and $c$'s on slower timescales. The weight
matrices $H, G, F, W^{\mbox{\scriptsize in}}, W^{\mbox{\scriptsize
    out}}$ are identical in all modules. $F, W^{\mbox{\scriptsize
    in}}$ are created randomly at design time and remain unchanged.
$H$ and $W^{\mbox{\scriptsize out}}$ are trained on samples of clean
``prototype'' patterns in an initial pattern loading procedure. $G$ is
first created randomly as $G^\ast$ and then is regularized using white
noise (all detail in Section \ref{secDocHierarchical}). 
 
The effective input signal $u_{[l]}(n+1)$ to $\mathcal{M}_{[l]}$ is thus
a mixture mediated by a ``trust'' variable $\tau_{[l-1,l]}$ of a module-external
external input $y_{[l-1]}(n+1)$ and a module-internal input recreation
signal $H\,z_{[l]}(n)$.  On the bottom layer,
$\tau_{[01]} \equiv 0$ and $y_{[0]}(n) = p(n)$, that is, this layer has
no self-feedback input simulation and is entirely driven by the
external input signal $p(n)$. Higher modules receive the output
$y_{[l-1]}(n+1)$ of the respective lower module as their external
input.  Both input mix components  $y_{[l-1]}$ and $H\,z_{[l]}$ represent
partially denoised versions of the external pattern input. The
component $y_{[l-1]}$ from the module below will typically be noisier
than the component $Hz_{[l]}$ that is cycled back within the module,
because each module is supposed to de-noise the signal further in its
internal reservoir-conceptor feedback loop.   If $\tau_{[l-1,l]}$ were to be 1, the module
would be running in an autonomous pattern generation mode and would be
expected to re-generate a very clean version of a stored pattern --
which might however be a wrong one. If
$\tau_{[l-1,l]}$ were to be 0, the module would be running in an
entirely externally driven mode, with no ``cleaning'' in effect.  It
is crucial for the success of this system that these mixing weights
$\tau_{[l-1,l]}$ are appropriately set. They reflect a ``trust'' of the
system in its current hypothesis about the type of the driving
pattern $p$, hence I call them \emph{trust} variables.  I mention at
this point that when the external input $p$ changes from one pattern
type to another, the trust variables must quickly decrease in order to
temporarily admit the architecture to be in an altogether more
input-driven, and less self-generating, mode. All in all this
constitutes a bottom-up flow of information, whereby the raw input $p$
is cleaned stage-wise with an amount of cleaning determined by the
current trust of the system that it is applying the right conceptor to
effect the cleaning.

The output signals $y_{[l]}$ of the three modules are computed from
the reservoir states $r_{[l]}$ by output weights $W^{\mbox{\scriptsize
    out}}$, which are the same on all layers. These output weights are
initially trained in the standard supervised way of reservoir
computing to recover the input signal given to the reservoir from the
reservoir state.  The 3-rd layer output $y_{[3]}$ also is the ultimate
output of the entire architecture and should give a largely denoised
version of the external driver $p$.

Besides this bottom-up flow of information there is a top-down flow of
information. This top-down pathway affects the conception weight
vectors $c_{[l]}$ which are applied in each module. The guiding idea
here is that on the highest layer ($l = 3$ in our example), the
conceptor   $c_{[3]}$ is  of the form
\begin{equation}\label{eqArchcOr}
c_{[3]}(n) = \bigvee_{j=1,\ldots,4} \varphi(c^j, \gamma^j(n)),
\end{equation}
where $c^1, \ldots, c^4$ are \emph{prototype} conception weight
vectors corresponding to the four training patterns. These prototype
vectors are computed and stored at training time. In words, at the
highest layer the conception weight vector is constrained to be a
disjunction of aperture-adapted versions of the prototype conception
weight vectors. Imposing this constraint on the highest layer can be
regarded as inserting a qualitative bias in the ensuing classification
and denoising process. Adapting $c_{[3]}(n)$ amounts to adjusting the
aperture adaptation factors $\gamma^j(n)$.

At any time during processing external input, the current composition
of $c_{[3]}$ as a $\gamma^j$-weighted disjunction of the four
prototypes reflects the system's current \emph{hypothesis} about the
type of the current external input. This hypothesis is stagewise
passed downwards through the lower layers, again mediated by the trust
variables. 

This top-down  pathway is realized as follows. Assume that
$c_{[3]}(n)$ has been computed. In each of the two modules below ($l =
1, 2$), an (auto-)conception weight vector $c^{\mbox{\scriptsize
    aut}}_{[l]}(n)$ is computed by a module-internal execution of the
standard autoconception adaptation described in the previous section
(Equation (\ref{eqcAdapt})). To arrive at the effective conception
weight vector $c_{[l]}(n)$, this $c^{\mbox{\scriptsize aut}}_{[l]}(n)$
is then blended with the current conception weight vector
$c_{[l+1]}(n)$ from the next higher layer, again using the respective
trust variable as mixing coefficient:
\begin{equation}\label{eqArchMixc}
c_{[l]}(n) = (1-\tau_{[l,l+1]}(n)) \, c^{\mbox{\scriptsize aut}}_{[l]}(n)
+ \tau_{[l,l+1]}(n) \, c_{[l+1]}(n).
\end{equation}

In the demo task reported here, the raw input $p$ comes from either of
four sources $p^1, \ldots, p^4$ (the familiar two sines and 5-periodic
patterns), with additive noise. These four patterns are initially
stored in each of the modules $\mathcal{M}_{[l]}$ by training input
recreation weights $H$, as described in Section
\ref{secBiolPlausible}. The same $H$ is used in all layers.

In the exploitation phase (after patterns have been loaded into $H$,
output weights have been learnt, and prototype conceptors $c^j$ have
been learnt), the architecture is driven with a long input sequence
composed of intermittent periods where the current input is chosen
from the patterns $p^j$ in turn. While being driven with $p^j$, the
system must autonomously determine which of the four stored pattern is
currently driving it, assign trusts to this judgement, and accordingly
tune the degree of how strongly the overall processing mode is
autonomously generative (high degree of cleaning, high danger of
``hallucinating'') vs.\ passively input-driven (weak cleaning,
reliable coupling to external driver).

Summing up, the overall functioning of the trained architecture is
governed by two pathways of information flow,
\begin{itemize}
\item a bottom-up pathway where the external noisy input $p$ is
successively de-noised, 
\item a top-down pathway where hypotheses about the current pattern
type, expressed in terms of conception weight vectors, are passed
downwards,
\end{itemize}
and by two online adaptation processes,
\begin{itemize}
\item adjusting the trust variables $\tau_{[l-1,l]}$,  and
\item adjusting the conception weight vector $c_{[l]}$ in the top module.
\end{itemize}

I now describe the two adaptation processes in more detail. 

\emph{Adapting the trust variables.} Before I enter technicalities I
want to emphasize that here we are confronted with a fundamental
problem of information processing in situated intelligent agents
(``SIA'': animals, humans, robots). A SIA continuously has to ``make
sense'' of incoming sensor data, by matching them to the agent's
learnt/stored concepts. This is a multi-faceted task, which appears in
many different instantiations which likely require specialized
processing strategies. Examples include online speech understanding,
navigation, or visual scene interpretation. For the sake of this
discussion I will lump them all together and call them ``online data
interpretation'' (ODI) tasks. ODI tasks variously will involve
subtasks like de-noising, figure-ground separation, temporal
segmentation, attention control, or novelty detection. The demo
architecture described in this section only addresses de-noising and
temporal segmentation. In many cognitive architectures in the
literature, ODI tasks are addressed by maintaining an online
representation of a ``current best'' interpretation of the input data.
This representation is generated in ``higher'' levels of a processing
hierarchy and is used in a top-down fashion to assist lower levels,
for instance by way of providing statistical bias or predictions
(discussion in \cite{Clark12}). This top-down information then tunes
the processing in lower levels in some way that enables them to
extract from their respective bottom-up input specific features while
suppressing others -- generally speaking, by making them selective. An
inherent problem in such architectures is that the agent must not grow
overly confident in its top-down pre-conditioning of lower processing
layers.  In the extreme case of relying entirely on the current
interpretation of data (instead of on the input data themselves), the
SIA will be hallucinating. Conversely, the SIA will perform poorly
when it relies entirely on the input data: then it will not
``understand'' much of it, becoming unfocussed and overwhelmed by
noise. A good example is semantic speech or text understanding.
Linguistic research has suggested that fast \emph{forward inferences}
are involved which predispose the SIA to interpret next input in terms
of a current representation of semantic context (for instance,
\cite{Shastri99a}).  As long as this current interpretation is
appropriate, it enables fast semantic processing of new input; but
when it is inappropriate, it sends the SIA on erroneous tracks which
linguists call ``garden paths''. Generally speaking, for robust ODI an
agent should maintain a reliable measure of the degree of trust that
the SIA has in its current high-level interpretation. When the trust
level is high, the SIA will heavily tune lower levels by higher-level
interpretations (top-down dominance), while when trust levels are low,
it should operate in a bottom-up dominated mode.

Maintaining an adaptive measure of trust is thus a crucial objective
for an SIA. In Bayesian architectures (including Kalman filter
observers in control engineering), such a measure of trust is directly
provided by the posterior probability $p(\mbox{interpretation} \;|\;
\mbox{data})$.  A drawback here is that a number of potentially
complex probability distributions have to be learnt beforehand and may
need extensive training data, especially when prior hyperdistributions have
to be learnt instead of being  donated by an oracle.  In mixture of
predictive expert models (for instance \cite{WolpertKawato98}),
competing interpretation models are evaluated online in parallel and
are assigned relative trust levels according to how precisely they can
predict the current input. A problem that I see here is
computational cost, besides the biological implausibility of
executing numerous predictors in parallel. In adaptive
resonance theory \cite{Grossberg13}, the role of a trust measure
is filled by the ratio between the norm of a top-down pattern
interpretation over the norm of an input pattern; the functional
effects of this ratio for further processing depends on whether that
ratio is less or greater than a certain ``vigilance'' parameter.
Adaptive resonance theory however is primarily a static pattern
processing architecture not designed for online processing of temporal
data.

Returning to our demo architecture, here is how trust variables are
computed. They are based on auxiliary quantities
$\delta_{[l]}(n)$  which are computed within each module $l$. Intuitively,
$\delta_{[l]}(n)$  measures the (temporally smoothed) discrepancy between
the external input signal fed to the module and the self-generated,
conceptor-cleaned version of it. For layers $l > 1$ the external input signal
is the bottom-up passed output $y_{[l-1]}$ of the lower layer. The
conceptor-cleaned, module-generated version is the signal $H z_{[l]}(n)$
extracted from the conception-weighted feature space signal $z_{[l]}(n) =
 c_{[l]}(n) \,.\!\ast\, F' r_{[l]}(n)$ by the
input recreation weights $H$, where $r_{[l]}(n)$ is the reservoir state in
layer $l$. Applying exponential smoothing with smoothing rate $\sigma < 1$,
and normalizing by the likewise smoothed variance of $y_{[l-1]}$, gives
update equations 
\begin{eqnarray}
\bar{y}_{[l-1]}(n+1) & = &  \sigma \, \bar{y}_{[l-1]}(n) +
(1-\sigma)\, y_{[l-1]}(n+1),  
\mbox{ (running average)}\label{eqUpdatediscrepancy1}\\ 
\overline{\mbox{var}}\,y_{[l-1]}(n+1) & = & \nonumber\\
&& \!\!\!\!\!\!\!\!\!\!\!\!\!\!\!\sigma \,
\overline{\mbox{var}}\,y_{[l-1]}(n+1) + (1-\sigma)\, (y_{[l-1]}(n+1) -
\bar{y}_{[l-1]}(n+1))^2, \label{eqUpdatediscrepancy2}  \\
\delta_{[l]}(n+1) & = & \sigma \,\delta_{[l]}(n) +  (1-\sigma)\,
\frac{(y_{[l-1]}(n+1) - 
  Hz_{[l]}(n+1))^2}{\overline{\mbox{var}}\,y_{[l-1]}(n+1)}\label{eqUpdatediscrepancy} 
\end{eqnarray}
for the module-internal detected discrepancies $\delta_{[l]}$. In the bottom
module $\mathcal{M}_{[1]}$, the same procedure is applied to obtain
$\delta_{[1]}$ except that the module input is here the external driver
$p(n)$ instead the output $y_{[l-1]}(n)$ from the level below. 

From these three discrepancy signals $\delta_{[l]}(n)$ two trust variables
$\tau_{[12]}, \tau_{[23]}$ are derived. The intended semantics of
$\tau_{[l,l+1]}$ can be stated as ``measuring the degree by which the
discrepancy is reduced when going upwards from level $l$ to level
$l+1$''.  The rationale behind this is that when the currently active
conception weights in modules $l$ and $l+1$ are appropriate for the
current drive entering module $l$ from below (or from the outside when
$l = 1$), the discrepancy should \emph{decrease} when going from level $l$ to level
$l+1$, while if the the currently applied conception weights are the
wrong ones, the discrepancy should \emph{increase} when going
upwards. The  core of measuring trust  is thus the difference
$\delta_{[l]}(n) - \delta_{[l+1]}(n)$, or rather (since  we want the same
sensitivity across all levels of absolute values of $\delta$) the
difference  $\log(\delta_{[l]}(n)) - \log(\delta_{[l+1]}(n))$. 
Normalizing this to a range of $(0,1)$
by applying a logistic sigmoid with steepness $d_{[l,l+1]}$ finally gives
\begin{equation}\label{eqdefTrust}
\tau_{[l,l+1]}(n) = \left(1 + \left(\frac{\delta_{[l+1]}(n)} {\delta_{[l]}(n)}
\right)^{d_{[l,l+1]}}\right)^{-1}. 
\end{equation}
The steepness $d_{[l,l+1]}$ of the trust sigmoid is an important design
parameter, which currently I set manually. Stated in intuitive terms
it determines how ``decisively'' the system follows its own trust
judgement. It could be rightfully called a ``meta-trust'' variable,
and should itself be adaptive. Large values of this \emph{decisiveness} leads
the system to make fast decisions regarding the type of the current
driving input, at an increased risk of settling down prematurely on a
wrong decision. Low values of $d_{[l,l+1]}$ allow the system to take
more time for making a decision, consolidating information acquired
over longer periods of possibly very noisy and only weakly
pattern-characteristic input. My current view on the regulation of
decisiveness is that it  cannot be regulated on the sole
basis of the information contained in input data, but reflects 
higher cognitive capacities (connected to mental attitudes like
``doubt'', ``confidence'', or even ``stubbornness''...)
which are intrinsically not entirely data-dependent.

\emph{Adapting the top-level conception weight vectors $c_{[l]}$.} For
clarity of notation I will omit the level index $[l]$ in what
follows, assuming throughout $l = 3$. 
By equation
(\ref{eqArchcOr}), the effective conception weight vector used in the
top module will be constrained to be a disjunction
$c(n) = \bigvee_{j=1,\ldots,4} \varphi(c^j, \gamma^j(n))$,
where the $c^j$ are prototype conception weight vectors,
computed  at training time. 
Adapting $c(n)$ amounts to adjusting the apertures of the
disjunctive components $c^j$ via $\gamma^j(n)$. This is done
indirectly. 

The training of the prototype conception weights (and of the input
recreation matrix $H$ and of the readout weights $W^{\mbox{\scriptsize
    out}}$) is done with a single module that is driven by the clean
patterns $p^j$. Details of the training procedure are given in the
Experiments and Methods Section \ref{secDocHierarchical}.  The
prototype conception weight vectors can be written  as
\begin{displaymath}
c^j = E[(z^j)^{.\wedge 2}] \, .\!\ast \, ( E[(z^j)^{.\wedge 2}] +
\alpha^{-2})^{.\wedge -1}, 
\end{displaymath}
where $z^j(n) = c^j \, .\!\ast \, F'\, r^j(n)$ is the $M$-dimensional
signal fed back from the feature space to the reservoir while the
module is being driven with pattern $j$ during training, and the
aperture $\alpha$ is a design parameter. Technically, we do not
actually store the $c^j$ but their constituents $\alpha$ and the
corresponding mean signal energy vectors $E[(z^j)^{.\wedge 2}]$, the
latter of which are collected in an $M \times 4$ \emph{prototype}
matrix
\begin{equation}\label{eqZmatrix}
P = (E[(z^j_i)^2])_{i = 1,\ldots,M; \; j = 1,\ldots,4}.
\end{equation} 
I return to the conceptor adaptation dynamics in the top module at
exploitation time.  
Using results from  previous sections, equation (\ref{eqArchcOr})  can be re-written
as  
\begin{equation}\label{eqcn}
c(n) = \left(\sum_j   (\gamma^j(n))^{.\wedge 2} \, .\!\ast \,
  E[(z^j)^{.\wedge 2}]\right) \, .\!\ast \, \left(\sum_j
  (\gamma^j(n))^{.\wedge 2} \, .\!\ast \, E[(z^j)^{.\wedge 2}] +
  \alpha^{-2}\right)^{.\wedge -1}, 
\end{equation}
where the $+\alpha^{-2}$ operation is applied
component-wise to its argument vector.  The strategy for adapting
the factors $\gamma^j(n)$ is to minimize the loss function
\begin{equation}\label{eqGammaLoss}
\mathcal{L} \{ \gamma^1,\ldots,\gamma^4\} = \|\sum_j
  (\gamma^j)^{.\wedge 2} \,  
  E[(z^j)^{.\wedge 2}] - E[z^{.\wedge 2}]  \|^2, 
\end{equation}
where $z$ is the feature space output signal $z(n) = c(n) \, .\!\ast \, 
F' r(n)$ available during exploitation time in the top module.  In
words, the adaptation of $c$ aims at finding a weighted disjunction of
prototype vectors which optimally matches the currently observed mean
 energies of the $z$ signal.

It is straightforward to derive a
stochastic gradient descent adaptation rule for minimizing the loss
(\ref{eqGammaLoss}). Let $\gamma = (\gamma^1,\ldots,\gamma^4)$ be the
row vector made from the $\gamma^j$, and let $\cdot.^2$ denote
element-wise squaring of a vector.  Then
\begin{equation}\label{eqAdaptGamma}
\gamma(n+1)  =  \gamma(n) + \lambda_\gamma \; \left(z(n+1)^{.\wedge 2} - P\,
(\gamma'(n))^{.\wedge 2}\right)' \; P\,\mbox{diag}(\gamma(n))  
\end{equation}
implements the stochastic gradient of $\mathcal{L}$ with respect to
$\gamma$, where $\lambda_\gamma$ is an adaptation rate. In fact I do
not use this formula as is, but add two helper mechanisms, effectively
carrying out
\begin{eqnarray}
\gamma^\ast(n+1)  & = & \gamma(n) + \nonumber\\ 
 && \!\!\!\!\!\!\!\!\!\!\!\!\!\!\!\!\!\!\!\!\lambda_\gamma  \left( \left(z(n+1)^{.\wedge 2} - P\,
(\gamma'(n))^{.\wedge 2}\right)' \; P\,\mbox{diag}(\gamma(n)) + d\,(1/2 - \gamma(n))\right)\label{eqGammaAdapt1}\\
 \gamma(n+1) & = & \gamma^\ast(n+1) / \mbox{sum}(\gamma^\ast(n+1)).
\label{eqGammaAdapt2}
\end{eqnarray}
The addition of the term $d\,(1/2 - \gamma(n))$ pulls the $\gamma^j$
away from the possible extremes $0$ and $1$ toward $1/2$ with a
\emph{drift} force $d$, which is a design parameter. This is helpful
to escape from extreme values (notice that if $\gamma^j(n) = 0$, then
$\gamma^j$ would forever remain trapped at that value in the absence of
the drift force). The normalization (\ref{eqGammaAdapt2}) to a unit
sum of the $\gamma^j$ greatly reduces adaptation jitter. I found both
amendments crucial for a reliable performance of the $\gamma$
adaptation.

Given the $\gamma(n)$ vector, the top-module $c(n)$ is obtained by
$$c(n) = \left(P\,(\gamma'(n))^{.\wedge 2}\right) \, .\!\ast \,
\left(P\,(\gamma'(n))^{.\wedge 2} + \alpha^{-2}  \right)^{.\wedge
  -1}.$$

\begin{figure}[htbp]
\center
\includegraphics[width=140mm]{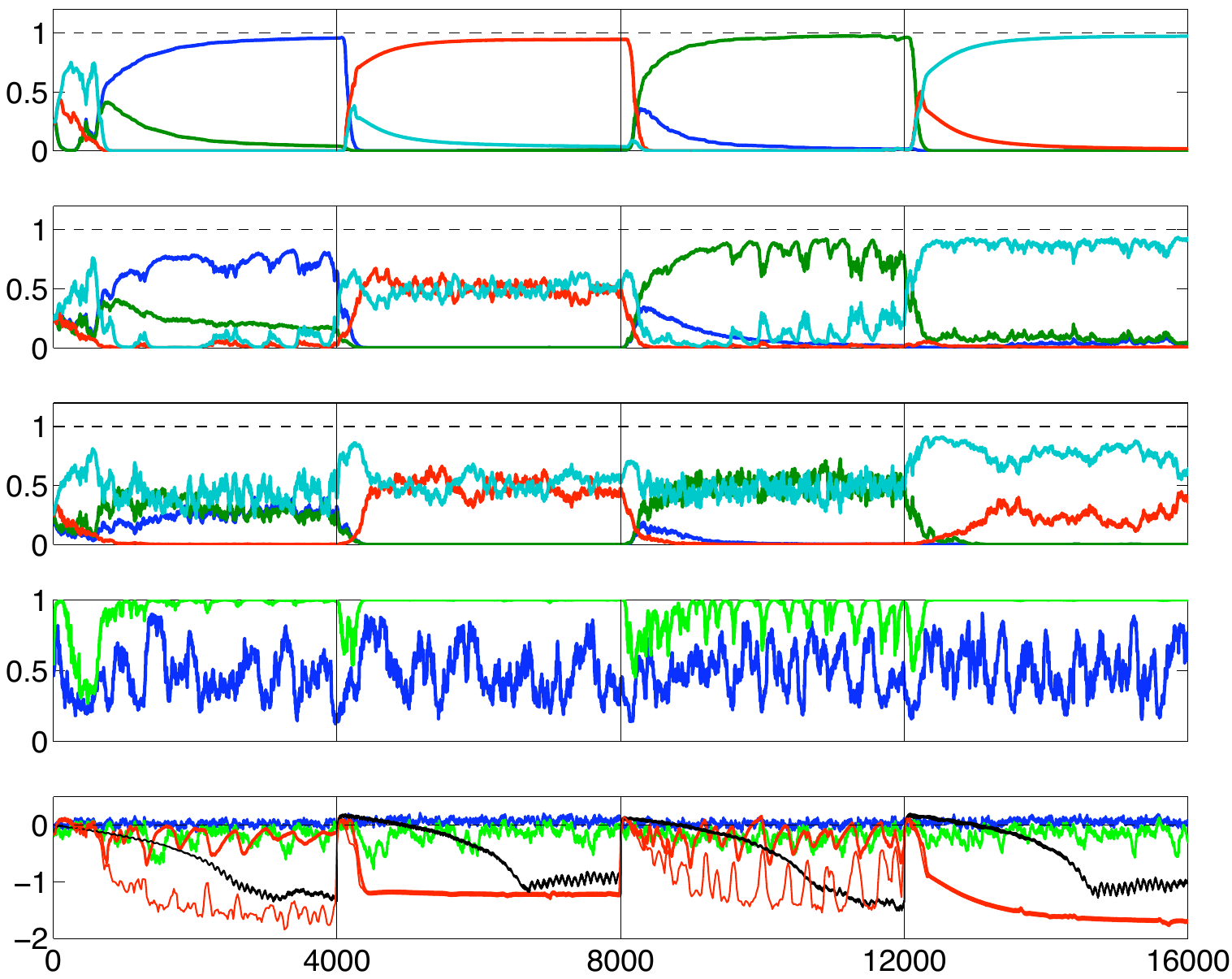}\\
\vspace{4mm}
\includegraphics[width=135mm]{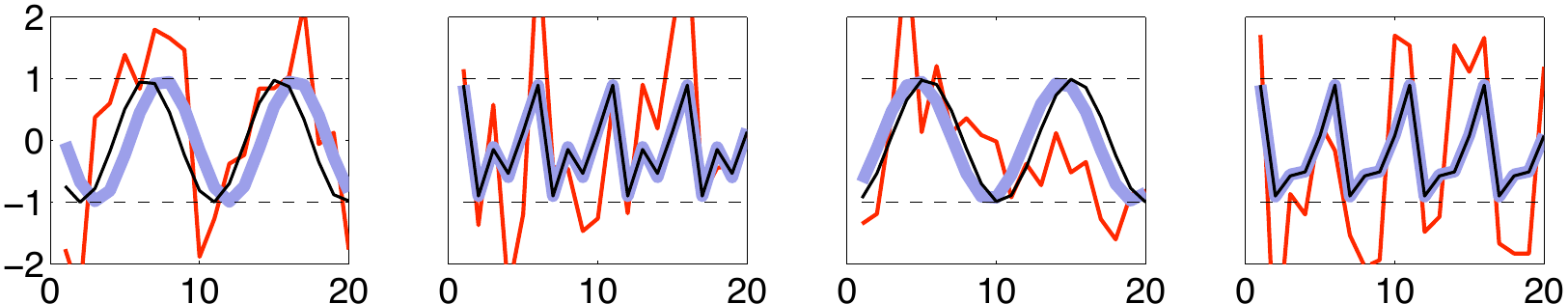}
\caption{Denoising and
  classification of prototype patterns. Noisy patterns
  were given as input in the order $p^1, p^3, p^2, p^4$ for 4000
  timesteps each.  Top row: evolution of  $\gamma^1$ (blue),
  $\gamma^2$ (green), $\gamma^3$ (red), $\gamma^4$ (cyan) in top layer module. Rows 2
  and 3: same in modules 2 and 1. Fourth row:  trust variables
  $\tau_{[12]}$ (blue) and $\tau_{[23]}$ (green).  Fifth row:
  NRMSEs for reconstructed signals $y_{[1]}$ (blue), $y_{[2]}$ (green) and
  $y_{[3]}$ (red). Black line shows the linear filter reference
  NRMSE. Thin red line: NRMSE of phase-aligned $y_{[3]}$. The
  plotting scale is logarithmic base 10. Bottom: pattern
  reconstruction snapshots from the last 20 timesteps in each pattern
  presentation block, showing the noisy input (red),
  the layer-3 output $y_{[3]}$ (thick gray) and the clean signal (thin black). For
  explanation see text.}
\label{figcArch}
\end{figure}

\emph{Simulation Experiment 1: Online Classification and De-Noising.}
Please consult Figure \ref{figcArch} for a graphical display of this
experiment. Details (training, initalization, parameter settings) are
provided in the Experiments and Methods section \ref{secDocHierarchical}. 

The trained 3-layer architecture was driven with a
16,000 step input signal composed of four blocks of 4000 steps
each. In these blocks, the input was generated from the patterns $p^1,
p^3, p^2, p^4$ in turn (black lines in bottom row
of Figure  \ref{figcArch}), with additive Gaussian noise scaled such that
a signal-to-noise ratio of 1/2 was obtained (red lines in bottom row
of Figure  \ref{figcArch}). The 3-layer architecture was run for the
16,000 steps without external intervention. 

The evolution of the four $\gamma^j$ weights in the top layer
represent the system's classification hypotheses concerning the type of the
current driver (top row in Figure \ref{figcArch}). In all four blocks,
the correct decision is reached after an initial ``re-thinking''
episode. The trust variable $\tau_{[23]}$ quickly approaches 1 after
taking a drop at the block beginnings (green line in fourth row of
Figure). This drop allows the system to partially pass through the external
driver signal up to the top module, de-stabilizing the hypothesis
established in the preceding block. The trust variable   $\tau_{[12]}$
(blue line in fourth row) oscillates more irregularly but also takes
its steepest drops at the block beginnings. 

For diagnostic purposes, $\gamma^j_{[l]}$ weights were also computed
on the lower layers $l = 2,3$, using (\ref{eqGammaAdapt1}) and
(\ref{eqGammaAdapt2}). These quantities, which were not entering the
system's processing, are indicators of the ``classification belief
states'' in lower modules (second and third
row in the Figure). A stagewise consolidation of these hypotheses can
be observed as one passes upwards through the layers. 

The four patterns fall in two natural classes, ``sinewaves'' and
``period-5''. Inspecting the top-layer $\gamma^j$ it can be seen that
within each block, the two hypothesis indicators associated with the
``wrong'' class are quickly suppressed to almost zero, while the two
indicators of the current driver's class quickly dominate the picture
(summing to close to 1) but take a while to level out to
their relative final values. Since the top-level $c(n)$ is formally a
$\gamma^j$-weighted disjunction of the four prototype $c^j$ conception
vectors, what happens in the first block (for instance) can also be
rephrased as, ``after the initial re-thinking, the system is confident
that the current driver is $p^1$ OR $p^3$, while it is quite sure that
it is NOT ($p^2$ OR $p^4$)''. Another way to look at the same
phenomena is to say, ``it is easy for the system to quickly decide between
the two classes, but it takes more time to distinguish the rather
similar patterns within each class''. 

The fifth row in Figure  \ref{figcArch} shows the log10 NRMSE (running
smoothed average) of the three module outputs $y_{[l]}(n)$ with
respect to a clean version of the driver (thick lines; blue =
$y_{[1]}$, green = $y_{[2]}$, red = $y_{[3]}$). For the 5-periodic
patterns (blocks 2 and 4) there is a large increase in accuracy from
$y_{[1]}$ to $y_{[2]}$ to $y_{[3]}$. For the sinewave patterns this
is not the case, especially not in the first block. The reason is that
the re-generated sines $y_{[2]}$ and $y_{[3]}$ are not perfectly
phase-aligned to the clean version of the driver. This has to be
expected because such relative phase shifts are typical for coupled
oscillator systems; each module can be regarded as an
oscillator. After optimal phase-alignment (details in the Experiments
and Methods section), the top-level sine re-generation matches the
clean driver very accurately (thin red line). The 5-periodic signal
behaves differently in this respect. Mathematically, an
5-periodic discrete-time dynamical system attractor is not an oscillation but
a  fixed point of the 5-fold iterated map, not admitting anything like
a gradual phase shift. 

As a baseline reference, I also trained a linear transversal filter
(Wiener filter, see \cite{FarhangBoroujeny98} for a textbook treatment) on
the task of predicting the next clean input value (details in the Experiments
and Methods section). The length of this filter was 2600, matching
the number of trained parameters in the conceptor architecture ($P$
has 500 $\ast$ 4 learnt parameters, $H$ has 500, $W^{\mbox{\scriptsize
  out}}$ has 100). The smoothed log10 NRMSE of this linear predictor is
plotted as a black line. It naturally can reach its best prediction levels only
after 2600 steps in each block, much more slowly than the conceptor
architecture. Furthermore, the ultimate accuracy is inferior for all
four patterns.   

\emph{Simulation Experiment 2: Tracking Signal Morphs.}   Our
architecture can be characterized as a signal
cleaning-and-interpretation systems which guides itselft by allowing
top-down hypotheses to make lower processing layers selective. An
inherent problem in such systems is that that they may erroneously
lock themselves on false hypotheses. Top-down hypotheses  are
self-reinforcing to a certain degree because
they cause lower layers to filter out data components that do not
agree with the hypothesis -- which is the essence of de-noising after
all. 

In order to test how our architecture fares with respect to this
``self-locking fallacy'', I re-ran the simulation with an  input sequence that 
was organized as a linear morph from $p^1$ to $p^2$ in the first 4000 steps
(linearly ramping up the sine frequency), then in the next block back
to $p^1$; this was followed by a morph from $p^3$ to $p^4$ and back
again. The task now is to keep track of the morph mixture in the
top-level $\gamma^j$. This is a greatly more difficult task than the previous one
because the system does not have to just decide between 4 patterns,
but has to keep track of minute changes in relative mixtures. The
signal-to-noise ratio of the external input was kept at 0.5. The outcome reveals an
interesting qualitative difference in how the system copes with the
sine morph as opposed to the 5-periodic morph. As can be seen in
Figure \ref{figcArchMorph}, the highest-layer hypothesis indicators
$\gamma^j$ can track the frequency morph of the sines (albeit with a
lag), but get caught in a constant hypothesis for the 5-period morph.

This once again illustrates that irrational-period sines are treated
qualitatively differently from integer-periodic signals in conceptor
systems. I cannot offer a mathematical analysis, only an intuition.
In the sinewave tracking, the overall architecture can be described as
a chain of three coupled oscillators, where the bottom oscillator is
externally driven by a frequency-ramping sine. In such a driven chain
of coupled oscillators, one can expect either chaos, the occurrence of
natural harmonics, or frequency-locking across the elements of the
chain. Chaos and harmonics are ruled out in our architecture because
the prototype conceptors and the loading of two related basic sines
prevent it. Only frequency-locking remains as an option, which is
indeed what we find. The 5-periodic morph cannot benefit from this
oscillation entrainment. The minute differences in shape between the two
involved prototypes do not stand out strongly enough from the noise
background to induce a noticable decline in the trust variable
$\tau_{[23]}$: once established, a single hypothesis persists. 

On a side note it is interesting to notice that the linear filter that
was used as a baseline cannot at all cope with the frequency sweep,
but for the 5-periodic morph it performs as well as in the previous
simulation. Both effects can easily be deduced from the nature of such
filters. 

Only when I used a much cleaner input signal (signal-to-noise ratio of 10),
and after the decisiveness $d$ was reduced to $0.5$, it became
possible for the system to also track the 5-period pattern morph,
albeit less precisely than it could track the sines  (not
shown).

\begin{figure}[htbp]
\center
\includegraphics[width=140mm]{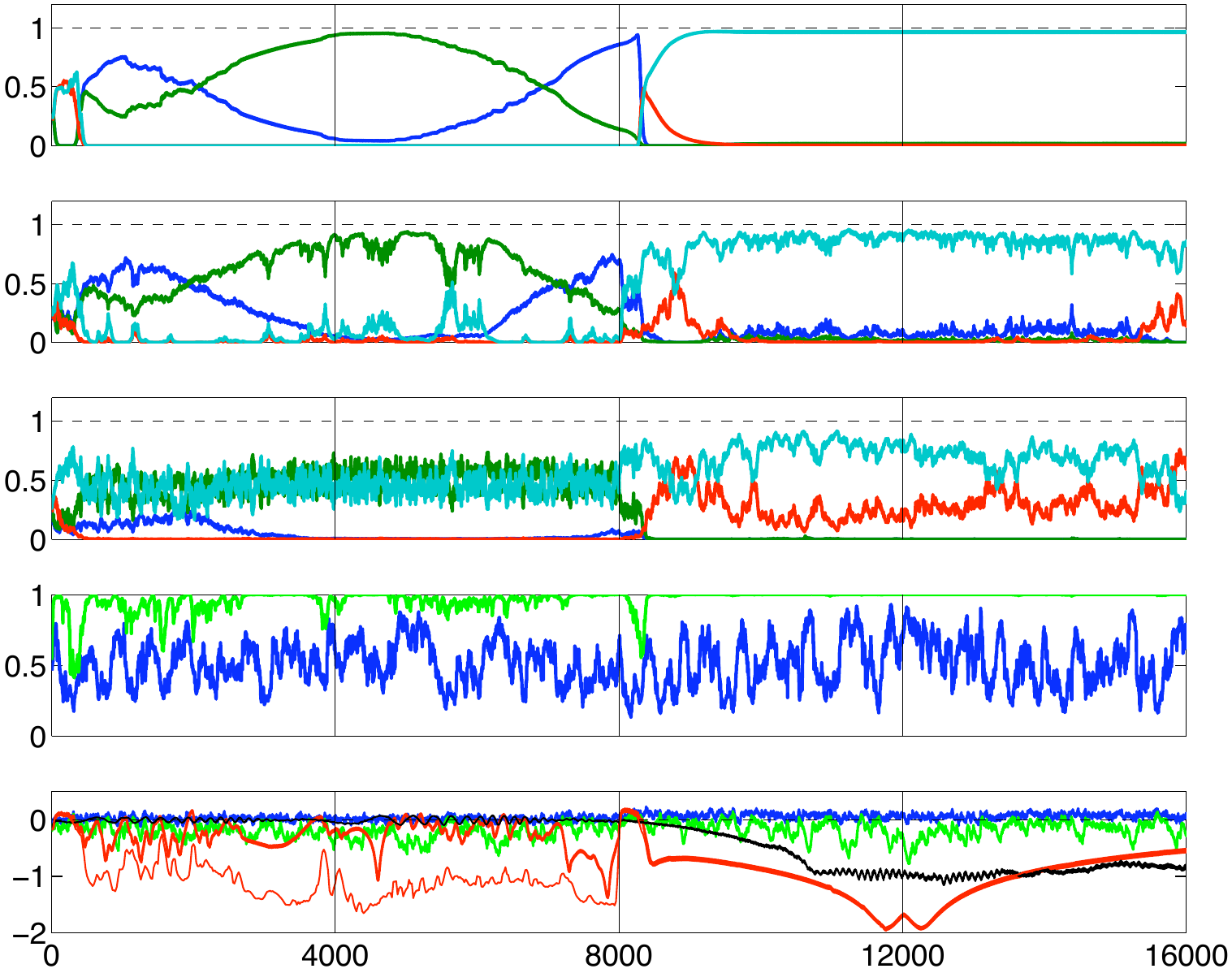}\\
\vspace{2mm}
\includegraphics[width=135mm]{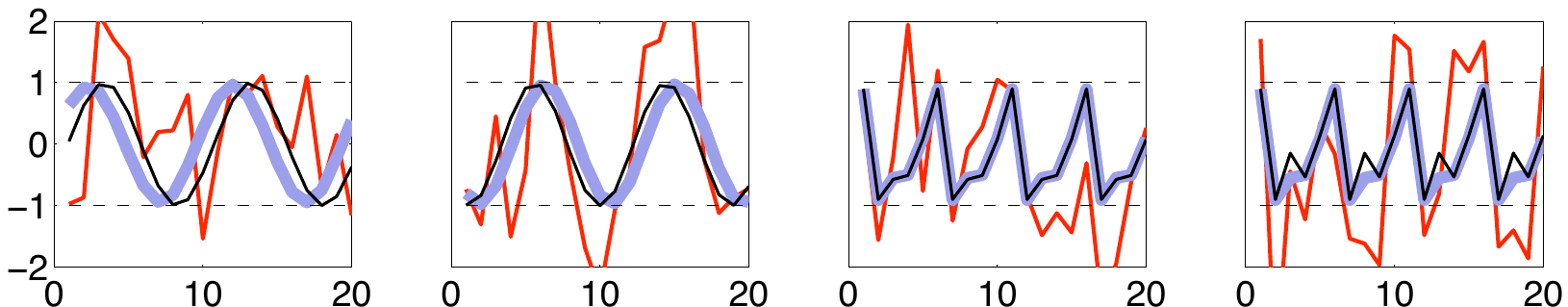}
\caption{Morph-tracking.  First 4000 steps: morphing from
  $p^1$ to $p^2$, back again in next 4000 steps; steps 8000 -- 16,000:
  morphing from $p^3$ to $p^4$ and back again. Figure layout same as
  in Figure \ref{figcArch}.}
\label{figcArchMorph}
\end{figure}

\emph{Variants and Extensions.} When I first experimented with
architectures of the kind proposed above, I computed the
module-internal conception weight vectors $c_{[l]}^{\mbox{\scriptsize
    aut}}$ (compare Equation (\ref{eqArchMixc})) on the two lower
levels not via the autoconception mechanism, but in a way that was
similar to how I computed the top-level conception weight vector
$c_{[3]}$, that is, optimizing a fit to a disjunction of the four
prototypes. Abstractly speaking, this meant that  a powerful piece of
prior information,
namely of knowing that the driver was one of the four prototypes, was
inserted in all processing layers. This led to a better system
performance than what I reported above (especially, faster decisions
in the sense of faster convergence of the $\gamma_{[3]}$). However I
subsequently renounced this ``trick'' because the differences in
performance were only slight, and from a cognitive modelling
perspective I found it more appealing to insert such a valuable
prior only in the top layer (motto: the retina does not conceptually understand
what it sees).

Inspecting again the top row in Figure \ref{figcArch}, one finds 
fast initial decision between the alternatives ``pattern 1 or  2''
versus ``pattern 3 or 4'', followed by a much slower
differentation within these two classes. This suggests architecture
variants where all layers are informed by priors of the kind as in
Equation (\ref{eqArchcOr}), that is, the local conceptor on a layer is
constrained to an aperture-weighted disjunction of a finite number of
prototype conceptors. However, the number of prototype conceptors
would shrink as one goes upwards in the hierarchy. The reduction in
number would be effected by merging  several
distinct prototype conception vectors $c^{j_1},\ldots,c^{j_k}$ in
layer $l$ into a single prototype vector $c^{j} = \bigvee
\{c^{j_1},\ldots,c^{j_k}\}$. In terms of classical AI
knowledge representation formalisms this would mean to implement an
abstraction hierarchy. A further refinement that suggests itself would
be to install a top-down processing pathway by which the current
hypothesis on layer $l+1$ selects which
finer-grained disjunction of prototypes on layer $l$ is chosen. For
instance, when $c_{[l+1]}(n) = c^j$ and $c^j = \bigvee
\{c^{j_1},\ldots,c^{j_k}\}$, then the conception weight vector
$c_{[l]}(n)$ is constrained to be of the form
$\bigvee_{i=1,\ldots,k}\varphi(c^{j_i},
\gamma^{j_i}_{[l]}(n))$. This remains for future work. 

The architecture presented above is replete with ad-hoc design
decisions. Numerous details could have been realized
differently. There is no  unified theory which could inform a
system designer what are the ``right'' design decisions. A complete SIA
architecture must provide a plethora of dynamical mechanisms for
learning, adaptation, stabilization, control, attention and so forth, and each of
them in multiple versions tailored to different subtasks and temporal
scales. I do not see even a theoretical possibility for an
overarching, principled theory which could afford us with rigorous
design principles for all of these. The hierarchical conceptor
architecture presented here is far from realizing a complete SIA
system, but repercussions of that under-constrainedness of design
already show. 

\emph{Discussion.} Hierarchical neural learning architectures for
pattern recognition have been proposed in many variants (examples:
\cite{LeCunetal98,Gravesetal09,Friston05,HintonSalakhutdinov06,GedeonArathorn07,Zibneretal11}),
albeit almost always for static patterns. The only example of
hierarchical neural architectures for temporal pattern recognition
that I am aware of are the localist-connectionistic SHRUTI networks
for text understanding \cite{Shastri99a}. Inherently temporal
hierarchical pattern classification is however realized in standard
hidden-Markov-model (HMM) based models for speech recognition. 

There is one
common characteristic across all of these hierarchical recognition
systems (neural or otherwise, static or temporal). This shared trait is that
when one goes upward through the processing layers, increasingly
 ``global'' or ``coarse-grained'' or ``compound''
features are extracted (for instance, local edge detection in early
visual processing leading through several stages to object
recognition in the highest layer). While the concrete nature of
this layer-wise integration of information differs between approaches,
at any rate there is change of represented categories across layers.
For the sake of discussion, let me refer to this  as the
\emph{feature integration}  principle. 

From the point of view of logic-based knowldege representation,
another important trait is shared by hierarchical pattern recognition
systems: abstraction. The desired highest-level output is a class
labelling of the input data. The recognition architecture has to be
able to \emph{generalize} from the particular input instance. This
abstraction function is not explicitly implemented in the layers of
standard neural (or HMM) recognizers. In rule-based decision-tree
classification systems (textbook: \cite{Mitchell97}), which can also
be regarded as hierarchical recognition systems, the hierarchical
levels however directly implement a series of class abstractions. I
will refer to the abstraction aspect of pattern recognition as the
\emph{categorical abstraction}  principle.

The conceptor-based architecture presented in this section implements
categorical abstraction through the $\gamma$ variables in the
highest layer. They yield (graded) class judgements similar to what is
delivered by 
the class indicator variables in the top layers of typical neural
pattern recognizers.  

The conceptor-based architecture  is
different from the typical neural recognition systems in that it does
not implement the feature integration principle.  As one
progresses upwards through the layers, always the same dynamic item is
represented, namely, the current periodic pattern, albeit in
increasingly denoised versions. I will call this the \emph{pattern
  integrity} principle. The pattern integrity principle is inherently
conflicting with the feature integration principle. 

By decades of theoretical research and successful pattern recognition
applications, we have become accustomed to the feature integration
principle. I want to argue that the pattern integrity principle
has some cognitive plausibility and should be considered when one
designs SIA architectures.
 
Consider the example of listening to a familiar piece of music from a
CD player in a noisy party environment. The  listener is
capable of two things. Firstly, s/he can classify the piece of music,
for instance by naming the title or the band. This corresponds to
performing categorical abstraction, and this is what standard pattern
recognition architectures would aim for. But secondly, the listener
can also overtly sing (or whistle or hum) along with the melody, or
s/he can mentally entrain to the melody. This overt or covert
accompaniment has strong de-noising characteristics -- party talk
fragments are filtered out in the ``mental tracing'' of the melody.
Furthermore, the mental trace is temporally entrained to the source
signal, and captures much temporal and dynamical detail
(single-note-level of accuracy, stressed versus unstressed beats,
etc). That is an indication of pattern integrity.

Another example of pattern integrity: viewing the face of a
conversation partner during a person-to-person meeting, say Anne
meeting Tom. Throughout the conversation, Anne knows that the visual
impression of Tom's face is indeed \emph{Tom}'s face: categorical
abstraction is happening. But also, just
like a listener to noise-overlaid music can trace online the clean
melody, Anne maintains a ``clean video'' representation of Tom's face
as it undergoes a succession of facial expressions and head motions.
Anne experiences more of Tom's face than just the top-level
abstraction that it is Tom's; and this online experience is entrained
to Anne's visual input stream.

Pattern integrity could be said to be realized in standard
hierarchical neural architectures to the extent that they are
\emph{generative}. Generative models afford of mechanisms by which
example instances of a recognition class can be actively produced by
the system. Prime examples are the architectures of adaptive resonance
theory \cite{Grossberg13}, the Boltzmann Machine \cite{Ackleyetal85}
and the Restricted Boltzmann Machine / Deep Belief Networks
\cite{HintonSalakhutdinov06}. In systems of this type, the
reconstruction of pattern instances occurs (only) in the input layer,
which can be made to ``confabulate'' or ``hallucinate'' (both terms
are used as technical terms in the concerned literature) pattern
instances when primed with the right bias from higher layers. 

Projecting
such architectures to the human brain (a daring enterprise) and
returning to the two examples above, this would correspond to
re-generating melodies in the early auditory cortex or facial
expressions in the early visual cortex (or even in the retina). But I
do not find this a convincing model of what happens in a human brain.
Certainly I am not a neuroscientist and not qualified to make
scientific claims here. My doubts rest on introspection (forbidden! I
know) and on a computational argument. Introspection: when I am
mentally humming along with a melody at a party, I still do
\emph{hear} the partytalk -- I dare say my early auditory modules keep
on being excited by the entire auditory input signal. I don't feel
like I was hallucinating a clean version of the piece of music, making
up an auditory reality that consists only of clean music. But I
do not \emph{listen} to the talk noise, I listen only to the music
components of the auditory signal. The reconstruction of a clean
version of the music happens -- as far as I can trust my introspection
-- ``higher up'' in my brain's hierarchy, closer to the quarters where
consciously controllable cognition resides. The computational
argument: generative models, such as the ones mentioned, cannot (in
their current versions at least)  generate clean versions of
noisy input patterns while the input is presented. They either produce
a high-level classification response while being exposed to input
(bottom-up processing mode), or they generate patterns in their
lowest layer while being primed to a particular class in their
highest layer (top-down mode). They can't do both at the same time. But
humans can: while being exposed to input, a cleaned-up version of the
input is being maintained. Furthermore, humans (and the conceptor
architecture) can operate in an online-entrained mode when driven by
temporal data, while almost all existing recognition architectures in
machine learning are designed for static patterns. 

Unfortunately I cannot offer a clear definition of ``pattern
integrity''. An aspect of pattern integrity that I find important if
not defining is that some temporal and spatial detail of a recognized
pattern is preserved across processing layers.  Even at the highest
layers, a ``complete'' representation of the pattern should be
available. This seems to agree with cognitive theories positing that
humans represent concepts by \emph{prototypes}, and more specifically,
by \emph{exemplars} (critical discussion \cite{Lakoff99}). However,
these cognitive theories relate to empirical findings on human
classification of static, not temporal, patterns.  I am aware that I
am vague. One reason for this is that we lack a scientific
terminology, mathematical models, and standard sets of examples for
discussing phenomena connected with pattern integrity. All I can bring to
the table at this point is just a new architecture that has some
extravagant processing characteristics. This is, I hope,
relevant, but it is premature to connect this in any detail to
empirical cognitive phenomena.

\subsection{Toward a Formal Marriage of Dynamics with Logic}\label{secLogic}

In this subsection I assume a basic acquaintance of the reader with
Boolean and first-order predicate logic.

So far, I have established that conceptor matrices can be combined
with (almost) Boolean operations, and can be ordered by (a version of)
abstraction. In this subsection I explore a way to extend these
observations into a formal ``conceptor logic''.

Before I describe the formal apparatus, I will comment on how I will
be understanding the notions of ``concept'' and ``logic''.  Such a
preliminary clarification is necessary because these two terms are
interpreted quite differently in different contexts.

\emph{``Concepts'' in the cognitive sciences.} I start with a quote
from a recent survey on research on concepts and categories in the
cognitive sciences \cite{MedinRips05}: \emph{``The concept of concepts
  is difficult to define, but no one doubts that concepts are
  fundamental to mental life and human communication.  Cognitive
  scientists generally agree that a concept is a mental representation
  that picks out a set of entities, or a category.  That is, concepts
  \emph{refer}, and what they refer to are categories.  It is also
  commonly assumed that category membership is not arbitrary but
  rather a principled matter.  What goes into a category belongs there
  by virtue of some law-like regularities.  But beyond these sparse
  facts, the concept CONCEPT is up for grabs.'' } Within this research
tradition, one early strand \cite{Quillian67,CollinsQuillian69}
posited that the overall organization of a human's conceptual
representations, his/her \emph{semantic memory}, can be formally well
captured by AI representation formalisms  called
\emph{semantic networks} in later years. In semantic network
formalisms, concepts are ordered in abstraction hierarchies, where a
more abstract concept refers to a more comprehensive category. In
subsequent research this formally clear-cut way of defining and
organizing concepts largely dissolved under the impact of
multi-faceted empirical findings. Among other things, it turned out
that human concepts are graded, adaptive, and depend on features which
evolve by learning. Such findings led to a diversity of enriched
 models of human concepts and their organization (my
favourites: \cite{Lakoff87,Drescher91,LakoffNunez00}), and many
fundamental questions remain controversial. Still, across all
diversity and dispute, the basic conception of concepts spelled out in
the initial quote remains largely intact, namely that concepts are mental
representations of categories, and categories are defined
\emph{extensionally} as a set of ``entities''. The nature
of these entities is however ``up to grabs''. For instance, the
concept named ``Blue'' might be referring to the set of blue physical
objects, to a set of wavelengths of light, or to a set of sensory
experiences, depending on the epistemic approach that is taken.

\emph{``Concepts'' in logic formalisms.}  I first note that the word
``concept'' is not commonly used in logics. However, it is quite clear
what elements of logical systems are equated with concepts when such
systems are employed as models of semantic memory in cognitive
science, or as knowledge representation frameworks in AI.  There is a
large variety of logic formalisms, but almost all of them employ typed
symbols, specifically unary predicate symbols, relation
symbols of higher arity, constant symbols, and function symbols.  In the
model-theoretic view on logic, such symbols become extensionally
\emph{interpreted} by sets of elements defined over the domain set of
a set-theoretic model.  Unary predicate symbols become
interpreted by sets of elements; $n$-ary relation symbols become
interpreted by $n$-tuples of such elements; function symbols by sets
of argument-value pairs; constant symbols by individual elements. A
logic \emph{theory} uses a fixed set of such symbols called the
theory's \emph{signature}. Within a theory, the interpretation of the
signature symbols becomes constrained by the axioms of the theory. In
AI knowledge representation systems, this set of axioms can be very
large, forming a \emph{world model} and \emph{situation model}
(sometimes called ``T-Box'' and ``A-Box''). In the parlance of
logic-oriented AI, the extension of unary predicate symbols are often
called \emph{classes} instead of ``categories''. 

In AI applications,
the world model is often implemented in the structure of a
\emph{semantic network} \cite{Lehmann92}, where the classes are
represented by nodes labelled by predicate symbols. These nodes are
arranged in a hierarchy with more abstract class nodes in
higher levels. This allows the computer program to exploit
inheritance of properties and relations down the hierarchy, reducing
storage requirements and directly enabling many elementary inferences.
Class nodes in semantic networks
can be laterally linked by relation links, which are labelled by
 relation symbols. At the bottom of
such a hierarchy one may locate \emph{individual} nodes labelled by
constant symbols. A cognitive scientist employing such a semantic
network representation would consider the class nodes, individual
nodes, and relation links as computer implementations or formal models
of class concepts, individual concepts, and relational concepts,
respectively. Also, semantic network specification languages are
sometimes called \emph{concept description languages} in AI
programming. On this background, I will understand the symbols
contained in a logic signature as names of concepts. 

Furthermore, a logical \emph{expression} $\varphi[x_1,\ldots,x_n]$
containing $n$ free (first-order) variables can be interpreted by the
set of all $n$-tuples satisfying this expression.
$\varphi[x_1,\ldots,x_n]$ thus defines an $n$-ary relation. For
example, $\varphi[x] = \mbox{\sc Fruit}(x) \wedge \mbox{\sc Yellow}(x)
\wedge \mbox{\sc Longish}(x) \wedge \mbox{\sc Curved}(x)$ would
represent a class (seems to be the class of bananas). Quite generally,
logical expressions formed according to the syntax of a logic
formalism can build representations of new concepts from given ones.

There is an important difference between how ``concepts'' are viewed
in cognitive modeling versus logic-based AI. In the latter field,
concepts are typically \emph{named} by symbols, and the formal
treatment of semantics is based on a reference relationship between
the symbols of a signature and their interpretations. However, even in
logic-based knowledge representation formalisms there can be un-named
concepts which are formally represented as logic expressions with free
variables, as for instance the banana formula above. In cognitive
science, concepts are not primarily or necessarily named, although a
concept can be optionally labelled with a name. Cognitive modeling can
deal with conceptual systems that have not a single symbol, for
instance when modeling animal cognition. By contrast, AI-style logic
modeling typically is strongly relying on symbols (the only exception
being mathematical theories built on the empty signature; this is of
interest only for intra-mathematical investigations).

\emph{Remarks on ``Logics''.} In writing this paragraph, I follow the
leads of the PhD thesis \cite{Rabe08} of Florian Rabe which gives a comprehensive
and illuminating account of today's world of formal logic research.
The field of mathematical logics has grown and diversified enormously
in the last three decades. While formal logic historically has been
developed within and for pure mathematics, much of this recent boost
was driven by demands from theoretical computer science, AI, and
semantic web technologies. This has led to a cosmos populated by a
multitude of ``logics'' which sometimes differ from each other even in
basic premises of what, actually, qualifies a formal system as a
``logic''. In turn, this situation has led to \emph{meta-logical}
research, where one develops formal \emph{logical frameworks} in order to
systematically categorize and compare different logics. 

Among the existing such logical frameworks, I choose the framework of
\emph{institutions} \cite{GoguenBurstall92}, because it has been
devised as an abstraction of model-theoretic accounts of logics, which
allows me to connect quite directly to concepts, categories, and the
semantic reference link between these two. Put briefly, a formal
system qualifies as a logic within this framework if it can be
formulated as an institution. The framework of institutions is quite
general: all logics used in AI, linguistics and theoretical cognitive
sciences can be characterized as institutions.

The framework of institutions uses tools from category theory. In this
section I do not assume that the reader is familiar with category
theory, and therefore will give only an intuitive account of how
``conceptor logic'' can be cast as an institution. A full categorical
treatment is given in Section \ref{sec:ClogicCategoryTheory}. 

An institution is made of three main components, familiar from the
model theory of standard logics like first-order predicate logic:
\begin{enumerate}
\item a collection $\mathbf{Sign}$ of \emph{signatures}, where each
signature $\Sigma$ is a set of \emph{symbols}, 
\item for each signature $\Sigma$, a set $Sen(\Sigma)$ of
$\Sigma$-\emph{sentences} that can be formed using the symbols of $\Sigma$, 
\item again for each signature $\Sigma$, a collection $M\!od(\Sigma)$ of
$\Sigma$-\emph{models}, where a $\Sigma$-model is a mathematical
structure in which the symbols from $\Sigma$ are interpreted.   
\end{enumerate}
Furthermore, for every signature $\Sigma$  there is a \emph{model
  relation} $\models_\Sigma \;\subseteq M\!od(\Sigma) \times
  Sen(\Sigma)$. For a  $\Sigma$-model $m$ and a $\Sigma$-sentence $\chi$,
  we write infix notation $m \models_\Sigma \chi$ for $(m,\chi) \in\;
  \models_\Sigma$, and say ``$m$ is a model of $\chi$'', with the
  understanding  that the sentence $\chi$ makes a true statement about
  $m$.

The
relationsships between the main elements of an 
institution   can be visualized as in Figure \ref{figInstitution}.

\begin{figure}[htb]
  \center \includegraphics[width=35mm]{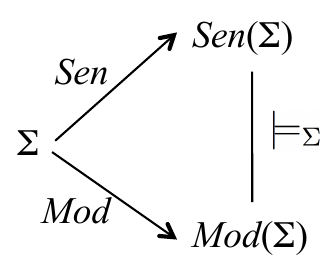}
  \caption{How the elements of  an institution relate to each other.
  For explanation see text.} 
  \label{figInstitution}
\end{figure}

The full definition of an institution includes a
mechanism for symbol re-naming. The intuitive picture is the following. If a
mathematician or an AI engineer writes down a set of axioms, expressed
as sentences  in a logic, the choice of symbols should be of no
concern whatsoever. As Hilbert allegedly put it, the mathematical theory of
geometry should remain intact if instead of ``points, lines, surfaces''
one would speak of ``tables, chairs, beer mugs''. In the framework of
institutions this is reflected by formalizing how a signature
$\Sigma$ may be transformed into another signature $\Sigma'$ by a
\emph{signature morphism} $\phi: \Sigma \to \Sigma'$, and how signature morphisms are
extended to sentences (by re-naming symbols in a sentence according to
the signature morphism) and to models (by interpreting the re-named
symbols by the same elements of a model that were previously used for
interpreting the original symbols). Then, if $m', \chi'$ denote the
re-named model $m$ and sentence $\chi$, an institution essentially demands that
$m    \models_\Sigma \chi$ if and only if $m'  \models_{\Sigma'} \chi'$.

For example, first-order logic (FOL) can be cast as an institution by
taking for $\mathbf{Sign}$ the class of all FOL signatures, that is the class of
all sets containing typed predicate, relation, function and constant
symbols; $Sen$ maps a signature $\Sigma$ to the set of all closed
(that is, having no free variables) $\Sigma$-expressions (usually
called \emph{sentences}); $M\!od$ assigns to each signature the class of
all set-theoretic $\Sigma$-structures; and $\models$ is the satisfaction relation of
FOL (also called model relation). For another example, Boolean logic
can be interpreted as an institution in several ways, for instance by
declaring  $\mathbf{Sign}$ as the class of all totally ordered countable sets
(the elements of which would be seen as Boolean variables); for each
signature  $\Sigma$  of Boolean variables,
$Sen(\Sigma)$ is the set of all Boolean expressions
$\varphi[X_{i_1},\ldots, X_{i_n}]$ over $\Sigma$ and $M\!od(\Sigma)$ is the set
of all truth value assignments $\tau: \Sigma \to \{T,F\}$ to the
Boolean variables in $\Sigma$; and $\tau \models_\Sigma \varphi$ if
$\varphi$ evaluates to $T$ under the assignment $\tau$.

In an institution, one can define \emph{logical entailment} between
$\Sigma$-sentences in the familiar way, by declaring that $\chi$
logically entails $\chi'$ (where $\chi, \chi'$ are $\Sigma$-sentences) if
and only if for all $\Sigma$-models $m$ it holds that
$m \models_\Sigma \chi$ implies $m
\models_\Sigma \chi'$. By a standard abuse of notation, this is also
written as $\chi \models_\Sigma \chi'$ or $\chi \models \chi'$.

I will sketch two entirely different approaches to define a ``conceptor
logic''. The first follows in the footsteps of familiar logics.
Conceptors can be named by arbitrary symbols, sentences are built by
an inductive procedure which specifies how more complex sentences can
be constructed from simpler ones by similar syntax rules as in
first-order logic, and models are designated as certain mathematical
structures built up from named conceptors. This leads to a logic that
essentially represents a version of first-order logic constrained to
conceptor domains. It would be a logic useful for mathematicians to
investigate ``logical'' characteristics of conceptor mathematics,
especially whether there are complete calculi that allow one to
systematically prove all true facts concerning conceptors. I call such
logics \emph{extrinsic conceptor logics}. Extrinsic conceptor logics
are tools for mathematicians to reason \emph{about} conceptors. A
particular extrinsic conceptor logic as an institution is detailed in
Section \ref{sec:ClogicCategoryTheory}.

The other approach aims at a conceptor logic that, instead of being a
tool for mathematicians to reason \emph{about} conceptors, is a model
of how a situated intelligent agent does ``logical reasoning''
\emph{with} conceptors. I call this \emph{intrinsic conceptor
  logic} (ICL). An ICL has a number of  unconventional properties:

\begin{itemize}
\item An ICL should function as a model of a situated agent's
conceptor-based information processing. Agents are bound to
differ widely in their structure and their concrete lifetime
learning histories. Therefore I do not attempt to design a general
``fits-all-agents'' ICL. Instead, for every single, concrete agent
life history there will be an ICL, \emph{the} private ICL of \emph{that} agent
life. 
\item An agent with a personal learning history is bound to develop
its private  ``logic'' over time. The ICL of an agent life thus becomes a
dynamical system in its own right. The framework of institutions was
not intended by its  designers to model temporally evolving
objects. Specifying an institution such that it can be considered a
dynamical system leads to some particularly unconventional
characteristics of an agent life ICL. Specifically, signatures
become time-varying objects, and signature morphisms (recall that
these model the ``renaming'' of symbols) are used to capture the
temporal evolution of signatures.  
\end{itemize}

An agent's lifetime ICL is formalized differently according to whether
the agent is based on matrix conceptors or random feature
conceptors. Here I work out only the second case.

In the following outline I  use the concrete three-layer
de-noising and classification architecture from Section
\ref{secHierarchicalArchitecture} as a reference example to fill the
abstract components of ICL with life. Even more concretely, I
use the specific ``lifetime history'' of the 16000-step adaptation run
illustrated in Figure \ref{figcArch} as demonstration example. 
For simplicity I will refer to that particular de-noising
and classification architecture run as ``DCA''.

Here is a simplified sketch of the main
components of an agent's lifetime ICL (full  treatment in  Section
\ref{sec:ClogicCategoryTheory}):

\begin{enumerate}
    \item An ICL is designed to model a particular agent lifetime
  history. A specification of such an ICL requires that a
  formal model of such an \emph{agent life} is available beforehand.
  The core part of an agent life model $\mathcal{AL}$ is a  set
  of $m$ conceptor adaptation sequences $\{a_1(n),\ldots, a_m(n)\}$,
  where each $a_i(n)$ is an $M$-dimensional conception weight vector. It is
  up to the modeler's discretion which conceptors in a modeled agent
  become included in the agent life model $\mathcal{AL}$. In the DCA example
  I choose the four prototype conception weight vectors $c^1,\ldots,
  c^4$ and the two auto-adapted $c_{[l]}^{\mbox{\scriptsize aut}}$ on
  layers $l = 1,2$. In this example, the core constituent of the
  agent life model $\mathcal{AL}$ is thus the set of $m = 6$ conceptor
  adaptation trajectories $c^1(n),\ldots, c^4(n),
  c_{[1]}^{\mbox{\scriptsize aut}}(n), c_{[2]}^{\mbox{\scriptsize
      aut}}(n)$, where $1 \leq n \leq 16000$. The first four
  trajectories $c^1(n),\ldots, c^4(n)$ are constant over time because
  these prototype conceptors are not adapted; the last two evolve over
  time. Another part of an agent life model is the \emph{lifetime} $T$, which is
  just the interval of timepoints $n$ for which the adaptation
  sequences $a_i(n)$ are defined. In the DCA example, $T =
  (1,2,\ldots,16000)$.  
    \item A signature is a finite non-empty set $\Sigma^{(n)} = \{A_1^{(n)},
  \ldots, A_m^{(n)} \}$ of $m$ time-indexed symbols $A_i$. For every
  $n \in T$ there is a signature $\Sigma^{(n)}$.

 \emph{DCA     example:} In the ICL of this example agent life, the collection
  $\mathbf{Sign}$ of signatures is made of 16000 signatures $\{C_1^{(n)},
  \ldots,C_4^{(n)}, A_1^{(n)}, A_2^{(n)} \}$ containing six symbols
  each, with the understanding that the first four symbols refer to
  the prototype conceptors $c^1(n),\ldots, c^4(n)$ and the last two
  refer to the auto-adapted conceptors $ c_{[1]}^{\mbox{\scriptsize
  aut}}(n), c_{[2]}^{\mbox{\scriptsize aut}}(n)$.
    \item For every pair $\Sigma^{(n+k)}, \Sigma^{(n)}$ of signatures,
  where $k \geq 0$, there is a signature morphism $\phi^{(n+k,n)}:
  \Sigma^{(n+k)} \to \Sigma^{(n)}$ which maps $A^{(n+k)}_i$ to
  $A^{(n)}_i$. These signature morphisms introduce a time arrow into
  $\mathbf{Sign}$. This time arrow ``points backwards'', leading from
  later times $n+k$ to earlier times $n$.  There is a good reason for
  this backward direction. Logic is all about describing facts. In a
  historically evolving system, facts $\chi^{(n+k)}$ established at
  some later time $n+k$ can be explained in terms of facts
  $\zeta^{(n)}$ at
  preceding times $n$, but not vice versa. Motto: ``the future can be
  explained in terms of the past, but the past cannot be reduced to
  facts from the future''. Signature morphisms are a technical vehicle
  to re-formulate descriptions of facts. They must point backwards in
  time in order to allow facts at later times to become re-expressed
  in terms of facts stated for earlier times. Figure
  \ref{figICLSigMorph} illustrates the signatures and their morphisms
  in an ICL. 

\begin{figure}[htb]
\center
\includegraphics[width=80 mm]{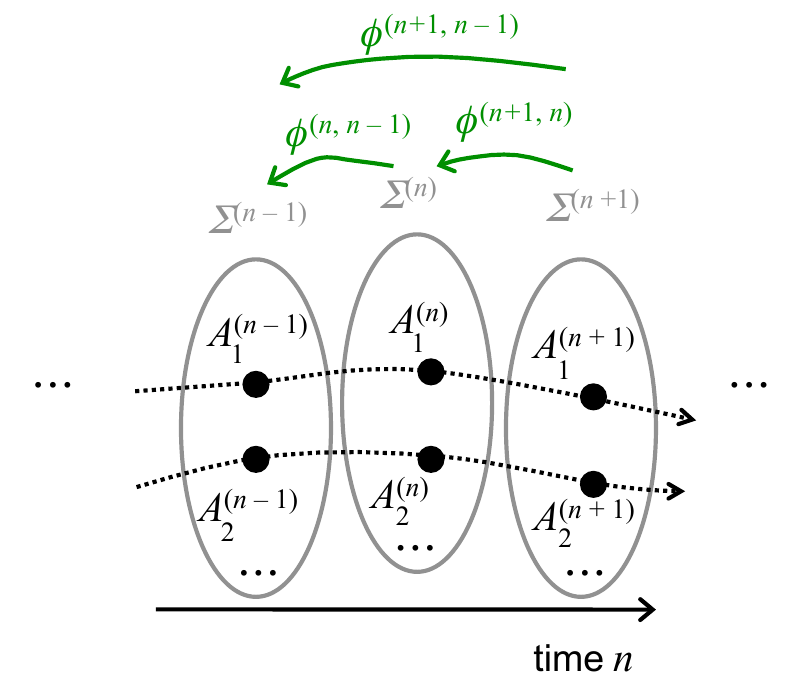}
\caption{Signatures and their morphisms in an ICL (schematic). For
  explanation see text.}
\label{figICLSigMorph}
\end{figure}  

\item Given a signature $\Sigma^{(n)} = \{A_1^{(n)},
  \ldots, A_m^{(n)} \}$, the set of sentences $Sen( \Sigma^{(n)})$ which can
  be expressed with the symbols of this signature is the set of
  syntactic expressions defined inductively by the following
  rules (incomplete, full treatment in next subsection):
\begin{enumerate}
\item $A_1^{(n)}, \ldots, A_m^{(n)}$  are sentences in
$Sen(\Sigma^{(n)})$.
\item For $k \geq 0$ such that $n,n+k \in T$, for $A_i^{(n)} \in
\Sigma^{(n)}$,   $\delta^{(n)}_k\,A_i^{(n)}$ is in
$Sen(\Sigma^{(n)})$.
  \item If $\zeta, \xi \in Sen(\Sigma^{(n)})$, then $(\zeta \vee \xi),
(\zeta \wedge \xi), \neg \zeta \in Sen(\Sigma^{(n)})$. 
\item If $\zeta \in Sen(\Sigma^{(n)})$, then $\varphi(\zeta, \gamma)
\in Sen(\Sigma^{(n)})$ for every  $\gamma \in
[0,\infty]$ (this captures 
aperture adaptation).
  \item If $\zeta, \xi \in Sen(\Sigma^{(n)})$ and $0 \leq b \leq 1$,
then $\beta_b(\zeta, \xi) \in Sen(\Sigma^{(n)})$
(this will take care of linear blends $ b\zeta + (1-b)\xi$). 
\end{enumerate}
In words, sentences express how new conceptors can be built from
existing ones by Boolean operations, aperture adaptation, and linear
blends. The ``seed'' set for these inductive constructions is provided by the
conceptors that can be directly identified by the symbols in
$\Sigma^{(n)}$. 

The sentences of form $\delta^{(n)}_k\,A_i^{(n)}$ deserve a special
comment. The operators $\delta^{(n)}_k$ are time evolution operators.
A sentence $\delta^{(n)}_k\,A_i^{(n)}$ will be made to refer to the
conceptor version $a_i(n+k)$ at time $n+k$ which has evolved from
$a_i(n)$.

  \item For every time $n$, the set $M\!od(\Sigma^{(n)})$ of
$\Sigma^{(n)}$-models is the set $\mathbf{Z}$ of $M$-dimensional
nonnegative vectors. 

\emph{Remarks:} (i) The idea for these models is
that they represent mean energy vectors $E[z^{.\wedge .2}]$ of feature
space states. (ii) The set of models $M\!od(\Sigma^{(n)})$ is the same
for every signature $\Sigma^{(n)}$.

 \emph{DCA example:} Such feature
space signal energy vectors occur at various places in the DCA, for
instance in Equations (\ref{eqZmatrix}), (\ref{eqcn}), and conception
weight vectors which appear in the DCA evolution are all defined or adapted in
one way or other on the basis of such feature space signal energy vectors. 

\item Every $\Sigma^{(n)}$-sentence $\chi$ is associated with a concrete conception
weight vector $\iota(\chi)$ by means of the following inductive
definition: 
 \begin{enumerate}
\item $\iota( A_i^{(n)}) = a_i(n)$.
\item $\iota(\delta_k^{(n)} A_i^{(n)}) = a_i(n+k)$.
\item Case  $\chi = (\zeta \vee \xi)$: $\iota(\chi) = \iota(\zeta)
\vee \iota(\xi)$ (compare Definition
\ref{def:Booleancvecs}). 
\item Case $\chi = (\zeta \wedge \xi)$: $\iota(\chi) = \iota(\zeta)
\wedge \iota(\xi)$.
\item Case $\chi = \neg \zeta$: $\iota(\chi) = \neg \iota(\zeta)$. 
\item Case $\chi =  \varphi(\zeta, \gamma)$: $\iota(\chi) =
\varphi(\iota(\zeta), \gamma)$ (compare Definition
\ref{def:Apadaptcvecs}).
\item Case $\chi = \beta_b(\zeta, \xi)$:  $\iota(\chi) =  b\,
\iota(\zeta) + (1-b) \iota(\xi)$. 
\end{enumerate}
\emph{Remark:} This statement of the interpretation operator $\iota$
is suggestive only. The rigorous definition (given in the next
section) involves additional nontrivial mechanisms to establish the connection
between the symbol $A_i^{(n)}$ and the concrete conceptor version
$a_i(n)$ in the agent life. Here I simply appeal to the reader's
understanding that symbol $A_i$ refers to object $a_i$. 

\item For $z^{.\wedge 2} \in \mathbf{Z}$ and $\chi \in
Sen(\Sigma^{(n)})$, the model relationship is defined by 
\begin{equation}\label{eqSemModels}
z^{.\wedge
  2} \models_{\Sigma^{(n)}} \chi \quad \mbox{iff}\quad    z^{.\wedge 2} \,.\!\ast\, (z^{.\wedge 2} + 1)^{.\wedge
  -1} \leq \iota(\chi).
\end{equation}
\emph{Remark:} This definition in essence just repeats how a conception
weight vector is derived from a feature space signal energy vector.
\end{enumerate}

When all category-theoretical details are filled in which I have omitted
here, one obtains a formal definition of an institution which
represents \emph{the} ICL of an agent life $\mathcal{AL}$. It can be shown that in an
ICL, for all $\zeta,\xi \in Sen(\Sigma^{(n)})$ it holds that
$$\zeta \models_{\Sigma^{(n)}} \xi \quad \mbox{iff} \quad \iota(\zeta)
\leq \iota(\xi).$$
By virtue of this fact, logical entailment becomes \emph{decidable} in
an ICL: if one wishes to determine whether $\xi$ is implied by
$\zeta$, one can effectively compute the vectors $\iota(\zeta),
\iota(\xi)$ and then check in constant time whether $\iota(\zeta)
\leq \iota(\xi)$. 

Returning to the DCA example (with lifetime history shown in Figure
\ref{figcArch}), its ICL identiefies over time the four prototype
conceptors $c^1,\ldots,c^4$ and the two auto-adapted conceptors
$c_{[1]}^{\mbox{\scriptsize auto}}, c_{[2]}^{\mbox{\scriptsize auto}}$
by temporally evolving symbols  $\{C_1^{(n)}, \ldots, C_4^{(n)},$ $A_1^{(n)}, A_2^{(n)} \}$. All other conceptors that are computed in this
architecture can be defined in terms of these six ones. For instance,
the top-level conceptor $c_{[3]}(n)$ can be expressed in terms of the
identifiable four prototype conceptors by $c_{[3]}(n) =
\bigvee_{j=1,\ldots,4} \varphi(c^j, \gamma^j(n))$ by combining the
operations of  disjunction
and aperture adaptation. In ICL syntax this construction would be
expressible by a $\Sigma^{(n)}$ sentence, for instance by
$$(((\varphi(C_1^{(n)},\gamma^1(n)) \vee
\varphi(C_2^{(n)},\gamma^2(n))) \vee \varphi(C_3^{(n)},\gamma^3(n)))
\vee \varphi(C_4^{(n)},\gamma^4(n))).$$

A typical  adaptation objective of a conception vector
$c(n)$ occurring in an agent life is to minimize a loss of the
form (see Definition \ref{eqcAdaptObjective})
$$E_z[\| z - c(n) \,.\!\ast \, z \|^2] +
\alpha^{-2}\,\|c(n)\|^2,$$
or equivalently, the objective is to converge to
$$c(n) = E[\alpha^2 z^{.\wedge 2}] \,.\!\ast\, (E[\alpha^2 z^{.\wedge
  2}] + 1)^{.\wedge -1}.$$
This can be
re-expressed in ICL terminology as ``adapt $c(n)$ such that $\alpha^2 E[z^{.\wedge
  2}] \models_{\Sigma^{(n)}} \chi_{c(n)}$, and such that not
  $z^{.\wedge  2} \models_{\Sigma^{(n)}} \chi_{c(n)}$ for any
  $z^{.\wedge  2} > \alpha^2 E[z^{.\wedge
  2}]$'' (here $\chi_{c(n)}$ is an adhoc notation
for an ICL sentence $\chi_{c(n)} \in Sen(\Sigma^{(n)})$ specifying
$c(n)$).  In more abstract terms, the typical adaptation of random
  feature conceptors can be understood as an attempt to converge
  toward the conceptor that is maximally  $\models$-specific under a
  certain constraint.  
  
  \emph{Discussion.}  I started this section by a rehearsal of how the
  notion of ``concept'' is understood in cognitive science and
  logic-based AI formalisms. According to this understanding, a
  concept \emph{refers} to a category (terminology of cognitive
  science); or a class symbol or logical expression with free
  variables is \emph{interpreted by} its set-theoretical extension
  (logic terminology). Usually, but not necessarily, the
  concepts/logical expressions are regarded as belonging to an
  ``ontological'' domain that is different from the domain of their
  respective referents. For instance, consider a human maintaining a
  concept named {\sf cow} in his/her mind. Then many cognitive
  scientists would identifiy the category that is referred to by this
  concept with the some set of physical cows. Similarly, an AI expert
  system set up as a farm management system would contain a symbol
  {\sf cow} in its signature, and this symbol would be deemed to refer
  to a collection of physical cows. In both cases, the concept /
  symbolic expression {\sf cow} is ontologically different from a set
  of physical cows.  However, both in cognitive science and AI,
  concepts / symbolic expressions are sometimes brought together with
  their referents much more closely. In some perspectives taken in
  cognitive science, concepts are posited to refer to other mental
  items, for instance to sensory perceptions. In most current AI proof
  calculi (``inference engines''), models of symbolic expressions are
  created which are assembled not from external physical objects but
  from symbolic expressions (``Herbrand universe'' constructions).
  Symbols from a signature $\Sigma$ then refer to sets of
  $\Sigma$-terms. In sum, fixing the ontological nature of referents
  is ultimately left to the modeling scientist in cognitive science or
  AI.
  
  In contrast, ICL is committed to one particular view on the semantic
  relationship: $\Sigma^{(n)}$-sentences are always describing
  conception weight vectors, and 
  refer to neural activation energy vectors $z^{.\wedge 2}$. In the
  case of matrix conceptor based agents, $\Sigma^{(n)}$-sentences
  describe conceptor matrices and refer to neural activation
  correlation matrices $R$ by the following variant of (\ref{eqSemModels}):

\begin{equation}\label{eqSemModelsMatrix}
R \models_{\Sigma^{(n)}} \chi \quad \mbox{iff}\quad    R (R + I)^{-1}
\leq \iota(\chi). 
\end{equation} 

In Figure \ref{figSemantics} I try to visualize this difference
between the classical, extensional view on symbols and their
referents, and the view adopted by ICL. This figure contrasts how
classical logicians and cognitive scientists would usually model an
agent's representation of farm livestock, as opposed to how ICL
renders that situation.  The semantic relation is here established
between the physical world on the one side, and symbols and logical
expressions on the other side.  The world is idealized as a set of
individuals (individual  animals in this example), and
symbols for concepts (predicate symbols in logic) are semantically
interpreted by sets of individuals. In the farmlife example, a
logician might introduce a symbol {\sf lifestock} which would denote
the set of all economically relevant animals grown in farms, and one
might introduce another symbol {\sf poultry} to denote the subset of
all feathered such animals. The operator that creates ``meaning'' for
concept symbols is the grouping of individuals into sets (the bold
 ``$\{\,\,\}$'' in Figure \ref{figSemantics}).

With conceptors, the semantic relation connects neural activity
patterns triggered by perceiving animals on the one side, with
conceptors acting on neural dynamics on the other side. The core
operator that creates meaning is the condensation of the incoming data
into a neural activation energy pattern $z^{.\wedge 2}$ (or correlation matrix $R$
for matrix conceptors) from which conceptors are generated via
the fundamental construction $c = E[z^{.\wedge 2}] \,.\!\ast\,
(E[z^{.\wedge 2}]+1)^{.\wedge -1}$ or $C = R(R+I)^{-1}$ (Figure
\ref{figSemantics} depicts the latter case).

\begin{figure}[htb]
\center
\includegraphics[width=140mm]{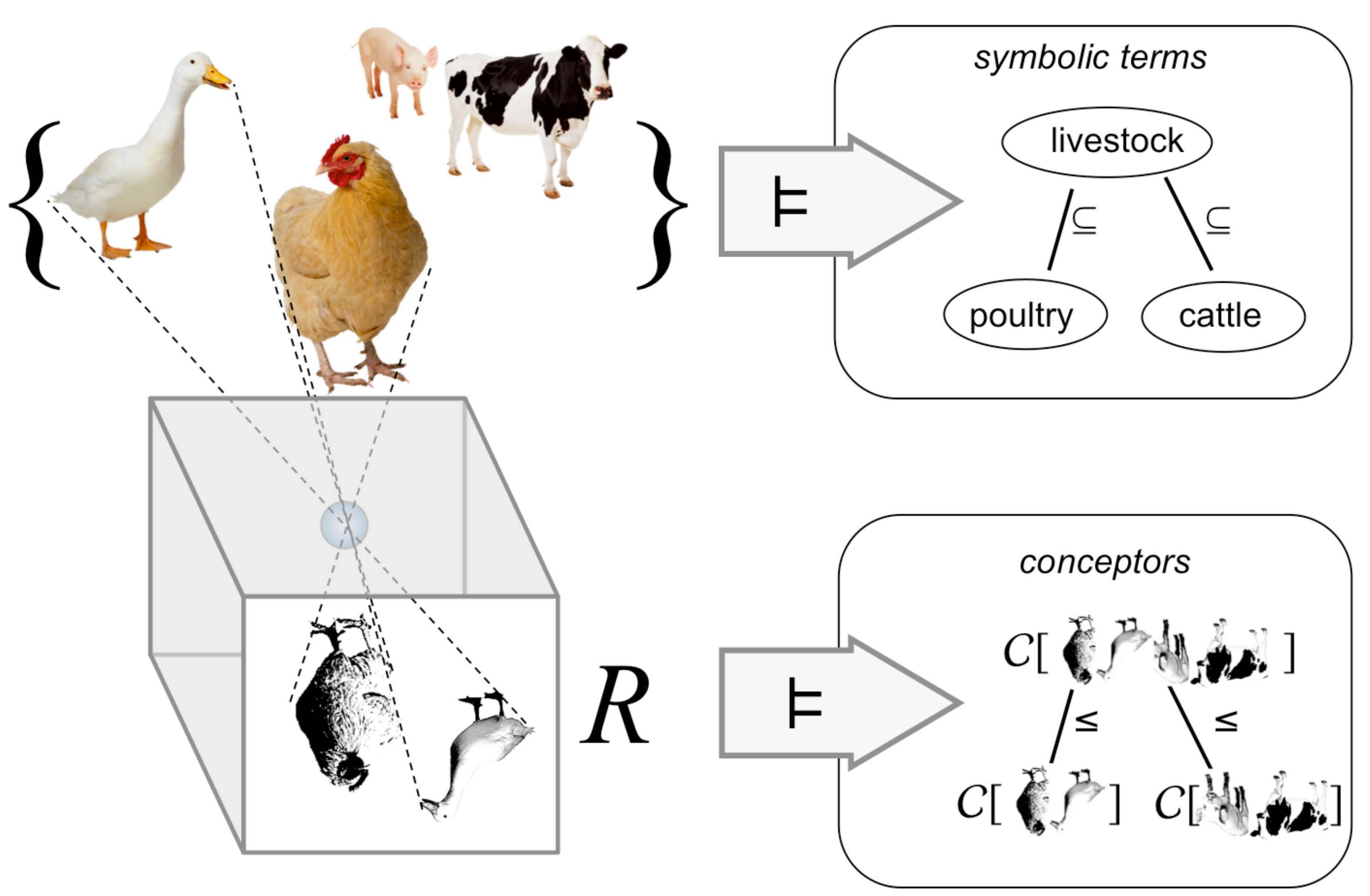}
\caption{Contrasting the extensional semantics of classical knowledge
  representation formalisms (upper half of graphics) with the
  system-internal neurodynamical semantics of conceptors (lower half).
  In both modeling approaches, abstraction hierarchies of ``concepts''
  arise. For
  explanation see text.}
\label{figSemantics}
\end{figure}

ICL, as presented here, cannot claim to be a model of all ``logical
reasoning'' in a neural agent. Specifically, humans sometimes engage
in reasoning activities which are very similar to how syntactic logic
calculi are executed in automated theorem proving.  Such activities
include the build-up and traversal of search trees,
creating and testing hypotheses, variable binding and renaming, and
more. A standard example is the step-by-step exploration of move
options done by a human chess novice. ICL is not designed to capture
such conscious combinatorial logical reasoning. Rather, ICL is intended to
capture the automated aspects of neural information processing of
a situated agent, where incoming (sensor) information is immediately 
transformed into perceptions and maybe situation representations in a
tight dynamical coupling with the external driving signals.

 The material presented in
this and the next section is purely theoretical and offers no computational add-on
benefits over the material presented in earlier sections. There are three reasons why
nonetheless I invested the effort of defininig ICLs:

\begin{itemize}
    \item By casting conceptor logic rigorously as an institution,
  I wanted to substantiate my claim that conceptors are ``logical'' in
  nature, beyond a mere appeal to the intuition that anything
  admitting Boolean operations is logic. 
    \item The institutional definition given here provides a
  consistent formal picture of the semantics of conceptors.  A
  conceptor $c$ identified by an ICL sentence $\chi_c$ ``means''
  neural activation energy vectors $z^{.\wedge 2}$. Conceptors and
  their meanings are both neural objects of the same mathematical
  format, $M$-dimensional nonnegative vectors. Having a clear view on
  this circumstance helps to relate conceptors to the notions of
  concepts and their referents, which are so far from being fully
  understood in the cognitive sciences.
  
    \item Some of the design ideas that went into casting ICLs as
  institutions may be of more general interest for mathematical logic
  research. Specifically, making signatures to evolve over time -- and
  hence, turn an institution into a dynamical system -- might be found
  a mechanism worth considering in scenarios, unconnected with
  conceptor theory or neural networks, where one wants to analyse
  complex dynamical systems by means of formal logics.
\end{itemize} 

\subsection{Conceptor Logic as  Institutions: Category-Theoretical
  Detail}\label{sec:ClogicCategoryTheory} 

In this section I provide a formal specification of conceptor logic as
an institution. This section addresses only readers with a dedicated
interest in formal logic. I assume that the reader is familiar with the
institution framework for representing logics (introduced in
\cite{GoguenBurstall92} and explained in much more detail in Section 2
in \cite{Rabe08}) and with basic elements of category theory. I first
repeat almost verbatim the categorical definition of an institution
from \cite{GoguenBurstall92}.

\begin{definition}\label{defInstitutionCatTh}
An \emph{institution} $\mathcal{I}$ consists of
\begin{enumerate}
\item a  category $\mathbf{Sign}$, whose objects $\Sigma$ are called
\emph{signatures} and whose arrows are called \emph{signature morphisms},
\item a functor $Sen : \mathbf{Sign} \to \mathbf{Set}$, giving for
each signature a set whose elements are called \emph{sentences} over
that signature,
\item a functor $M\!od : \mathbf{Sign} \to
\mathbf{Cat}^{\mbox{\scriptsize \emph{op}}}$, giving for each signature
$\Sigma$ a category $M\!od(\Sigma)$ whose objects are called $\Sigma$-\emph{models},
 and whose arrows are called $\Sigma$-(model) \emph{morphisms}, and
\item a relation $\models_\Sigma \; \subseteq \;  {M\!od}(\Sigma) 
\times Sen(\Sigma)$ for each $\Sigma \in  \mathbf{Sign} $, called $\Sigma$-\emph{satisfaction},
\end{enumerate}
\noindent such that for each morphism $\phi: \Sigma_1 \to \Sigma_2$ in
$\mathbf{Sign}$, the \emph{Satisfaction Condition}
\begin{equation}\label{eqSatisfactionCondition}
m_2 \models_{\Sigma_2} Sen(\phi)(\chi_1) \quad \mbox{iff} \quad
{M\!od}(\phi)(m_2)   \models_{\Sigma_1} \chi_1
\end{equation}
holds for each $m_2 \in {M\!od}(\Sigma_2) $ and each $\chi_1 \in
Sen(\Sigma_1)$. 
\end{definition}
The interrelations of these items are visualized in Figure
\ref{figInstitutionCatTheory}. 

\begin{figure}[htb]
\center
\includegraphics[width=80 mm]{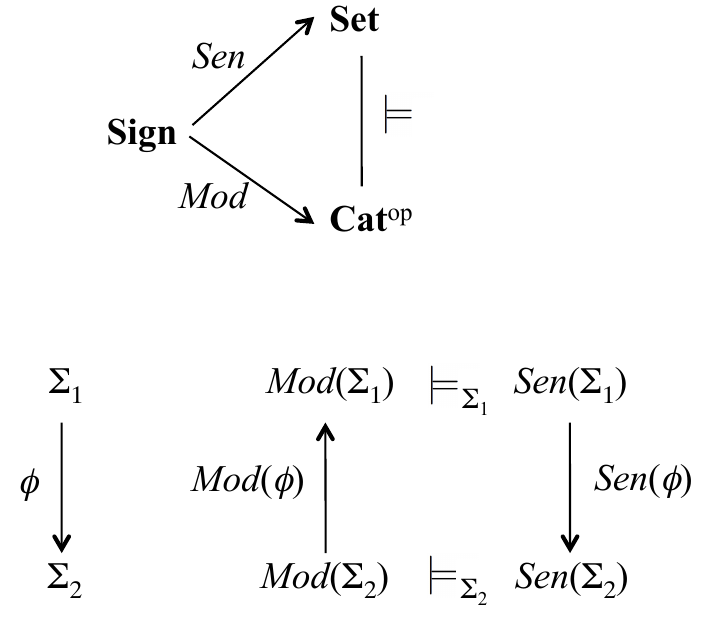}
\caption{Relationships between the constituents of an institution (redrawn from \cite{GoguenBurstall92}).}
\label{figInstitutionCatTheory}
\end{figure}

\emph{Remarks:}
\begin{enumerate}
    \item The morphisms in $\mathbf{Sign}$ are the categorical model of
  re-naming the symbols in a logic. The essence of the entire
  apparatus given in Definition \ref{defInstitutionCatTh} is to
  capture the condition
  that the model-theoretic semantics of a logic is invariant to
  renamings of symbols, or, as Goguen and Burstall state it, ``Truth
  is invariant under change of notation''.
    \item The intuition behind $\Sigma$-model-morphisms, that is, maps
  $\mu: I_1^\Sigma \to I_2^\Sigma$, where $I_1^\Sigma, I_2^\Sigma$ are
  two $\Sigma$-models, is that $\mu$ is an embedding of $I_1^\Sigma$ in
  $I_2^\Sigma$. If we take first-order logic as an example, with
  $I_1^\Sigma, I_2^\Sigma$  being two $\Sigma$-structures, then $\mu:
  I_1^\Sigma \to I_2^\Sigma$ would be a map from the domain of
  $I_1^\Sigma$ to the domain of $I_2^\Sigma$ which preserves functional
  and relational relationships specified under the interpretations of
  $I_1^\Sigma$ and $I_2^\Sigma$.
  \item In their original 1992 paper \cite{GoguenBurstall92}, the
authors show how a number of standard logics can be represented as
institutions. In the time that has passed since then, institutions have
become an important ``workhorse''  for software specification in
computer science and for semantic knowledge
management systems in AI, especially for managing mathematical
knowledge, and several families of programming toolboxes have been
built on  institutions (overview in \cite{Rabe08}). Alongside with the
model-theoretic spirit of institutions, this proven
usefulness of institutions has motivated me to adopt them as a logical
framework for conceptor logic.
\end{enumerate}

Logical entailment between sentences is defined in institutions  in
the traditional way:

\begin{definition}\label{defEntailment}
Let $\chi_1, \chi_2 \in Sen(\Sigma)$. Then $\chi_1$ \emph{entails} $\chi_2$,
written $\chi_1 \models_{\Sigma} \chi_2$, if for all $m \in Ob(M\!od(\Sigma))$
it holds that $m \models_{\Sigma} \chi_1 \;\rightarrow m \models_{\Sigma} \chi_2$.
\end{definition}

Institutions are flexible and offering many ways for
defining logics. I will frame two entirely different
kinds of conceptor logics. The first kind follows the intuitions
behind the familiar first-order predicate logic, and should function as a
formal tool for mathematicians to reason about (and with)
conceptors. Since it looks at conceptors ``from the outside'' I will
call it \emph{extrinsic conceptor logic} (ECL). Although ECL 
follows the footsteps of familiar logics in many respects, in some
aspects it deviates from tradition. The other kind aims at
modeling the ``logical'' operations that an intelligent neural agent
can perform 
whose ``brain'' implements conceptors. I find this the more
interesting formalization; certainly it is the more exotic one. I
will call it \emph{intrinsic conceptor logic} (ICL).

\emph{Extrinsic conceptor logic.} I first give an intuitive outline. I
treat only the case of matrix-based conceptors. 
An ECL concerns conceptors of a fixed dimension $N$ and their logical
interrelationships, so one should more precisely speak of
$N$-dimensional ECL.  I assume some $N$ is fixed.  Sentences of ECL
should enable a mathematician to talk about conceptors in a similar
way as familiar predicate logics allow a mathematician to describe
facts about other mathematical objects. For example, ``for all
conceptors $X$, $Y$ it holds that $X \wedge Y \leq X$ and $X \wedge Y
\leq Y$'' should be formalizable as an ECL sentence. A little
notational hurdle arises because Boolean operations
appear in two roles: as operators acting on conceptors (the ``$\wedge$''
in the sentence above), and as constituents of the logic language
(the ``and'' in that sentence). To keep these two roles notationally
apart, I will use $\mbox{AND}, \mbox{OR}, \mbox{NOT}$ (allowing infix
notation) for the role as operators, and $\wedge, \vee, \neg$ for the
logic language. The above sentence would then  be formally
written as ``$\forall x \forall y \, (x \mbox{ AND } y \leq x) \wedge (x
\mbox{ AND } y \leq y)$''. 

The definition of signatures and ECL-sentences in many respects
follows standard customs (with significant simplifications to be
explained after the definition) and is the same for any conceptor
dimension $N$:

\begin{definition}\label{defSyntaxECL}
Let $\mbox{Var} = \{x_1, x_2,...\}$ be a fixed countable indexed set of \emph{variables}. 
\begin{enumerate}
\item (ECL-signatures) The objects (signatures) of $\mbox{\bf Sign}$ are all countable
sets, whose elements are called \emph{symbols}. For signatures
$\Sigma_1, \Sigma_2$, the set of morphisms 
$\mbox{hom}\,(\Sigma_1,\Sigma_2)$ is the set of all functions $\phi:
\Sigma_1 \to \Sigma_2$.
\item (ECL-terms) Given a signature $\Sigma$, the set of
$\Sigma$-\emph{terms} is defined inductively by
\begin{enumerate}
\item Every variable $x_i$, every symbol $a \in \Sigma$, and $I$ is a
$\Sigma$-term.
\item For $\Sigma$-terms $t_1, t_2$ and $\gamma \in [0,\infty]$, the
following are $\Sigma$-terms: $\mbox{\emph{NOT }}t_1$, $(t_1 \mbox{\emph{ AND }}
t_2)$, $(t_1 \mbox{\emph{ OR }} t_2)$, and $\varphi(t_1,\gamma)$.
\end{enumerate}  
\item (ECL-expressions) Given a signature $\Sigma$, the set
$Exp(\Sigma)$ of
$\Sigma$-\emph{expressions} is defined inductively by
\begin{enumerate}
\item If $t_1, t_2$ are $\Sigma$-terms, then $t_1 \leq t_2$ is a
$\Sigma$-expression. 
\item If $e_1, e_2$ are $\Sigma$-expressions, and $x_i$ a variable,
then the following are $\Sigma$-expressions: $\neg e_1$, $(e_1 \wedge
e_2)$, $(e_1 \vee e_2)$, $\forall x_i \, e_1$.
\end{enumerate}
\item (ECL-sentences) A  $\Sigma$-expression that contains no free
variables is a  $\Sigma$-\emph{sentence} (free occurrence of
variables to be defined as usual, omitted here.) 
\end{enumerate} 
\end{definition}

Given a $\Sigma$-morphism $\phi: \Sigma_1 \to \Sigma_2$, its image
$\mbox{\sl Sen}(\phi)$ under the functor $\mbox{\sl Sen}$ is the map which
sends every $\Sigma_1$-sentence $\chi_1$ to the $\Sigma_2$-sentence $\chi_2$
obtained from $\chi_1$ by replacing all occurrences of $\Sigma_1$ symbols in
$\chi_1$ by their images under $\phi$. I omit the obvious inductive
definition of this replacement construction.

Notes:
\begin{itemize}
    \item ECL only has a single sort of symbols with arity 0, namely constant
  symbols (which will be made to refer to conceptors later). This
  renders the categorical treatment of ECL much simpler than it is for
  logics with sorted symbols of varying arities.
\item The operator symbols $\mbox{NOT}$, $\mbox{AND}$, $\mbox{OR}$,
the parametrized operation symbol $\varphi(\cdot, \gamma)$ and
the relation symbol $\leq$ are not made part of signatures, but become
universal elements in the construction of sentences. 
\end{itemize}

The models of ECL are quite simple. For a signature $\Sigma$, the
objects of $\mbox{\emph{Mod}}(\Sigma)$ are the sets of $\Sigma$-indexed
$N$-dimensional conceptor matrices
\begin{displaymath}
\mbox{\emph{Ob}}(\mbox{\emph{Mod}}(\Sigma)) = \{m \subset
\mathcal{C}_{N \times N} \times \Sigma \;|\; \forall \sigma \in
\Sigma\;\exists^{=1} C \in \mathcal{C}_{N \times N}: (C,\sigma) \in m   \}
\end{displaymath}
where $\mathcal{C}_{N \times N}$ is the set of all $N$-dimensional
conceptor matrices. The objects of $\mbox{\emph{Mod}}(\Sigma)$ are
thus the graph sets of the functions from $\Sigma$ to the set of $N$-dimensional
conceptor matrices.   The model morphisms of  $\mbox{\emph{Mod}}(\Sigma)$ are
canonically given by the index-preserving maps
\begin{displaymath}
\mbox{\emph{hom}}(\{(C_1,\sigma)\}, \{(C_2,\sigma)\}) =
\{\mu: \{(C_1,\sigma)\} \to \{(C_2,\sigma)\} \;|\; \mu: (C_1,\sigma)
\mapsto (C_2,\sigma)\}. 
\end{displaymath}
 Clearly, $\mbox{\emph{hom}}(\{(C,\sigma)\}, \{(C',\sigma)\})$
 contains exactly one element.

Given a signature morphism $\Sigma_1 \stackrel{\phi}{\to} \Sigma_2$, then
$\mbox{\emph{Mod}}(\phi)$ is defined to be a map from
$\mbox{\emph{Mod}}(\Sigma_2)$ to  $\mbox{\emph{Mod}}(\Sigma_1)$ as
follows. For a $\Sigma_2$-model $m_2 = \{(C_2,\sigma_2)\}$  let $[\![
\sigma_2 ]\!]^{m_2}$ denote the interpretation of
$\sigma_2$ in $m_2$, that is, $[\![
\sigma_2 ]\!]^{m_2}$ is the conceptor matrix $C_2$ for which $(C_2,
\sigma_2) \in m_2$. Then $\mbox{\emph{Mod}}(\phi)$ assigns to to $m_2$ the
$\Sigma_1$-model   $m_1 = \mbox{\emph{Mod}}(\phi)(m_2) = \{([\![
\phi(\sigma_1) ]\!]^{m_2}, \sigma_1)\} \in
\mbox{\emph{Mod}}(\Sigma_1)$.

The model relations $\models_\Sigma$ are defined in the same way as in
the familiar first-order logic. Omitting some detail, here is how:

\begin{definition}
Preliminaries: A map $\beta: \mbox{Var} \to \mathcal{C}_{N \times N}$ is
called a \emph{variable assignment}. $\mathcal{B}$ is the set of all
variable assignments. We denote by $\beta
\frac{C}{x_i}$ the variable assignment that is identical to $\beta$
except that $x_i$ is mapped to $C$. A $\Sigma$-\emph{interpretation}
is a pair $\mathcal{I} = (m,\beta)$ consisting of a $\Sigma$-model $m$ and a
variable assignment $\beta$. By  $\mathcal{I}\frac{C}{x_i}$ we denote
the interpretation $(m,\beta\frac{C}{x_i})$. For a $\Sigma$-term $t$,
the interpretation $\mathcal{I}(t) \in \mathcal{C}_{N \times N}$ is
defined in the obvious way. Then  $\models_\Sigma^\ast \; \subseteq \;
({M\!od}(\Sigma) \times \mathcal{B}) \times Exp(\Sigma)$ is defined
inductively by
\begin{enumerate}
\item $\mathcal{I} \models_\Sigma^\ast t_1 \leq t_2 \quad$ iff
$ \quad\mathcal{I}(t_1) \leq \mathcal{I}(t_2)$,
\item  $\mathcal{I} \models_\Sigma^\ast \neg e \quad$ iff  $ \quad
\mbox{not}\;\;\mathcal{I} 
\models_\Sigma^\ast e$,
\item $\mathcal{I} \models_\Sigma^\ast (e_1 \wedge e_2)  \quad$ iff  $ \quad\mathcal{I}
\models_\Sigma^\ast e_1$ and $\mathcal{I}
\models_\Sigma^\ast e_2$,
 \item $\mathcal{I} \models_\Sigma^\ast (e_1 \vee e_2)  \quad$ iff  $ \quad\mathcal{I}
\models_\Sigma^\ast e_1$ or $\mathcal{I}
\models_\Sigma^\ast e_2$,
\item $\mathcal{I} \models_\Sigma^\ast \forall x_i \, e \quad$ iff $
\quad$ for all
$C \in \mathcal{C}_{N \times N}$ it holds that  $\mathcal{I}\frac{C}{x_i} \models_\Sigma^\ast  e$.
\end{enumerate}
  $\models_\Sigma$ then is the restriction of  $\models_\Sigma^\ast$
  on sentences. 
\end{definition} 

This completes the definition of ECL as an institution. The
satisfaction condition obviously holds. While in many respects ECL
follows the role model of  first-order logic, the associated model
theory is much more restricted in that only $N$-dimensional conceptors
are admitted as interpretations of symbols. The natural next step
would be to design calculi for ECL and investigate whether this logic
is complete or even decidable. Clarity on this point would amount to
an insight in the computational tractability of knowledge
representation based on matrix conceptors with Boolean and aperture
adaptation opertors.

\emph{Intrinsic conceptor logic.} I want to present ICL as a model of
the ``logics'' which might unfold \emph{inside} a neural agent. All
constituents of ICL should be realizable in terms of neurodynamical
processes, giving a logic not for reasoning \emph{about} conceptors, but
\emph{with} conceptors. 

Taking the idea of placing ``logics'' \emph{inside} an agent
seriously has a number of  consequences which lead  quite far
away from traditional intuitions about ``logics'':

\begin{itemize}
    \item Different agents may have different logics. I will therefore
  not try to define a general ICL that would fit any neural agent.
  Instead every concrete agent with a concrete lifetime learning
  history will need his/her/its own individual conceptor logic.  I
  will use the signal de-noising and classification architecture from
  Section \ref{secHierarchicalArchitecture} as an example ``agent''
  and describe how an ICL can be formulated as an institution for this
  particular case. Some general design principles will however become
  clear from this case study.
    \item Conceptors are all about temporal processes, learning and
  adaptation. An agent's private ICL will have to possess an eminently
  dynamical character. Concepts will change their meaning over time in
  an agent. This ``personal history dynamics'' is quintessential for
  modeling an agent and should become reflected in making an ICL a
  dynamical object itself -- as opposed to introducing time through
  descriptive syntactical elements in an otherwise static logic, like
  it is traditionally done by means of modal operators or axioms
  describing a timeline. In my proposal of ICLs, time enters the
  picture through the central constituent of an institution,
  signature morphisms. These maps between signatures (all commanded by
  the same agent) will model time, and an agent's lifetime history of
  adaptation will be modeled by an evolution of signatures. Where the
  original core motif for casting logics as institutions was that
  ``truth is invariant under change of notation''
  (\cite{GoguenBurstall92}), the main point of ICLs could be
  contrasted as ``concepts and their meaning change with time''. The role of signature
  morphisms in ICLs is fundamentally different in ICLs compared to customary
  formalizations of logics. In the latter, signature changes should
  leave meaning invariant; in the former, adaptive changes in
  conceptors are reflected by temporally indexed
  changes in signature. 
\item Making an ICL private to an agent  implies that the model
relation $\models$ becomes agent-specific. An ICL cannot be specified
as an abstract object in isolation. Before it can be defined, one
first needs to have a formal model of a particular agent with a
particular lifelong adaptation history. 
\end{itemize}

In sum, an ICL (formalized as institution) itself becomes a dynamical
system, defined relative to an existing (conceptor-based) neural agent
with a particular adaptation history. The ``state space'' of an ICL
will be the set of signatures. A ``trajectory'' of the temporal
evolution of an ICL will essentially be a sequence of signatures,
enriched with  information pertaining to forming
sentences and models. For an illustration, assume that a
neural agent adapts two random feature conceptors
$a(n), b(n)$.  These are named by two \emph{temporally indexed}
symbols $A^{(n)}, B^{(n)}$. A signature will be a timeslice of these,
$\Sigma^{(n)} = \{A^{(n)}, B^{(n)}\}$. For every pair of integer timepoints
$(n+k, n)$ (where $k  \geq 0$) there will be a signature morphism
$\phi^{(n+k,n)}: \Sigma^{(n+k)} \to \Sigma^{(n)}$. The (strong) reason
why signature morphisms point backwards in time will become clear
later. Figure \ref{figICLSigMorph} visualizes the components of this
example. The dotted lines connecting the $A^{(n)}_i$ are
suggestive graphical hints that the symbols  $A^{(n)}_i$ all name the
``same'' conceptor $a_i$. How this
``sameness of identity over time'' can be captured in the institution formalism
will become clear presently.

Formal definitions of ICLs will vary depending on what kind of
conceptors are used (for instance, matrix or random feature based), or
whether time is taken to be discrete or continuous. I give a
definition for discrete-time, random feature conceptor based ICLs.

Because ICLs will be  models of an agent's private logic which evolves
over the agent's lifetime, the
definition of an ICL is stated relative to an agent's lifetime
conceptor adaptation history. The only
property of such an agent that is needed for defining an ICL is the
existence of temporally adapted conceptors owned by the agent. Putting
this into a formal definition:

\begin{definition}\label{defICLagent}
An \emph{agent life} (here: $M$-dimensional
  random feature conceptor based, discrete time) is a structure
  $\mathcal{AL} = (T, \Sigma, \iota_{\mathcal{AL}})$, where
\begin{enumerate}
\item $T \subseteq \mathbb{Z}$ is an interval (finite
  or infinite) of the integers, the \emph{lifetime} of   $\mathcal{AL}$,
\item $\Sigma = \{A_1,\ldots, A_m\}$ is a finite nonempty set of
\emph{conceptor identifiers}, 
\item $\iota_{\mathcal{AL}}: \Sigma \times T \to [0,1]^M, (A_i, n) \mapsto a_i(n)$
assigns to every time point and conceptor identifier an
\emph{adaptation version}
$a_i(n)$ of the conceptor identified by the symbol $A_i$.  
\end{enumerate}
\end{definition}

As an example of an agent consider the signal de-noising and
classification architecture (DCA) presented in Section
\ref{secHierarchicalArchitecture}, with a ``life'' being the concrete
16000-step adaptation run illustrated in Figure \ref{figcArch}. In
this example, the lifetime is $T = \{1,\ldots,16000\}$. I will
identify by symbols the four prototype conceptors $c^1,\ldots,c^4$ and
the two auto-adapted conceptors $c_{[1]}^{\mbox{\scriptsize auto}},
c_{[2]}^{\mbox{\scriptsize auto}}$.  Accordingly I choose $\Sigma$ to
be $ \{C_1, \ldots, C_4, A_1, A_2 \}$.  The map $\iota_{\mathcal{AL}}$
is constant in time for the four protype conceptors:
$\iota_{\mathcal{AL}}(n,C_j) = c^j$ for all $n \in T, j = 1,\ldots,4$.
For the remaining two conceptors, $\iota_{\mathcal{AL}}(n,A_i) =
c_{[i]}^{\mbox{\scriptsize auto}}(n)$.

The stage is now prepared to spell out the definition of an agent's
lifetime ICL (for the case of an agent based on $M$-dimensional
  random feature conceptor  and discrete time):

\begin{definition}\label{defICL}
  The \emph{intrinsic conceptor logic (ICL) of an agent life}$\mathcal{AL} =
  (T,  \Sigma, \iota_{\mathcal{AL}})$ is an institution
  whose components obey the following conditions:
\begin{enumerate}
    \item The objects (signatures) of $\;\mathbf{Sign}$ are the
  pairs $\Sigma^{(n)} = (\{A_1^{(n)}, \ldots, A_m^{(n)} \},\sigma^{(n)})$, where $n
  \in T$, and $\sigma^{(n)}: \Sigma \to \{A_1^{(n)}, \ldots,
  A_m^{(n)}\}$ is a bijection.
  
  {\bf DCA example:} The lifetime of this example is $T =
  \{1,\ldots,16000\}$. A signature $\Sigma^{(n)} = (\{C_1^{(n)} ,
  \ldots, C_4^{(n)} , A_1^{(n)} , A_2^{(n)} \}, \sigma^{(n)})$ at time $n \in T$ 
  will later be employed to denote some of the conceptors in the DCA
  in their adapted versions at time $n$. These conceptor adaptation
  versions will thus become identifiable by symbols from
  $\Sigma^{(n)}$. For $\sigma^{(n)}$ I take the natural projection
  $C_j \mapsto  C_j^{(n)}, A_i \mapsto A_i^{(n)}$. 
  
    \item For every $n, n + k \in T$ (where $k \geq 0$),
  $\phi^{(n+k,n)}: \Sigma^{(n+k)} \to \Sigma^{(n)}, A_i^{(n+k)}
  \mapsto (\sigma^{(n)} \circ (\sigma^{(n+k)})^{-1})( A_i^{(n+k)})$
  is a morphism in $\;\mathbf{Sign}$.  There are no other morphisms in
  $\mathbf{Sign}$ besides these. \emph{Remark:} At first sight this
  might seem unneccessarily complicated. Why not simply require
  $\phi^{(n+k,n)}:  A_i^{(n+k)} \mapsto  A_i^{(n)}$? The reason is
  that the set of symbols $\{A_1^{(n)},\ldots A_m^{(n)}\}$ of
  $\Sigma^{(n)}$ is just that, a set of symbols. That over time
  $A_i^{(n)}$ should correspond to $A_i^{(n+k)}$ is only visually
  suggested to us, the mathematicians, by the chosen notation for
  these symbols, but by no means does it actually
  follow from that notation. 

  \item $Sen(\Sigma^{(n)})$ is inductively defined as follows:
\begin{enumerate}
\item $A_1^{(n)}, \ldots, A_m^{(n)}$ and $I$ and $0$ are sentences in
$Sen(\Sigma^{(n)})$. 
\item For $k \geq 0$ such that $n,n+k \in T$, for $A_i^{(n)} \in
\Sigma^{(n)}$,   $\delta^{(n)}_k\,A_i^{(n)}$ is in
$Sen(\Sigma^{(n)})$. \emph{Remark:} the $\delta$ operators capture the
temporal adaptation of conceptors. The symbol $A_i^{(n)}$ will be used
to denote a conceptor $a_i(n)$ in its adaptation version at time $n$,
and the sentence
$\delta^{(n)}_k\,A_i^{(n)}$ will be made to refer to $a_i(n+k)$. 
  \item If $\zeta, \xi \in Sen(\Sigma^{(n)})$, then $(\zeta \vee \xi),
(\zeta \wedge \xi), \neg \zeta \in Sen(\Sigma^{(n)})$. \emph{Remark:}
unlike in ECL there is no need for a notational distinction between
$\wedge$ and $\mbox{\emph{AND}}$ etc.
\item If $\zeta \in Sen(\Sigma^{(n)})$, then $\varphi(\zeta, \gamma)
\in Sen(\Sigma^{(n)})$ for every  $\gamma \in
[0,\infty]$ (this captures 
aperture adaptation).
  \item If $\zeta, \xi \in Sen(\Sigma^{(n)})$ and $0 \leq b \leq 1$,
then $\beta_b(\zeta, \xi) \in Sen(\Sigma^{(n)})$
(this will take care of linear blends $ b\zeta + (1-b)\xi$). 
\end{enumerate}
\emph{Remark:} Including $I$ and $0$ in the sentence syntax is a
convenience item. $0$ could be defined in terms of any $A_i^{(n)}$ by
$0 \stackrel{\wedge}{=} (\varphi(A_i^{(n)},\infty) \wedge \neg
\varphi(A_i^{(n)},\infty))$, and $I$ by $I  \stackrel{\wedge}{=} \neg
0$. Likewise, $\vee$ (or $\wedge$) could be dismissed because it can be expressed in
terms of $\wedge$ and $\neg$ ($\vee$ and $\neg$, respectively). 
\item For a signature morphism  $\phi^{(n+k,n)}: \Sigma^{(n+k)} \to
\Sigma^{(n)}$, $Sen(\phi^{(n+k,n)}): Sen(\Sigma^{(n+k)}) \to
Sen(\Sigma^{(n)})$ is the map defined inductively as follows:
\begin{enumerate}
\item $Sen(\phi^{(n+k,n)}): I \mapsto I, 0 \mapsto 0$.
\item $Sen(\phi^{(n+k,n)}): A_i^{(n+k)} \mapsto \delta^{(n)}_k\,
\phi^{(n+k,n)}(A_i^{(n+k)})$.  \emph{Remark 1:} When we use the
natural projections $\sigma^{(n)}: A_i \mapsto A_i^{(n)}$, this rule
could be more simply written as $Sen(\phi^{(n+k,n)}): A_i^{(n+k)}
\mapsto \delta^{(n)}_k\, A_i^{(n)}$.  \emph{Remark 2:} This is the
pivotal point in this entire definition, and the point where the
difference to customary views on logics comes to the surface most
conspicuously. Usually signature morphisms act on sentences by simply
re-naming all signature symbols that occur in a sentence. The
structure of a sentence remains unaffected, in agreement with the
motto ``truth is invariant under change of notation''. By contrast,
here a signature symbol $A_i^{(n+k)}$ is replaced by an temporal
change operator term $\delta^{(n)}_k\, A_i^{(n)}$, reflecting the new
motto ``meaning changes with time''. The fact that $\phi^{(n+k,n)}$
leads from $A_i^{(n+k)}$ to $A_i^{(n)}$ establishes ``sameness of
identity over time'' between $A_i^{(n+k)}$ and $A_i^{(n)}$. Usually
one would formally express sameness of identity of some mathematical
entity by using the same symbol to name that entity at different time
points. Here different symbols are used, and thus another mechanism
has to be found in order to establish that an entity named by
different symbols at different times remains ``the same''. The dotted
``identity'' lines in Figure \ref{figICLSigMorph} are fixed by the
signature morphisms $\phi^{(n+k,n)}$, not by using the same symbol
over time. \emph{Remark 3:} At this point it also becomes clear why
the signature morphisms $\phi^{(n+k,n)}: \Sigma^{(n+k)} \to
\Sigma^{(n)}$ lead backwards in time. A conceptor $a_i(n+k)$ in its
time-($n+k$) version can be expressed in terms of the earlier version
$a_i(n)$ with the aid of the temporal evolution operator $\delta$, but
in general an earlier version $a_i(n)$ cannot be expressed in terms of
a later $a_i(n+k)$. This reflects the fact that, seen as a trajectory
of an input-driven dynamical system, an agent life is (typically)
irreversible. To put it into everyday language, ``the future can be
explained from the past, but not vice versa''. 
\item $Sen(\phi^{(n+k,n)}): \delta^{(n+k)}_l \, A_i^{(n+k)} \mapsto
\delta^{(n)}_{(k+l)}\, \phi^{(n+k,n)}(A_i^{(n+k)})$.
\item For  $\zeta, \xi \in Sen(\Sigma^{(n+k)})$, put
\begin{eqnarray*}
Sen(\phi^{(n+k,n)}):&&\\
 (\zeta \vee \xi) & \mapsto & 
(Sen(\phi^{(n+k,n)})(\zeta) \vee Sen(\phi^{(n+k,n)})(\xi)),\\
(\zeta \wedge \xi)  &\mapsto & (Sen(\phi^{(n+k,n)})(\zeta) \wedge
Sen(\phi^{(n+k,n)})(\xi)),\\ 
\neg \zeta & \mapsto & \neg 
Sen(\phi^{(n+k,n)})(\zeta),\\ 
 \varphi(\zeta, \gamma) & \mapsto &
\varphi(Sen(\phi^{(n+k,n)})(\zeta), \gamma),\\ 
\beta_b(\zeta, \xi) & \mapsto & \beta_b(Sen(\phi^{(n+k,n)})(\zeta),
Sen(\phi^{(n+k,n)})(\xi)).
\end{eqnarray*} 
\end{enumerate}

  \item For every signature $\Sigma^{(n)} \in \mathbf{Sign}$, $M\!od(\Sigma^{(n)})$
is always the same category $\mathbf{Z}$ with objects  all
non-negative $M$-dimensional vectors $z^{.\wedge 2}$. There are no
model morphisms except the identity morphisms $z^{.\wedge 2} \stackrel{id}{\to} z^{.\wedge 2}$. 

\item For every morphism $\phi^{(n+k,n)} \in \mathbf{Sign}$,
$M\!od(\phi^{(n+k,n)})$ is the identity morphism of $\mathbf{Z}$.

\item As a preparation for defining the model relationships
$\models_{\Sigma^{(n)}}$ we assign by induction to every sentence $\chi \in
Sen(\Sigma^{(n)})$ an $M$-dimensional conception weight vector
$\iota(\chi)$ as follows:
\begin{enumerate}
\item $\iota(I) = (1,\ldots,1)'$ and $\iota(0) = (0,\ldots,0)'$.
\item $\iota( A_i^{(n)}) = \iota_{\mathcal{AL}}(n, (\sigma^{(n)})^{-1}
A_i^{(n)})$.
\item $\iota(\delta_k^{(n)} A_i^{(n)}) =  \iota_{\mathcal{AL}}(n+k,
(\sigma^{(n)})^{-1} A_i^{(n)})$.
\item Case  $\chi = (\zeta \vee \xi)$: $\iota(\chi) = \iota(\zeta)
\vee \iota(\xi)$ (compare Definition
\ref{def:Booleancvecs}). 
\item Case $\chi = (\zeta \wedge \xi)$: $\iota(\chi) = \iota(\zeta)
\wedge \iota(\xi)$.
\item Case $\chi = \neg \zeta$: $\iota(\chi) = \neg \iota(\zeta)$. 
\item Case $\chi =  \varphi(\zeta, \gamma)$: $\iota(\chi) =
\varphi(\iota(\zeta), \gamma)$ (compare Definition
\ref{def:Apadaptcvecs}).
\item Case $\chi = \beta_b(\zeta, \xi)$:  $\iota(\chi) =  b\,
\iota(\zeta) + (1-b) \iota(\xi)$. 
\end{enumerate}

\item For $z^{.\wedge 2} \in \mathbf{Z} = M\!od(\Sigma^{(n)})$ and $\chi \in
Sen(\Sigma^{(n)})$, the model relationship is defined by 
$$z^{.\wedge  2} \models_{\Sigma^{(n)}} \chi \quad \mbox{iff} \quad
z^{.\wedge 2} \,.\!\ast\, (z^{.\wedge 2} + 1)^{.\wedge  -1} \leq
\iota(\chi).$$ 
\end{enumerate}
\end{definition}

The satisfaction condition (\ref{eqSatisfactionCondition}) requires
that for $\chi^{(n+k)} \in Sen(\Sigma^{(n+k)})$ and $z^{.\wedge 2} \in
M\!od(\Sigma^{(n)})$ it holds that
\begin{displaymath}
z^{.\wedge 2} \models_{\Sigma^{(n)}} Sen(\phi^{(n+k,n)})(\chi^{(n+k)}
) \;\; \mbox{iff} \;\; M\!od(\phi^{(n+k,n)})(z^{.\wedge 2})
\models_{\Sigma^{(n+k)}} \chi^{(n+k)},
\end{displaymath}
which is equivalent to
\begin{equation}\label{eqSatCon}
z^{.\wedge 2} \models_{\Sigma^{(n)}} Sen(\phi^{(n+k,n)})(\chi^{(n+k)}
) \;\; \mbox{iff} \;\; z^{.\wedge 2}
\models_{\Sigma^{(n+k)}} \chi^{(n+k)}
\end{equation}
because $ M\!od(\phi^{(n+k,n)})$ is the identity morphism on
$\mathbf{Z}$. This follows directly from the following fact:

\begin{lemma}\label{lemmaSatCon}
$\iota\left( Sen(\phi^{(n+k,n)})(\chi^{(n+k)})\right) = \iota(\chi^{(n+k)})$, 
\end{lemma}
\noindent which in turn can be established by an obvious induction on the
structure of sentences, where the crucial steps are the cases (i)
$\chi^{(n+k)} = A_i^{(n+k)}$ and (ii) $\chi^{(n+k)} = \delta_l^{(n+k)}
A_i^{(n+k)}$. 

In case (i), $\iota(Sen(\phi^{(n+k,n)})(A_i^{(n+k)})) =
\iota(\delta_k^{(n)} \phi^{(n+k,n)} (A_i^{(n+k)})) = \iota(\delta_k^{(n)}
(\sigma^{(n)} \circ (\sigma^{(n+k)})^{-1}) (A_i^{(n+k)})) =
\iota_\mathcal{AL}(n+k, (\sigma^{(n)}))^{-1} (\sigma^{(n)} \circ
(\sigma^{(n+k)})^{-1}) (A_i^{(n+k)})) =  \iota_\mathcal{AL}(n+k,
(\sigma^{(n+k)})^{-1} (A_i^{(n+k)})) = \iota(A_i^{(n+k)})$. 

In case
(ii), conclude $\iota(Sen(\phi^{(n+k,n)})( \delta_l^{(n+k)}
A_i^{(n+k)})) = \iota(\delta^{(n)}_{(k+l)}\,
\phi^{(n+k,n)}(A_i^{(n+k)})) =$\\ $\iota(\delta^{(n)}_{(k+l)}\,(\sigma^{(n)} \circ (\sigma^{(n+k)})^{-1})
(A_i^{(n+k)})) =  \iota_{\mathcal{AL}}(n+k+l, (\sigma^{(n)}))^{-1} (\sigma^{(n)} \circ
(\sigma^{(n+k)})^{-1}) (A_i^{(n+k)})) =  \iota_{\mathcal{AL}}(n+k+l,
(\sigma^{(n+k)})^{-1} (A_i^{(n+k)})) =
\iota(\delta_l^{(n+k)}\,A_i^{(n+k)})$.

An important difference between ``traditional'' logics and the ICL of
an agent $\mathcal{AL}$ concerns different intuitions about semantics.
Taking first-order logic as an example of a traditional logic, the
``meaning'' of a first-order sentence $\chi$ with signature $\Sigma$
is the class of all of its models. Whether a set is a model of $\chi$
depends on how the symbols from $\Sigma$ are extensionally interpreted
over that set.  First-order logic by itself does not prescribe how the
symbols of a signature have to be interpreted over some domain. In
contrast, an ICL is defined with respect to a concrete agent
$\mathcal{AL}$, which in turn uniquely fixes how the symbols from an
ICL signature $\Sigma^{(n)}$ \emph{must} be interpreted -- this is the
essence of points \emph{7. (a -- c)} in Definition \ref{defICL}.

Logical entailment in an ICL coincides with abstraction of conceptors:

\begin{proposition}\label{propEntailICL}
In an ICL, for all $\zeta,\xi \in Sen(\Sigma^{(n)})$ it holds that 
\begin{equation}\label{eqEntailICL}
\zeta \models_{\Sigma^{(n)}} \xi \quad \mbox{iff} \quad \iota(\zeta) \leq \iota(\xi).
\end{equation}
\end{proposition}

The simple proof is given in Section \ref{proofRentailment}. In an
agent's lifetime ICL, for any sentence  $\chi \in Sen(\Sigma^{(n)})$
the concrete interpretation  $\iota(\chi) \in [0,1]^M$ can be effectively computed
via the rules stated in Nr.\ \emph{7} in Definition \ref{defICL},
provided one has access to the identifiable conceptors $a_i(n)$ in the
agent's life. Since for two vectors $a,b \in [0,1]^M$ it can be
effectively checked whether $a \leq b$, it is decidable whether
$\iota(\zeta) \leq \iota(\xi)$. An ICL is therefore decidable. 

Seen from a categorical point of view, an ICL is a particularly
small-size institution. Since the (only) category in the image of
$M\!od$ is a set, 
we can regard the functor $M\!od$ as having codomain
$\mathbf{Set}^{\mbox{\scriptsize op}}$ instead of
$\mathbf{Cat}^{\mbox{\scriptsize op}}$. Also the category
$\mathbf{Sign}$ is small, that is, a set. Altogether, an ICL institution
nowhere needs proper classes.

The ICL definition I gave here is very elementary. It could easily be
augmented in various natural ways, for instance by admitting
permutations and/or projections of conceptor vector components as
model morphisms in $\mathbf{Z}$, or allowing conceptors of different
dimensions in the makeup of an agent life. Likewise it is
straightforward to spell out definitions for an agent life and its ICL
for matrix-based conceptors. In the latter case, the referents of
sentences are correlation matrices $R$, and the defining equation for
the model relationship appears as

$$R \models_{\Sigma^{(n)}} \chi \quad \mbox{iff}  \quad R (R + I)^{ -1} \leq \iota(\chi).$$

  In traditional logics and their applications, an important role is
  played by \emph{calculi}. A calculus is a set of syntactic
  transformation rules operating on sentences (and expressions
  containing free variables) which allows one to derive purely
  syntactical proofs of logical entailment statements $\zeta \models
  \xi$.  Fundamental properties of familiar logics, completeness and
  decidability in particular, are defined in terms of calculi. While
  it may be possible and mathematically interesting to design
  syntactical calculi for ICLs, they are not needed because $\zeta
  \models \xi$ is decidable via the semantic equivalent $\iota(\zeta)
  \leq \iota(\xi)$. Furthermore, the ICL decision procedure is
  computationally very effective: only time $O(1)$ is needed
  (admitting a parallel execution) to determine whether some
  conception vector is at most as large as another in all components.
  This may help to explain why humans can so quickly carry out many
  concept subsumption judgements (``this looks like a cow to me'').

\textbf{A note added to revision 4:} Formal investigations into
establishing a formal conceptor logic have
in the meantime been made by theoretical computer scientists who are
much better qualified than I am \cite{Mossakowskietal19}.

\subsection{Final Summary and Outlook}\label{secOutlook}

Abstracting from all technical detail, here is a summary account of
the conceptor approach:

\begin{description}
\item[From neural dynamics to conceptors.] Conceptors capture the
shape of a neural state cloud by a positive semi-definite operator.
\item[From conceptors to neural dynamics.] Inserting a conceptor into
a neurodynamical state update loop allows to select and stabilize a
previously stored neural activation pattern. 
\item[Matrix conceptors] are useful in machine learning applications
and as mathematical tools for analysing patterns emerging in nonlinear
neural dynamics.
\item[Random feature conceptors] are not biologically apriori implausible and
can be neurally coded by single neurons.
\item[Autoconceptor] adaptation dynamics leads to content-addressable
neural memories of dynamical patterns and to signal filtering and
classification systems.
\item[Boolean operations and the abstraction ordering on conceptors]
establish a bi-directional connection between logic and neural
dynamics. 
\item[From static to dynamic models.] Conceptors allow to understand
and control dynamical patterns in scenarios that previously have been
mostly restricted to static patterns. Specifically this concerns
content-addressable memories, morphing, ordering concepts in logically
structured abstraction hierarchies, and top-down hypothesis control in
hierarchical architectures.   
\end{description}

The study of conceptors is at an early stage. 
There are numerous natural directions for next steps in 
conceptor research:

\begin{description}
\item[Affine conceptor maps.] Conceptors constrain reservoir states by
applying a positive semi-definite map. This map is adapted to the
``soft-bounded'' 
linear subspace visited by a reservoir when it is driven through a
pattern. If the pattern-driven reservoir states do not have zero mean,
it seems natural to first subtract the mean before applying the
conceptor. If the mean is $\mu$, this would result in an update loop
of the form $x(n+1) = \mu + C (\tanh(...) - \mu)$. While this may be
expected to improve control characteristics of pattern re-generation,
it is not immediately clear how Boolean operations transfer to such
affine conceptors which are characterized by pairs $(C, \mu)$.
\item[Nonlinear conceptor filters.] Even more generally, one might
conceive of conceptors which take the form of nonlinear filters, for
instance instantiated as feedforward neural networks. Such filters $F$
could be trained on the same objective function as I used for
conceptors, namely minimizing $E[\|z - F(z)\|^2] +
\alpha^{-2}\|F\|^2$, where $\|F\|$ is a suitably chosen norm defined
for the filter. Like with affine conceptor maps, the pattern specificity of
such nonlinear conceptor filters would be greater than for our
standard matrix conceptors, but again it is not clear how logical
operations would extend to such filters. 
\item[Basic mathematical properties.] From a mathematics viewpoint,
there are some elementary questions about conceptors which should be
better understood, for instance:
\begin{itemize}
\item What is the relationship between Boolean operations, aperture
adaptation, and linear blending? in particular, can the latter be
expressed in terms of the two former?
\item Given a dimension $N$, what is the minimial number of ``basis'' conceptors
such that the transitive closure under Boolean operations, aperture
adaptation, and/or linear blending is the set of all $N$-dimensional
conceptors?
\item Are there normal forms for expressions which compose conceptors
by Boolean operations, aperture
adaptation, and/or linear blending?
\item Find conceptor analogs of standard structure-building
mathematical operations, especially products. These would be needed to
design architectures with several conceptor modules (likely of
different dimension) where the ``soft subspace constraining'' of the
overall dynamics works on the total architecture state
space. Presumably this leads to tensor variants of conceptors. This
may turn out to be a challenge because a mathematical theory of
``semi positive-definite tensors'' seems  to be only in its infancy
(compare \cite{Qi12}).   
\end{itemize}
\item[Neural realization of Boolean operations.] How can Boolean
operations be implemented in biologically not impossible neural
circuits? 
  \item[Complete analysis of autoconceptor adaptation.] The analysis
of autoconceptor adaptation in Section \ref{subsecCAD} is preliminary
and incomplete. It only characterizes certain aspects of fixed-point
solutions of this adaptation dynamics but remains ignorant about the
effects that the combined, nonlinear conceptor-reservoir update
dynamics may have when such a fixed-point solution is perturbed.
Specifically, this reservoir-conceptor interaction will have to be
taken into account in order to understand the dynamics in the center
manifold of fixed-point solutions. 
\item[Applications.] The usefulness of conceptors as a practical tool
for machine learning and as a modeling tool in the cognitive and
computational neurosciences will only be established by a suite of
successful applications. 
\end{description}

\newpage

\section{Documentation of Experiments and Methods}\label{secDocExp}

In this section I provide details of all simulation experiments
reported in Sections \ref{secOverview} and \ref{secTheoryDemos}.

\subsection{General Set-Up, Initial Demonstrations (Section
  \ref{secOverview} and Section
  \ref{sec:InitialDrivingDemo} -
  \ref{sec:RetrieveGeneric})}\label{secGeneralSetupExpDetail}

A reservoir with $N = 100$ neurons, plus one input unit and one
output unit  was created with a random input weight vector
$W^{\mbox{\scriptsize in}}$, a random bias $b$ and preliminary
reservoir weights $W^\ast$, to be run according to the update equations

\begin{eqnarray}
x(n+1) & = & \tanh(W^\ast\,x(n) + W^{\mbox{\scriptsize
    in}}\,p(n+1) + b),\label{somEq01_Sec3}\\
y(n) & = &  W^{\mbox{\scriptsize out}}\, x(n). \nonumber
\end{eqnarray}

Initially $W^{\mbox{\scriptsize out}}$ was left
undefined. The input weights were sampled from a normal distribution
$\mathcal{N}(0,1)$ and then rescaled by a factor of 1.5. The bias was
likewise sampled from $\mathcal{N}(0,1)$ and then rescaled by 0.2.  The
reservoir weight matrix $W^\ast$ was first created as a sparse random
matrix with an approximate density of 10\%, then scaled to obtain a
spectral radius (largest absolute eigenvalue) of 1.5. These scalings
are typical in the field of reservoir computing
\cite{LukoseviciusJaeger09} for networks to be employed in
signal-generation tasks.

For each of the four driving signals $p^j$ a training time series of
length 1500 was generated.  The reservoir was driven with these
$p^j(n)$ in turn, starting each run from a zero initial reservoir
state. This resulted in reservoir state responses $x^j(n)$.  The first
500 steps were discarded in order to exclude data influenced by the
arbitrary starting condition, leading to four 100-dimensional
reservoir state time series of length $L = 1000$, which were recorded
into four $100 \times 1000$ state collection matrices $X^j$, where $X^j(:,n) = x^j(n+500)$ ($j = 1,
\ldots,4$). Likewise, the corresponding driver signals were recorded
into four pattern collection (row) vectors $P^j$ of size $1 \times
1000$. In addition to this, a version $\tilde{X}^j$ of $X^j$ was built,
identical to $X^j$ except that it was delayed by one step:
$\tilde{X}^j(:,n) = x^j(n+499)$. These
collections were then concatenated to obtain $X = [X^1 | X^2 | X^3 |
X^4], \tilde{X} = [\tilde{X}^1 | \tilde{X}^2 | \tilde{X}^3 |
\tilde{X}^4], P = [P^1 | P^2 | P^3 | P^4]$.

The ``PC energy'' plots in Figure \ref{Fig1}  render the
singular values of the correlation matrices $X^j (X^j)' / L$. 

The output weights $W^{\mbox{\scriptsize out}}$ were computed as the
regularized Wiener-Hopf solution (also known as ridge regression, or
Tychonov regularization)

\begin{equation}
W^{\mbox{\scriptsize out}} = ((XX' + \varrho^{\mbox{\scriptsize out}}
 I_{N \times N})^{-1}\; X\,P')', \label{somEQ03}
\end{equation}

\noindent where the regularizer $\varrho^{\mbox{\scriptsize out}}$ was
 set to 0.01. 

Loading: After loading, the reservoir weights $W$ should lead to the approximate
 equality $W x^j(n) \approx W^\ast\,x^j(n) + W^{\mbox{\scriptsize
    in}}\,p^j(n+1)$, across all patterns $j$, which leads to the
 objective of minimizing the 
 squared error $\epsilon^j(n+1) = ((\tanh^{-1}(x^j(n+1)) - b) - W x^j(n))^2$, averaged  over all
 four $j$ and training time points. Writing $B$ for the $100 \times (4
 * 1000)$ matrix whose columns are all identical equal to $b$, this
 has the ridge
 regression solution

 \begin{equation}
W = ((\tilde{X}\tilde{X}' + \varrho^{\mbox{\scriptsize W}}
 I_{N \times N})^{-1} \; \tilde{X}\,(\tanh^{-1}(X) - B)' )', \label{somEQ04}
\end{equation}

\noindent where the regularizer $\varrho^{\mbox{\scriptsize W}}$ was
set to 0.0001. To assess the accuracy of the weight computations, the
training normalized root mean square error (NRMSE) was computed.  For
the readout weights, the NRMSE between  $y(n) = W^{\mbox{\scriptsize out}} x(n)$ and
the target $P$ was 0.00068. For
the reservoir weights, the average (over  reservoir neurons $i$, times
$n$ and
patterns $j$) NRMSE
between $W(i,:) x^j(n)$ and the target $W^\ast(i,:)\,x^j(n) + W^{\mbox{\scriptsize
    in}}(i)\,p^j(n+1)$ was 0.0011.

In order to determine the accuracy of fit between the original driving
signals $p^j$ and the network observation outputs $y^j(n) =
W^{\mbox{\scriptsize out}}\, C^j \, \tanh(W\,x(n-1) + b)$ in the
conceptor-constrained autonomous runs, the driver signals and the
$y^j$ signals were first interpolated with cubic splines
(oversampling by a factor of 20). Then a segment length 400 of the oversampled
driver (corresponding to 20 timesteps  before interpolation) was
shifted over the oversampled $y^j$ in search of a position of best
fit. This is necessary to compensate for the  indeterminate phaseshift
between the driver data and the network outputs. The NRMSEs given in
Figure \ref{Fig1} were calculated from the best-fit phaseshift
position, and the optimally phase-shifted version of $y^j$ was also
used for the plot.

\subsection{Aperture Adaptation  (Sections
  \ref{secApertureAdapt} and \ref{secApAdjustGuide})} \label{secExpDetailApertureAdapt}

\emph{Data generation.} For the R\"ossler attractor,
training time series were obtained from running simple Euler
approximations of the following ODEs:

\begin{eqnarray*}
\dot{x} & =& -(y + z)\\
\dot{y} & = & x + a \,y\\
\dot{z} & = & b + x \,z - c \,z,
\end{eqnarray*}

\noindent using  parameters $a = b = 0.2, c = 8$. The evolution of this system was
Euler approximated   with stepsize $1/200$ and the resulting discrete
time series was then subsampled by 150. The $x$ and $y$ coordinates
were assembled in a 2-dimensional driving sequence, where each of the
two channels was shifted/scaled to a range of $[0,1]$. For the Lorenz
attractor, the ODE

\begin{eqnarray*}
\dot{x} & = & \sigma(y - x)\\
\dot{y} & = & r\, x - y - x\,z\\
\dot{z} & = & x \,y - b \,z
\end{eqnarray*}

\noindent with $\sigma = 10, r = 28, b = 8/3$ was Euler-approximated
with stepsize $1/200$ and subsequent subsampling by 15. The $x$ and
$z$ coordinates were collected in a 2-dimensional driving sequence, again each
channel normalized to a range of $[0,1]$. The Mackey Glass timeseries
was obtained from the delay differential equation

\begin{displaymath}
\dot{x}(t) = \frac{\beta \, x(t-\tau)}{1 + x(t-\tau)^n} - \gamma\, x(t)
\end{displaymath}

\noindent with $\beta = 0.2, n = 10, \tau = 17, \gamma = 0.1$, a
customary setting when this attractor is used in neural network 
demonstrations. An Euler approximation with stepsize $1/10$ was
used. To obtain a 2-dim timeseries that could be fed to the reservoir
through the same two input channels as the other attractor data, pairs
$x(t), x(t - \tau)$ were combined into 2-dim vectors. Again, these two
signals were normalized to the $[0,1]$ range. The H\'{e}non attractor
is governed by the iterated map

\begin{eqnarray*}
x(n+1) & = & y(n) + 1 - a\,x(n)\\
y(n+1) & = & b \, x(n),
\end{eqnarray*}

\noindent where I used $a = 1.4, b = 0.3$. The two components were
filed into a 2-dim timeseries $(x(n),y(n))'$ with no further
subsampling, and again normalization to a range of $[0,1]$ in each
component. 

\emph{Reservoir setup.} A 500-unit reservoir RNN was created with a
normal distributed, 10\%-sparse weight matrix $W^\ast$ scaled to a spectral radius of $0.6$. The bias vector $b$
and input weights $W^{\mbox{\scriptsize in}}$ (sized $400 \times 2$ for
two input channels) were sampled from standard normal distribution and
then scaled by $0.4$ and $1.2$, respectively. These scaling parameters
were found by a (very coarse) manual optimization of the performance of the pattern
storing process. The network size was chosen large enough to warrant a
robust trainability of the four chaotic patterns. Repeated executions
of the experiment with different randomly initialized weights (not
documented) showed no significant differences. 

\emph{Pattern storing.} The $W^\ast$ reservoir was
driven, in turn, by 2500 timesteps of each of the four chaotic
timeseries. The first 500 steps were discarded to account for initial
reservoir state washout, and the remaining 4 $\times$ 2000 reservoir
states were collected in a $500 \times 8000$ matrix $X$. From this,
the new reservoir weights $W$ were computed as in (\ref{somEQ04}),
with a regularizer $\varrho^{\mbox{\scriptsize W}} = 0.0001$.  The
readout weights were computed as in (\ref{somEQ03}) with regularizer
$\varrho^{\mbox{\scriptsize out}} = 0.01$. The average NRMSEs obtained
for the reservoir and readout weights were $0.0082$ and $0.013$,
respectively.

\emph{Computing conceptors.} From each of the four $n = 2000$ step
reservoir state sequences $X$ recorded in the storing procedure,
obtained from driving the reservoir with one of the four chaotic
signals, a preliminary correlation matrix $\tilde{R} = X X' / 2000$
and its SVD $U\tilde{S}U' = \tilde{R}$ were computed. This correlation matrix was then used to obtain a
conceptor associated with the respective chaotic signal, using an
aperture $\alpha = 1$. From these unit-aperture conceptors, versions
with differing $\alpha$ were obtained through aperture adaptation per
(\ref{eqSemAlpha3}).

In passing I note that the overall stability and parameter robustness
of this simulation can be much improved if small singular values in
$\tilde{S}$ (for instance, with values smaller than 1e-06) are zeroed,
obtaining a clipped $S$, from which a ``cleaned-up'' correlation
matrix $R = U S U'$ would be computed. This would lead to a range of
well-working apertures spanning three orders of magnitude (not
shown). I did not do this in
the reported simulation in order to illustrate the effects of too
large apertures; these effects would be partly suppressed when the
spectrum of $R$ is clipped.

\emph{Pattern retrieval and plotting.} The loaded network was run
using the conceptor-constrained update rule $x(n+1) = C \;
\tanh(W\,x(n) + W^{\mbox{\scriptsize in}}\,p(n+1) + b)$ with various
$C = \varphi(C_\#, \gamma_{\#,i})$ ($\# = \mbox{R, L, MG, H}, i =
1,\ldots,5$) for 800 steps each time, of which the first 100 were
discarded to account for initial state washout. The delay embedding
plots in Figure \ref{figChaosClover} were generated from the remaining
700 steps. Embedding delays of 2, 2, 3, 1 respectively were used for plotting
the four attractors.

For each of the four 6-panel blocks in Figure \ref{figChaosClover},
the five aperture adaptation factors $\gamma_{\#,i}$ were determined in the
following way. First, by visual inspection, the middle $\gamma_{\#,3}$
was determined to fall in the trough center of the attenuation plot of
Fig.\ \ref{figBlindout} {\bf A}. Then the remaining $\gamma_{\#,1},
\gamma_{\#,2},\gamma_{\#,4},\gamma_{\#,5}$ were set in a way that (i) the
entire $\gamma$ sequence was a geometrical progression, and (ii) that
the plot obtained from the first $\gamma_{\#,1}$ was visually strongly corrupted.

\subsection{Memory Management, Demo 1 (Section
  \ref{subsec:memManageDemo1})}\label{secDetailmemmanageDemo1} 

\emph{Data:} The patterns were either sines
sampled at integer fractions of $2\pi$ or random periodic patterns,
always scaled to a range of $[-0.9, 0.9]$. Period lengths were picked
randomly between 3 and 15.  The length of pattern signals used for
loading was $L = 100$, with an additional washout of the same length.

\emph{Parameters:} Reservoir size $N = 100$,
$W^\ast$ sparse with density approximately 10\%, weights sampled
from a standard normal distribution, rescaled to a
spectral radius of 1.5. The input weights $W^{\mbox{\scriptsize in}}$
were non-sparse, sampled from a standard normal distribution, scaled
by 1.5. The bias $b$ was sampled from a standard normal distribution
and scaled by 0.25. Ridge regression coefficients 0.01 and 0.001 for
  computing the increments to $W^{\mbox{\scriptsize out}}$ and $D$,
  respectively. Data from 100 reservoir
  update steps were used for computing conceptors and regressions. 
  
  \emph{Computing test errors:} The conceptors $C^j$ obtained during
  the loading procedure were used with an aperture $\alpha = 1000$ by
  running the reservoir via $\mathbf{x}(n+1) = C^j\, \tanh(W^\ast\,
  \mathbf{x}(n) + D\, \mathbf{x}(n) + \mathbf{b})$, starting from
  random initial states. After a washout period of 200 steps, the network outputs
  $y^j(n)$ were recorded for 50 steps. A 20-step portion of the
  original driver pattern was interpolated (supersampling rate 10) and
  shifted over a likewise interpolation version of these outputs. The
  best matching shift position was used for plotting and NRMSE
  computation. Such shift-search for a good fit is necessary because
  the autonomous runs are not phase-synchronized with the original
  drivers.

\emph{Non-incremental loading:} For the standard non-incremental
loading experiment that served as comparison and yielded the results
marked by dots in the left panel of Figure \ref{figMemManOverview},
the same parameters (scalings, aperture) were used as for the
incremental loading procedure. 

\subsection{Memory Management, Demo 2 (Section
  \ref{subsec:memManageDemo2})}\label{secDetailmemmanageDemo2} 

 The set-up is the same as for the
  integer-periodic basic demo. Scalings: spectral radius of $W^\ast$:
  1.5; $W^{\mbox{\scriptsize in}}$: 1.5; $b$: 1.0. Aperture: 10. Ridge
  regression coefficients 0.02 both for increments to
  $W^{\mbox{\scriptsize out}}$ and $D$. Data collection runlengths in
 loading: 500 (plus 200 washout). Test runlength: 800 (plus 200
  washout). Test output was compared to
  20-step sample from original driver after interpolation and shift as
  described for the  integer-periodic basic demo.

\subsection{Memory Management, Close-Up Inspection (Section
  \ref{subsecIncLoadSurvey})}\label{secDetailMemManageCloseup} 

\emph{Integer-periodic patterns, period length 6 (condition IP6):}
Scalings of reservoirs: spectral radius of $W^\ast$: 1.5;
$W^{\mbox{\scriptsize in}}$: 1.5; $b$: 0.5. Aperture: 1000. Ridge
regression coefficients 0.01 for $W^{\mbox{\scriptsize out}}$ and
0.001 for
$D$. Data collection runlengths in loading: 100 (plus 100 washout) per
pattern. Pattern recall tests were started from random network
states and recorded for 50 steps after a washout of 200. Recall NRMSEs
were computed as described in Section \ref{secDetailmemmanageDemo1}. 

\emph{Irrational-period patterns from a parametric family (condition PF):}
Scalings of reservoirs: spectral radius of $W^\ast$: 1.5;
$W^{\mbox{\scriptsize in}}$: 1.5; $b$: 1.0. Aperture: 10. Ridge
regression coefficients 0.02 both for $W^{\mbox{\scriptsize out}}$ and
$D$. Data collection runlengths in loading: 100 (plus 100 washout) per
pattern. Pattern recall tests were started from random network
states and recorded for 80 steps after a washout of 200.

\emph{Integer-periodic patterns, period 3  (condition IP3):} Same as
in condition IP6.

\subsection{Memory Management, Arbitrary Patterns (Section
  \ref{subsecIncLoadMixed})}\label{secDetailMemManageArbitrary} 

Spectral radius of $W^\ast$: 1.5; $W^{\mbox{\scriptsize
      in}}$: 1.5; $b$: 0.25. Apertures of ``raw'' conceptors $C^m$
      (before ``rectangularization'') were set to
  10 for integer-periodic patterns and to 1.2 for the quasiperiodic
  patterns.  Ridge regression coefficients: 0.001 for increments to
  $W^{\mbox{\scriptsize out}}$ and 0.005 for increments to $D$. Data
  collection runlengths in loading: 500 (plus 100 washout). Test
  runlength: 200 (plus 100 washout), compared to 20-step sample from
      original driver as described in Section
      \ref{secDetailmemmanageDemo1}. --- For the standard
      non-incremental loading procedure that served as comparison
      (dots in Figure \ref{fig16_Rev1}, left panel), manually optimized
      conceptor apertures of 1000 | 10 were used for the
      integer-periodic | quasi-periodic patterns.

\subsection{Content-Addressable Memory  (Section
  \ref{secAutoCMemExample})} \label{secContentAddressableExperiment}

\emph{Network setup.} Reservoir network matrices $W$ were sampled
sparsely (10\% nonzero weights) from a normal distribution, then
scaled to a spectral radius of 1.5 in all experiments of this section.
Reservoir size was $N = 100$ for all period-4 experiments and the
unrelated patterns experiment, and $N = 200$ for all experiments that
used mixtures-of-sines patterns. For all experiments in the section,
input weights $W^{\mbox{\scriptsize in}}$ were randomly sampled from
the standard normal distribution and rescaled by a factor of 1.5. The
bias vector $b$ was likewise sampled from the standard normal
distribution and rescaled by 0.5. These scaling parameters had been
determined by a coarse manual search for a well-working configuration
when this suite experiments was set up. The experiments are remarkably
insensitive to these parameters.

\emph{Storing patterns.} The storage procedure was  set up identically
for all experiments in this section. The reservoir was driven by the
$k$ loading patterns in turn for $l = 50$ steps (period-4 and
unrelated patterns) or $l = 500$ steps (mix-of-sines) time steps, plus
a preceding 100 step initial washout. The observed network states
$x(n)$ were concatenated into a $N \times kl$ sized state collection
matrix $X$, and the one-step earlier states $x(n-1)$ into a matrix
$\tilde{X}$ of same size. The  driver pattern signals were
concatenated into a $kl$ sized row vector $P$. The readout
weights $W^{\mbox{\scriptsize out}}$ were then obtained by ridge
regression via $W^{\mbox{\scriptsize out}} = ((XX' + 0.01\,I)^{-1}\;
XP')'$, and $D$ by $D = ((\tilde{X} \tilde{X}' + 0.001\,I)^{-1}\;
\tilde{X} (W^{\mbox{\scriptsize in}}\,P)')'$. 

\emph{Quality measurements.} After the cueing, and at the end of the
recall period (or at the ends of the three interim intervals for some
of the experiments), the  current conceptor $C$ was tested for
retrieval accuracy as follows. Starting from the current network state
$x(n)$, the reservoir network was run for 550 steps, constrained by
$C$. The first 50 steps served as washout and were discarded. The
states $x$ from the last 500 steps were transformed to patterns by
applying $W^{\mbox{\scriptsize out}}$, yielding a 500-step pattern
reconstruction. This was interpolated with cubic splines and then
sampled at double resolution, leading to a 999-step pattern
reconstruction $\tilde{y}$. 

A 20-step template sample of the original pattern was similarly
interpolated-resampled and then  passed
over $\tilde{y}$, detecting the 
best-fitting position where the the NRMSE between the target template
and $\tilde{y}$ was minimal (this shift-search accomodated for unknown
phase shifts between the target template and $\tilde{y}$). This
minimal NRMSE was returned as measurement result.

\emph{Irrational-period sines.} This simulation was done exactly as
the one before, using twelve irrational-period sines as reference
patterns.

\subsection{The Japanese Vowels Classification (Section
  \ref{subsec:JapVow})}\label{subsecJapVowExp}

\emph{Network setup.}  In each of the 50 trials, the weights in a fully connected,
$10 \times 10$ reservoir weight matrix, a 12-dimensional input weight
vector, a 10-dimensional bias vector, and a 10-dimensional start state
were first sampled from a normal
distribution, then rescaled to a spectral radius of 1.2 for $W$, and
by factors of 0.2, 1, 1 for $W^{\mbox{\scriptsize in}}, b,
x_{\mbox{\scriptsize start}}$ respectively. 

\emph{Numerical determination of best aperture.} To determine
$\gamma_i^+$, the quantities $\| \varphi(\tilde{C}^+_i, 2^g)
\|^2_{\mbox{\scriptsize fro}}$ were computed for $g = 0, 1, \ldots,
8$. These values were interpolated on a 0.01 raster with cubic
splines, the support point $g_{\mbox{\scriptsize max}}$ of the maximum
of the interpolation curve was detected, returning $\gamma_i^+ =
2^{g_{\mbox{\scriptsize max}}}$.

\emph{Linear classifier training.} The linear classifier that serves
as a baseline comparison was designed in essentially the same way as
the Echo State Networks based classifiers which  in
\cite{Jaegeretal07} yielded zero test misclassifications (when
combining 1,000 such classifiers made from 4-unit networks) and 2 test
misclassifications (when a single such classifier was based on a 1,000
unit reservoir), respectively. Thus, linear classifiers based on
reservoir responses outperform all other reported methods on this
benchmark and therefore provide a substantive baseline information. 

In detail, the linear classifier was learnt from 270 training data
pairs of the form $(z, y_{\mbox{\scriptsize teacher}})$, where the $z$
were the same 88-dimensional vectors used for constructing conceptors,
and the $y_{\mbox{\scriptsize teacher}}$ were 9-dimensional, binary
speaker indicator vectors with a ``1'' in the position of the speaker
of $z$. The classifier consists in a $9 \times 88$ sized weight matrix
$V$, and the cost function was the quadratic error $\| V z -
y_{\mbox{\scriptsize teacher}}\|^2$. The classifier weight matrix $V$
which minimized this cost function on average over all training
samples was computed by linear
regression with Tychonov regularization, also known as ridge
regression \cite{Verstraeten09}. The Tychonov parameter which
determines the degree of regularization was determined by a grid
search over a 5-fold
cross-validation on the training data. Across the 50 trials it was
found to vary quite widely in a range between 0.0001 and 0.25; in a
separate auxiliary investigation it was also found that the effect
of variation of the regularizer within this range was very small and
the training of the linear classifier can therefore be considered
robust.  

In testing, $V$ was used to determine classification decisions by
computing $y_{\mbox{\scriptsize test}} = V z_{\mbox{\scriptsize
    test}}$ and opting for the index of the largest entry in
$y_{\mbox{\scriptsize test}}$ as the speaker.

\subsection{Conceptor Dynamics Based on RFC Conceptors  (Section
  \ref{secBiolPlausible})}\label{secExpBiolPlausible}

\emph{Network setup.} In both experiments reported in this section, the
same reservoir made from $N = 100$ units was used. $F$ and $G$ were
full matrices with entries first sampled from the standard normal
distribution. Then they were both scaled by an identical factor $a$
such that the product $a^2\,G\,F$ attained a spectral radius of 1.4.
The input weight vector $W^{\mbox{\scriptsize in}}$ was sampled from
the standard normal distribution and then scaled by 1.2. The bias $b$
was likewise sampled from the normal distribution and then scaled by
0.2.

These values were determined by coarse manual search, where the main
guiding criterion was recall accuracy. The settings were rather robust
in the first experiment which used stored $c^j$. The spectral radius
could be \emph{individually} varied from 0.6 to 1.45, the input weight
scaling from 0.3 to 1.5, the bias scaling from 0.1 to 0.4, and the
aperture from 3 to 8.5, while always keeping the final recall NRMSEs
for all four patterns below 0.1.  Furthermore, much larger
\emph{combined} variations of these scalings were also possible (not
documented).

In the second experiment with content-addressed recall, the functional
parameter range was much narrower. Individual parameter variation
beyond $\pm$5\% was disruptive. Specifically, I observed a close
inverse coupling between spectral radius and aperture: if one of the
two was raised, the other had to be lowered.

\emph{Loading procedure.} The loading procedure is described in some
detail in the report text.  The mean NRMSEs on training
data was 0.00081 for recomputing $G$, 0.0011 for $H$, and 0.0029 for
$W^{\mbox{\scriptsize out}}$. The mean absolute size of matrix
elements in $G$ was 0.021, about a third of the mean absolute size of
elements of $G^\ast$.

\emph{Computing NRMSEs.} The NRMSE comparison between the re-generated
patterns at the end of the $c$ adaptation and the original drivers was
done in the same way as reported on earlier occasions (Section
\ref{secContentAddressableExperiment}), that is, invoking spline
interpolation of the comparison patterns and optimal phase-alignment.

\subsection{Hierarchical Classification and Filtering Architecture
  (Section
  \ref{secHierarchicalArchitecture})}\label{secDocHierarchical}

\emph{Module setup.} The three modules are identical copies of each
other. The reservoir had $N = 100$ units and the feature space had a
dimension of $M = 500$. The input weight matrix $W^{\mbox{\scriptsize
    in}}$ was sampled from the standard normal distribution and
rescaled by a factor of 1.2. The reservoir-featurespace projection and
backprojection matrices $F, G^\ast$ (sized $N \times M$)  were first sampled from
the standard normal distribution, then linearly rescaled by a common
factor such that the $N \times N$ matrix $G^\ast F'$ (which functionally corresponds to
an internal reservoir weight matrix) had a spectral radius of 1.4. The
bias $b$ was likewise sampled from the standard normal distribution
and then scaled by 0.2. 

A regularization procedure was then applied to $G^\ast$ to give $G$ as
follows. The preliminary module was driven per
$$z(n+1) = F'r(n), \quad r(n+1) = \tanh(G^\ast z(n+1) +
W^{\mbox{\scriptsize in}} u(n) + b),$$
with an i.i.d. input signal
$u(n)$ sampled uniformly from $[-1,1]$, for 1600 steps (after
discarding an initial washout). The values obtained for $z(n+1)$ were
collected as columns in a $M \times 1600$ matrix $Z$. The final $G$ was
then computed by a ridge regression with a regularizer $a = 0.1$ by
$$G  =  ((ZZ' + aI)^{-1}\,Z\,(G^\ast Z)')'.$$ 
In words, $G$ should behave as the initially sampled $G^\ast$ in a
randomly driven module, but do so with minimized weight sizes. This
regularization was found to be important for a stable working of the
final architecture. 

\emph{Training.} The input recreation weights $H$ (size $1 \times M$)
and $W^{\mbox{\scriptsize out}}$ (size $1 \times N$) were trained by
driving a single module with the four target patterns, as follows. The
module was driven with clean $p^1, \ldots, p^4$ in turn, with
auto-adaptation of conception weights $c$ activated:
\begin{eqnarray*}
z(n+1) & = & c(n) \, .\!\ast \, F'\, r(n),\\
r(n+1) & = & \tanh(G\, z(n+1) + W^{\mbox{\scriptsize in}}\, p^j(n) +
b),\\
c(n+1) & = & c(n) + \lambda_c \left(\left(z(n+1) - c(n)\, .\!\ast \, z(n+1)\right)\,
  .\!\ast \, z(n+1)  - \alpha^{-2}\,c(n)   \right), 
\end{eqnarray*}
where the $c$ adaptation rate was set to $\lambda_c = 0.5$, and an
aperture $\alpha = 8$ was used. After discarding initial washouts in
each of the four driving conditions 
(long enough for $c(n)$ to stabilize), 400 reservoir state vectors
$r(n+1)$, 400 $z$ vectors $z(n)$ and 400 input values $p^j(n)$ were collected for $j =
1,\ldots,4$, and collected column-wise in matrices $R$ (size $N
\times 1600$), $Z$ (size $M \times 1600$) and $Q$ (size $1 \times
1600$), respectively. In addition, the 400-step submatrices $Z^j$ of $Z$
containing the $z$-responses of the module when driven with $p^j$ were
registered separately. 

The output weights were then computed by ridge regression on the
objective to recover $p^j(n)$ from $r(n+1)$ by
$$W^{\mbox{\scriptsize out}} = \left((RR' + aI)^{-1}\,RQ' \right)',$$
using a regularizer $a = 0.1$. In a similar way, the input recreation
weights were obtained as $$H = \left((ZZ' + aI)^{-1}\,ZQ' \right)'$$
with a regularizer $a = 0.1$ again. The training NRMSEs for
$W^{\mbox{\scriptsize out}}$ and $H$ were 0.0018 and 0.0042,
respectively.

The  $M \times 4$ prototype matrix $P$ was  computed as
follows. First, a preliminary version $P^\ast$ was constructed whose
$j$-the column vector was the mean of the element-wise squared column
vectors in $Z^j$. The four column vectors of $P^\ast$ were then
normalized such that the norm of each of them was the mean of the
norms of columns in    $P^\ast$. This gave $P$. This normalization is
important for the performance of the architecture, because without it
the optimization criterion (\ref{eqGammaLoss}) would systematically
lead to smaller values for those $\gamma^j$ that are associated with
smaller-norm columns in $P$. 

\emph{Baseline linear filter.} The transversal filter that served as a
baseline was a row vector $w$ of size 2600. It was computed to
minimize the loss function
$$\mathcal{L}(w) = \frac{1}{4}\,\sum_{j=1}^4  E[p^j(n+1) -
(p^j(n-2600+1),\ldots, p^j(n))^2] + a^2 \|w\|^2,$$
where $p^j(n)$ were clean versions of the four patterns. 400 
timesteps per pattern were used for training, and $a$ was set to
1.0. The setting of $a$ was very robust. Changing $a$ in either
direction by factors of 100 changed the resulting test NRMSEs at
levels below the plotting accuracy in Figure \ref{figcArch}. 

\emph{Parameter settings in testing.}  The adaptation rate
$\lambda_\gamma$ was set to 0.002 for the classification simulation
and to 0.004 for the morph-tracking case study. The other global
control parameters were identical in both simulations: trust smoothing
rate $\sigma = 0.99$, decisiveness $d_{[12]} = d_{[23]} = 8$, drift $d
= 0.01$, $c^{\mbox{\scriptsize aut}}_{[l]}$ adaptation rate $\lambda =
0.5$ (compare Equation (\ref{eqcAdapt}); I used the same adaptation
rate $\lambda_i \equiv \lambda$ for all of the 500 feature units).

\emph{Computing and plotting running NRMSE estimates.}  For the NRMSE
plots in the fifth rows in Figures \ref{figcArch} and
\ref{figcArchMorph}, a running estimate of the NRMSE between the
module outputs $y_{[l]}$ and the clean input patterns $p$ (unknown to
the system) was computed as follows. A running estimate
$\overline{\mbox{var}}\,p(n)$ of the variance of the clean pattern was
maintained like it was done for $\overline{\mbox{var}}\,y_{[l]}(n)$ in
(\ref{eqUpdatediscrepancy1}) and (\ref{eqUpdatediscrepancy2}), using
an exponential 
smoothing rate of $\sigma =  0.95$. Then the running NRMSE was
computed by another exponential smoothing per
$$\overline{\mbox{nrmse}}\,y_{[l]}(n+1) =
\sigma\,\overline{\mbox{nrmse}}\,y_{[l]}(n) + (1-\sigma)\,
\left(\frac{(p(n+1) -
    y_{[l]}(n+1))^{2}}{\overline{\mbox{var}}\,p(n+1)}\right)^{1/2}.$$ 
The running NRMSE for the baseline transversal filter were obtained in
a similar fashion.

\newpage

\section{Proofs and Algorithms}\label{secProofs}

\subsection{Proof of Proposition \ref{propCompConceptor} (Section \ref{sec:RetrieveGeneric})}\label{secProofPropcompconceptor}

\emph{Claim 1}. We first re-write the minimization quantity, using $R
= E[xx']$:
\begin{eqnarray*}
\lefteqn{E[\| x - Cx \|^2] +
\alpha^{-2}\,\|C\|^2_{\mbox{\scriptsize {fro}}} = }\\
& = & E[\mbox{tr}\, (x'(I - C')(I - C) x)] +
\alpha^{-2}\,\|C\|^2_{\mbox{\scriptsize {fro}}}\\
& = & \mbox{tr}\, ((I - C')(I - C)R +
\alpha^{-2}\,C'C)\\ 
& = & \mbox{tr}\, ( R - C'R - CR +
C'C(R + \alpha^{-2}\,I)) \\
& = & \sum_{i=1,\ldots,N} e'_i ( R - C'R - CR + C'C(R + \alpha^{-2}\,I)) e_i.
\end{eqnarray*}
This quantity is quadratic in the parameters of $C$ and non-negative.
Because $\mbox{tr}\,C'C(R + \alpha^{-2}\,I) = \mbox{tr}\,C(R +
\alpha^{-2}\,I) C'$ and $R + \alpha^{-2}\,I$ is positive definite,
$\mbox{tr}\,C(R + \alpha^{-2}\,I) C'$ is positive definite in the
$N^2$-dimensional space of $C$ elements. Therefore $E[\| x - Cx \|^2]
+ \alpha^{-2}\,\|C\|^2_{\mbox{\scriptsize {fro}}}$ has a unique minimum
in $C$ space. To locate it we compute the derivative of the $i$-th
component of this sum with respect to the entry $C(k,l) = C_{kl}$:
\begin{eqnarray*}
\lefteqn{\frac{\partial} {\partial C_{kl}} \; e'_i ( R - C'R - CR +
C'C(R + \alpha^{-2}\,I)) e_i =}  \\
& = & - 2\,R_{kl} + \partial/\partial C_{kl} \; \sum_{j,a=1,\ldots,N}
C_{ij}\, C_{ja}\, A_{ai}\\
& = & - 2\,R_{kl} +  \sum_{a=1,\ldots,N} \partial/\partial C_{kl} \;
C_{kj}\, C_{ka}\, A_{ai}\\
& = & - 2\,R_{kl} +  \sum_{a=1,\ldots,N} (C_{ka}\, A_{al} +
C_{ki}\,A_{li})\\
& = &  - 2\,R_{kl} + (CA)_{kl} + N \, C_{ki}\,A_{il}.
\end{eqnarray*} 
where we used the  abbreviation $A = R +
\alpha^{-2}\,I$, observing in the last line that $A = A'$. Summing
over $i$ yields
\begin{displaymath}
\frac{\partial} {\partial C_{kl}} \; E[\| x - Cx \|^2] +
\alpha^{-2}\,\|C\|^2_{\mbox{\scriptsize \emph{fro}}} = -2\,N\,R_{kl} +
2N\,(CA)_{kl},  
\end{displaymath}
which in matrix form is
 \begin{displaymath}
\frac{\partial} {\partial C} \; E[\| x - Cx \|^2] +
\alpha^{-2}\,\|C\|^2_{\mbox{\scriptsize \emph{fro}}} = -2\,N\, R +
2N\,C( R + \alpha^{-2}\,I),
\end{displaymath}
where we re-inserted the expression for $A$. Setting this to zero
 yields the  claim \emph{1.} stated in the proposition. 

The
subclaim that $R(R + \alpha^{-2}I)^{-1} = (R + \alpha^{-2}I)^{-1}\,R$
can be easily seen when $R$ is written by its SVD $R = U\Sigma U'$:
$U\Sigma U'(U\Sigma U' + \alpha^{-2}I)^{-1} = U\Sigma U'(U(\Sigma+\alpha^{-2}I)U')^{-1} =
U\Sigma U'U(\Sigma+\alpha^{-2}I)^{-1}U' = U\Sigma(\Sigma+\alpha^{-2}I)^{-1}U' =
U(\Sigma+\alpha^{-2}I)^{-1}\Sigma U' = ... =  (R + \alpha^{-2}I)^{-1}\,R$. This
also proves claim \emph{2.}

\emph{Claims 3.--5.} can be derived from the first claim by elementary arguments.

\subsection{Proof of Proposition \ref{propANDDef} (Section
  \ref{secFormalBooleanDef})} \label{secProofAndDef}

 $C_\delta^{-1}$ can be written as $U S_\delta^{-1}
U'$, where the diagonal of $S_\delta^{-1}$ is $(s_1^{-1},\ldots,
s_l^{-1},\delta^{-1},\ldots,\delta^{-1})'$. Putting $e
= \delta^{-1}$ and $S_e := (s_1^{-1},\ldots,
s_l^{-1},e,\ldots,e)'$, and similarly
$T_e = (t_1^{-1},\ldots,
t_m^{-1},$ $e,\ldots,e)'$, we can express the
limit $ \lim_{\delta \to 0}(C_\delta^{-1} + B_\delta^{-1} - I)^{-1}$
equivalently as $\lim_{e \to \infty} (US_e U' + V
T_e V' - I)^{-1}$. Note that $US_e U' + V
T_e V' - I$ is invertible for sufficiently large
$e$. Let $U_{>l}$ be the $N \times (N - l)$ submatrix of $U$
made from the last $N-l$ columns of $U$ (spanning the
null space of $C$), and let $V_{>m}$ be $N \times (N - m)$ submatrix of $V$
made from the last $N-m$ columns of $V$. The $N \times N$ matrix
$U_{>l} (U_{>l})' + V_{>m} (V_{>m})'$ is positive semidefinite. Let
$W\Sigma W'$ be its SVD, with singular values in $\Sigma$ in
descending order. Noting that $C^\dagger = U \mbox{diag}(s_1^{-1},\ldots,
s_l^{-1},0,\ldots,0)' U'$, and $B^\dagger = V \mbox{diag}(t_1^{-1},\ldots,
t_m^{-1},0,\ldots,0)' V'$,  we can rewrite

\begin{eqnarray}
\lim_{\delta \to 0} C_\delta \wedge B_\delta & = & \lim_{e
  \to \infty} (US_e U' + V T_e V' - I)^{-1} \nonumber\\
 & = & \lim_{e \to \infty} (C^\dagger + B^\dagger +
  e \, W \Sigma W' - I)^{-1}\nonumber\\
& = & W \,\left(\lim_{e \to \infty} (W'C^\dagger W + W'
  B^\dagger W + e \Sigma - I)^{-1}\right) \,W'. \label{eqtrans}
\end{eqnarray}  

If $\Sigma$ is invertible, clearly (\ref{eqtrans}) evaluates to the
zero matrix. We proceed to consider the case of   non-invertible $\Sigma =
\mbox{diag}(\sigma_1,\ldots, \sigma_k,0,\ldots, 0)$, where $0 \leq k <
N$. 

We derive two  auxiliary claims. Let $W_{>k}$ be the $N \times
(N-k)$ submatrix of $W$ made from the last $N-k$ columns.
\emph{Claim 1:} the $(N-k) \times (N-k)$ matrix $A = (W_{>k})' \,
(C^\dagger + B^\dagger - I)\,W_{>k}$ is invertible.  We rewrite

\begin{equation}\label{eqproof1-1}
A =  (W_{>k})' C^\dagger W_{>k} + (W_{>k})' B^\dagger W_{>k} -
I_{(N-k) \times (N-k)},  
\end{equation}

\noindent and analyse $(W_{>k})' C^\dagger W_{>k}$. 
It holds that $(W_{>k})' U_{>l} = 0_{(N-k)\times (N-l)}$, because

\renewcommand{\arraystretch}{0.5}
\begin{eqnarray}
&
W \,\left(\begin{array}{c@{\hspace{1mm}}c@{\hspace{1mm}}c@{\hspace{1mm}}c@{\hspace{1mm}}c@{\hspace{1mm}}c}\sigma_1
    &&&&&\\& \ddots &&&&\\&& \sigma_k &&&\\&&& 0 &&\\ &&&& \ddots  &\\
    &&&&& 0 \end{array} \right)\,W' & =  \quad U_{>l} (U_{>l})' + V_{>m}
(V_{>m})' \nonumber\\
\Longrightarrow
 & \left( \begin{array}{c@{\hspace{1mm}}c@{\hspace{1mm}}c@{\hspace{1mm}}c@{\hspace{1mm}}c@{\hspace{1mm}}c}\sigma_1
     &&&&&\\& \ddots &&&&\\&&     \sigma_k &&&\\&&& 0 &&\\ &&&& \ddots
     &\\ &&&&& 0 \end{array}  \right) & =  \quad W' U_{>l} (U_{>l})' W + W'
 V_{>m} (V_{>m})' W \nonumber\\
 \Longrightarrow  &  0_{(N-k)\times (N-k)} & = \quad (W_{>k})' U_{>l}
 (U_{>l})' W_{>k} +    (W_{>k})' V_{>m} (V_{>m})' W_{>k}\nonumber\\ 
 \Longrightarrow  &  (
W_{>k})' U_{>l} & = \quad 0_{(N-k) \times (N-l)}.\label{eqproof1-2}
\end{eqnarray}

Let $U_{\leq l}$ be the $N\times l$ submatrix of $U$ made of the first $l$
columns. Because of (\ref{eqproof1-2}) it follows that

\begin{eqnarray}\label{eqproof1-3}
(W_{>k})' C^\dagger W_{>k} & = & (W_{>k})'  U \,\left(\begin{array}{c@{\hspace{1mm}}c@{\hspace{1mm}}c@{\hspace{1mm}}c@{\hspace{1mm}}c@{\hspace{1mm}}c}s_1^{-1}
    &&&&&\\& \ddots &&&&\\&& s_l^{-1} &&&\\&&& 0 &&\\ &&&& \ddots  &\\
    &&&&& 0 \end{array} \right)\, U' W_{>k} \nonumber\\
& = & (W_{>k})' U_{\leq l} \left( \begin{array}{c@{\hspace{1mm}}c@{\hspace{1mm}}c}s_1^{-1}
     &&\\& \ddots &\\&&     s_l^{-1} \end{array}  \right) (U_{\leq
    l})' W_{>k} 
\end{eqnarray}

The rows of the $(N-k) \times l$ sized matrix $(W_{>k})' U_{\leq l}$
are orthonormal, which also follows from (\ref{eqproof1-2}). The
Cauchy interlacing theorem (stated in Section \ref{subsecCAD} in
another context) then implies that all singular values of $(W_{>k})'
C^\dagger W_{>k}$ are greater or equal to $\min\{s_1^{-1},\ldots,
s_l^{-1}\}$. Since all $s_i$ are smaller or equal to $1$, all singular
values of $(W_{>k})'
C^\dagger W_{>k}$ are greater or equal to 1. The singular values of
the matrix  $(W_{>k})'
C^\dagger W_{>k} - 1/2 \, I_{(N-k)\times (N-k)}$ are therefore greater
or equal to $1/2$. Specifically, $(W_{>k})'
C^\dagger W_{>k} - 1/2 \, I_{(N-k)\times (N-k)}$ is positive
definite. By a similar argument, $(W_{>k})'
B^\dagger W_{>k} - 1/2 \, I_{(N-k)\times (N-k)}$ is positive
definite. The matrix $A$ from  Claim 1 is therefore revealed as the
sum of two positive definite matrices, and hence is invertible. 

\emph{Claim 2:} If $M$ is a symmetric $N \times
N$ matrix, and the right lower principal submatrix $M_{>k>k} =
M(k+1:N,k+1:N)$ is invertible, and $\Sigma =
\mbox{diag}(\sigma_1,\ldots, \sigma_k,0,\ldots, 0)$ with all $\sigma_i
> 0$, then 

\renewcommand{\arraystretch}{0.8}
\begin{equation}\label{eqproof1-4}
\lim_{e \to \infty} (M + e \Sigma)^{-1} = \left(\begin{array}{cc}0 &
    0\\ 0 & M_{>k>k}^{-1} \end{array}  \right). 
\end{equation}

We exploit the following elementary block representation of the
inverse of a symmetric matrix (e.g.\ \cite{Bernstein09}, fact 2.17.3):

\begin{displaymath}
\left(\begin{array}{c@{\hspace{1mm}}c}X & Y\\ Y' & Z
    \end{array} \right)^{-1} =
  \left(\begin{array}{cc} V^{-1} & - V^{-1}YZ^{-1}\\ -Z^{-1}Y'V^{-1} & 
Z^{-1}Y'V^{-1}Y Z^{-1}+Z^{-1}  \end{array} \right),
\end{displaymath}

\noindent where $Z$ is assumed to be invertible and $V = (X - YZ^{-1}Y')$ is
assumed to be invertible. Block-structuring $M + e\Sigma$ analogously to this
representation, where $M_{>k>k}$ is identified with $Z$, then easily leads
to the claim (\ref{eqproof1-4}).

Applying Claims 1 and 2 to the limit expression $\lim_{e \to \infty}
(W'C^\dagger W + W' B^\dagger W + e \Sigma - I)^{-1}$ in
(\ref{eqtrans}), where the matrix $M$ in Claim 2 is identified with
$W'C^\dagger W + W' B^\dagger W - I$ and the matrix $Z$ from Claim 2
is identified with the matrix $A = (W_{>k})' \, (C^\dagger + B^\dagger
- I)\,W_{>k}$ from Claim 1, yields
\begin{displaymath}
\lim_{e \to \infty}
(W'C^\dagger W + W' B^\dagger W + e \Sigma - I)^{-1} 
=  \left(\begin{array}{cc}0 & 0\\ 0 & \left((W_{>k})' \, (C^\dagger + B^\dagger
- I)\,W_{>k}\right)^{-1}  \end{array} \right) 
\end{displaymath}
\noindent which, combined with (\ref{eqtrans}), leads to 
\begin{equation}\label{eqproof1-5a}
\lim_{\delta \to 0} C_\delta \wedge B_\delta  =  
W_{>k} \, \left((W_{>k})' \, (C^\dagger + B^\dagger
- I)\,W_{>k}\right)^{-1} (W_{>k})'.
\end{equation}
$W_{>k}$ is an $N \times (N-k)$ size matrix whose columns are
orthonormal. Its range is
 \begin{eqnarray}
\mathcal{R}(W_{>k}) & = & \mathcal{N}(W\Sigma W') \quad = \quad
\mathcal{N}(U_{>l} (U_{>l})' + V_{>m} (V_{>m})') \nonumber\\
& = & \mathcal{N}((U_{>l} (U_{>l})') \cap \mathcal{N}(V_{>m}
(V_{>m})') \quad = \quad \mathcal{N}( (U_{>l})') \cap
\mathcal{N}((V_{>m})')\nonumber\\ 
& = & \mathcal{R}(U_{>l})^\perp \cap \mathcal{R}(V_{>m})^\perp \quad =
\quad \mathcal{N}(C)^\perp \cap \mathcal{N}(B)^\perp\nonumber\\
& = & \mathcal{R}(C') \cap \mathcal{R}(B') \quad = \quad
\mathcal{R}(C) \cap \mathcal{R}(B), \label{eqproof1-9}
\end{eqnarray}
that is, $W_{>k}$ is a matrix whose columns form an orthonormal
basis of $\mathcal{R}(C) \cap  \mathcal{R}(B)$. 
  
Let $\mathbf{B}_{\mathcal{R}(C) \cap \mathcal{R}(B)}$ be any matrix
whose columns form an orthonormal basis of $\mathcal{R}(C) \cap
\mathcal{R}(B)$. Then there exists a unique orthonormal matrix $T$ of size
$(N-k)\times (N-k)$ such that $\mathbf{B}_{\mathcal{R}(C) \cap
  \mathcal{R}(B)} = W_{>k} \, T$. It holds that 
\begin{eqnarray}
\lefteqn{W_{>k} \, \left((W_{>k})' \, (C^\dagger + B^\dagger
- I)\,W_{>k}\right)^{-1} (W_{>k})' = }\nonumber\\
& = & W_{>k}T \, \left((W_{>k}T)' \, (C^\dagger + B^\dagger
- I)\,W_{>k}T\right)^{-1} (W_{>k}T)' \nonumber\\
& = & \mathbf{B}_{\mathcal{R}(C) \cap \mathcal{R}(B)} \, \left(\mathbf{B}'_{\mathcal{R}(C) \cap \mathcal{R}(B)} \, (C^\dagger + B^\dagger
- I)\,\mathbf{B}_{\mathcal{R}(C) \cap \mathcal{R}(B)}\right)^{-1}\,\mathbf{B}_{\mathcal{R}(C) \cap \mathcal{R}(B)}', \label{eqCompANDinproof}
\end{eqnarray}
 which gives the final form of the claim in the
  proposition.

For showing equivalence of (\ref{eqCompANDgenComp_1}) with
(\ref{eqCompANDinproof}), I exploit a fact known in
matrix theory (\cite{Bernstein09},  Fact 6.4.16): for two
real matrices $X, Y$ of sizes $n \times m, m \times l$ the following
two condition are equivalent: (i) $(X\,Y)^\dagger = Y^\dagger \,
X^\dagger$, and (ii) $\mathcal{R}(X'\,X\,Y) \subseteq \mathcal{R}(Y)$
and  $\mathcal{R}(Y\,Y'\,X') \subseteq \mathcal{R}(X')$. Observing
that $\mathbf{P}_{\mathcal{R}(C) \cap \mathcal{R}(B)} =
\mathbf{B}_{\mathcal{R}(C) \cap
  \mathcal{R}(B)}\mathbf{B}_{\mathcal{R}(C) \cap \mathcal{R}(B)}'$ and
that $\mathbf{B}_{\mathcal{R}(C) \cap \mathcal{R}(B)}^\dagger =
\mathbf{B}_{\mathcal{R}(C) \cap \mathcal{R}(B)}'$, setting $X =
\mathbf{B}_{\mathcal{R}(C) \cap \mathcal{R}(B)}$ and $Y =
\mathbf{B}_{\mathcal{R}(C) \cap \mathcal{R}(B)}' \, (C^\dagger + B^\dagger
- I) \mathbf{P}_{\mathcal{R}(C) \cap \mathcal{R}(B)}$ in
(\ref{eqCompANDgenComp_1}) yields
\begin{eqnarray*}
\lefteqn{\left(\mathbf{P}_{\mathcal{R}(C) \cap \mathcal{R}(B)} \, (C^\dagger + B^\dagger
- I)\,\mathbf{P}_{\mathcal{R}(C) \cap \mathcal{R}(B)}\right)^{-1} =}
\\
& & \left(\mathbf{B}_{\mathcal{R}(C) \cap \mathcal{R}(B)}' \, (C^\dagger + B^\dagger
- I)\,\mathbf{P}_{\mathcal{R}(C) \cap
  \mathcal{R}(B)}\right)^{\dagger} \, \mathbf{B}_{\mathcal{R}(C) \cap
\mathcal{R}(B)}',
\end{eqnarray*}
where condition (ii) from the abovementioned fact is
easily verified. In a second, entirely analog step one can pull apart
$\left(\mathbf{B}_{\mathcal{R}(C) \cap \mathcal{R}(B)}' \, (C^\dagger + B^\dagger
- I)\,\mathbf{P}_{\mathcal{R}(C) \cap
  \mathcal{R}(B)}\right)^{\dagger}$ into  $$\mathbf{B}_{\mathcal{R}(C) \cap
  \mathcal{R}(B)} \, \left(\mathbf{B}_{\mathcal{R}(C) \cap \mathcal{R}(B)}' \, (C^\dagger + B^\dagger
- I)\,\mathbf{B}_{\mathcal{R}(C) \cap
  \mathcal{R}(B)}\right)^{\dagger}.$$

{\bf Algorithm for Computing }$\mathbf{B}_{\mathcal{R}(C)
    \cap \mathcal{R}(B)}.$ Re-using ideas from this proof, a  basis matrix
$\mathbf{B}_{\mathcal{R}(C) \cap \mathcal{R}(B)}$ can be computed as
follows:

\begin{enumerate}
\item Compute the SVDs $C = U \mbox{diag}(s_1,\ldots, s_l,0,\ldots,0)
U'$ and $B = V  \mbox{diag}(t_1,\ldots, t_m,$ $0,\ldots,0) V'$. 
\item Let $U_{>l}$ be the submatrix of $U$ made from the last $N-l$
columns in $U$, and similarly let $V_{>m}$ consist of the  last $N-m$
columns in $V$. 
\item Compute the SVD $U_{>l} (U_{>l})' + V_{>m} (V_{>m})' = W\Sigma
W'$, where \\$\Sigma = \mbox{diag}(\sigma_1,\ldots, \sigma_k, 0, \ldots,
0)$. 
\item Let $W_{>k}$ be the submatrix of $W$ consisting of the last
$N-k$ columns of $W$. Then $\mathbf{B}_{\mathcal{R}(C) \cap \mathcal{R}(B)} =
W_{>k}$.
\end{enumerate}

\subsection{Proof of Proposition \ref{propLimitOr} (Section
  \ref{secFormalBooleanDef})}. \label{secProofpropLimitOr}

By Definition \ref{def:finalBoolean}, Proposition \ref{propANDDef},
and Equation (\ref{eq:CompConceptor}),
\begin{eqnarray*}
C \vee B & = & \neg\,(\neg \, C \,\wedge \, \neg\, B) = I - \lim_{\delta
  \downarrow 0}\left((\neg C)_\delta^{-1} + (\neg B)_\delta^{-1} -
  I\right)^{-1}\\
& = & I - \lim_{\delta
  \downarrow 0}\left((I - C^{(\delta)})^{-1} + (I - B^{(\delta)})^{-1} -
  I\right)^{-1}\\
& = & I - \lim_{\delta
  \downarrow 0}\left((I - R_C^{(\delta)}(R_C^{(\delta)}+I)^{-1})^{-1} + (I - R_B^{(\delta)}(R_B^{(\delta)}+I)^{-1} -
  I\right)^{-1}.
\end{eqnarray*}
It is easy to check that $(I - A(A+I)^{-1})^{-1} = I + A$ holds for any
positive semidefinite matrix $A$. Therefore, 
\begin{displaymath}
C \vee B = I - \lim_{\delta
  \downarrow 0}\left(R_C^{(\delta)} + R_B^{(\delta)} + I  \right)^{-1}.
\end{displaymath}
Furthermore, for positive semidefinite $A, B$ it generally holds that
$I - (A + B + I)^{-1} = (A+B)(A+B+I)^{-1}$, and hence
 \begin{displaymath}
C \vee B = \lim_{\delta \downarrow 0}(R_C^{(\delta)} +
R_B^{(\delta)})\,(R_C^{(\delta)} + R_B^{(\delta)} + I)^{-1}.
\end{displaymath}
 
\subsection{Proof of Proposition \ref{propdeMorganAND} (Section
  \ref{secFormalBooleanDef})} \label{secProofpropdeMorganAND}
 
Using (\ref{eqCompANDgenComp}), Proposition \ref{propLimitOr} and that
fact $A^{(\delta)} = R_A^{(\delta)}\,(R_A^{(\delta)} + I)^{-1}$ holds
for any conceptor $A$ (which entails $I - A^{(\delta)} =
(R_A^{(\delta)} + I)^{-1}$), we
derive the claim as follows:
\begin{eqnarray*}
C \wedge B & = & \lim_{\delta \downarrow 0} \left(C_\delta^{-1} +
B_\delta^{-1} - I\right)^{-1}\\
& = & \lim_{\delta \downarrow 0} \left((\neg \neg C)_\delta^{-1} +
(\neg \neg B)_\delta^{-1} - I \right)^{-1}\\
& = & \lim_{\delta \downarrow 0} \left((\neg (\neg C)^{(\delta)})^{-1} +
(\neg (\neg B)^{(\delta)})^{-1} - I \right)^{-1}\\ 
 & = & \lim_{\delta \downarrow 0} \left((I -(\neg C)^{(\delta)})^{-1} +
 (I - (\neg B)^{(\delta)})^{-1} - I \right)^{-1}\\
 & = & \lim_{\delta \downarrow 0} \left((R_{\neg C}^{(\delta)} + I) +
 (R_{\neg B}^{(\delta)}) + I) - I \right)^{-1}\\
 & = & \lim_{\delta \downarrow 0} (R_{\neg C}^{(\delta)} + R_{\neg
   B}^{(\delta)} + I)^{-1}\\
 & = & I - \lim_{\delta \downarrow 0}(I - (R_{\neg C}^{(\delta)} + R_{\neg
   B}^{(\delta)} + I)^{-1})\\
 & = & \neg (\neg C \vee \neg B).
\end{eqnarray*}

\subsection{Proof of Proposition \ref{propSpaces} (Section
  \ref{secFactsSubspaces})} \label{secProofpropSpaces} 

\emph{Claims 1. -- 3.} are elementary. 

\emph{Claim 4a:} $\mathcal{R}(C \wedge B) = \mathcal{R}(C) \cap
\mathcal{R}(B)$. By definition we have 
\begin{displaymath}C \wedge B = \mathbf{B}_{\mathcal{R}(C) \cap \mathcal{R}(B)} \, \left(
  \mathbf{B}'_{\mathcal{R}(C) \cap \mathcal{R}(B)} (C^\dagger +
  B^\dagger - I) \mathbf{B}_{\mathcal{R}(C) \cap \mathcal{R}(B)}
\right)^{-1} \, \mathbf{B}'_{\mathcal{R}(C) \cap
  \mathcal{R}(B)}.
\end{displaymath}
 Since $\mathbf{B}'_{\mathcal{R}(C) \cap
  \mathcal{R}(B)} (C^\dagger +   B^\dagger - I)
\mathbf{B}_{\mathcal{R}(C) \cap \mathcal{R}(B)}$ is  invertible and $\mathcal{R}(\mathbf{B}'_{\mathcal{R}(C) \cap
  \mathcal{R}(B)}) = \mathbb{R}^{\mbox{\scriptsize dim}(\mathcal{R}(C)\cap
  \mathcal{R}(B))}$, we have $\mathcal{R}\left((\mathbf{B}'_{\mathcal{R}(C) \cap
  \mathcal{R}(B)} (C^\dagger +   B^\dagger - I)
\mathbf{B}_{\mathcal{R}(C) \cap \mathcal{R}(B)})^{-1} \,\mathbf{B}'_{\mathcal{R}(C) \cap
  \mathcal{R}(B)}  \right) = \mathbb{R}^{\mbox{\scriptsize dim}(\mathcal{R}(C)\cap
  \mathcal{R}(B))}$. From this it follows that $\mathcal{R}(C \wedge
B) = \mathcal{R}(\mathbf{B}_{\mathcal{R}(C) \cap
  \mathcal{R}(B)}) = \mathcal{R}(C) \cap \mathcal{R}(B)$.

\emph{Claim 5a:} $\mathcal{I}(C \wedge B) = \mathcal{I}(C) \cap
\mathcal{I}(B)$. We have, by definition, 
\begin{displaymath}C \wedge B = \mathbf{B}_{\mathcal{R}(C) \cap \mathcal{R}(B)} \, \left(
  \mathbf{B}'_{\mathcal{R}(C) \cap \mathcal{R}(B)} (C^\dagger +
  B^\dagger - I) \mathbf{B}_{\mathcal{R}(C) \cap \mathcal{R}(B)}
\right)^{-1} \, \mathbf{B}'_{\mathcal{R}(C) \cap
  \mathcal{R}(B)}.
\end{displaymath}
For shorter notation put $\mathbf{B} = \mathbf{B}_{\mathcal{R}(C) \cap
  \mathcal{R} (B)}$ and $X = \mathbf{B}'_{\mathcal{R}(C) \cap
  \mathcal{R}(B)} (C^\dagger + B^\dagger - I)
\mathbf{B}_{\mathcal{R}(C) \cap \mathcal{R}(B)}$. Note (from proof of
  Proposition \ref{propANDDef}) that $X$ is invertible. We characterize the
  unit eigenvectors of $C \wedge B$. It holds that $\mathbf{B} X^{-1}
  \mathbf{B}'\, x = x$ if and only if $\mathbf{B} X
  \mathbf{B}'\, x = x$. We need to show that the conjunction $Cx = x$ and $Bx =
  x$ is equivalent to $(C \wedge B)\,x = \mathbf{B} X^{-1}
  \mathbf{B}\, x = x$. 
  
  First assume that $Cx = x$ and $Bx = x$. This implies $x \in
  \mathcal{R}(C) \cap \mathcal{R}(B)$ and $C^\dagger \,x = x$ and
  $B^\dagger \,x = x$, and hence $\mathbf{P}_{\mathcal{R}(C) \cap
    \mathcal{R}(B)}(C^\dagger + B^\dagger - I)
  \mathbf{P}_{\mathcal{R}(C) \cap \mathcal{R}(B)}\, x = x$. But
  $\mathbf{P}_{\mathcal{R}(C) \cap \mathcal{R}(B)}(C^\dagger +
  B^\dagger - I) \mathbf{P}_{\mathcal{R}(C) \cap \mathcal{R}(B)} =
  \mathbf{B} X \mathbf{B}'$, thus $(C \wedge B)\,x = x$. 
  
  Now assume conversely that not $Cx = x$ or not $Bx = x$. 

Case 1: $x
  \notin \mathcal{R}(C) \cap \mathcal{R}(B)$. Then
  $\mathbf{P}_{\mathcal{R}(C) \cap \mathcal{R}(B)}(C^\dagger +
  B^\dagger - I) \mathbf{P}_{\mathcal{R}(C) \cap \mathcal{R}(B)}\, x
  \neq x$ and hence $\mathbf{B} X \mathbf{B}'\, x \neq x$, which
  implies $(C \wedge B)\,x \neq x$. 

Case 2: $x \in \mathcal{R}(C) \cap
  \mathcal{R}(B)$. We first show an auxiliary claim: $\| (C^\dagger +
  B^\dagger - I)\,x \| > \| x\|$. Let $C_0 = C^\dagger - CC^\dagger, B_0 = B^\dagger -
  BB^\dagger$. $C_0$ and $B_0$ are positive semidefinite because the
  nonzero singular values of $C^\dagger, B^\dagger$ are greater or
  equal to 1. Furthermore, 
  $CC^\dagger \,x = BB^\dagger \, x = x$. Thus, $(C^\dagger +
  B^\dagger)\,x = 2 I x + C_0 x + B_0 x$, i.e.\ $(C^\dagger +
  B^\dagger - I)\,x = I x + C_0 x + B_0 x$. From not $Cx = x$ or not $Bx
  = x$ it follows that $C_0 x \neq 0$ or $B_0 x \neq 0$. We infer

\begin{eqnarray*}
C_0 x \neq 0 \mbox{ or } B_0 x \neq 0 & \Longrightarrow & x'C_0x > 0
\mbox{ or } x' B_0 x > 0\\
& \Longrightarrow & x'(C_0 + B_0)x > 0 \quad \Longrightarrow \quad x'(C_0 + B_0)^2x > 0.
\end{eqnarray*}

This implies $\|I x + C_0 x + B_0 x  \|^2 = \|x\|^2 + 2x'(C_0 + B_0)x
+ x'(C_0 + B_0)^2x > \|x\|^2$, or equivalently,  $\| (C^\dagger +
  B^\dagger - I)\,x \| > \| x\|$, the auxiliary claim.

Since $\mathbf{P}_{\mathcal{R}(C)
    \cap \mathcal{R}(B)}$ preserves vector norm on $\mathcal{R}(C)
  \cap \mathcal{R}(B)$ and $\mathcal{R}(C^\dagger + B^\dagger - I)
  \mathbf{P}_{\mathcal{R}(C) \cap \mathcal{R}(B)} \subseteq
  \mathcal{R}(C) \cap \mathcal{R}(B)$, it follows that $\|
  \mathbf{P}_{\mathcal{R}(C) \cap \mathcal{R}(B)} (C^\dagger +
  B^\dagger - I) \mathbf{P}_{\mathcal{R}(C) \cap \mathcal{R}(B)} \, x
  \| > \| x\|$, hence $(C \wedge B)\,x \neq x$. 

Altogether we have that the conjunction $Cx = x$ and $Bx = x$ is
equivalent to $(C \wedge B)\,x = x$, which is equivalent to the claim. 

\emph{Claim 6a:} $\mathcal{N}(C \wedge B) = \mathcal{N}(C) +
\mathcal{N}(B)$. This follows from \emph{4a} by $\mathcal{N}(C \wedge
B) = (\mathcal{R}(C \wedge B))^\perp = (\mathcal{R}(C) \cap
\mathcal{R}(B))^\perp = (\mathcal{N}(C)^\perp \cap
\mathcal{N}(B)^\perp)^\perp =  \mathcal{N}(C) +
\mathcal{N}(B)$. 

\emph{Claims 4b, 5b, 6b:} The second statements in \emph{4., 5., 6.}
follow from the first statements and \emph{3.}, exploiting de Morgan's
rule. 

\emph{Claim 7} follows from Equation (\ref{eqpropadapt}).

\emph{Claim 8:} Let $A = A \wedge C$. Then claim \emph{4.}  implies
$\mathcal{R}(A) \cap \mathcal{R}(C) = \mathcal{R}(A)$. By Proposition
\ref{propANDDef} we can write
\begin{displaymath}
 A \wedge C = \mathbf{B}_{\mathcal{R}(A)} \, \left(
  \mathbf{B}'_{\mathcal{R}(A)} \,(C^\dagger +
  B^\dagger - I) \,\mathbf{B}_{\mathcal{R}(A)}
\right)^{-1} \, \mathbf{B}'_{\mathcal{R}(A) }.
\end{displaymath}

Let $A = U S U'$ be the SVD of $A$, and assume $A$ has rank $k \leq
N$, that is, exactly the first $k$ singular values in $S$ are
nonzero. Let $U_k$ be the $N \times k$ matrix consisting of the first $k$
columns of $U$. It holds that $U_k = \mathbf{B}_{\mathcal{R}(A)}$. We obtain

\begin{eqnarray}
\lefteqn{S = U' A U = } \nonumber\\
& = & U' \, U_k \, \left(U'_k (A^\dagger + C^\dagger - I)\, U_k
\right)^{-1} U'_k \, U \nonumber \\
& = & I_k \, \left(U'_k (A^\dagger + C^\dagger - I)\, U_k
\right)^{-1} I'_k, \label{eqProof7dot8_1} 
\end{eqnarray}
where $I_k$ is the $N \times k$ matrix consisting of the first $k$
columns of $I$. Let $S_k$ be the $k \times k$ upper left submatrix of
$S$. Then $S_k = \left(U'_k (A^\dagger + C^\dagger - I)\, U_k
\right)^{-1}$ and 
\begin{eqnarray*}
S_k^{-1} & = & U'_k (A^\dagger + C^\dagger - I)\, U_k\\
& = & U'_k \, A^\dagger \, U_k + U'_k \, ( C^\dagger - I) \, U_k\\
& = & S_k^{-1} +   U'_k \, ( C^\dagger - I) \, U_k, 
\end{eqnarray*}
hence $U'_k \, ( C^\dagger - I) \, U_k = 0_{k \times k}$ or
equivalently, $U'_k \,  C^\dagger  \, U_k = I_{k \times k}$. This
implies $\mathcal{R}(A) \subseteq \mathcal{I}(C)$. 

Conversely, assume $\mathcal{R}(A) \subseteq \mathcal{I}(C)$.
Going
through the above line of arguments in reverse order establishes 
again $S = I_k \, \left(U'_k (A^\dagger + C^\dagger - I)\, U_k
\right)^{-1} I'_k$ which implies
\begin{displaymath}
A = \mathbf{B}_{\mathcal{R}(A)} \, \left(
  \mathbf{B}'_{\mathcal{R}(A)} \,(C^\dagger +
  B^\dagger - I) \,\mathbf{B}_{\mathcal{R}(A)}
\right)^{-1} \, \mathbf{B}'_{\mathcal{R}(A) }. 
\end{displaymath}
$\mathcal{R}(A) \subseteq \mathcal{I}(C)$ implies
$\mathcal{R}(A) \subseteq \mathcal{R}(C)$, which leads to 
\begin{eqnarray*}
A & = & \mathbf{B}_{\mathcal{R}(A) \cap \mathcal{R}(C)} \, \left(
  \mathbf{B}'_{\mathcal{R}(A) \cap \mathcal{R}(C)} \,(C^\dagger +
  B^\dagger - I) \,\mathbf{B}_{\mathcal{R}(A) \cap \mathcal{R}(C)}
\right)^{-1} \, \mathbf{B}'_{\mathcal{R}(A)  \cap \mathcal{R}(C)}\\
& = & A \wedge C.
\end{eqnarray*}

The dual $A = A \vee C \; \Leftrightarrow \mathcal{I}(A)^\perp
\subseteq \mathcal{N}(C)$ is easily obtained from $A = A \wedge C \;
\Leftrightarrow \mathcal{R}(A) \subseteq \mathcal{I}(C)$ by applying
de Morgan's rules and claim \emph{3.}.

\emph{Claims 9 -- 13} follow from Equation (\ref{eqpropadapt}). For \emph{Claim
  14.}\ use \emph{11.}\ and \emph{4.}\ and \emph{6.} to first
  establish that $\mathcal{R}(H \wedge G) = \mathcal{I}(H) \cap
  \mathcal{I}(G)$ and $\mathcal{N}(H \wedge G) = (\mathcal{I}(H) \cap
  \mathcal{I}(G))^\perp$, from which \emph{14.}\ follows. \emph{Claim
  15}\ is analog.

\subsection{Proof of Proposition \ref{propBooleanAperture} (Section
  \ref{secBooleanAperture})} \label{secProofpropBooleanAperture}

 \emph{Claim 1:} $\neg \varphi(C,\gamma) = \varphi(\neg C,
 \gamma^{-1})$.  Notation: All of the matrices $C, \neg C,
 \varphi(C,\gamma),$ $\neg \varphi(C,\gamma), \varphi(\neg C,
 \gamma^{-1})$ are positive semidefinite and have SVDs with identical
 principal component matrix $U$. For any matrix $X$ among these, let
 $U S^X U'$ be its SVD. We write $s_i^X$ for the $i$th singular
 value in $S^X$. We have to show that  $s_i^{\neg
 \varphi(C,\gamma)} = s_i^{\varphi(\neg C,
 \gamma^{-1})}$. In the derivations below, we use various facts from
 Proposition \ref{propSpaces} and Equation (\ref{eqpropadapt}) without
 explicit reference.

\emph{Case }$0 < \gamma < \infty, 0 < s_i^{C} < 1$:
\begin{eqnarray*}
s_i^{\neg \varphi(C,\gamma)} & = & 1 - \frac{s_i^C}{s_i^C +
  \gamma^{-2}\,(1-s_i^C)} \quad = \quad \frac{1-s_i^C}{(1-s_i^C) +
  \gamma^2\,s_i^C}\\
& = & \frac{s_i^{\neg C}}{s_i^{\neg C} + (\gamma^{-1})^{-2}\, (1 -
  s_i^{\neg C})} \quad = \quad s_i^{\varphi(\neg C, \gamma^{-1})}.
\end{eqnarray*}

\emph{Case }$0 < \gamma < \infty, s_i^{C} = 0$: Using $s_i^{C} = 0
\Leftrightarrow s_i^{\neg C} = 1$ we have
\begin{displaymath}
s_i^{\neg \varphi(C,\gamma)} = 1 - s_i^{\varphi(C,\gamma)} = 1 = s_i^{\varphi(\neg C, \gamma^{-1})}.
\end{displaymath}

\emph{Case }$0 < \gamma < \infty, s_i^{C} = 1$: dual of previous case. 

\emph{Case }$ \gamma = 0$: We show $\neg \varphi(C,\gamma) = \varphi(\neg C,
 \gamma^{-1})$ directly. 
\begin{eqnarray*}
\neg \varphi(C,\gamma) & = & \neg \varphi(C,0) = I - \varphi(C,0) = I - \mathbf{P}_{\mathcal{I}(C)} =
\mathbf{P}_{\mathcal{I}(C)^\perp}\\
& = & \mathbf{P}_{\mathcal{N}(\neg C)^\perp} = \varphi(\neg C,
\infty) = \varphi(\neg C, \gamma^{-1}).
\end{eqnarray*}

\emph{Case }$ \gamma = \infty$: the dual analog.

 \emph{Claim 2:}  $\varphi(C,\gamma) \vee \varphi(B, \gamma) = \varphi(C \vee B,
\gamma)$. 

\emph{Case }$0 < \gamma < \infty$: Using concepts and notation from
Proposition \ref{propLimitOr}, it is easy to check that  any
conceptor $A$ can be written as 
\begin{equation}\label{eqproof4-1}
A = \lim_{\delta \downarrow 0}\, R_A^{(\delta)}( R_A^{(\delta)} + I)^{-1},
\end{equation}
and its aperture adapted versions as
\begin{equation}\label{eqproof4-2}
\varphi(A,\gamma) = \lim_{\delta \downarrow 0}\,
R_A^{(\delta)}( R_A^{(\delta)} + \gamma^{-2}I)^{-1}.
\end{equation}
Using Proposition \ref{propLimitOr} and (\ref{eqproof4-1}) we thus have
\begin{eqnarray}
C \vee B & = & \lim_{\delta \downarrow 0}\,(R_C^{(\delta)} +
R_B^{(\delta)})\,(R_C^{(\delta)} + R_B^{(\delta)} + I)^{-1}\label{eqproof4-3}\\
& = &  \lim_{\delta \downarrow 0}\, R_{C \vee B}^{(\delta)}( R_{C \vee
B}^{(\delta)} + I)^{-1}.\label{eqproof4-4}
\end{eqnarray}
Furthermore, again by Proposition \ref{propLimitOr} and by
(\ref{eqproof4-2}), 
\begin{equation}\label{eqproof4-5}
\varphi(C,\gamma) \vee \varphi(B,\gamma) = \lim_{\delta \downarrow 0}\,(R_{\varphi(C,
  \gamma )}^{(\delta)} +
R_{\varphi(B,\gamma)}^{(\delta)})\,(R_{\varphi(C,
  \gamma )}^{(\delta)} + R_{\varphi(B,\gamma)}^{(\delta)} + I)^{-1}
\end{equation}
and 
\begin{equation}\label{eqproof4-6}
\varphi(C \vee B,\gamma)  = \lim_{\delta \downarrow
  0}\, (R_{C \vee B}^{(\delta)})(R_{C \vee
  B}^{(\delta)} + \gamma^{-2}I)^{-1}.
\end{equation}
Using (\ref{eqSemAlpha0c1}), it follows for any conceptor $A$ that 
\begin{equation}\label{eqproof4-7}
R_{\varphi(A,\gamma)}^{(\delta)} = \gamma^2 \,
R_A^{(\delta / (\delta + \gamma^{-2}(1-\delta)))}.
\end{equation}
Applying this to (\ref{eqproof4-5}) and observing that $\lim_{\delta
  \downarrow 0} \delta = 0 = \lim_{\delta
  \downarrow 0} \delta / (\delta + \gamma^{-2}(1-\delta))$ yields
\begin{eqnarray}
\varphi(C,\gamma) \vee \varphi(B,\gamma) & = & \lim_{\delta \downarrow
  0}\,(\gamma^2 R_{C}^{(\delta)} +
\gamma^2 R_{B}^{(\delta)})\,(\gamma^2 R_C^{(\delta)} + \gamma^2
  R_{B}^{(\delta)} + I)^{-1}\nonumber\\
& = & \lim_{\delta \downarrow
  0}\,(R_{C}^{(\delta)} +
 R_{B}^{(\delta)})\,(R_C^{(\delta)} + 
  R_{B}^{(\delta)} + \gamma^{-2}I)^{-1}.\label{eqproof4-8}
\end{eqnarray}
We now exploit the following auxiliary fact which can
be checked by elementary means: If
$(X^{(\delta)})_{\delta}$ is a $\delta$-indexed family of positive semidefinite matrices whose
eigenvectors are identical for different $\delta$, and similarly the
members of the 
familiy $(Y^{(\delta)})_{\delta}$ have identical eigenvectors, and if
the limits $\lim_{\delta \downarrow 0}
X^{(\delta)}(X^{(\delta)}+I)^{-1}$, $\lim_{\delta \downarrow 0}
Y^{(\delta)}(Y^{(\delta)}+I)^{-1}$ exist and are equal, then the
limits  $\lim_{\delta \downarrow 0}
X^{(\delta)}(X^{(\delta)}+ \gamma^{-2} I)^{-1}$, $\lim_{\delta \downarrow 0}
Y^{(\delta)}(Y^{(\delta)}+ \gamma^{-2} I)^{-1}$ exist and are equal, too.
Putting $X^{(\delta)} = R_{C}^{(\delta)} + R_{B}^{(\delta)}$ and
$Y^{(\delta)} = R_{C \vee B}^{(\delta)}$, combining
(\ref{eqproof4-3}), (\ref{eqproof4-4}), (\ref{eqproof4-6}) and
(\ref{eqproof4-8}) with this auxiliary fact yields $\varphi(C \vee B, \gamma) =
\varphi(C,\gamma) \vee \varphi(B,\gamma)$. 

Case $\gamma = 0$: Using various findings from Proposition \ref{propSpaces} we
have

\begin{eqnarray*}
\varphi(C,0) \vee \varphi(B, 0) & = & \mathbf{P}_{\mathcal{I}(C)}
\vee \mathbf{P}_{\mathcal{I}(B)} \quad = \quad
\mathbf{P}_{\mathcal{I}(\mathbf{P}_{\mathcal{I}(C)})\, + \,
  \mathcal{I}(\mathbf{P}_{\mathcal{I}(B)})} \\
& = &  \mathbf{P}_{\mathcal{I}(C) + \mathcal{I}(B)} \quad = \quad
\mathbf{P}_{\mathcal{I}(C \vee B)} \quad = \quad \varphi(C \vee B, 0).
\end{eqnarray*}

Case $\gamma = \infty$:

\begin{eqnarray*}
\varphi(C,\infty) \vee \varphi(B, \infty) & = & \mathbf{P}_{\mathcal{R}(C)}
\vee \mathbf{P}_{\mathcal{R}(B)} \quad = \quad 
\mathbf{P}_{\mathcal{I}(\mathbf{P}_{\mathcal{R}(C)})\, + \,
  \mathcal{I}(\mathbf{P}_{\mathcal{R}(B)})} \\
& = &  \mathbf{P}_{\mathcal{R}(C)\, +\, \mathcal{R}(B)}
 \quad =
\quad  \mathbf{P}_{\mathcal{R}(C \vee B)} \quad = \quad
\varphi(C \vee B, \, \infty).
\end{eqnarray*}

 \emph{Claim 3:}  $\varphi(C,\gamma) \wedge \varphi(B, \gamma) = \varphi(C \wedge B,
\gamma)$: follows from Claims \emph{1.} and \emph{2.} with de
Morgan's law.  

\emph{Claim 4:} $\varphi(C,\gamma) \vee \varphi(C, \beta) = \varphi(C,
\sqrt{\gamma^{2}+\beta^{2}})$:

Case $0 < \gamma, \beta < \infty$: Using Proposition
\ref{propSpaces} and Equations (\ref{eqproof4-7}), (\ref{eqproof4-3}),
(\ref{eqproof4-2}), we obtain
\begin{eqnarray*}
\varphi(C,\gamma) \vee \varphi(C, \beta) & = & \lim_{\delta \downarrow
  0}\, (R_{\varphi(C,\gamma)}^{(\delta)} +
R_{\varphi(C, \beta)}^{(\delta)})\,(R_{\varphi(C,\gamma)}^{(\delta)} +
  R_{\varphi(C, \beta)}^{(\delta)} + I)^{-1}\\
& = & \lim_{\delta \downarrow
  0}\, (\gamma^2 R_{C}^{(\delta / (\delta + \gamma^{-2}(1-\delta)))} + \beta^2
R_{C}^{(\delta / (\delta + \beta^{-2}(1-\delta)))}) \, \cdot\\
& & \quad \quad \cdot \, (\gamma^2
  R_C^{(\delta  / (\delta + \gamma^{-2}(1-\delta)))} + \beta^2
  R_{C}^{(\delta  / (\delta + \beta^{-2}(1-\delta)))} + I)^{-1}\\
& \stackrel{(*)}{=} & \lim_{\delta \downarrow
  0}\, (\gamma^2 + \beta^2) R_C^{(\delta)} ((\gamma^2 + \beta^2)
  R_C^{(\delta)} + I)^{-1} \\
& = & \lim_{\delta \downarrow
  0}\, R_C^{(\delta)} (  R_C^{(\delta)} + (\gamma^2 +
  \beta^2)^{-1}\,I)^{-1} \\
& = & \varphi(C,
\sqrt{\gamma^{2}+\beta^{2}}), 
\end{eqnarray*}
where in step (*) we exploit the fact that the singular values of
$R_C^{(\delta  / (\delta + \gamma^{-2}(1-\delta)))}$ corresponding to
eigenvectors whose eigenvalues in $C$ are less than unity are
identical to the singular values of $R_C^{(\delta)}$ at the analog
positions. 

Case $ \gamma = 0, 0 < \beta < \infty$: Using Proposition
\ref{propLimitOr}, facts from Proposition
\ref{propSpaces}, and Equation (\ref{eqproof4-1}), we obtain
\begin{eqnarray*}
\varphi(C,0) \vee \varphi(C, \beta) & = & \mathbf{P}_{\mathcal{I}(C)}
\vee \varphi(C, \beta)\\
& = & \lim_{\delta \downarrow  0}\, (R_{\mathbf{P}_{\mathcal{I}(C)}}^{(\delta)} +
R_{\varphi(C, \beta)}^{(\delta)})\,(R_{\mathbf{P}_{\mathcal{I}(C)}}^{(\delta)} +
  R_{\varphi(C, \beta)}^{(\delta)} + I)^{-1}\\
& \stackrel{(*)}{=} & \lim_{\delta \downarrow
  0}\, R_{\varphi(C, \beta)}^{(\delta / (2 - \delta))}\,
  (R_{\varphi(C, \beta)}^{(\delta / (2 - \delta))} + I)^{-1} \; = \; \lim_{\delta \downarrow  0}\, R_{\varphi(C, \beta)}^{(\delta)}\,
  (R_{\varphi(C, \beta)}^{(\delta /)} + I)^{-1}\\
& = & \varphi(C, \beta) \;\; = \; \; \varphi(C, \sqrt{0^2 + \beta^2}),
\end{eqnarray*}
where step (*) is obtained by observing $\mathcal{I}(C) =
\mathcal{I}(\varphi(C,\beta)) $ and applying the definition  of
$R^{(\delta)}_A$ given in the statement of Proposition
\ref{propLimitOr}.

Case $ \gamma = \infty, 0 < \beta < \infty$: the dual analog to the
previous case:
\begin{eqnarray*}
\varphi(C,\infty) \vee \varphi(C, \beta) & = & \mathbf{P}_{\mathcal{R}(C)}
\vee \varphi(C, \beta)\\
& = & \lim_{\delta \downarrow  0}\, (R_{\mathbf{P}_{\mathcal{R}(C)}}^{(\delta)} +
R_{\varphi(C, \beta)}^{(\delta)})\,(R_{\mathbf{P}_{\mathcal{R}(C)}}^{(\delta)} +
  R_{\varphi(C, \beta)}^{(\delta)} + I)^{-1}\\
& \stackrel{(*)}{=} &  \lim_{\delta \downarrow  0}\, R_{\mathbf{P}_{\mathcal{R}(C)}}^{(\delta)}\,
  (R_{\mathbf{P}_{\mathcal{R}(C)}}^{(\delta /)} + I)^{-1}\\
& = & \mathbf{P}_{\mathcal{R}(C)} \; = \; \varphi(C,\infty) = \varphi(C, \sqrt{\infty^2 + \beta^2}),
\end{eqnarray*}
where in step (*) I have omitted obvious intermediate calculations.

The cases $0 < \gamma < \infty, \beta \in \{0,\infty\}$ are symmetric
to cases already treated, and the cases $ \gamma, \beta \in \{0,\infty
\}$ are obvious.

\emph{Claim 5:} $\varphi(C,\gamma) \wedge \varphi(C, \beta) = \varphi(C,
(\gamma^{-2}+\beta^{-2})^{-2})$: an easy exercise of applying de
Morgan's rule in conjunction with Claims \emph{1.} and \emph{4.}

\subsection{Proof of Proposition \ref{propBooleanElementaryLaws}
  (Section \ref{secLogicLaws})} \label{secProofpropBooleanElementaryLaws}

\emph{1. De Morgan's rules:} By Definition \ref{def:finalBoolean} and
Proposition \ref{propdeMorganAND}. 
\emph{2. Associativity:} From Equations (\ref{eqproof4-3}) and
(\ref{eqproof4-4}) it follows that for any conceptors $B,C$ it holds
that $\lim_{\delta \downarrow 0}R^{(\delta)}_{B \vee C} = \lim_{\delta
  \downarrow 0} R^{(\delta)}_B +  R^{(\delta)}_C$. Employing this
fact and using Proposition \ref{propLimitOr} yields associativity of
OR. Applying de Morgan's law then transfers associativity to AND.
\emph{3. Commutativity and 4. double negation} are clear. 
\emph{5. Neutrality:} Neutrality of $I$: Observing that
$\mathcal{R}(C) \cap \mathcal{R}(I) = \mathcal{R}(C)$ and $I^\dagger =
I$, starting from the definition of $\vee$ we obtain
\begin{eqnarray*}
C \vee I & = & (\mathbf{P}_{\mathcal{R}(C)} C^\dagger
\mathbf{P}_{\mathcal{R}(C)})^\dagger \\
& = & (\mathbf{P}_{\mathcal{R}(C^\dagger)} C^\dagger
\mathbf{P}_{\mathcal{R}(C^\dagger)})^\dagger \quad \mbox{(by Prop.\
  \ref{propSpaces} Nr.\ 2)}\\
& = & (C^\dagger)^\dagger\\
& = & C.
\end{eqnarray*}
Neutrality of $0$ can be obtained from neutrality of $I$ via de
Morgan's rules. 
 
\emph{6. Globality:} $C \wedge 0 = 0$ follows immediately from the
Definition of $\wedge$ given in \ref{def:finalBoolean}, observing that
$\mathbf{P}_{\mathcal{R}(C) \cap \mathcal{R}(0)} = 0$. The dual $C
\vee I = I$ is obtained by applying de Morgan's rule on $C \wedge 0 =
0$.

\emph{7. and
8. weighted absorptions} follows from Proposition
\ref{propBooleanAperture} items \emph{4.} and \emph{5.}

\subsection{Proof of Proposition \ref{propPseudoabsorbtion} (Section
  \ref{secLogicLaws})} \label{secProofpropPseudoabsorbtion}

Let $\leq$ denote the well-known \emph{L\"{o}wner} ordering on the set
of real $N \times N$ matrices defined by $A \leq B$ if $B - A$ is
positive semidefinite. Note that a matrix $C$ is a conceptor matrix if
and only if $0 \leq C \leq I$. I first show the following

\begin{lemma}\label{lemmaLoewner1}
Let $A \leq B$. Then $A^\dagger \geq \mathbf{P}_{\mathcal{R}(A)}\,
B^\dagger \,\mathbf{P}_{\mathcal{R}(A)}$.
\end{lemma}

\emph{Proof of Lemma.} $A^\dagger$ and $B^\dagger$ can be written as
\begin{equation}\label{eqLemmaL1}
A^\dagger = \lim_{\delta \to 0} \mathbf{P}_{\mathcal{R}(A)} (A +
\delta I)^{-1} \mathbf{P}_{\mathcal{R}(A)} \mbox{ and } B^\dagger =
\lim_{\delta \to 0} \mathbf{P}_{\mathcal{R}(B)} (B + 
\delta I)^{-1} \mathbf{P}_{\mathcal{R}(B)}.
\end{equation}
From $A \leq B$ it follows that $\mathcal{R}(A) \subseteq
\mathcal{R}(B)$, that is, $\mathbf{P}_{\mathcal{R}(A)}
\mathbf{P}_{\mathcal{R}(B)} = \mathbf{P}_{\mathcal{R}(A)}$, which in turn yields
\begin{equation}\label{eqLemmaL2}
\mathbf{P}_{\mathcal{R}(A)}\,B^\dagger \, \mathbf{P}_{\mathcal{R}(A)}
= \lim_{\delta \to 0} \mathbf{P}_{\mathcal{R}(A)} (B +
\delta I)^{-1} \mathbf{P}_{\mathcal{R}(A)}.
\end{equation}
$A \leq B$ entails $A + \delta I \leq B + \delta I$, which is
equivalent to $(A + \delta I)^{-1} \geq (B + \delta I)^{-1}$ (see
\cite{Bernstein09}, fact 8.21.11), which implies
\begin{displaymath}
 \mathbf{P}_{\mathcal{R}(A)} (B + \delta I)^{-1} \mathbf{P}_{\mathcal{R}(A)} \leq
 \mathbf{P}_{\mathcal{R}(A)} (A + \delta I)^{-1} \mathbf{P}_{\mathcal{R}(A)},
\end{displaymath}
see Proposition 8.1.2 (xii) in \cite{Bernstein09}. Taking the limits
(\ref{eqLemmaL1}) and (\ref{eqLemmaL2}) leads to the claim of the
lemma (see fact 8.10.1 in \cite{Bernstein09}).

\emph{Proof of Claim 1.} Let $A, B$ be conceptor matrices of size $N
\times N$.  According to Proposition \ref{propLoewnerBoolean} (which is proven
    independently of the results stated in Proposition
    \ref{propPseudoabsorbtion}), it holds that $A \leq A \vee B$,
    which combined with the lemma above establishes 
\begin{displaymath}
 \mathbf{P}_{\mathcal{R}(A)}  (A^\dagger - (A \vee B)^\dagger)
 \mathbf{P}_{\mathcal{R}(A)} \geq 0, 
\end{displaymath}
from which it follows that $ I + \mathbf{P}_{\mathcal{R}(A)}
(A^\dagger - (A \vee B)^\dagger)  \mathbf{P}_{\mathcal{R}(A)}$ is
positive semidefinite with all singular values greater or equal to
one. Therefore, $ \mathbf{P}_{\mathcal{R}(A)}(I + 
A^\dagger - (A \vee B)^\dagger)  \mathbf{P}_{\mathcal{R}(A)}$ is
positive semidefinite with all nonzero singular values  greater or equal to
one. Hence $C  = \left(\mathbf{P}_{\mathcal{R}(A)}
  \left(I + A^\dagger - (A \vee B)^\dagger \right)
  \mathbf{P}_{\mathcal{R}(A) } \right)^{\dagger}$ is a conceptor
matrix. It is furthermore obvious that $\mathcal{R}(C) =
\mathcal{R}(A)$.

From $A \leq A \vee B$ it follows that $\mathcal{R}(A) \subseteq
\mathcal{R}(A \vee B)$, which together with $\mathcal{R}(C) =
\mathcal{R}(A)$ leads to $\mathcal{R}(A \vee B) \cap \mathcal{R}(C) =
\mathcal{R}(A)$. Exploiting this fact, starting from the definition of
AND in Def.\ \ref{def:finalBoolean}, we conclude
\begin{eqnarray*}
(A \vee B) \wedge C & = & \left(\mathbf{P}_{\mathcal{R}(A \vee B) \cap
    \mathcal{R}(C) } \left((A \vee B)^\dagger + C^\dagger - I \right)
    \mathbf{P}_{\mathcal{R}(A \vee B) \cap 
    \mathcal{R}(C) }  \right)^\dagger \\
& = &  \left(\mathbf{P}_{\mathcal{R}(A)} \left((A \vee B)^\dagger +
    C^\dagger - I \right) \mathbf{P}_{\mathcal{R}(A) }
    \right)^\dagger \\ 
& = &  \left(\mathbf{P}_{\mathcal{R}(A)} \left((A \vee B)^\dagger +
    \mathbf{P}_{\mathcal{R}(A)}
  \left(I + A^\dagger - (A \vee B)^\dagger \right)
  \mathbf{P}_{\mathcal{R}(A) } - I \right) \mathbf{P}_{\mathcal{R}(A) }
    \right)^\dagger \\
& = & (\mathbf{P}_{\mathcal{R}(A) } A^\dagger
    \mathbf{P}_{\mathcal{R}(A) })^\dagger \;\; = \;\; A.
\end{eqnarray*}

\emph{Proof of Claim 2.} This claim is the Boolean dual to claim 1 and
can be straightforwardly derived by transformation from claim 1, using
de Morgan's rules and observing that $\mathcal{R}(\neg A) =
\mathcal{I}(A)^\perp$ (see Prop.\ \ref{propSpaces} item 3), and that
$\neg A = I - A$.

%% \item Let $A, B \in \mathcal{C}_N$ and $B \leq A$. Then 
%% \begin{displaymath}
%% C = \mathbf{P}_{\mathcal{R}(B)}\, (B^\dagger -
%% \mathbf{P}_{\mathcal{R}(B)}\, A^\dagger\, \mathbf{P}_{\mathcal{R}(B)}
%% + I)^{-1}\,\mathbf{P}_{\mathcal{R}(B)} 
%% \end{displaymath}
%% is a conceptor matrix and
%% \begin{displaymath}
%% B = A \wedge C.
%% \end{displaymath}
%% \item Let again $A, B \in \mathcal{C}_N$ and $A \leq B$. Then 
%% \begin{displaymath}
%% C = I  - \left(\mathbf{P}_{\mathcal{I}(A)^\perp} \left(I +
%%     (I-A)^\dagger - (I - (A \wedge B))^\dagger \right)
%%     \mathbf{P}_{\mathcal{I}(A)^\perp} \right)^\dagger 
%% \end{displaymath}
%% is a conceptor matrix and
%% \begin{displaymath}
%% B = A \vee C.
%% \end{displaymath}
%% \item If for $A, B, C \in \mathcal{C}_N$ it holds that $A \wedge C =
%% B$, then $B \leq A$.
%% \item If for $A, B, C \in \mathcal{C}_N$ it holds that $A \vee C =
%% B$, then $A \leq B$.
%% \item For $A \in \mathcal{A}_N$ and $\gamma \geq 1$ it holds that $A
%% \leq \varphi(A, \gamma)$; for  $\gamma \leq 1$ it holds that
%% $\varphi(A, \gamma) \leq A$.
%% \end{enumerate}
%% \end{proposition}

\subsection{Proof of Proposition \ref{propBasicAbstraction} (Section
  \ref{secAbstraction})} \label{secProofpropBasicAbstraction}

\emph{Claim 1:} $0 \leq A$ is equivalent to $A$ being positive
semidefinite, and for a positive semidefinite matrix $A$, the
condition $A \leq I$ is equivalent to all singular values of $A$ being
at most one. Both together yield the claim. 

\emph{Claim 2:} Follows from claim 1.

\emph{Claim 3:} $A \leq B$ iff $-A \geq -B$ iff $I-A \geq I-B$, which
is the same as $\neg A \geq \neg B$. 

\emph{Claim 4:} We first show an auxiliary, general fact:

\begin{lemma}\label{lemmaLoewner2}
Let $X$ be positive semidefinite, and $\mathbf{P}$ a projector
matrix. Then $$(\mathbf{P} (\mathbf{P} X \mathbf{P} +
I)^{-1}\mathbf{P})^\dagger = \mathbf{P} X \mathbf{P} + \mathbf{P}.$$
\end{lemma}

\emph{Proof of Lemma.} Let $U = \mathcal{R}(\mathbf{P})$ be the
projection space of $\mathbf{P}$. It is clear that
$\mathbf{P}X\mathbf{P} + I: U \to U$ and $\mathbf{P}X\mathbf{P} + I:
U^\perp \to U^\perp$, hence $\mathbf{P} X \mathbf{P} + I$ is a
bijection on $U$. Also $\mathbf{P}$ is a bijection on $U$. We now call
upon the following well-known property of the pseudoinverse (see
\cite{Bernstein09}, fact 6.4.16):

\emph{For matrices $K,L$ of compatible sizes it holds that
  $(KL)^\dagger = L^\dagger K^\dagger$ if and only if $\mathcal{R}(K'KL)
  \subseteq \mathcal{R}(L)$ and $\mathcal{R}(LL'K)
  \subseteq \mathcal{R}(K')$.}

Observing that $\mathbf{P}$ and  $\mathbf{P} X \mathbf{P} + I$ are
bijections on $U = \mathcal{R}(\mathbf{P})$, a twofold application of
the mentioned fact yields $(\mathbf{P} (\mathbf{P} X \mathbf{P} +
I)^{-1}\mathbf{P})^\dagger = \mathbf{P}^\dagger (\mathbf{P} X \mathbf{P} +
I)  \mathbf{P}^\dagger =  \mathbf{P} X \mathbf{P} + \mathbf{P}$ which
completes the proof of the lemma.

Now let $A, B \in \mathcal{C}_N$ and $B \leq A$. By Lemma
\ref{lemmaLoewner1}, $B^\dagger - \mathbf{P}_{\mathcal{R}(B)}\,
A^\dagger\, \mathbf{P}_{\mathcal{R}(B)}$ is positive semidefinite,
hence $B^\dagger - \mathbf{P}_{\mathcal{R}(B)}\, A^\dagger\,
\mathbf{P}_{\mathcal{R}(B)} + I$ is positive definite with singular
values greater or equal to one, hence invertible. The singular values
of $(B^\dagger - \mathbf{P}_{\mathcal{R}(B)}\, A^\dagger\,
\mathbf{P}_{\mathcal{R}(B)} + I)^{-1}$ are thus at most one, hence $C
= \mathbf{P}_{\mathcal{R}(B)}\, (B^\dagger -
\mathbf{P}_{\mathcal{R}(B)}\, A^\dagger\, \mathbf{P}_{\mathcal{R}(B)}
+ I)^{-1}\,\mathbf{P}_{\mathcal{R}(B)} $ is a conceptor
matrix. Obviously $\mathcal{R}(C) = \mathcal{R}(B)$.

Using these findings and 
Lemma \ref{lemmaLoewner2} we can now infer
\begin{eqnarray*}
A \wedge C & = & \left(\mathbf{P}_{\mathcal{R}(A) \cap
    \mathcal{R}(C)}\,(A^\dagger + C^\dagger - I) \,\mathbf{P}_{\mathcal{R}(A) \cap
    \mathcal{R}(C)}  \right)^\dagger \\
& = & \left(\mathbf{P}_{\mathcal{R}(B)}\,(A^\dagger +
    \left(\mathbf{P}_{\mathcal{R}(B)}\, (B^\dagger -
    \mathbf{P}_{\mathcal{R}(B)}\, A^\dagger\,
    \mathbf{P}_{\mathcal{R}(B)} + I)^{-1}\,\mathbf{P}_{\mathcal{R}(B)} 
    \right)^\dagger - I) \,\mathbf{P}_{\mathcal{R}(B)}
    \right)^\dagger \\
& = & \left(\mathbf{P}_{\mathcal{R}(B)}\,\left(A^\dagger +
    \mathbf{P}_{\mathcal{R}(B)} \,(B^\dagger -
    \mathbf{P}_{\mathcal{R}(B)} \,A^\dagger
    \mathbf{P}_{\mathcal{R}(B)} ) \,\mathbf{P}_{\mathcal{R}(B)}
    \right)  \,\mathbf{P}_{\mathcal{R}(B)}  \right)^\dagger \\
& = & B.
\end{eqnarray*}

\emph{Claim 5:} This claim is the Boolean dual to the previous
claim. It follows by a straightforward transformation applying de
Morgan's rules and observing that $\mathcal{R}(\neg B) =
\mathcal{I}(B)^\perp$ (see Prop.\ \ref{propSpaces} item 3), and that
$\neg A = I - A$.  

\emph{Claim 6:} Let $A \wedge C = B$. Using the notation and claim
from Prop.\ \ref{propANDDef} rewrite  $A \wedge C = \lim_{\delta \to
  0} (C_\delta^{-1} + A_\delta^{-1} - I)^{-1}$. Similarly, obviously
we can also rewrite $A =  \lim_{\delta \to  0} A_\delta$. Since
$C_\delta^{-1} \geq I$, conclude
\begin{eqnarray*}
& & C_\delta^{-1} + A_\delta^{-1} - I \geq  A_\delta^{-1} \\
& \Longleftrightarrow & (C_\delta^{-1} + A_\delta^{-1} - I)^{-1} \leq
A_\delta \\
& \Longrightarrow & \lim_{\delta \to  0} (C_\delta^{-1} + A_\delta^{-1} -
I)^{-1} \leq \lim_{\delta \to  0} A_\delta\\
& \Longleftrightarrow & A \wedge C \leq A,
\end{eqnarray*}
where use is made of the fact that taking limits preserves $\leq$ (see
\cite{Bernstein09} fact 8.10.1).

\emph{Claim 7:} Using the result from the previous claim, infer $A
\vee C = B \Longrightarrow \neg A \wedge \neg C = \neg B
\Longrightarrow \neg A \geq \neg B \Longrightarrow A \leq B$.

\emph{Claim 8:}  Let $\gamma = \sqrt{1 + \beta^2} \geq 1$, where
$\beta \geq 0$. By
Proposition \ref{propBooleanAperture} \emph{4.} we get
$\varphi(A,\gamma) = \varphi(A,1) \vee \varphi(A,\beta) = A \vee
\varphi(A, \beta)$, hence $A \leq \varphi(A,\gamma)$. The dual
version is obtained from this result  by using 
Proposition \ref{propIterateApAdapt}: let $\gamma \leq 1$, hence
$\gamma^{-1} \geq 1$. Then $\varphi(A,\gamma) \leq
\varphi(\varphi(A,\gamma), \gamma^{-1}) = A$.

\emph{Claim 9:} If $\gamma = 0$, then from Proposition \ref{propapadapt} it is
clear that $\varphi(A,0)$ is the
projector matrix on $\mathcal{I}(A)$ and  $\varphi(B,0)$ is the
projector matrix on $\mathcal{I}(B)$. From $A \leq B$ and the fact
that $A$ and $B$ do not have singular values exceeding 1 it is clear
that $\mathcal{I}(A) \subseteq \mathcal{I}(B)$, thus
$\varphi(A,0) \leq \varphi(B,0)$. 

If $\gamma = \infty$, proceed in an analog way and use Proposition
\ref{propapadapt} to conclude that $\varphi(A,\infty),
\varphi(B,\infty)$ are the projectors on $\mathcal{R}(A),
\mathcal{R}(B)$ and apply that $A \leq B$ implies $\mathcal{R}(A)
\subseteq \mathcal{R}(B)$.

It remains to treat the case $0 < \gamma < \infty$. Assume $A \leq B$,
that is, there exists a positive semidefinite matrix $D$ such that $A
+ D = B$. Clearly $D$ cannot have singular values exceeding one, so $D$
is a conceptor matrix. For (small) $\delta > 0$, let $A^{(\delta)} =
(1-\delta) A$. Then $A^{(\delta)}$ can be written as
$A^{(\delta)} = R^{(\delta)}(R^{(\delta)} + I)^{-1}$ for a positive
semidefinite $R^{(\delta)}$, and it holds that
$$A = \lim_{\delta \to 0} A^{(\delta)},$$
and furthermore 
$$A^{(\delta)} \leq A^{(\delta')} \leq A \;\; \mbox{for } \delta \geq \delta'.$$
Similarly, let $D^{(\delta)} = (1-\delta)D$, with $D^{(\delta)} =
Q^{(\delta)}(Q^{(\delta)} + I)^{-1}$, and observe again 
$$D = \lim_{\delta \to 0} D^{(\delta)} \quad \mbox{and} \quad
D^{(\delta)} \leq D^{(\delta')} \leq D \;\; \mbox{for } \delta \geq
\delta'.$$ 
Finally, define ${B}^{(\delta)} = (1-\delta)B$, where ${B}^{(\delta)} =
P^{(\delta)}(P^{(\delta)}+I)^{-1}$.  Then 
$$B = \lim_{\delta \to 0} {B}^{(\delta)} \quad \mbox{and} \quad
{B}^{(\delta)} \leq {B}^{(\delta')} \leq B \;\; \mbox{for } \delta \geq
\delta' \quad \mbox{and} \quad {B}^{(\delta)} = A^{(\delta)} + D^{(\delta)}.$$ 
Because of ${B}^{(\delta)} = A^{(\delta)} + D^{(\delta)}$ we have 
\begin{equation}\label{eqproofadapt2}
R^{(\delta)}(R^{(\delta)} + I)^{-1} \leq P^{(\delta)}(P^{(\delta)}+I)^{-1}.
\end{equation}
We next state a lemma which is of interest in its own right too.
\begin{lemma}\label{lemmaProofAdapt1}
For correlation matrices $R, P$ of same size it holds that
\begin{equation}\label{eqproofadapt3}
R(R+I)^{-1} \leq P(P+I)^{-1} \quad \mbox{iff} \quad R \leq P.
\end{equation}
\end{lemma}
\emph{Proof of Lemma.} Assume $R(R+I)^{-1} \leq P(P+I)^{-1}$. By claim
\emph{4.}\ of this proposition, there is a conceptor matrix $C$ such
that $P(P+I)^{-1} = R(R+I)^{-1} \vee C$. Since $P(P+I)^{-1} < I$, $C$
has no unit singular values and thus can be written as $S(S+I)^{-1}$,
where $S$ is a correlation matrix. Therefore, $P(P+I)^{-1} =
R(R+I)^{-1} \vee S(S+I)^{-1} = (R+S)(R+S+I)^{-1}$, hence $P = R+S$,
that is, $R \leq P$.

Next assume $R \leq P$, that is, $P = R+S$ for a correlation matrix
$S$. This implies $P(P+I)^{-1} = (R+S)(R+S+I)^{-1} = R(R+I)^{-1} \vee
S(S+I)^{-1}$. By claim \emph{6.}\ of this proposition, $R(R+I)^{-1}
\leq P(P+I)^{-1}$ follows. This concludes the proof of the lemma.

Combining this lemma with (\ref{eqproofadapt2}) and the obvious fact
that $R^{(\delta)} \leq P^{(\delta)}$ if and only if $\gamma^2\,
R^{(\delta)} \leq \gamma^2\, P^{(\delta)}$ yields
\begin{equation}\label{eqproofadapt4}
\gamma^2 R^{(\delta)}(\gamma^2 R^{(\delta)} + I)^{-1} \leq \gamma^2
P^{(\delta)}(\gamma^2 P^{(\delta)}+I)^{-1}.
\end{equation}

Another requisite auxiliary  fact is contained in the next

\begin{lemma}\label{lemmaProofAdapt2}
  Let $0 < \gamma < \infty$. If $A = USU' = \lim_{\delta \to 0}
  R^{(\delta)} (R^{(\delta)} + I )^{-1}$ and for all $\delta$,
  $R^{(\delta)}$ has a SVD $R^{(\delta)} = U\Sigma^{(\delta)}U'$, then
  $\varphi(A,\gamma) = \lim_{\delta \to 0}
  \gamma^2 R^{(\delta)} (\gamma^2 R^{(\delta)} + I )^{-1}$.
\end{lemma}
\emph{Proof of Lemma.} Since all $R^{(\delta)}$ (and hence, all
$R^{(\delta)} (R^{(\delta)} + I )^{-1}$ and $ \gamma^2 R^{(\delta)}
(\gamma^2 R^{(\delta)} + I )^{-1}$) have the same eigenvectors as $A$,
it suffices to show the convergence claim on the level of individual
singular values of the concerned matrices. Let $s, s_{\gamma},
\sigma^{(\delta)}, s^{(\delta)}, s^{(\delta)}_\gamma$ denote a
singular value of $A, \varphi(A,\gamma), R^{(\delta)}, R^{(\delta)}
(R^{(\delta)} + I )^{-1}, \gamma^2 R^{(\delta)} (\gamma^2 R^{(\delta)}
+ I )^{-1}$, respectively (all these versions referring to the same
eigenvector in $U$). For convenience I restate from Proposition
\ref{propapadapt} that
\begin{displaymath}
s_{\gamma} = \left\{\begin{array}{ll}
s / (s + \gamma^{-2}(1-s)) & \quad \mbox{for }\;\; 0 < s < 1, \\
0 & \quad \mbox{for  }\;\; s = 0,\\
1 & \quad \mbox{for  }\;\; s = 1.\\
  \end{array}    \right. 
\end{displaymath} 
It holds that $s^{(\delta)} = \sigma^{(\delta)} / (\sigma^{(\delta)} +
1)$ and $\lim_{\delta \to 0} s^{(\delta)} = s$, and similarly
$s^{(\delta)}_\gamma = \gamma^2 \sigma^{(\delta)} / (\gamma^2 \sigma^{(\delta)} +
1)$. It needs to be shown that $\lim_{\delta \to 0}
s^{(\delta)}_\gamma = s_\gamma$. 

\noindent \emph{Case $s = 0$}: 
\begin{eqnarray*}
s = 0 & \Longrightarrow & \lim_{\delta \to 0} s^{(\delta)} = 0 \;\;
\Longrightarrow \;\; \lim_{\delta \to 0} \sigma^{(\delta)} = 0\\
& \Longrightarrow & \lim_{\delta \to 0} s^{(\delta)}_\gamma = 0 .
\end{eqnarray*}
\emph{Case $s = 1$}: 
\begin{eqnarray*}
s = 1 & \Longrightarrow & \lim_{\delta \to 0} s^{(\delta)} = 1 \;\;
\Longrightarrow \;\; \lim_{\delta \to 0} \sigma^{(\delta)} = \infty\\
& \Longrightarrow & \lim_{\delta \to 0} s^{(\delta)}_\gamma = 1.
\end{eqnarray*}
\emph{Case $0 < s < 1$}: 
\begin{eqnarray*}
s =  \lim_{\delta \to 0} s^{(\delta)}  & \Longrightarrow & s = 
\lim_{\delta \to 0}  \sigma^{(\delta)} / (\sigma^{(\delta)} +
1) \;\;
 \Longrightarrow \;\;  \lim_{\delta \to 0} \sigma^{(\delta)} =
 s/(1-s)\\
 & \Longrightarrow & \lim_{\delta \to 0} s^{(\delta)}_\gamma =
 \frac{\gamma^2 s / (1-s)}{\gamma^2 s / (1-s) + 1} \; = \; s / (s +
 \gamma^{-2}(1-s)). 
\end{eqnarray*}
This concludes the proof of the lemma.

After these preparations, we can finalize the proof of claim
\emph{9.}\ as follows. From Lemma \ref{lemmaProofAdapt2} we know that
$$\varphi(A,\gamma) = \lim_{\delta \to 0} \gamma^2 R^{(\delta)}
(\gamma^2 R^{(\delta)} + I )^{-1}$$
and
$$\varphi(B,\gamma) = \lim_{\delta \to 0} \gamma^2 P^{(\delta)}
(\gamma^2 P^{(\delta)} + I )^{-1}.$$
From (\ref{eqproofadapt4}) and the fact that $\leq$ is preserved under
limits we obtain $\varphi(A,\gamma) \leq \varphi(B,\gamma)$.

\subsection{Proof of Proposition \ref{prop1} (Section
  \ref{subsecCAD})}\label{secProofprop1} 

We use the following notation for matrix-vector transforms. We sort
the entries of an $N \times N$ matrix $M$ into an $N^2$-dimensional
vector $vec\,M$ row-wise (!). That is, $vec\,M(\mu) =$ \\$
M(\lceil\mu / N\rceil, mod_1(\mu, N))$, where the ceiling
$\lceil x \rceil$ of a real number $x$ is the smallest integer greater
or equal to $x$, and $mod_1(\mu, N)$ is the modulus function except for
arguments of the form $(l k, k)$, where we replace the standard value
$mod(l m, m) = 0$ by $mod_1(l m, m) = m$. Conversely, $M(i,j) =
vec\,M((i-1)N + j)$. 

The Jacobian $J_C$ can thus be written as a $N^2
\times N^2$ matrix $J_C(\mu,\nu) = \partial \,vec\,\dot{C}(\mu) /
\partial \,vec\,C(\nu)$. The natural
parametrization of matrices $C$ by their matrix elements does not lend
itself easily to an eigenvalue analysis. Assuming that a reference solution
$C_0 = U S U'$ is fixed, any $N \times N$ matrix $C$ is uniquely
represented by a parameter matrix $P$ through $C = C(P) = U \, (S + P)
\, U'$, with $C(P) = C_0$ if and only if $P = 0$. Conversely, any
parameter matrix $P$ yields a unique $C(P)$.

Now we consider the Jacobian $J_P(\mu, \nu) =  \partial \,vec \;\dot{P}_C(\mu) /
\partial \,vec \;P_C(\nu)$.  By using that (i) $\dot{P} =
U'\dot{C}U$, (ii) $vec\,(X'YX) = (X \otimes X)'\,vec\,Y$ for
square matrices $X, Y$, and (iii) $\partial\, A \dot{x} / \partial\, A
x = A (\partial\, \dot{x} / \partial\, x) A^{-1}$ for invertible $A$,
(iv) $(U \otimes U)^{-1} = (U \otimes U)'$, 
one obtains that $J_P = (U \otimes U)' J_C (U \otimes U)$ and hence $J_P$
and $J_C$ have the same eigenvalues. 

Using $C_0 = USU'$ and $\dot{C} = (I - C) C D C' - \alpha^{-2} C$ and the
fact that diagonal entries in $S$ of index greater than $k$ are zero, yields

\begin{eqnarray}
\left.\frac{\partial \, \dot{p}_{lm}}{\partial \, p_{ij}}\right|_{P = 0} &
= & e'_l\; (I_{ij}U'DUS + SU'_kDUI_{ji} \nonumber\\
& & - I_{ij}SU'_kDUS - SI_{ij}U'DUS - S^2U'_kDUI_{ji} - \alpha^{-2}
I_{ij})\; e_m, \label{somEq1}
\end{eqnarray}

\noindent where $e_l$ is the $l$-th unit vector and $I_{ij} =
e_i\,e'_j$. Depending on how $l, m, i, j$ relate to
$k$ and to each other, calculating (\ref{somEq1}) leads to numerous
case distinctions. Each of the cases concerns entries in a specific
 subarea of $J_P$. These subareas are depicted in Fig.
 \ref{somFig1}, which shows $J_P$ in an instance with $N = 5, k = 3$.

\begin{figure}[htb]
\center
\includegraphics[width=80 mm]{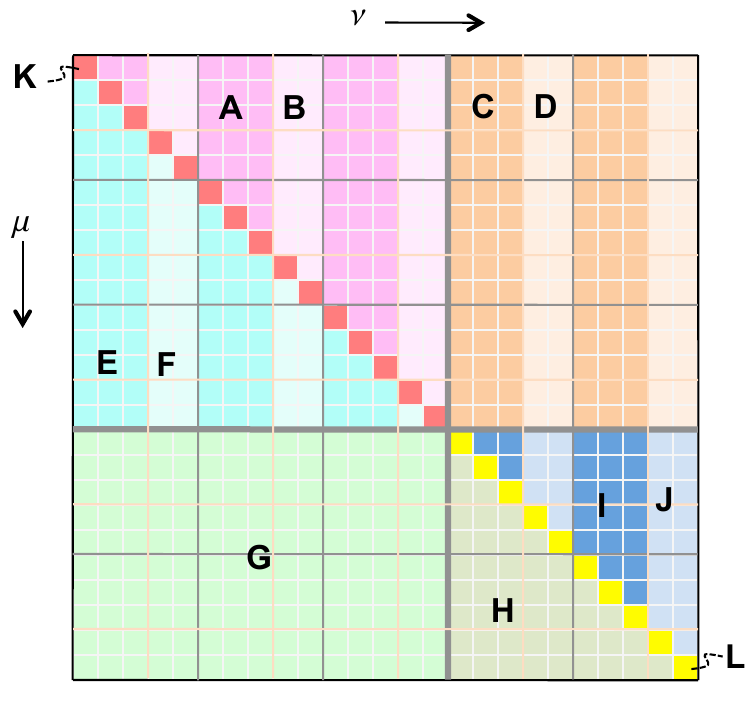}
\caption{Main case distinction areas for computing values in the
  matrix $J_P$. An instance 
  with $N = 5, k = 3$ is shown. Areas are denoted by A, ..., L; same
  color = same area. $J_P$ has size $N^2 \times N^2$. Its structure is largely
  organized by a $kN \times kN$-dimensional and a $(N-k)N \times
  (N-k)N$ submatrix on the diagonal (areas ABEFK and HIJL,
  respectively). Column/row indices are denoted by $\mu, \nu$.
  Area specifications: A: $\mu < \nu, \nu \leq kN,
  \mbox{mod}_1(\nu, N) \leq k.$ B: $\mu < \nu, \nu \leq kN,
  \mbox{mod}_1(\nu, N) > k.$ C: $\nu > kN, \mu \leq kN,
  \mbox{mod}_1(\nu, N) \leq k.$ D: $\nu > kN, \mu \leq kN,
  \mbox{mod}_1(\nu, N) > k.$ E: $\mu > \nu, \mu \leq kN,
  \mbox{mod}_1(\nu, N) \leq k.$ F: $\mu > \nu, \mu \leq kN,
  \mbox{mod}_1(\nu, N) > k.$ G: $\mu > kN, \nu \leq kN.$ H:
  $\nu > kN, \nu < \mu.$ I: $\mu > kN, \mu < \nu,
  \mbox{mod}_1(\nu, N) \leq k.$ J: $\mu > kN, \mu < \nu,
  \mbox{mod}_1(\nu, N) > k.$ K: $\mu = \nu \leq kN.$ L:
  $\mu = \nu > kN.$}
\label{somFig1}
\end{figure}

I will demonstrate in detail only two of these cases (subareas A and B in
Fig. \ref{somFig1}) and summarize the results of the others (calculations are
mechanical). 

The case A concerns all entries $J_P(\mu, \nu)$ with $\mu < \nu, \nu \leq kN,  \mbox{mod}_1(\nu, N) \leq k.$ 
Translating indices
$\mu, \nu$ back to indices $l, m, i, j$ via $J_P(\mu, \nu) =
\partial \,vec \,\dot{P}_C(\mu)\, / \,\partial \,vec \, P_C(\nu)
= $ \\$\partial \, \dot{p}_{(\lceil \mu /N \rceil, mod_1\,(\mu,N))} \,
/ \, \partial \, p_{(\lceil \nu /N \rceil, mod_1\,(\nu,N))} =
\partial \, \dot{p}_{lm} \, / \, \partial p_{ij} $ yields conditions
(i) $i \leq k$ (from $i = \lceil \nu / N \rceil$ and $\nu \leq kN$), (ii) $j
\leq k$ (from $j = mod_1(\nu, N) \leq k$) and (iii.a) $l < i$ or (iii.b)
$l = i \wedge m < j$ (from $\mu < \nu$).

I first treat the subcase (i), (ii), (iii.a). Since $l \neq i$ one has
$e'_l \, I_{ij} = 0$ and eqn.\ (\ref{somEq1}) reduces to the terms
starting with $S$, leading to

\begin{eqnarray}
\left.\frac{\partial \, \dot{p}_{lm}}{\partial \, p_{ij}}\right|_{P = 0} &
= & e'_l\; (SU'_kDUI_{ji}  - SI_{ij}U'DUS - S^2U'_kDUI_{ji} )\; e_m
\nonumber\\
& = & s_l u'_l D u_j \delta_{im} - s_l e'_l I_{ij}U'DUSe_m - s_l^2
u'_l D u_j \delta_{im} \nonumber\\
& = & s_l u'_l D u_j \delta_{im} - s_l^2 u'_l D u_j \delta_{im}\nonumber\\
& = & \left\{ \begin{array}{lll} 0, & \mbox{if } i \neq m & \mbox{(subcase A1)} \nonumber\\ 
0, & \mbox{if } i = m, j \neq l & \mbox{(A2)}\\
\alpha^{-2},  & \mbox{if } i = m, j = l &\mbox{(A3)}, \end{array}   \right.\label{somEq2}
\end{eqnarray}

\noindent where $u_l$ is the $l$-th column in $U$ and $\delta_{im} = 1$
if and only if $i = m$ (else $0$) is the Kronecker delta. The value
$\alpha^{-2}$ noted
for subcase A3 is obtained through $(s_l - s_l^2) u'_l D u_l = (s_l -
s_l^2) \tilde{d}_l = \alpha^{-2}$. Note that
since $l = i \leq k$ in subcase A3 it holds that $s_l > 1/2$. 

Next, in the subcase (i), (ii), (iii.b) one has 

\begin{eqnarray}
\left.\frac{\partial \, \dot{p}_{lm}}{\partial \, p_{ij}}\right|_{P = 0} &
= & u'_j D u_m s_m + s_l u'_l D u_j \delta_{im} - s_j u'_j D u_m s_m
\nonumber\\ 
& & - s_l u'_j D u_m s_m - s_l^2 u'_l D u_j \delta_{im} - \alpha^{-2} e'_j e_m \nonumber\\
& = & (s_l - s_l^2) u'_l D u_j \delta_{im}  \quad \mbox{(since }u'_j D u_m
  = 0 \mbox{ and } j \neq m \mbox{)}\nonumber\\
& = & 0, \quad \mbox{(A4)}\label{somEq3}
\end{eqnarray}

\noindent because assuming $i \neq m$ or $j \neq l$ each null the
last expression, and $i = m \wedge j = l$ is impossible because
condition (iii.b) would imply $m = j$ contrary to (iii.b).  

The case B concerns all entries $J_P(\mu, \nu)$ with $\mu <
\nu, \nu \leq kN, mod_1(\nu, N) > k.$ Like in case A above,
this yields conditions on the P-matrix indices: (i) $i \leq k$, (ii) $j > k$, 
(iii.a) $l < i$ or (iii.b) $l = i \wedge m < j$.

Again we first treat the subcase (i), (ii), (iii.a). Since $l \neq i$ one has
$e'_l \, I_{ij} = 0$ and eqn.\ (\ref{somEq1}) reduces to the terms
starting with $S$, leading to

\begin{eqnarray}
\left.\frac{\partial \, \dot{p}_{lm}}{\partial \, p_{ij}}\right|_{P = 0} &
= & e'_l\; (SU'_kDUI_{ji}  - SI_{ij}U'DUS - S^2U'_kDUI_{ji} )\; e_m
\nonumber\\
& = & s_l u'_l D u_j \delta_{im} - s_l e'_l I_{ij}U'DUSe_m - s_l^2
u'_l D u_j \delta_{im} \nonumber\\
& = & s_l u'_l D u_j \delta_{im} - s_l^2 u'_l D u_j \delta_{im}\nonumber\\
& = & \left\{ \begin{array}{lll} 0, & \mbox{if } i \neq m & \mbox{(B1)} \nonumber\\ 
(s_l - s_l^2)u'_l D u_j  & \mbox{if } i = m & \mbox{(B2).} \end{array}   \right.\label{somEq4}
\end{eqnarray}

\noindent where $u_l$ is the $l$-th column in $U$ and $\delta_{im} = 1$
if and only if $i = m$ (else $0$) is the Kronecker delta. Note that
since $l < i \leq k$ it holds that $s_l > 1/2$. 

In the subcase (i), (ii), (iii.b) from (\ref{somEq1}) one obtains

\begin{eqnarray}
\left.\frac{\partial \, \dot{p}_{lm}}{\partial \, p_{ij}}\right|_{P = 0} &
= & u'_j D u_m s_m + s_l u'_l D u_j \delta_{im} - s_j u'_j D u_m
s_m\nonumber\\
& & - s_l u'_j D u_m s_m - s_l^2 u'_l D u_j \delta_{im} - \alpha^{-2} e'_j
e_m \nonumber\\
& = & s_m (1 - s_l) u'_j D u_m  + (s_l - s_l^2) u'_l D u_j \delta_{im} \nonumber\\
& = & \left\{ \begin{array}{lll}  s_m (1 - s_l) u'_j D u_m & \mbox{if }
    i \neq m \mbox{ and } m \leq k & \mbox{(B3)}\nonumber\\ 
0 & \mbox{if } i \neq m \mbox{ and } m > k & \mbox{(B4)}\\
s_m (1 - s_l) u'_j D u_m  + (s_l - s_l^2) u'_l D u_j & \mbox{if } i = m &\mbox{(B5)},  \end{array}   \right.\label{somEq5}
\end{eqnarray}
 
\noindent where in the step from the first to the second line one
exploits $j > k$, hence $s_j = 0$; and $m < j$, hence $e'_j e_m = 0$. Note that
since $l = i \leq k$ it holds that $s_l > 0$; that $m > k$ implies
$s_m = 0$ and that $m = i \leq k$
implies $s_m > 0$.

Most of the other cases C -- L listed in Fig.\ \ref{somFig1} divide
into subcases like A and B. The calculations are similar to the ones
above and involve no new ideas. Table \ref{somTable1} collects all
findings. It only shows subcases for nonzero entries of $J_P$. Fig.
\ref{somFig2} depicts the locations of these nonzero areas.

\setlength{\tabcolsep}{2mm}
\renewcommand{\arraystretch}{1.3}
%\begin{footnotesize}
%\begin{centering}
\begin{table}[htbp]
\begin{tabular}{|p{1.5cm}|p{5cm}|p{6cm}|}\hline  
{\bf Subcase} & {\bf Index range }    & {\bf Cell value}\\\hline\hline
A3 & $l = j \leq k$; $i = m \leq k$; & $\alpha^{-2}$\\\hline
B2 & $l < i = m \leq k < j$ & $(s_l - s_l^2)\,u'_lDu_j$\\\hline
B3 & $l = i  \leq k < j$; $m \leq j$; $m \neq i$ &  $s_m (1 - s_l) u'_j D u_m$\\\hline
B5 & $l = i  \leq k < j$; $m \leq j$; $m = i$ & $s_l (1 - s_l) u'_j D u_l  + (s_l - s_l^2) u'_l D u_j$\\\hline
C1 & $l = j \leq k$; $i = m > k$  & $\alpha^{-2}$\\\hline
D1 & $l \leq k$; $i, j > k$; $i = m$ & $(s_l - s_l^2)\,u'_lDu_j$\\
\hline
E1 & $i = m < l = j \leq k$ & $\alpha^{-2}$  \\\hline
F1 & $i = m < l \leq k < j$ & $(s_l - s_l^2)\,u'_lDu_j$\\\hline
J1 & $i = l > k$; $m \leq k < j$ & $s_m u'_j D u_m$\\\hline
K1 & $i = j = l = m \leq k$ & $\alpha^{-2} (1 - 2s_l) / (1 -
    s_l)$ \\\hline
K2 & $i = l$; $j = m > k$; $l \neq m$ & $-\alpha^{-2}$ \\\hline
K3 & $i = l$; $j = m \leq k$; $l \neq m$ & $-\alpha^{-2} \,s_l / (1 -
  s_m) $\\\hline
L1 & $i = l > k$; $m = j > k$ & $-\alpha^{-2}$\\\hline
\end{tabular}
\caption{Values in nonzero areas of the Jacobian
  $J_P$.}\label{somTable1}
 \end{table}
%\end{centering}
%\end{footnotesize}

\begin{figure}[htb]
\center
\includegraphics[width=130 mm]{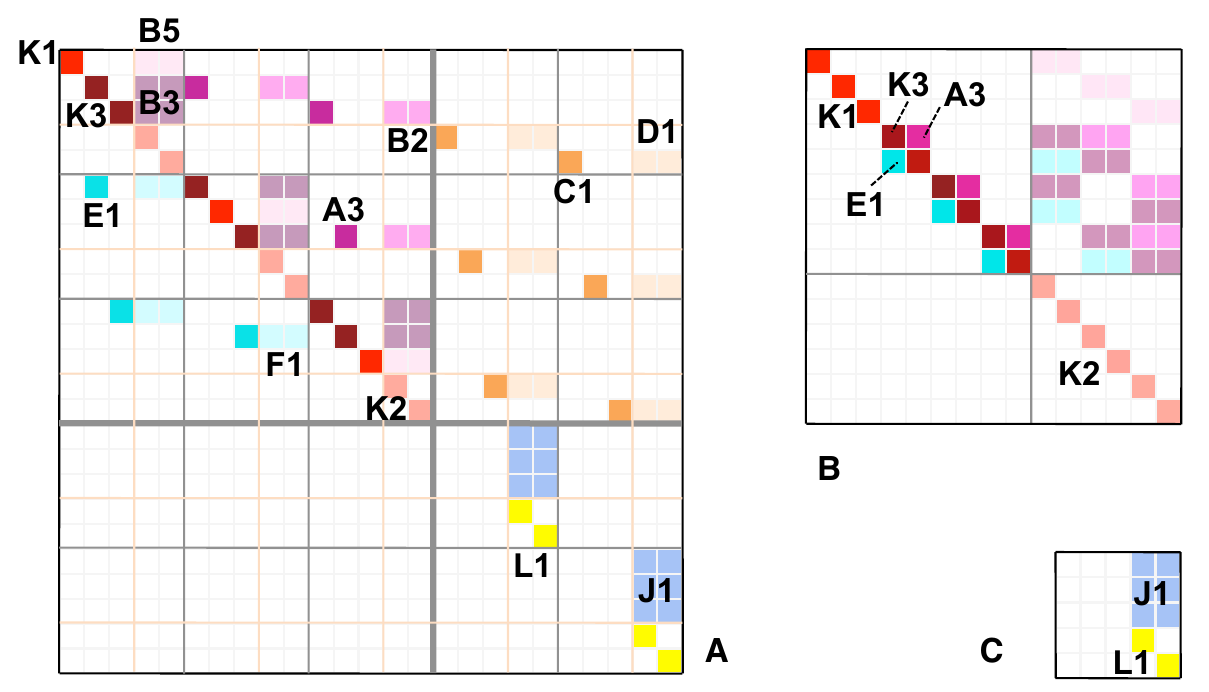}
\caption{{\bf A.} Nonzero areas in the Jacobian
  matrix $J_P$. An instance with $N = 5, k = 3$ is shown. Areas
  denotations correspond to Table 1. Same color = same area. Values:
  areas A3, C1, E1: $\alpha^{-2}$; B2, D1, F1: $(s_l - s_l^2)\,u'_lDu_j$;
  B3: $s_m (1 - s_l) u'_j D u_m$; B5: $s_l (1 - s_l) u'_j D u_l + (s_l
  - s_l^2) u'_l D u_j$; J1: $s_m u'_j D u_m$; K1: $\alpha^{-2} (1 - 2s_l) /
  (1 - s_l)$; K2, L1: $-\alpha^{-2}$; K3: $-\alpha^{-2} \,s_l / (1 - s_m) $.
  {\bf B.} The left upper principal submatrix re-arranged by
  simultaneous row/column permutations. {\bf C.} One of the $N
\times N$ submatrices $C_r$ from the diagonal of the $(N-k)N \times
(N-k)N$ right bottom submatrix of $J_P$. For explanations see text.}
\label{somFig2}
\end{figure}

The eigenvalues of $J_P$ are now readily obtained. First observe that
the eigenvalues of a matrix with block structure

\begin{displaymath}
\left(\begin{array}{cc}K & L \\ 0 & M\end{array} \right),
\end{displaymath}

\noindent where $K$ and $M$ are square, 
are the eigenvalues collected from the principal submatrices
$K$ and $M$. In $J_P$ we therefore only need to consider the leading
$kN \times kN$ and the trailing $(N-k)N \times (N-k)N$ submatrices;
call them $K$ and $M$. 

$M$ is upper triangular, its eigenvalues are
therefore its diagonal values. We thus can collect from $M$ $(N-k)k$
times eigenvalues 0 and $(N-k)^2$ times eigenvalues $-\alpha^{-2}$. 

By simultaneous permutations of rows and columns (which leave
eigenvalues unchanged) in $K$ we can bring it to the form
$K_{\mbox{\scriptsize perm}}$ shown in Fig. \ref{somFig2}{\bf B}. A
block structure argument as before informs us that the eigenvalues of
$K_{\mbox{\scriptsize perm}}$ fall into three groups. There are
$k(N-k)$ eigenvalues $-\alpha^{-2}$ (corresponding to the lower right
diagonal submatrix of $K_{\mbox{\scriptsize perm}}$, denoted as area
K2), $k$ eigenvalues $\alpha^{-2} (1-2s_l)/(1-s_l)$, where $l = 1,\ldots,
k$ earned from the $k$ leading diagonal elements of
$K_{\mbox{\scriptsize perm}}$ (stemming from the K1 entries in $J_P$),
plus there are the eigenvalues of $k(k-1)/2$ twodimensional
submatrices, each of which is of the form

\begin{displaymath}
K_{l,m} = - \alpha^{-2} \left(\begin{array}{cc}s_l / (1 - s_m) & 1 \\ 1 & s_m
    / (1 - s_l)\end{array} \right), 
\end{displaymath}

\noindent where $l = 1, \ldots, k$; $m < l$. By solving the
    characteristic polynomial of $K_{l,m}$ its eigenvalues are
    obtained as 

\begin{equation}\label{somEq6}
\lambda_{1,2} = \frac{\alpha^{-2}}{2} \left( \frac{s_l}{s_m - 1}  +
  \frac{s_m}{s_l - 1} \pm \sqrt{ \left( \frac{s_l}{s_m - 1}  -
      \frac{s_m}{s_l - 1} \right)^2 + 4}  \right). 
\end{equation}

Summarizing, the  eigenvalues of $J_P$ are constituted by the
following multiset:

\begin{enumerate}
\item $k(N-k)$ instances of 0,
\item $N(N-k)$ instances of $-\alpha^{-2}$, 
\item $k$ eigenvalues $\alpha^{-2} (1-2s_l)/(1-s_l)$, where $l = 1,\ldots,
k$,
\item $k(k-1)$ eigenvalues which come in pairs of the form given in
Eqn.\ (\ref{somEq6}).
\end{enumerate}

%% \subsection{Proof of Proposition \ref{propFsimulatesC}, Claim 4 (Section
%%   \ref{secBiolPlausible})} 

%%  Let $(I - 2F)R - \alpha^{-2}F = 0$ and $C = (I - F)^{-1}F$, where
%% $R = E[xx']$. We have to show that $(I-C)R -  \alpha^{-2}C = 0$, or
%% equivalently, inserting $C = (I - F)^{-1}F$, that
%% \begin{displaymath}
%% (I -  (I - F)^{-1}F)\,R - \alpha^{-2}(I - F)^{-1}F = 0.
%% \end{displaymath}
%% Inserting $F = R\,(2R + \alpha^{-2}I)^{-1}$ (claim 1.\ from the
%% proposition), and rearranging terms, this becomes
%% \begin{equation}\label{eqProofProp161}
%% R\,(R+\alpha^{-2}I)^{-1} - \left(I -  R\,(R+\alpha^{-2}I)^{-1}
%% \right)^{-1}\,R\,(R+\alpha^{-2}I)^{-1} = 0.
%% \end{equation}
%% Now exploit that  $R$ is positive semidefinite with
%% SVD $R = U\Sigma U'$. The condition (\ref{eqProofProp161})  is
%% equivalent to 
%% \begin{displaymath}
%% U\,\left(\Sigma\,(\Sigma+\alpha^{-2}I)^{-1} - \left(I -  \Sigma\,(\Sigma+\alpha^{-2}I)^{-1}
%% \right)^{-1}\,\Sigma\,(\Sigma+\alpha^{-2}I)^{-1}\right)\,U' = 0, 
%% \end{displaymath}
%% which can be easily verified using commutativity of products of
%% diagonal matrices. 

\subsection{Proof of Proposition \ref{propEntailICL} (Section
  \ref{sec:ClogicCategoryTheory})} \label{proofRentailment} 

\begin{eqnarray*}
\zeta \models_{\Sigma^{(n)}} \xi & \mbox{iff} & \forall z^{.\wedge 2}: z^{.\wedge 2}
\models_{\Sigma^{(n)}} \zeta \rightarrow z^{.\wedge 2}
\models_{\Sigma^{(n)}} \xi\\
& \mbox{iff} & \forall z^{.\wedge 2}: z^{.\wedge 2} \,.\!\ast\,
(z^{.\wedge 2} + 1)^{-1} \leq \iota(\zeta) \rightarrow z^{.\wedge 2} \,.\!\ast\,
(z^{.\wedge 2} + 1)^{-1} \leq \iota(\xi)\\
& \mbox{iff} & \iota(\zeta) \leq \iota(\xi),
\end{eqnarray*}
where the last step rests on the fact that all vector components of
$\zeta, \xi$ are at most 1 and that the set of vectors of the form
$z^{.\wedge 2} \,.\!\ast\, (z^{.\wedge 2} + 1)^{-1}$ is the set of nonnegative
vectors with components less than 1.

\newpage

\addcontentsline{toc}{section}{References}

%\bibliography{../../lit}

\end{document}